\documentclass{article} %

\usepackage{etoolbox}
\newcommand{\arxiv}[1]{\iftoggle{colt}{}{#1}}
\newcommand{\colt}[1]{\iftoggle{colt}{#1}{}}
\newtoggle{colt}
\global\togglefalse{colt}

\colt{
  \usepackage{times} 
}

\arxiv{  
	\usepackage{natbib}
	\bibliographystyle{plainnat}
	\bibpunct{(}{)}{;}{a}{,}{,}
}

\arxiv{
	\usepackage[letterpaper, left=1in, right=1in, top=1in,
        bottom=1in]{geometry}
        \usepackage{hyperref}
        \hypersetup{colorlinks,citecolor=blue!70!black,linkcolor =
          blue!70!black}       %
        \usepackage{parskip}
      }

\arxiv{
  \usepackage[dvipsnames]{xcolor}         %
}
\colt{
  \PassOptionsToPackage{dvipsnames}{xcolor}
}

\usepackage[utf8]{inputenc} %
\usepackage[T1]{fontenc}    %
\usepackage{url}            %
\usepackage{booktabs}       %
\usepackage{amsfonts}       %
\usepackage{nicefrac}       %
\usepackage{microtype}      %
\usepackage{amsthm}
\usepackage{mathtools}
\usepackage{amsmath}
\usepackage{bm}
\usepackage{bbm}
\usepackage{amssymb}
\usepackage{MnSymbol} %
\usepackage{makecell}
\usepackage{enumitem}
\usepackage{comment}
\usepackage{wrapfig}
\usepackage{colortbl}
\usepackage{setspace}
\usepackage{transparent}
\usepackage{inconsolata}
\usepackage[scaled=.90]{helvet}
\usepackage{xspace}
\usepackage{verbatim}
\usepackage{algorithm}
\usepackage[noend]{algpseudocode}
\newcommand{\multiline}[1]{\parbox[t]{\dimexpr\linewidth-\algorithmicindent}{#1}}
\usepackage{tablefootnote}
\usepackage{pifont}

\usepackage[nameinlink,capitalize]{cleveref}

\makeatletter
\newcommand{\neutralize}[1]{\expandafter\let\csname c@#1\endcsname\count@}
\makeatother

\usepackage{thmtools}
\declaretheorem[name=Theorem,parent=section]{theorem}
\declaretheorem[name=Lemma,parent=section]{lemma}
\declaretheorem[name=Corollary,parent=section]{corollary}

\declaretheorem[name=Assumption, parent=section]{assumption}
\declaretheorem[name=Condition, parent=section]{condition}

\declaretheorem[name=Remark,parent=section]{remark}
\declaretheorem[name=Proposition, parent=section]{proposition}

\usepackage{crossreftools}
\makeatletter
\renewenvironment{proof}[1][Proof]%
{%
	\par\noindent{\bfseries\upshape {#1.}\ }%
}%
{\qed\newline}
\makeatother

\theoremstyle{plain}
\newtheorem{definition}[theorem]{Definition}

\usepackage{xpatch}
\xpatchcmd{\proof}{\itshape}{\normalfont\proofnameformat}{}{}
\newcommand{\proofnameformat}{\bfseries}

\renewcommand{\eqref}[1]{\texorpdfstring{\hyperref[#1]{(\ref*{#1})}}{(\ref*{#1})}}
\crefformat{equation}{#2Eq.\,(#1)#3}
\Crefformat{equation}{#2Eq.\,(#1)#3}

\Crefformat{figure}{#2Figure~#1#3}
\Crefformat{assumption}{#2Assumption~#1#3}

\Crefname{assumption}{Assumption}{Assumptions}

\def\ddefloop#1{\ifx\ddefloop#1\else\ddef{#1}\expandafter\ddefloop\fi}
\def\ddef#1{\expandafter\def\csname bb#1\endcsname{\ensuremath{\mathbb{#1}}}}
\ddefloop ABCDEFGHIJKLMNOPQRSTUVWXYZ\ddefloop
\def\ddefloop#1{\ifx\ddefloop#1\else\ddef{#1}\expandafter\ddefloop\fi}
\def\ddef#1{\expandafter\def\csname b#1\endcsname{\ensuremath{\mathbf{#1}}}}
\ddefloop ABCDEFGHIJKLMNOPQRSTUVWXYZ\ddefloop
\def\ddef#1{\expandafter\def\csname sf#1\endcsname{\ensuremath{\mathsf{#1}}}}
\ddefloop ABCDEFGHIJKLMNOPQRSTUVWXYZ\ddefloop
\def\ddef#1{\expandafter\def\csname c#1\endcsname{\ensuremath{\mathcal{#1}}}}
\ddefloop ABCDEFGHIJKLMNOPQRSTUVWXYZ\ddefloop
\def\ddef#1{\expandafter\def\csname h#1\endcsname{\ensuremath{\widehat{#1}}}}
\ddefloop ABCDEFGHIJKLMNOPQRSTUVWXYZ\ddefloop
\def\ddef#1{\expandafter\def\csname hc#1\endcsname{\ensuremath{\widehat{\mathcal{#1}}}}}
\ddefloop ABCDEFGHIJKLMNOPQRSTUVWXYZ\ddefloop
\def\ddef#1{\expandafter\def\csname t#1\endcsname{\ensuremath{\widetilde{#1}}}}
\ddefloop ABCDEFGHIJKLMNOPQRSTUVWXYZ\ddefloop
\def\ddef#1{\expandafter\def\csname tc#1\endcsname{\ensuremath{\widetilde{\mathcal{#1}}}}}
\ddefloop ABCDEFGHIJKLMNOPQRSTUVWXYZ\ddefloop
\def\ddefloop#1{\ifx\ddefloop#1\else\ddef{#1}\expandafter\ddefloop\fi}
\def\ddef#1{\expandafter\def\csname scr#1\endcsname{\ensuremath{\mathscr{#1}}}}
\ddefloop ABCDEFGHIJKLMNOPQRSTUVWXYZ\ddefloop

\newcommand{\veps}{\varepsilon}

\DeclareMathOperator*{\argmin}{arg\,min} %
\DeclareMathOperator*{\argmax}{arg\,max}

\def\ddef#1{\expandafter\def\csname b#1\endcsname{\ensuremath{\mb{#1}}}}
\ddefloop abcdeghijklmnopqrstuvwxyz\ddefloop

\newcommand{\sss}[1]{{\scriptscriptstyle#1}}
\newcommand{\ind}[1]{^{\sss{(#1)}}}

\DeclarePairedDelimiter{\abs}{\lvert}{\rvert} %
\DeclarePairedDelimiter{\brk}{[}{]}

\DeclarePairedDelimiter{\nrm}{\|}{\|}

\let\P\undefined
\DeclareMathOperator{\En}{\mathbb{E}}

\DeclareMathOperator{\P}{P}

\DeclareMathOperator{\proj}{Proj}

\newcommand{\mb}[1]{\boldsymbol{#1}}
\renewcommand{\bm}[1]{\boldsymbol{#1}}
\newcommand{\wt}[1]{\widetilde{#1}}
\newcommand{\wh}[1]{\widehat{#1}}

\let\underbar\undefined

\makeatletter
\let\save@mathaccent\mathaccent
\newcommand*\if@single[3]{%
	\setbox0\hbox{${\mathaccent"0362{#1}}^H$}%
	\setbox2\hbox{${\mathaccent"0362{\kern0pt#1}}^H$}%
	\ifdim\ht0=\ht2 #3\else #2\fi
}
\newcommand*\rel@kern[1]{\kern#1\dimexpr\macc@kerna}
\newcommand*\widebar[1]{\@ifnextchar^{{\wide@bar{#1}{0}}}{\wide@bar{#1}{1}}}
\newcommand*\underbar[1]{\@ifnextchar_{{\under@bar{#1}{0}}}{\under@bar{#1}{1}}}
\newcommand*\wide@bar[2]{\if@single{#1}{\wide@bar@{#1}{#2}{1}}{\wide@bar@{#1}{#2}{2}}}
\newcommand*\under@bar[2]{\if@single{#1}{\under@bar@{#1}{#2}{1}}{\under@bar@{#1}{#2}{2}}}
\newcommand*\wide@bar@[3]{%
	\begingroup
	\def\mathaccent##1##2{%
		\let\mathaccent\save@mathaccent
		\if#32 \let\macc@nucleus\first@char \fi
		\setbox\z@\hbox{$\macc@style{\macc@nucleus}_{}$}%
		\setbox\tw@\hbox{$\macc@style{\macc@nucleus}{}_{}$}%
		\dimen@\wd\tw@
		\advance\dimen@-\wd\z@
		\divide\dimen@ 3
		\@tempdima\wd\tw@
		\advance\@tempdima-\scriptspace
		\divide\@tempdima 10
		\advance\dimen@-\@tempdima
		\ifdim\dimen@>\z@ \dimen@0pt\fi
		\rel@kern{0.6}\kern-\dimen@
		\if#31
		\overline{\rel@kern{-0.6}\kern\dimen@\macc@nucleus\rel@kern{0.4}\kern\dimen@}%
		\advance\dimen@0.4\dimexpr\macc@kerna
		\let\final@kern#2%
		\ifdim\dimen@<\z@ \let\final@kern1\fi
		\if\final@kern1 \kern-\dimen@\fi
		\else
		\overline{\rel@kern{-0.6}\kern\dimen@#1}%
		\fi
	}%
	\macc@depth\@ne
	\let\math@bgroup\@empty \let\math@egroup\macc@set@skewchar
	\mathsurround\z@ \frozen@everymath{\mathgroup\macc@group\relax}%
	\macc@set@skewchar\relax
	\let\mathaccentV\macc@nested@a
	\if#31
	\macc@nested@a\relax111{#1}%
	\else
	\def\gobble@till@marker##1\endmarker{}%
	\futurelet\first@char\gobble@till@marker#1\endmarker
	\ifcat\noexpand\first@char A\else
	\def\first@char{}%
	\fi
	\macc@nested@a\relax111{\first@char}%
	\fi
	\endgroup
}
\newcommand*\under@bar@[3]{%
	\begingroup
	\def\mathaccent##1##2{%
		\let\mathaccent\save@mathaccent
		\if#32 \let\macc@nucleus\first@char \fi
		\setbox\z@\hbox{$\macc@style{\macc@nucleus}_{}$}%
		\setbox\tw@\hbox{$\macc@style{\macc@nucleus}{}_{}$}%
		\dimen@\wd\tw@
		\advance\dimen@-\wd\z@
		\divide\dimen@ 3
		\@tempdima\wd\tw@
		\advance\@tempdima-\scriptspace
		\divide\@tempdima 10
		\advance\dimen@-\@tempdima
		\ifdim\dimen@>\z@ \dimen@0pt\fi
		\rel@kern{0.6}\kern-\dimen@
		\if#31
		\underline{\rel@kern{-0.6}\kern\dimen@\macc@nucleus\rel@kern{0.4}\kern\dimen@}%
		\advance\dimen@0.4\dimexpr\macc@kerna
		\let\final@kern#2%
		\ifdim\dimen@<\z@ \let\final@kern1\fi
		\if\final@kern1 \kern-\dimen@\fi
		\else
		\underline{\rel@kern{-0.6}\kern\dimen@#1}%
		\fi
	}%
	\macc@depth\@ne
	\let\math@bgroup\@empty \let\math@egroup\macc@set@skewchar
	\mathsurround\z@ \frozen@everymath{\mathgroup\macc@group\relax}%
	\macc@set@skewchar\relax
	\let\mathaccentV\macc@nested@a
	\if#31
	\macc@nested@a\relax111{#1}%
	\else
	\def\gobble@till@marker##1\endmarker{}%
	\futurelet\first@char\gobble@till@marker#1\endmarker
	\ifcat\noexpand\first@char A\else
	\def\first@char{}%
	\fi
	\macc@nested@a\relax111{\first@char}%
	\fi
	\endgroup
}
\makeatother

\newcommand{\alghyperref}[1]{\hyperref[#1]{Alg.~\ref*{#1}}}

\newcommand{\s}{\mathfrak{s}}

\newcommand{\pa}{\parallel}

\newcommand{\Proj}{\mathrm{Proj}}

\newcommand{\W}{W}

\newcommand{\sfrak}{\mathfrak{s}}

\newcommand{\evale}{\mathrm{eval}}

\newcommand{\initd}{\rho_\mathrm{init}}

\newcommand{\thetahat}{\hat{\theta}}

\newcommand{\thetabar}{\bar{\theta}}

\newcommand{\inner}[2]{\langle #1,#2\rangle}
\newcommand{\tr}{\mathrm{Tr}}

\newcommand{\pibar}{\bar{\pi}}

\newcommand{\pd}{\mathbb{S}^{d\times d}_{++}}
\newcommand{\phih}{\phi}
\renewcommand{\bo}{\texttt{b}}
\newcommand{\re}{\texttt{r}}
\newcommand{\ldef}{\vcentcolon=}

\newcommand{\pihat}{\wh{\pi}}
\newcommand{\pistar}{\pi^{\star}}

\let\oldparagraph\paragraph
\renewcommand{\paragraph}[1]{\oldparagraph{#1.}}

\newcommand{\reg}{\texttt{reg}}

\newcommand{\reals}{\mathbb{R}}

\renewcommand{\P}{\mathbb{P}}

\newcommand{\E}{\mathbb{E}}

\newcommand{\nn}{\nonumber} 
\newcommand{\ldotst}{%
	\mathinner{{\ldotp}{\ldotp}}%
}

\newcommand{\unifa}{\pi_\texttt{unif}}

\newcommand{\unif}{\texttt{unif}}

\renewcommand{\a}{\bm{a}}

\newcommand{\x}{\bm{x}}

\newcommand{\supp}{\mathrm{supp}\,}

\newcommand{\wtilde}[1]{\widetilde{#1}}
\newcommand{\wbar}[1]{\widebar{#1}}

\newcommand{\mainalg}{\texttt{Optimistic}\text{-}\texttt{PSDP}}

\newcommand{\Pim}{\Pi}

\newcommand{\traj}{\texttt{traj}}

\renewcommand{\emptyset}{\varnothing}

\newcommand{\algcommentlight}[1]{\textcolor{blue!70!black}{\transparent{0.5}\footnotesize{\texttt{\textbf{//\hspace{2pt}#1}}}}}

\newcommand{\algcommentbiglight}[1]{\textcolor{blue!70!black}{\transparent{0.5}\footnotesize{\texttt{\textbf{/* #1~*/}}}}}

\newcommand{\approxleq}{\lesssim}

\newcommand{\bigoht}{\wt{O}}

\newcommand{\poly}{\mathrm{poly}}
\newcommand{\polylog}{\mathrm{polylog}}

\newcommand{\ee}{\mathbb{E}}

\newcommand{\norm}[1]{\|#1 \|}

\newcommand{\diag}{\mathrm{diag}}
\newcommand{\op}{\mathrm{op}}

\newcommand{\sgn}{\texttt{sgn}}

\newcommand{\bench}{\texttt{bench}}

\newcommand{\framework}{RLLS\xspace}

\newcommand{\Qstar}{Q^\star}
\newcommand{\Vstar}{V^\star}
\newcommand{\Qhat}{\widehat{Q}}
\newcommand{\vphib}{\bar{\varphi}}

\newcommand{\w}{\bm{w}}
\newcommand{\varphih}{\varphi}

\newcommand{\cDhat}{\widehat{\cD}}
\newcommand{\des}{\texttt{D}}

\newcommand{\conc}{\mathrm{hoff}}

\newcommand{\base}{\texttt{Base}}
\renewcommand{\bench}{\texttt{Bench}}
\newcommand{\phiello}{\phi}
\newcommand{\phiell}{\phi}

\newcommand{\varphiell}{\varphi}
\newcommand{\vepsd}{\veps_\texttt{disc}}
\newcommand{\rghi}{\rg}
\newcommand{\pih}{\pi}
\newcommand{\phiho}{\phi}
\newcommand{\rgh}{\rg}
\newcommand{\rgk}{\rg}
\newcommand{\rgell}{\rg}
\newcommand{\pistarh}{\pistar}
\newcommand{\pihatho}{\pihat_{h-1}}
\newcommand{\pihatell}{\pihat_\ell}
\newcommand{\pihath}{\pihat_h}
\newcommand{\piho}{\pi_{h-1}}
\newcommand{\rg}{\mathrm{Rg}^{\texttt{D}}}

\newcommand{\vphibell}{\vphib}
\newcommand{\frakc}{\mathfrak{c}}
\newcommand{\bsigma}{\bm{\sigma}}

\newcommand{\disc}{\texttt{Discretize}}

\Crefformat{line}{#2Line #1#3}

\makeatletter
\let\OldStatex\Statex
\renewcommand{\Statex}[1][3]{%
  \setlength\@tempdima{\algorithmicindent}%
  \OldStatex\hskip\dimexpr#1\@tempdima\relax}
\makeatother

\newcommand{\paragraphi}[1]{\par\noindent\emph{#1.}}
\newcommand{\fakepar}[1]{\paragraph{#1}}

\usepackage[suppress]{color-edits}

\addauthor{zm}{red}
\newcommand{\zm}[1]{\zmcomment{#1}}
\usepackage{autonum}

\arxiv{
  \title{Sample and Oracle Efficient Reinforcement Learning for MDPs with Linearly-Realizable Value Functions}
  \author{Zakaria Mhammedi\\{\small \texttt{mhammedi@google.com}}}
\date{}
\date{}
}

\begin{document}
	
	\maketitle
	
	\begin{abstract}
Designing sample-efficient and computationally feasible reinforcement learning (RL) algorithms is particularly challenging in environments with large or infinite state and action spaces. In this paper, we advance this effort by presenting an efficient algorithm for Markov Decision Processes (MDPs) where the state-action value function of any policy is linear in a given feature map. This challenging setting can model environments with infinite states and actions, strictly generalizes classic linear MDPs, and currently lacks a computationally efficient algorithm under online access to the MDP. Specifically, we introduce a new RL algorithm that efficiently finds a near-optimal policy in this setting, using a number of episodes and calls to a cost-sensitive classification (CSC) oracle that are both polynomial in the problem parameters. Notably, our CSC oracle can be efficiently implemented when the feature dimension is constant, representing a clear improvement over state-of-the-art methods, which require solving non-convex problems with horizon-many variables and can incur computational costs that are exponential in the horizon.

 	\end{abstract}

        \colt{
        }

	 		\tableofcontents

	\section{Introduction}
	\label{sec:intro}

The field of reinforcement learning (RL) is advancing rapidly, making the development of sample-efficient algorithms increasingly important as the dimensionality of modern problems continues to grow. Traditional RL methods used in practice often lack sample complexity guarantees, meaning there are no bounds on the number of interactions with the environment needed to find a near-optimal policy \zm{cite}. Designing RL algorithms with provable guarantees is particularly challenging, especially in applications involving large state and action spaces. While theoretical algorithms that offer such guarantees exist, they often rely on strong assumptions about the environment and are typically computationally infeasible for practical use. In this paper, we take a step toward bridging this gap by focusing on a classical environment assumption that has been widely studied in recent years in the RL context, yet has lacked computationally efficient algorithms.

Specifically, we consider MDPs where the state-action value functions of any policy are linear in some known feature map, which we refer to as linearly  $Q^\pi$-realizable MDPs \citep{lattimore2020learning,du2019good}. This assumption is particularly interesting because it does not impose direct constraints on the dynamics of the MDP, unlike classic linear MDPs \citep{jin2020provably, yang2019sample,yang2020reinforcement}, where the transition operator is assumed to have a low-rank structure. However, with less structure to exploit, learning in linearly $Q^\pi$-realizable MDPs is significantly more challenging. Consequently, these MDPs have primarily been studied in settings where a local simulator is available, allowing the learner to revisit previously encountered state-action pairs \citep{hao2022confident,weisz2021query,li2021sample,yin2022efficient,weisz2022confident}. While sample- and computationally efficient algorithms are known in the local simulator setting, it was not until recently that \cite{weisz2024online} developed a sample-efficient algorithm for linearly $Q^\pi$-realizable MDPs without relying on a local simulator, though this approach is not computationally efficient. \cite{weisz2024online} left open the question of whether a computationally efficient algorithm could be developed for this setting.

\paragraph{Contributions} In this paper, we present an RL algorithm that finds a near-optimal policy in linearly  $Q^\pi$-realizable MDPs, using a number of episodes and calls to a cost-sensitive classification (CSC) oracle over policies that are both polynomial in the problem parameters. Additionally, we show that, due to the nature of the policy class involved, our CSC oracle can be efficiently implemented when the feature dimension is \emph{constant}.\footnote{This means that the computational complexity is polynomial in all parameters except for the feature dimension, where it can be exponential.} This represents a clear improvement over the algorithm by \cite{weisz2024online}, which relies on \emph{global optimism} and requires solving non-convex optimization problems with horizon-many variables that can lead to a computational cost exponential in the horizon. %

Although cost-sensitive classification is NP-hard in the worst case, it can be reduced to binary classification \citep{beygelzimer2009error,langford2005sensitive}, a problem for which many practical algorithms are available and which forms the foundation of empirical machine learning. %
It is also worth noting that certain algorithms based on global optimism such as \texttt{OLIVE} \citep{jiang2017contextual} (which is similar to the algorithm in \citep{weisz2024online}) are known to be incompatible with Oracle-efficient implementations for various common RL Oracles, including the CSC Oracle considered in this paper. This highlights a separation between these computationally intractable algorithms and our approach, which can be implemented in an Oracle-efficient manner.

\paragraph{Related works}
Our work builds on a substantial body of literature in reinforcement learning (RL) with linear function approximation, a foundational framework for understanding which structural assumptions enable the design of sample- and computationally-efficient RL algorithms. A prominent example is the linear MDP framework \citep{jin2020provably}, where the transition operator is assumed to have a low-rank structure. Linear MDPs have been studied extensively, with sample-efficient algorithms developed for both offline and online settings. In contrast, as mentioned earlier, the linear $Q^\pi$-assumption, which generalizes the linear MDP assumption, has mostly been explored in the presence of a local simulator \citep{hao2022confident, weisz2021query, li2021sample, yin2022efficient, weisz2022confident} due to the inherent challenges of the setting.

In the offline RL setting, where the learner is provided with a fixed dataset and cannot interact with the MDP to generate new trajectories, \cite{wang2021exponential} established an exponential-in-horizon sample complexity lower bound for learning in linearly $Q^\pi$-realizable MDPs, even under a feature coverage assumption. Later, \cite{tkachuk2024trajectory} showed that with a stronger concentrability assumption \citep{foster2021statistical}, linearly $Q^\pi$-realizable MDPs become statistically tractable in the offline setting, though no computationally efficient algorithm was provided. In the online RL setting, which is the focus of this paper, \cite{weisz2021exponential} developed a sample-efficient algorithm, but it, too, lacks computational efficiency.

In \cref{sec:additional_related}, we present further related works in the context of RL with linear function approximation. This paper extends the ongoing exploration of structural assumptions that allow for both sample- and computationally-efficient algorithms in RL. We believe our techniques will inspire more efficient algorithms for planning and RL with general function approximation, beyond the linear $Q^\pi$-setting addressed in this paper. 

\paragraph{Paper organization}
The remainder of this paper is organized as follows. In \cref{sec:setup}, we formally introduce the setup considered in this paper and provide preliminary results. \cref{sec:main} presents the main result, focusing on the sample and computational complexity of our algorithm, \mainalg{}. In \cref{sec:alg}, we explore the algorithm in detail, offering both high-level descriptions and key insights into its design. Throughout, we emphasize the challenges inherent in linearly realizable MDPs and discuss the strategies we employ to address them. \cref{sec:sketch} outlines the proof of the main theorem. Finally, in \cref{sec:conclusion}, we discuss the limitations of our approach and explore remaining open questions.

\colt{\vspace{-14pt}}

	\section{Setup and Preliminaries}
	\label{sec:setup}
In \cref{sec:setting}, we outline the reinforcement learning (RL) setting considered in this paper and introduce the notation used throughout. In \cref{sec:prelimrange}, we present key structural results for linearly $Q^\pi$-realizable MDPs, which are essential to our analysis. Finally, in \cref{sec:benchmark}, we describe our computational oracle and its properties.
\subsection{Online Reinforcement Learning in Linearly Realizable MDPs}
\label{sec:setting} 
A Markov
Decision Process (MDP) is a tuple $\cM = ( \cX, \cA, P, \initd, R, H)$, where
$\cX$ is a (large/potentially infinite) state space, $\cA$ is the
action space, which we assume is finite and abbreviate $A=\abs{\cA}$, $H \in \bbN$ is the horizon, $R: \cX \times \cA \rightarrow [0,1]$ is the reward function,  $P
: \cX \times \cA \rightarrow \Delta(\cX)$ is the transition distribution, and $\initd \in \Delta(\cX)$ is the initial state distribution. A (Markovian)
policy $\pi$ is a map $\pi : \cX
\rightarrow \Delta(\cA)$; we use $\Pim$ to denote the set of
all such maps. When a policy is executed, it
generates a trajectory $(\bx_1,\ba_1,\br_1), \dots, (\bx_H, \ba_H, 
\br_h)$ via the process $\ba_h \sim \pi(\bx_h), \br_h \sim
R(\bx_h,\ba_h), \bx_{h+1} \sim P(\cdot\mid\bx_h,\ba_h)$,
initialized from $\bx_1 \sim \initd$ (we use
$\bx_{H+1}$ to denote a terminal state with zero
reward). We write $\bbP^{\pi}\brk*{\cdot}$ and $\En^{\pi}\brk*{\cdot}$
to denote the law and expectation under this process. We assume (following, e.g., \citep{jiang2017contextual, mhammedi2023efficient,weisz2024online}) that the MDP $\cM$ is \emph{layered} so that $\cX = \cX_1\cup \dots\cup  \cX_H$ for $\cX_i \cap \cX_j=\emptyset$ for all $i\neq j$, where $\cX_h\subseteq \cX$ is the subset of states in $\cX$ that are reachable at layer $h\in[H]$.\footnote{\label{foot:layered}On its own, the layered assumption comes with no loss of generality as one can always augment the state by adding information about the current layer. However, for linearly $Q^\pi$-realizable MDPs (the main assumption we make in this paper), layering \emph{does} comes with some loss generality. Tackling linearly $Q^\pi$-realizable MDPs without the layering assumption is an interesting open problem.} 

The goal in an MDP is to find a policy $\pihat \in \Pi$ such that 
\begin{align}
J(\pistar) - \En\brk*{ J(\pihat)} \leq \veps, \label{eq:goal}
\end{align}
 where $J(\pi) \coloneqq \bbE^\pi\brk[\big]{
	\sum_{h=1}^H \br_h}$ is expected reward of policy $\pi$, and $\pistar \in \argmax_{\pi\in \Pi}J(\pi)$ is the optimal policy. Since the state space $\cX$ can be very large or even infinite, achieving the goal in \eqref{eq:goal} efficiently can be very challenging in general without any additional assumptions on the MDP. In this paper, we assume that the MDP is \emph{linear} $Q^\pi$-\emph{realizable.}

  \paragraph{Linearly $Q^\pi$-realizable MDPs} To present our main assumption, we need to define the state-action value functions (or~$Q$-functions); for a policy $\pi\in \Pi$, the $Q$-function $Q_h:\cX_h \times \cA \rightarrow \reals$ at layer $h\in[H]$ is defined as 
\begin{align}
 Q_h^\pi(x,a) \coloneqq \bbE^\pi\brk*{\sum_{\ell=h}^H \br_{\ell} \mid \bx_h=x,\ba_h=a}, \label{eq:laugh}
\end{align}
for all $(x,a)\in \cX_h\times \cA$. Our main assumption in this paper asserts that the $Q$-functions are \emph{linear} with respect to some \emph{known} feature map $\phi: \cX\times \cA\rightarrow \reals^d$.
\begin{assumption}[Lin-$Q^\pi$]
  \label{assum:linearqpi} We assume that for all $h\in[H]$, $\pi \in \Pi$, there exists $\theta^\pi_h\in \bbB_d(H)\subseteq \reals^d$ such that\footnote{Technically, a more natural assumption is to posit the existence of $\theta^\pi_h \in \bbB_d(cH)$ for some $c \geq 1$. We set $c=1$ to simplify the presentation and avoid carrying it throughout. The final sample complexity would include factors of $\poly(c)$, which we omit.} 
  \begin{align}
  \forall (x,a)\in \cX_h\times \cA,\quad Q_h^\pi(x,a) = \phih(x,a)^\top \theta^\pi_h. \label{eq:lin}
  \end{align} 
  Furthermore, the feature map $\phi$ is known to the learner and satisfies $\|\phi(x,a)\|\leq 1$, for all $(x,a)\in \cX\times \cA$.
\end{assumption}
Results of this paper can easily be extended to the setting where the linear $Q^\pi$ assumption holds approximately; that is, when there is a $\veps_\texttt{lin}>0$ such that for all $h\in[H]$ and $\pi \in \Pi$, there exists $\theta^\pi_h\in \bbB_d(H)$ such that 
  \begin{align}
  \forall (x,a)\in \cX_h\times \cA,\quad \|Q_h^\pi(x,a) - \phih(x,a)^\top \theta^\pi_h\| \leq \veps_\texttt{lin}. \label{eq:linapprox}
  \end{align}
  However, to simplify the presentation, we focus on the exact linear $Q^\pi$ assumption in \cref{assum:linearqpi}. 
  
  We refer to an MDP that satisfies \cref{assum:linearqpi} as \emph{linearly $Q^\pi$-realizable}. Linearly $Q^\pi$-realizable MDPs are particularly interesting because, unlike many other common assumptions in the RL context, the assumption in \eqref{eq:lin} (or \eqref{eq:linapprox}) does not directly impose any constraints on the dynamics (which are captured by the transition operator $P$) of the MDP. For instance, in classic linear MDPs, the transition operator $P$ is typically assumed to have a low-rank structure.%

\paragraph{Online Reinforcement Learning} To achieve the goal in \eqref{eq:goal}, we consider the standard \emph{online} reinforcement learning framework, where the learner/algorithm repeatedly interacts with an (unknown) MDP\footnote{By an unknown MDP, we mean that the learner does not know the transition operator $P$ or the reward function $R$.} by executing a policy and observing the resulting trajectory, with the goal of maximizing the total reward. Formally, for each episode $\tau\in\brk{N_\texttt{episodes}}$, the learner
selects a policy $\pi\ind{\tau} = \{\pi_h\ind{\tau}\}_{h=1}^H$, executes it
in the underlying MDP $\cM$ and observes the trajectory
$\{(\bx_h\ind{\tau},\ba_h\ind{\tau},\br_h\ind{\tau})\}_{h=1}^H$. The goal is to achieve \eqref{eq:goal} with as few episodes of interaction with the MDP as possible; we say that an algorithm is \emph{sample-efficient} if the number of trajectories it requires to find $\pihat$ satisfying \eqref{eq:goal} under \cref{assum:linearqpi} is $N_\text{episodes} = \poly(d, A, H, \veps^{-1})$ (where $d$ is the dimension of the feature $\phi$) with no dependence on the state space $\cX$, which can be very large or infinite.

As mentioned earlier, \cite{weisz2024online} has already developed a sample-efficient, though not computationally efficient, algorithm for linearly $Q^\pi$-realizable MDPs. In this paper, we introduce a sample-efficient algorithm for this setting that makes a polynomial number of calls to a cost-sensitive classification oracle. Moreover, we show that this oracle can be implemented efficiently when the feature dimension is constant. The results of this paper bring us one step closer to understanding the computational complexity of sample-efficient online reinforcement learning in linearly $Q^\pi$-realizable MDPs.

\paragraph{Additional notation} In what follows, we let $V_h^\pi(x) \coloneqq \bbE^\pi\brk*{\sum_{\ell=h}^H \br_{\ell} \mid \br_h=x}$ denote the state value function at layer $h\in[H]$.
We denote by $\pi^\star\arxiv{ = \{\pi^\star_h\}_{h=1}^H}$ the optimal
deterministic policy that maximizes $Q^{\pistar}$\arxiv{ at all states}, and
write $\Qstar\ldef{}Q^{\pistar}$ and $\Vstar\ldef{}V^{\pistar}$. We denote by 
\begin{align}
  \Theta_h \coloneqq \{\theta^\pi_h \in \reals^d \mid \pi \in \Pi\}, \label{eq:Thetah}
\end{align}
the set of all parameter vectors corresponding to state-action value functions, where $\theta^\pi_h$ is as in \cref{assum:linearqpi}.
To simplify notation, we let $\varphi(\cdot,a,a') \coloneqq \phi(\cdot,a)-\phi(\cdot,a')$, for $a,a'\in \cA$, and for any matrix $W\in \reals^{d\times d}$ define:
\begin{align}
\varphiell(x; W)   = \W (\phi(x,a_x) - \phi(x,a_x')), \quad  \text{with}
\quad         (a_x,a_x')   \in \argmax_{(a,a')\in \cA^2} \|\W (\phi(x,a) - \phi(x,a'))\|. \label{eq:varphi0}
\end{align}
For any $m,n \in\mathbb{N}$, we denote by $[m\ldotst{}n]$
the integer interval $\{m,\dots, n\}$. For any sequence of objects $o_1, o_2,\dots$, we define $o_{m:n}\coloneqq (o_{i})_{i\in[m \ldotst n]}$. We also let $[n]\coloneqq [1\ldotst{}n]$. %
We use $\bigoht(\cdot)$ to denote a bound up to factors polylogarithmic in
parameters appearing in the expression. We use the notation $a \propto b$ to mean that there are absolute constants $c,C>0$ such that $c a\leq b\leq C a$. We define $\unifa\in\Pim$ as the random policy that
selects actions in $\cA$ uniformly\arxiv{ at random at each layer}. For $1\leq t\leq h\leq H$ and any pair of policies $\pi,\pi' \in \Pi$, we define $\pi \circ_t \pi'\in\Pi$ as the policy given by $(\pi \circ_t \pi')(x_{\ell}) = \pi(x_{\ell})$ for all $\ell<t$ and $(\pi \circ_t \pi')(x_{\ell}) = \pi'(x_{\ell})$ for all $\ell \in [t\ldotst H]$. For the analysis only, we let $\x_0$ denote a fictitious state such that $\phi(\x_0,a)=0$, for all $a\in \cA$. 

We use $\nrm*{\cdot}$ to denote the Euclidean norm in $\reals^d$, and let $\bbB_d(r)\subseteq\bbR^{d}$ denote the Euclidean ball of radius $r$; we drop the $d$ subscript when the dimension is clear from the context. We let $\|\cdot\|_{\op}$ denote the matrix Operator norm. For a positive semi-definite matrix $A\in \reals^{d\times d}$, we write $\|x\|_A \coloneqq \sqrt{x^\top A x}$ for $x\in \reals^d$. We denote by $\pd$ the set of positive definite matrices in $\reals^{d\times d}$. For a symmetric matrix $A\in \reals^{d\times d}$ and $c>0$, we denote by $\cS(A,c)$ the linear subspace spanned by the eigenvectors of $A$ corresponding to eigenvalues that are at least $c$. Given a subspace $S \subseteq \reals^{d}$, we denote by $\proj_{S}: \reals^d \rightarrow S$ the orthogonal projection operator onto $S$. We use $\dagger$ to denote the Moore-Penrose pseudo-inverse. Finally, we use $\otimes$ to denote the tensor product of vectors/matrices; for a vector $v\in \reals^d$ and matrix $M\in \reals^{d\times d}$, we let $v \otimes M\in \reals^{d\times d \times d}$ be the tensor that satisfies $(v \otimes M)_{i,j,k} = v_i M_{j,k}$, for all $i,j,k\in[d]$. 

\paragraph{Argmax tie-breaking} Whenever we write $\pi(\cdot) = \argmax_{a \in \cA} \phi(\cdot,a)^\top \theta$, for some $\theta \in \reals^d$, ties in the argmax are broken by choosing the action with the smallest index.

\colt{\framework acts as a happy medium between online RL, which is
  realistic but often intractable, and global simulator access (where the
  agent can query transitions for \emph{arbitrary}
state-action pairs) \citep{kearns1998finite,kakade2003sample,du2019good,yang2020reinforcement,lattimore2020learning}, which is
powerful, yet unrealistic.
}

\subsection{Range of States and Valid Preconditioning} 
\label{sec:prelimrange}
In this section, we introduce key concepts and structural results for linearly $Q^\pi$-realizable MDPs. Most of the results presented here are based on the works of \citep{weisz2024online,tkachuk2024trajectory}. The proofs for these statements can be found in \cref{sec:proofsstructural}.

Central to our analysis is the notation of \emph{range of states}.
\begin{definition}[Range of state]
  \label{def:range}
  For $h\in [H]$, the range of a state $x\in \cX_h$ is defined as
\begin{align}
\texttt{Rg}(x) \coloneqq \sup_{a,a'\in\cA}\sup_{\theta_h \in \Theta_h} \inner{\phih(x,a) - \phih(x,a')}{\theta_h}, \label{eq:range}
\end{align} 
where we recall that $\Theta_h$ is the set of all parameter vectors corresponding to $Q$-functions; see \eqref{eq:Thetah}.
\end{definition}
The key insight from \cite{weisz2024online} is that states with low range, such as those where $\texttt{Rg}(x)\leq O(\veps)$, are not particularly significant when it comes to learning an $O(\veps)$-optimal policy in a linearly $Q^\pi$-realizable MDP. More formally, for any two policies $\pi$ and $\pi'$, if we define $\tilde\pi(\cdot) = \mathbb{I}\{\texttt{Rg}(\cdot) \geq \veps\} \cdot \pi(\cdot) + \mathbb{I}\{\texttt{Rg}(\cdot) < \veps\}  \cdot \pi'(\cdot)$, then $J(\pi)\leq J(\tilde\pi)+O(\veps)$. This means that the actions a policy takes in low-range states have minimal impact. What is more, \cite{weisz2024online, tkachuk2024trajectory} show that if one has access to the range function $\texttt{Rg}$, there is a way in which one can ``skip'' over low-range states to reduce a linearly $Q^\pi$-realizable MDP to a linear MDP. 

The challenge, of course, is that the range function $\texttt{Rg}$ is not known to the learner, and we need to use some proxy for it. To introduce the proxy we use in this paper, we must first define a few key concepts. We begin with the notion of the \emph{design range} of a state, which restricts the supremum over $\theta$ in \eqref{eq:range} to an approximate design for the set $\Theta_h$.

\label{sec:prelims}
\begin{definition}[Approximate design]
  \label{def:approxdesign}
  Let $\cC= \{ c^z\}_{z\in \cZ}\subseteq \reals^d$ be a set indexed by an abstract set $\cZ$.
A distribution $\rho \in \Delta(\cZ)$ such that $|\supp \rho|<\infty$ is an approximate optimal design for $\cC$ if 
\begin{align}
  \sup_{c \in \cC} \|c\|^2_{\mathrm{G}(\rho)^{\dagger}} \leq 2d,\quad \text{where}\quad \mathrm{G}(\rho)\coloneqq \sum_{z \in \supp \rho} \rho(z) c^z (c^z)^\top.
\end{align}
\end{definition}
\paragraph{Approximate design for $\Theta_h$}
\label{para:special_paragraph}
For $h\in[H]$, let $\rho_h\in \Delta(\Pi)$ be an approximate optimal design for $\Theta_h=\{\theta^\pi_h \mid \pi \in \Pi\}$ with $|\supp \rho_h|\leq \tilde{d} \coloneqq 4 d \log \log d +16$; such a distribution $\rho_h$ is guaranteed to exist by \cite[Part (ii) of Lemma 3.9]{todd2016minimum}. For the rest of the paper, we fix such an approximate design $\rho_h$. With this, we now define the notation of design range.

\begin{definition}[Design range]
  \label{def:designrange}
  Let $h\in [H]$ and $\rho_h\in \Delta(\Pi)$ be the approximate optimal design just defined. The design range of $x\in \cX_h$ is defined as
  \begin{align}
    \rgh(x) \coloneqq \sup_{a,a'\in\cA}\max_{\pi \in \supp \rho_h} \inner{\phih(x,a) - \phih(x,a')}{\theta^{\pi}_h}. \label{eq:everybody}
  \end{align} 
  \end{definition}
The design range $\rg(\cdot)$ provides a good approximation of the range $\texttt{Rg}(\cdot)$ in the sense that $\rg(\cdot) \propto \texttt{Rg}(\cdot)$. In fact, we trivially have that $\rg(\cdot) \leq \texttt{Rg}(\cdot)$. And, as the next lemma shows, we also have $\texttt{Rg}(x)  \leq \sqrt{2d} \cdot \rgh(x)$. %
  \begin{lemma}[Restatement of Proposition 4.5 in \citep{weisz2024online}]
  \label{lem:rangerel}
  For $h\in[H]$ and $x\in \cX_h$, $\texttt{Rg}(x)  \leq \sqrt{2d} \cdot \rgh(x)$.
  \end{lemma} 
  We say that a function is \emph{admissible} if it is dominated by the design range.
  \begin{definition}[Admissibility]
    \label{def:admissible}
  For $h\in[H]$ and $\alpha >0$, a function $F: \cX_h \rightarrow \reals$ is $\alpha$-admissible if for all $x\in \cX_h$, $F(x)\leq \rgh(x)/\alpha$.
  \end{definition}
Admissibility plays a crucial role in the analysis of our algorithm. The following lemma, which restates \citep[Lemma 4.1]{weisz2024online}, shows that the conditional expectation of an admissible function is \emph{linear} in the feature map $\phi$. While in a linear MDP the conditional expectation of any function is linear in $\phi$, this is not always the case for linearly $Q^\pi$-realizable MDPs \citep{weisz2024online}
\begin{lemma}[Admissible realizability]
    \label{lem:admreal}
  For $h\in[H]$ and $\alpha >0$, if $f: \cX_h \rightarrow \reals$ is $\alpha$-admissible (\cref{def:admissible}), 
then for all $\ell \in[h-1]$ and $\tilde\pi \in \Pi$, there exists $\theta \in \bbB(4 \tilde d H/\alpha)$, where $\tilde{d}\coloneqq 4 d \log \log d + 16$, such that for all $(x,a)\in \cX_\ell\times \cA$, \[\E^{\tilde\pi}[f(\x_h)\mid\x_\ell=x,\a_\ell =a] = \phih(x,a)^\top \theta.\]
\end{lemma}
Algorithms for the classical MDP setting crucially rely on the linearity of conditional expectations to efficiently learn a near-optimal policy; this linearity enables effective exploration and planning. Our algorithm will also need to leverage this linearity. The challenge in our setting is that only admissible functions are guaranteed to have linear conditional expectations as reflected in \cref{lem:admreal}. The next lemma essentially shows how any non-admissible function can be transformed into an admissible one by ``ignoring'' low-range states, provided one has access to the design range function \(\rg(\cdot)\).

\begin{lemma}
  \label{lem:admissible}
  Let $\ell\in[H]$, $L>0$, $\gamma>0$, and $f: \cX_\ell\rightarrow [-L,L]$ be given, and let $F$ be the function defined as \begin{align}F(x) = \mathbb{I}\{\rgell(x) \geq \gamma \} \cdot f(x), 
  \end{align} 
  for all $x\in \cX_\ell$. Then, $F$ is $\alpha$-admissible with $\alpha = \gamma/L$; that is, for all $x\in \cX_\ell$, $F(x) \leq L \rgell(x)/\gamma$. 
  \end{lemma}
  \begin{proof}
    Fix $x\in \cX_\ell$. If $\rgell(x) <\gamma$, then $F(x)=0\leq \rgell(x)/\alpha$. Now, if $\rgell(x)\geq \gamma$, then $F(x) = f(x)\leq L \leq L\cdot \rgell(x)/\gamma$, which completes the proof. 
\end{proof}

Fortunately, as discussed right after \cref{def:range} in the prequel, the cost of ``ignoring'' low-range states is minimal when it comes to finding a near-optimal policy. The remaining challenge is that we do not have direct access to the range function $\rgell$ and need to rely on a proxy. Before introducing this proxy, we present a generalization of \cref{lem:admreal} that involves ``skipping'' over intermediate low-range states, which will be crucial to our analysis.\footnote{We thank Gell\'ert Weisz for a discussion that led to the realization of the lemma's result.} 
\begin{lemma}
  \label{lem:calm}
  Let $\ell \in[H]$, $L>0$, $\gamma>0$, and $f: \cX_\ell\rightarrow [-L,L]$ be given. Then, for any $h\in[\ell-1]$ and $\pi\in\Pi$, there exists $\theta\in \bbB(4 \tilde{d} H^2 L/\gamma)$, where $\tilde{d}\coloneqq 4 d \log \log d + 16$, such that for all $(x,a)\in \cX_h\times \cA$:
  \begin{align}
  \E^{\pi}\left[ \mathbb{I}\{\rgell(\x_\ell) \ge \gamma\} \prod_{k=h+1}^{\ell-1} \mathbb{I}\{\rgk(\x_k)<\gamma\} \cdot f(\x_\ell) \mid \x_h =x, \a_h =a \right] = \phih(x,a)^\top \theta.
  \end{align}
\end{lemma}

\paragraph{Proxy for the range} 
One more concept is required before we can introduce our proxy for the range.
\begin{definition}[Valid preconditioning]
  \label{def:precond}
  For $\nu>0$, a matrix $W_h$ is a valid $\nu$-preconditioning for layer $h\in[H]$ if there exists $k\in \mathbb{N}$ and vectors $(w_i)_{i \in[k]} \subset \reals^d$ such that $ W_h = \left(H^{-2} I + \sum_{i=1}^k w_i w_i^\top\right)^{-1/2}$ and for all $i \in [k]$
  \begin{gather}
    \sup_{\theta \in \Theta_h} |\theta^\top w_i|\leq 1, \qquad \left\|\left(H^{-2} I  + \sum_{j\in[i-1]} w_j w_j^\top\right)^{-1/2} w_i  \right\|^2  \geq \frac{1}{2}, \qquad \text{and} \qquad \|w_i\| \leq \nu^{-1}. 
  \end{gather}
  \end{definition}
  As reflected by the statement of the next lemma,  any valid preconditioning can be seen as parameterizing an ellipsoid that encapsulates the set $\Theta_h$; the more accurately this ellipsoid approximates $\Theta_h$, the better we can approximate $\rg(\cdot)$. 
  \begin{lemma}
    \label{lem:fanc}
     Let $h\in [H]$ be given. For $\nu >0$, let $W_h\in \reals^{d\times d}$ be a valid $\nu$-preconditioning for layer $h\in[H]$ (\cref{def:precond}). Then, we have 
     \begin{align}
        \sup_{\theta \in \Theta_h} \|\theta\|^2_{W_h^{-2}}  = \sup_{\theta \in \Theta_h} \|W_h^{-1}\theta\|^2 \leq 5 d \log(1 + 16 H^4 \nu^{-4}).
     \end{align}    
    \end{lemma}
  In light of this, given a valid precondining matrix $W_h$ and some small parameter $\mu>0$, we will essentially use the function $\|\varphi(\cdot;W_h)\|$, where $\varphi$ is as in \eqref{eq:varphi0}, as a proxy for the design range $\rgell$. The next lemma shows that a one-sided inequality always holds between the two. 
  \begin{lemma}
    \label{lem:subset}
    Let $\ell\in[H]$ be given. For $\nu>0$, let $W_\ell\in \reals^{d\times d}$ be a valid $\nu$-preconditioning for layer $\ell\in[H]$ (see \cref{def:precond}). Then, we have
    \begin{align}
     \forall x\in \cX_\ell, \quad  \rgh(x) \leq \sqrt{d_\nu }\cdot \|\varphi(x; W_\ell)\|,
    \end{align}
    where $d_\nu \coloneqq 5 d \log(1+16 H^4 \nu^{-4})$ and $\varphi(\cdot;W_\ell)$ is as in \eqref{eq:varphi0}.
    \end{lemma}
    The reverse side of the inequality does not necessarily hold; if it does, meaning that $\|\varphi(\cdot;W_\ell)\| \propto \rg(\cdot)$, then for any bounded function $f$ over $\cX_\ell$, the map $\mathbb{I}\{\|\varphi(\cdot;W_\ell)\|\geq \mu\} \cdot f(\cdot)$ would be admissible (as can be shown through a similar proof as that of \cref{lem:admissible}). However, even if $\|\varphi(\cdot;W_\ell)\|$ is not proportional to $\rg(\cdot)$, we can still use $\|\varphi(\cdot;W_\ell)\|$ as an effective proxy for the design range by adopting the following strategy, which we detail in the next section:
    \begin{enumerate}
        \item We use \(\|\varphi(\cdot;W_\ell)\|\) as a proxy for the design range until we encounter a function \(f\) such that the map \(g_f: x \mapsto \mathbb{I}\{\|\varphi(x;W_\ell)\|\geq \mu\} \cdot f(x)\) is not admissible. We can ``witness'' the non-admissibility of \(g_f\) if, for example, we identify a layer \(h \in [\ell-1]\) and a policy \(\pi\) for which \(\inf_{\theta}|\E^{\pi}[g_f(\x_\ell) - \phi(\x_h,\a_h)^\top \theta]|\) is too large—this quantity should be small for an admissible function (see \cref{lem:admreal}).
      
        \item When we witness non-admissibility, we can compute a non-zero preconditioning vector \(w_\ell\) (as detailed in the next section) such that the new matrix \((W^{-2}_\ell + w_\ell w_\ell^\top)^{-1/2}\) is also a valid preconditioning matrix. We then use this updated matrix in step 1 and repeat the process.
    \end{enumerate} The next lemma shows that we can update the preconditioning matrix at most \(\wtilde{O}(d)\) times; in other words, non-admissibility can only be witnessed \(\wtilde{O}(d)\) times. At a high level, as long as non-admissibility is not observed for certain functions of interest (such as bonus functions in our case), $\|\varphi(\cdot;W_h)\|$ will serve as a reliable proxy for the design range \(\rg(\cdot)\).
  \begin{lemma}[Length of a preconditioning]
    \label{lem:precond}
    For $\nu >0$, let $W_h\in \reals^{d\times d}$ be a valid $\nu$-preconditioning for layer $h\in[H]$, and let $k\in \mathbb{N}$ be the length of the corresponding sequence $(w_i)_{i \in[k]}$; see \cref{def:precond}. Then, $k\leq 4 d \log (1 + 16 \nu^{-4} H^4)$. 
    \end{lemma}

\subsection{Benchmark Policies and Computational Oracle}
\label{sec:benchmark}
In this subsection, we describe the computational Oracle our algorithm requires. For this, we need to introduce a class of benchmark policies.
  \paragraph{Benchmark policies}
  \label{par:bench} We consider the set $\Pi_\bench$ of benchmark policies such that $\pi\in \Pi_\bench$ if and only if there exist $\gamma>0$, $\pi',\pi'' \in \Pi_\base \coloneqq \left\{ x\mapsto \pi(x;\theta) = \argmax_{a\in \cA} \theta^\top \phih(x,a)  \mid  \theta \in \bbB(H)   \right\}$ and $\theta_{1}, \dots, \theta_{\tilde{d}}\in \bbB(H)$ with $\tilde{d} = 4 d \log \log d +16$ such that
  \begin{align}
      \pi(\cdot) = \mathbb{I}\left\{\max_{a,a'\in \cA, i\in[\tilde{d}]} \varphih(\cdot,a,a')^\top\theta_i \ge \gamma \right\} \cdot \pi'(\cdot) +  \mathbb{I}\left\{\max_{a,a'\in \cA,i\in[\tilde{d}]} \varphih(\cdot,a,a')^\top\theta_i < \gamma \right\} \cdot \pi''(\cdot), \label{eq:benchpol}
  \end{align}
  where $\varphi(\cdot,a,a') \coloneqq \phi(\cdot, a)- \phi(\cdot,a')$. Note that from the definition of the design range in \cref{def:designrange}, there exist $\theta_{1},\dots, \theta_{\tilde{d}}\in \bbB(H)$ such that $\rgh(\cdot) = \max_{a,a'\in \cA,i\in[\tilde{d}]}\inner{\phi(\cdot,a)-\phi(\cdot,a')}{\theta_i}$, and so we have that 
  \begin{align}
  \forall \pi',\pi''\in \Pi_\base,\forall \gamma>0,\quad  \mathbb{I}\{\rgh(\cdot)<\gamma\} \cdot \pi'(\cdot) + \mathbb{I}\{\rgh(\cdot)\ge \gamma \}
 \cdot \pi''(\cdot) \in \Pi_\bench. \label{eq:talk}
\end{align}
The \emph{growth function} of the policy class $\Pi_\bench$ at layer $h\in[H]$ is defined as 
\begin{align}
  \cG_h(\Pi_\bench, n) \coloneqq  \sup_{(x\ind{1},\dots, x\ind{n})\in \cX_h^n} \left|\left\{(\pi(x\ind{1}), \dots, \pi(x\ind{n})) \mid \pi \in \Pi_\bench \right\} \right|.
\end{align} 
The growth function of $\Pi_\bench$ is defined as $\cG(\Pi_\bench, n)\coloneqq \max_{h\in[H]} \cG_h(\Pi_\bench,n)$. The growth function is a key concept in the study of statistical generalization \citep{mohri2012}. Since we will be using the benchmark class \(\Pi_\bench\) to learn a good policy, it is essential to bound the growth function of \(\Pi_\bench\) to ensure the sample efficiency of our algorithm. The next lemma shows that the logarithm of the growth function of \(\Pi_\bench\) is polynomial in the problem parameters, which is sufficient to meet our sample efficiency requirements.
\begin{lemma}
  \label{lem:growth}
  For any $n\in \mathbb{N}$, the growth function $\cG(\Pi_\bench, n)$ is at most $(9^2 n A^2/d)^{(d+1)^2}$. 
\end{lemma}
We now describe our computational Oracle and how it uses the class $\Pi_\bench$.  
\paragraph{Computational Oracle} Our algorithm requires a \emph{Cost-Sensitive Classification} (CSC) Oracle over policies in $\Pi_\bench$ such that given any $n\in \mathbb{N}$, $h\in [H]$, and tuples $(c\ind{1},x\ind{1},a\ind{1}),\dots, (c\ind{n},x\ind{n}, a\ind{n})\in \reals \times \cX_h \times \cA$, the Oracle returns 
\begin{align}
  \pi' \in   \argmin_{\pi \in \Pi_\bench} \sum_{i=1}^n c\ind{i} \cdot \mathbb{I}\{\pi(x\ind{i}) =a\ind{i}\}.
\end{align}
Although cost-sensitive classification is NP-hard in the worst case \citep{dann2018oracle}, it can be simplified to binary classification \citep{beygelzimer2009error,langford2005sensitive}, a problem for which numerous practical algorithms are available, forming a foundation of empirical machine learning. Moreover, the CSC oracle has been successfully used in practical algorithms for contextual bandits, imitation learning, and structured prediction \citep{langford2007epoch,agarwal2014taming,ross2014reinforcement,chang2015learning}. Furthermore, we next show that, due to the nature of our benchmark policy class \(\Pi_\bench\), our CSC Oracle can be efficiently implemented when the feature dimension is constant.

\begin{lemma}[Restatement of \cref{lem:comp}]
  \label{lem:comp2}
      Let $n\in \mathbb{N}$, $h\in [H]$, and $(c\ind{1},x\ind{1},a\ind{1}),\dots, (c\ind{n},x\ind{n}, a\ind{n})\in \reals \times \cX_h \times \cA$ be given. Then, for the benchmark policy class $\Pi_\bench$ in \cref{par:bench}, it is possible to find 
      \begin{align}
        \pi' \in   \argmin_{\pi \in \Pi_\bench} \sum_{i=1}^n c\ind{i} \cdot \mathbb{I}\{\pi(x\ind{i}) =a\ind{i}\},
      \end{align}
      in $ O(\poly(n,d,A) \cdot(9n^2 A^2/d)^{(d+1)^2})$ time.
  \end{lemma}

       \section{Main Result: Guarantee of \mainalg}
       \label{sec:main}

\label{sec:guarantee}
The following is the main guarantee of \mainalg{} (the proof is in \cref{sec:fullproof}).
\begin{theorem}
\label{lem:say}
Let $\veps,\delta\in(0,1)$ be given and consider a call to $\mainalg(\Pi_\bench,\veps, \delta)$ (\cref{alg:ops-dp}) with $\Pi_\bench$ as in \cref{par:bench}. Then, with probability at least $1-2\delta$, we have 
\begin{align}
J(\pistar) - J(\pihat_{1:H}) \leq \veps,
\end{align}
where $\pihat_{1:H}$ is the policy returned by \cref{alg:ops-dp}. Further, the number of episodes \cref{alg:ops-dp} uses is at most $\poly(A,H,d,1/\veps, \log(1/\delta))$ and the number of calls to the CSC Oracle in \cref{sec:benchmark} is at most $\wtilde{O}(d^2 H^6)$. 
\end{theorem}
As desired, the sample complexity of \mainalg{} is polynomial in the problem parameters. However, we note that our approach incurs a \(\poly(A)\) factor in the sample complexity compared to the non-computationally efficient method of \cite{weisz2024online}, which is based on global optimism. It is unclear whether this factor can be eliminated when using a local optimism-based approach like ours.

\paragraph{Oracle complexity}
As mentioned earlier, our algorithm requires calls to a CSC Oracle (see \cref{sec:benchmark} for the Oracle's definition). As stated in \cref{lem:say}, the number of Oracle calls made by \mainalg{} does not depend on the desired suboptimality \(\veps\); crucial it does not grow with $O(1/\veps)$. %

\paragraph{Computational complexity and practicality}  
We now revisit a few key points regarding the complexity and practicality of the CSC Oracle (see \cref{sec:benchmark}):
\begin{itemize}
\item The policy optimization Oracle we require can be implemented efficiently when the feature dimension is constant (see \cref{lem:comp2}).
\item While implementing the Oracle is NP-hard in general (for non-constant feature dimension), it can be reduced to binary classification, allowing the use of well-established machine learning algorithms (see \cref{sec:benchmark}). %
\end{itemize} 

\paragraph{Comparison to previous algorithms} We note that our result strictly improves upon those of \cite{weisz2024online} in terms of computational complexity. The algorithm of \cite{weisz2024online} relies on global optimism, which involves solving non-convex optimization problems in \(\reals^{d H}\), leading to a computational complexity exponential the horizon in the worst case. In contrast, the computational complexity of \mainalg{} is polynomial in the horizon, as reflected by \cref{lem:say} and \cref{lem:comp2}. 

It is also worth noting that certain algorithms based on global optimism such as \texttt{OLIVE} \citep{jiang2017contextual} (which is similar to the algorithm in \citep{weisz2024online}) are known to be incompatible with Oracle-efficient implementation for various common RL Oracles, including the CSC Oracle considered in this paper (see \cite{dann2018oracle}). Therefore, \cref{lem:say} separates these computationally intractable algorithms from our algorithm. We leave open the question of whether an algorithm can be developed that is computationally efficient without relying on a computational oracle.

 \section{Algorithm, Intuition, and Challenges}
\label{sec:alg}
In this section, we describe our algorithm, \mainalg, offer some intuition behind its design (\cref{sec:highlevel}), and outline the challenges that motivated our approach (\cref{sec:challenge}). %

\begin{algorithm}[H]
    \caption{
	$\texttt{Optimistic}\text{-}\texttt{PSDP}$: 
	Optimistic Policy Search by Dynamic Programming. 
	}
	\label{alg:ops-dp}
	\begin{algorithmic}[1]
        \setstretch{1.2}
		\Statex[0] 	{\bfseries input:} Policy class $\Pi'$, suboptimality $\veps\in(0,1)$, confidence
		$\delta\in(0,1)$.
        \Statex[0] {\bfseries initialize:} $U\ind{1}_{h}\gets 0$, $W\ind{1}_{h}\gets H^{-2} I$, $\Psi\ind{1}_h\gets (\pi_\unif, 0)$ for all $h\in[0 \ldotst H]$, $\tau' \gets 1, J\gets 0$.
        \State Set parameters $T$, $n_\traj$, $\mu $, $\nu$, $\lambda$, and $\beta$ as in \eqref{eq:params} in \cref{sec:params}.
        \For{$t= 1, \dots, T$}
        \Statex[1] \algcommentbiglight{Fit optimistic values in a dynamic programming fashion.}
        \For{$h=H, \dots, 1$} \label{line:bigbang}
        \State Update $(\thetahat\ind{t}_h, w\ind{t,h}_{h+1:H}) \gets \texttt{FitOptValue}_{h}(\Psi\ind{t}_{h-1}, \pihat\ind{t}_{h+1:H} ,U\ind{t}_{h+1:H}, {W}\ind{t}_{h+1:H};\Pi', \mu,\nu,\lambda,\beta,\veps, \delta, n_\traj)$.  \label{line:fitoptval}
        \State Update $\W\ind{t}_{\ell} \gets ((\W\ind{t}_{\ell})^{-2}+ w\ind{t,h}_{\ell} (w\ind{t,h}_{\ell})^\top)^{-1/2}$, for all $\ell \in[h+1 \ldotst H]$. \label{line:updateprecond}
        \State Set $\pihat\ind{t}_h(\cdot) \gets \argmax_{a\in \cA} \phih(\cdot,a)^\top \thetahat\ind{t}_h$.\hfill \algcommentlight{Breaking ties by picking the smallest index action; see Sec.~\ref{sec:setting}} %
        \EndFor
        \Statex[1] \algcommentbiglight{Update the preconditioning matrices for the next iteration.}
        \State Update $W\ind{t+1}_{1:H} \gets {W}\ind{t}_{1:H}$. 
        \Statex[1] \algcommentbiglight{Compute new design direction and corresponding partial policy.}
        \For{$h=1,\dots, H$} \label{line:setinit}
        \State $(u\ind{t}_h, \tilde\pi_{1:h}\ind{t}, v\ind{t}_h)\gets \texttt{DesignDir}_h(\Psi\ind{t}_{0:h-1}, \pihat\ind{t}_{1:h}, U_h\ind{t}; \beta,n_\traj)$. \label{line:designdir}
        \State Set $U\ind{t+1}_h \gets U_h\ind{t} + u\ind{t}_{h} (u\ind{t}_{h})^\top$ and $\Psi\ind{t+1}_h \gets \Psi\ind{t}_h \cup \{(\tilde\pi\ind{t}_{1:h}, {v}\ind{t}_h)\}$. \label{line:updatePsi}
        \EndFor
       \Statex[1] \algcommentbiglight{Evalutating the policy $\pihat\ind{t}$.}
       \State Compute $J\ind{t} \gets \texttt{Evaluate}(\pihat_{1:H}\ind{t}, n_\traj)$.
       \If{$J < J\ind{t}$}
       \State Set $J \gets J\ind{t}$ and $\tau' \gets t$.
       \EndIf
       \EndFor
		\State \textbf{return} $\pihat_{1:H} =\pihat_{1:H}\ind{\tau'}$.
\end{algorithmic}
\end{algorithm}

\subsection{High-Level Algorithm Description and Intuition}
\label{sec:highlevel}
\label{sec:algorithm_desc}
Our main algorithm, \mainalg{} (\cref{alg:ops-dp}), builds on the classical Policy Search by Dynamic Programming (PSDP) algorithm (see, e.g., \cite{bagnell2003policy}) by incorporating bonuses into the rewards. In a nutshell, the algorithm learns a policy in a dynamic programming fashion by fitting optimistic value functions for each layer $h=H,\dots, 1$.\footnote{The approach of incorporating local optimism into \texttt{PSDP} has also been recently employed by \cite{golowich2024linear} for the linear Bellman complete setting.} It is well known in RL, that adding the right bonuses helps in driving exploration, which is what we use them for in our algorithm. \mainalg{} consists of three subroutines: \texttt{FitOptValue} (\cref{alg:fitoptvalue_p}), \texttt{DesignDir} (\cref{alg:fitbonus_p}), and \texttt{Evaluate} (\cref{alg:evaluate}). Note that \cref{alg:fitoptvalue_p} and \cref{alg:fitbonus_p} are simplified (asymptotic) versions of the full algorithms in \cref{sec:fullversions}; \cref{alg:fitoptvalue} and \cref{alg:fitbonus}, respectively. %

Before delving into the specifics of \mainalg, we first provide an overview of the key variables in \cref{alg:ops-dp}.

\subsubsection{Key Variables in \mainalg{} (\cref{alg:ops-dp})} \mainalg{} runs for $T= \wtilde{O}(d)$ iterations, where at each iteration $t\in[T]$, the algorithm maintains the following variables:
\begin{itemize}
    \item $\Psi\ind{t}_{h}\subset \Pi \times \reals^d$ consists of $t$ policy-vector pairs. The subroutines $\texttt{FitOptValue}_h$ and $\texttt{DesignDir}_h$ within \mainalg{} use the policies in $\Psi\ind{t}_{h}$ to generate trajectories up to layer $h$. Ideally, $\Psi\ind{t}_h$ should contain policies that provide good coverage over the state-action space at layer $h$. Intuitively, the set $\Psi\ind{t}_h$ plays the role of a core set of policies \citep{agarwal2020pc,zanette2021cautiously, du2019provably, wang2021exponential}.

    \item $U\ind{t}_h$ is a ``design matrix'' for layer $h$ consisting of the sum of outer products of (truncated) expected features vectors $\bar\phi^{\pi,v}_h \coloneqq \E^{\pi}[\mathbb{I}\{\phi(\x_h,\a_h)^\top v \geq 0\}\cdot \phi(\x_h,\a_h)]$, for $(\pi, v)\in \Psi\ind{t}_h$; that is,
    \begin{align}
        U\ind{t}_h \approx \sum_{(\pi,v)\in \Psi\ind{t}_h} \bar\phi^{\pi,v}_h (\bar\phi^{\pi,v}_h)^\top.  
    \end{align}
  In the sequel (see \cref{rem:peculiar}), we will elucidate the role of the indicator term $\mathbb{I}\{\phi(\x_h,\a_h)^\top v \geq 0\}$ in the definition of $\bar\phi^{\pi,v}_h$. The matrix $U\ind{t}_h$ is used to define bonus functions that are added to the rewards (as in the next bullet point).
    \item $W\ind{t}_h$ is a valid preconditioning (\cref{def:precond}) for all $t$ (with high probability). The matrix $W\ind{t}_h$ and the design matrix $U\ind{t}_h$ are used to define the bonus function $b\ind{t}_h: \cX_h \rightarrow [0,H]$ as 
    \begin{align}
        b_h\ind{t}(\cdot)= \min\left(H,\frac{\veps}{4H}\max_{a\in \cA} \| \phiell(\cdot ,a)\|_{(\beta I + U\ind{t}_h)^{-1}}\right) \cdot \mathbb{I}\{\|\varphiell(\cdot;W\ind{t}_h)\|\geq \mu\},\label{eq:probono}
    \end{align}
    for some parameter $\mu$ and suboptimality $\veps$.
    \item $\hat\theta\ind{t}_h$ is a parameter vector at layer $h$ used to approximate the optimistic value function at layer $h$: \begin{align}
        Q\ind{t}_h(x,a) \coloneqq Q^{\pihat\ind{t}}_h(x,a) + \E^{\pihat\ind{t}}\left[ \sum_{\ell=h}^H b_\ell\ind{t}(\x_\ell)  \mid \x_h = x, \a_h=a\right]. \label{eq:Qopt}
      \end{align}
      Essentially, \mainalg{} (or more specifically $\texttt{FitOptValue}_h$) ensures that $\phi(\x_h,\a_h)^\top\thetahat\ind{t}_h + b\ind{t}_h(\x_h) \approx  Q\ind{t}_h(\x_h,\a_h)$ in expectation under trajectories generated with policies in $\Psi\ind{t}_{h-1}$ (see \eqref{eq:linreg} below).
      \item $\pihat\ind{t}_{h}(\cdot)= \argmax_{a\in \cA}  \phi(\cdot,a)^\top\hat\theta\ind{t}_h$ represents the policy at layer $h$ in iteration $t$.
\end{itemize}
In each iteration $t \in [T]$, \mainalg{} computes the policies $\pihat\ind{t}_{1:H}$ in a dynamic programming fashion by fitting the optimistic value functions $(Q\ind{t}_h)$ at each layer $h$. The subroutine \texttt{FitOptValue} (informal version in \cref{alg:fitoptvalue_p}), which we describe next, is responsible for fitting these value functions.

\begin{algorithm}[H]
    \caption{Informal (asymptotic) version of
	$\texttt{FitOptValue}_{h}$ (\cref{alg:fitoptvalue}); for the formal version, all expectations are replaced by finite sample estimates, where the input $n$ represents the sample size.} 
	
	\label{alg:fitoptvalue_p}
	\begin{algorithmic}[1]
		\setstretch{1.2}
\Statex[0]{\bfseries input:}  \multiline{$h$, $\Psi_{h-1}$, $\pihat_{h+1:H}$, $U_{h+1:H}$, $W_{h+1:H}$, $\Pi'$, $\mu,\nu, \lambda,\beta,\veps, \delta,n$.}
	\Statex[0]{\bfseries initialize:} For all $\ell\in[h+1\ldotst H]$, $w_\ell\gets 0$. 
    \State Define $\veps'= 2 d c \veps_\reg^{\bo}(\tfrac{\delta}{6HT},\Pi', |\Psi_{h-1}|) + \frac{8cd\nu H A}{\mu \lambda}$, with $\veps_\reg^\bo$ be as in \eqref{eq:vepsb} and $c\coloneqq 20 d \log (1+ 16 H^4 \nu^{-4})$.  \label{line:def_p}
	\State Define bonuses $b_\ell(\cdot)= \min\left(H,\frac{\veps}{4H}\max_{a\in \cA} \| \phiell(\cdot ,a)\|_{(\beta I + U_\ell)^{-1}}\right) \cdot \mathbb{I}\{\|\varphiell(\cdot;W_\ell)\|\geq \mu\}$. \label{line:bonus_p}
    \Statex[0] \algcommentbiglight{Fitting the rewards and bonuses.}
\State Set $\Sigma_{h} \gets  \lambda I +  \sum_{(\pi,v)\in  \Psi_{h-1}}\E^{\pi\circ_h  \pi_\unif \circ_{h+1}\pihat_{h+1:H}} [\phih(\x_{h},\a_{h}) \phih(\x_{h},\a_{h})^\top]$. \label{line:sigmah_p}
\State Set $\thetahat^{\re}_{h} \gets \Sigma_{h}^{-1}\sum_{(\pi,v)\in \Psi_{h-1}}\E^{\pi\circ_h  \pi_\unif \circ_{h+1}\pihat_{h+1:H}} [\phih(\x_{h},\a_{h}) \cdot \sum_{\ell=h}^H \br_{\ell}]$. \label{line:thetare_p}
		\State Set $\thetahat^{\bo}_{h,\ell} \gets \Sigma_{h}^{-1}\sum_{(\pi,v)\in  \Psi_{h-1}}\E^{\pi\circ_h  \pi_\unif \circ_{h+1}\pihat_{h+1:H}} [\phih(\x_{h},\a_{h}) \cdot b_{\ell}(\x_{\ell})]$. \label{line:thetahl_p}
		\State Set $\thetahat_{h}\gets \thetahat_h^{\re} + \sum_{\ell=h+1}^H \thetahat_{h,\ell}^{\bo} \in \reals^d$. \label{line:theta_p}
		\Statex[0] \algcommentbiglight{Check the quality of the linear fit for the bonuses and compute new preconditioning vectors.}
	 \State Define $\Delta_{h, \ell}(\pi,v,\tilde\pi) \gets\E^{\pi\circ_h  \tilde\pi \circ_{h+1}\pihat_{h+1:H}} \left[\mathbb{I}\{\phiho(\x_{h-1},\a_{h-1})^\top v\geq 0\}  \left(b_\ell(\x_\ell) -   \phih(\x_{h},\a_{h})^\top \hat\theta^{\bo}_{h,\ell}\right)\right]$.
	\State Define $\cH \gets \{\ell \in [h+1\ldotst H]:  \max_{(\pi,v) \in \Psi_{h-1}} \max_{\tilde\pi\in \Pi'}|\Delta_{h,\ell}(\pi,v,\tilde\pi)|> \veps'   \}$. \hfill \algcommentlight{$\veps'$ as in \cref{line:def_p}} \label{eq:computation_p}
	\For{$\ell\in \cH$} \label{line:forloop_p}
		\State Define $B_\ell(\cdot) = b_\ell(\cdot)\cdot \|\varphiell(\cdot;W_\ell)\|^{-2}\cdot\varphiell(\cdot;W_\ell) \varphiell(\cdot;W_\ell)^\top \in \reals^{d\times d}$.
		\State 
		Set $\hat\vartheta^{\bo}_{h,\ell} \gets \Sigma_{h}^{-1}\sum_{(\pi,v)\in  \Psi_{h-1}}\E^{\pi\circ_h  \pi_\unif \circ_{h+1}\pihat_{h+1:H}} \left[\phih(\x_{h},\a_{h}) \otimes B_{\ell}(\x_{\ell})\right]\in \reals^{d\times d\times d}$. \label{line:varthetahl_p}
		\State Compute $((\pi_{h,\ell},v_{h,\ell}), \tilde\pi_{h,\ell})\in  \argmax_{((\pi,v),\tilde\pi)\in \Psi_{h-1}\times \Pi'}|\Delta_{h,\ell}(\pi,v,\tilde\pi)| $. \label{eq:piargmax_p}
		\State For $(\pi,v,\tilde\pi)= (\pi_{h,\ell},v_{h,\ell}, \tilde\pi_{h,\ell})$, compute \begin{align}
			z_\ell \gets \argmax_{z\in \bbB(1)} \left|\E^{\pi\circ_h  \tilde\pi \circ_{h+1}\pihat_{h+1:H}} \left[ \mathbb{I}\{\phiho(\x_{h-1},\a_{h-1})^\top v\geq 0\}  \cdot \left(z^\top B_\ell(\x_\ell) z - \hat\vartheta^{\bo}_{h,\ell}[\phih(\x_{h},\a_{h}),z,z]\right)\right]\right|.
		\end{align}
		\State Set $\tilde z_{\ell} \gets \mathrm{Proj}_{\cS(\W_{\ell},\nu)}(z_{\ell})$.
		\State $w_\ell \gets W_\ell^{-1} \tilde z_\ell$. \label{eq:dbelus_p}
		\EndFor
		\State \textbf{return} $(\thetahat_{h},w_{h+1:H})$.
\end{algorithmic}
\end{algorithm}

\subsubsection{\texttt{FitOptValue} (\cref{alg:fitoptvalue_p})}
\label{subsec:fitoptv} In each iteration $t \in [T]$, starting from $h = H$ and progressing down to $h = 1$, \mainalg{} invokes $\texttt{FitOptValue}_h$ with the input $(\Psi\ind{t}_{h-1}, \pihat\ind{t}_{h+1:H}, U\ind{t}_{h+1:H}, W\ind{t}_{h+1:H})$, returning the pair $(\thetahat\ind{t}_h, w\ind{t,h}_{h+1:H})$. The vectors $w\ind{t,h}_{h+1:H}$ are then used in \cref{line:updateprecond} of \cref{alg:ops-dp} to update the preconditioning matrices $(W\ind{t}_{h+1:H})$. The $\texttt{FitOptValue}_h$ subroutine ensures, with high probability, that $(W\ind{t}_{h})$ are valid preconditionings (see \cref{lem:fitoptvalue} and recall the definition of a valid preconditioning in \cref{def:precond}). Consequently, using \cref{lem:precond}—which bounds the maximum length of a sequence of non-zero preconditioning vectors—it can be shown that $w\ind{t,h}_{\ell}$ is non-zero only on $\wtilde{O}(d)$ iterations; these are the iterations where the preconditioning matrix $W\ind{t}_{\ell}$ actually changes (see \cref{line:updateprecond}). Intuitively, $W\ind{t}_\ell$ does not update too frequently.

On iterations $t$ where the preconditioning matrix is not updated (which is the case for most iterations as just argued), $\texttt{FitOptValue}_h$ ensures that $\phi(\cdot, \cdot)^\top\thetahat\ind{t}_h$ is a good approximation of the optimistic value function $Q\ind{t}_h$ in \eqref{eq:Qopt} in expectation under trajectories generated using policies in $\Psi\ind{t}_{h-1}$. More specifically, the subroutine $\texttt{FitOptValue}_h$ guarantees that on most iterations $t\in[T]$: for all $(\pi,v)\in \Psi\ind{t}_{h-1}$ and $\tilde\pi \in \Pi_\bench$ (see \cref{lem:fitoptvalue}),
\begin{align}
  \left|\E^{\pi \circ_h \tilde\pi}\left[\mathbb{I}\{\phiho(\x_{h-1},\a_{h-1})^\top v \geq 0\} \cdot (Q\ind{t}_h(\x_h,\a_h) - \Qhat\ind{t}_h(\x_h,\a_h) ) \right]\right| \leq O\left(\frac{\veps}{8 T H\sqrt{d}}\right), \label{eq:linreg}
\end{align} where $\Qhat\ind{t}_h(x,a) = \phih(x,a)^\top \thetahat\ind{t}_{h} + b\ind{t}_h(x)$, for all $(x,a)\in \cX_h\times \cA$, and $b\ind{t}_h$ is as in \eqref{eq:probono}. The bound in \eqref{eq:linreg} is a core result for the analysis of \mainalg{}. Note that this bound is weaker than the typical least-squares error bounds in reinforcement learning, which bound the squared approximation error of the (optimistic) value function \(Q\ind{t}_h\). However, this bound suffices for our purposes. In our setting, bounding the squared approximation error is not feasible because the bonus term in the definition of $Q\ind{t}_h$ in \eqref{eq:Qopt} is not necessarily linear in the feature vector $\phi(x,a)$; though the term $Q^{\pihat\ind{t}}_h(x,a)$ in the definition of $Q\ind{t}_h$ is linear in $\phi(x,a)$ thanks to \cref{assum:linearqpi}, the second ``bonus'' term $\E^{\pihat\ind{t}}\left[ \sum_{\ell=h}^H b_\ell\ind{t}(\x_\ell)  \mid h = x, \a_h=a\right]$ is not necessarily linear. As will become clearer in the sequel, it is precisely due to this non-linearity that we need \eqref{eq:linreg} to hold for all $\tilde\pi \in \Pi_\bench$; this is also why we require a CSC Oracle over policies in \texttt{FitOptValue} (see \cref{eq:piargmax} in \cref{alg:fitoptvalue}---the full version of $\texttt{FitOptValue}$). We comment on the CSC Oracle calls within \texttt{FitOptValue}{} in \cref{rem:oracle} below. In \cref{rem:peculiar}, we also comment on the presence of the peculiar term $\mathbb{I}\{\phiho(\x_{h-1},\a_{h-1})^\top v \geq 0\}$ in \eqref{eq:linreg}.

The reason \eqref{eq:linreg} is possible at all is because we have multiplied the bonuses by $\mathbb{I}\{\|\varphi(\cdot;W_\ell)\|\geq \mu\}$ (see \eqref{eq:probono}), where, as noted in \cref{sec:prelimrange}, we use $\|\varphi(\cdot;W_\ell)\|$ as a proxy for the design range $\rg(\cdot)$. From \cref{lem:admissible} in \cref{sec:prelimrange}, we know that for any $\gamma>0$ and function $f$, the map $g_f: x\mapsto \mathbb{I}\{\rg(x)\geq \gamma\}\cdot f(\cdot)$ is admissible, which in turn means that the conditional expectation $\E^{\pi}[g_f(\x_\ell)\mid \x_h=x,\a_h = a]$ is linear in $\phi(x,a)$ for any policy $\pi$. And, as already mentioned in \cref{sec:prelimrange}, even though $\|\varphi(\cdot; W_\ell)\|$ may not be a good proxy for the range function at the start of the algorithm, the algorithm updates the preconditioning matrix $W_\ell$ each time we ``witness'' non-admissibility of $\mathbb{I}\{\|\varphi(\cdot;W_\ell)\|\geq \mu\} \cdot f(\cdot)$ for some $f$. And, after each update, $\|\varphi(\cdot; W_\ell)\|$ becomes a better proxy for $\rg(\cdot)$.

 More specifically, we show (see \cref{sec:linearfit} or the proof of \cref{lem:decoy-find-new} for the guarantee of \texttt{FitOptValue}) that for any $\ell\in [h+1\ldotst H]$, any function $f:\cX_\ell \rightarrow [-L,L]$ for $L>0$, and any policy $\pi\in\Pi$, if $W_\ell$ is a valid preconditioning for layer $\ell$, then for some small parameters $\mu, \lambda>0$, one of the following holds: 
\begin{enumerate}
    \item  The discrepancy $\left|\E^\pi\left[ \mathbb{I}\{\|\varphi(\x_\ell; W_\ell)\| \geq \mu\} \cdot f(\x_\ell) - \phi(\x_h,\a_h)^\top \hat\theta^W  \right] \right|$ is small for the standard $\lambda$-regularized least-squares parameter \[\hat\theta^W \approx \Sigma^{-1}_\lambda \E^{\pi}[\mathbb{I}\{\|\varphi(\x_\ell; W_\ell)\| \geq \mu\}  \cdot f(\x_\ell)\cdot \phi(\x_h,\a_h)] \in \reals^d,\] where $\Sigma_\lambda \coloneqq \lambda I + \E^{\pi}[\phi(\x_h,\a_h)\phi(\x_h,\a_h)^\top]$; or 
\item It is possible to compute a non-zero vector $w_\ell\in \reals^d$ such that $(W_\ell^{-2} + w_\ell w_\ell^\top)^{-1/2}$ remains a valid preconditioning matrix for layer $\ell$. Specifically, by letting \[\hat\vartheta^W \coloneqq \Sigma^{-1}_\lambda \E^{\pi}\left[\mathbb{I}\{\|\varphi(\x_\ell; W_\ell)\|\geq \mu\} \cdot f(\x_\ell) \cdot \phi(\x_h,\a_h) \otimes \frac{\varphi(\x_\ell; W_\ell) \varphi(\x_\ell; W_\ell)^\top}{\|\varphi(\x_\ell; W_\ell)\|^2}\right]\in \reals^{d \times d \times d},\] the preconditioning vector $w_\ell$ can be computed by first solving the eigenvalue problems \begin{align} 
    z^{\pm} \in \argmax_{z\in \bbB(1)} \pm \left( z^\top \E^\pi\left[ \mathbb{I}\{\|\varphi(\x_\ell; W_\ell)\|\geq \mu\} \cdot f(\x_\ell) \cdot \frac{\varphi(\x_\ell; W_\ell) \varphi(\x_\ell; W_\ell)^\top}{\|\varphi(\x_\ell; W_\ell)\|^2} \right] z - \hat\vartheta^W[\E^{\pi}[\phi(\x_h,\a_h)], z, z]\right), \label{eq:eigen}\end{align} 
then setting $w_\ell = \proj_{\cS(W_\ell,\nu)}(\bar{z})$ where $\bar{z}$ is the vector in $\{z^-, z^+\}$ with the highest absolute objective value in the previous display. 
\end{enumerate}
In light of \cref{lem:precond}, by repeatedly testing if the discrepancy \(\left|\E^\pi\left[f(\x_\ell) \cdot \mathbb{I}\{\|\varphi(\x_\ell; W_\ell)\|\geq \mu\} - \phi(\x_h,\a_h)^\top \hat\theta^W  \right] \right|\) is small and updating the preconditioning matrix $W_\ell$ if it is not, it is possible to achieve a good linear fit of $f(\cdot) \cdot \mathbb{I}\{\|\varphi(\cdot; W_\ell)\|\geq \mu\}$ (in expectation under $\pi$) after at most $\wtilde{O}(d)$ updates to the matrix $W_\ell$. %

In the context of $\texttt{FitOptValue}_h$, $f$ plays the role of the ``untruncated'' bonus $H \wedge \frac{\veps}{4H}\max_{a\in \cA} \|\phiell(\cdot ,a)\|_{(\beta I + U_\ell)^{-1}}$. By calling $\texttt{FitOptValue}_h$ and updating the preconditioning matrix $\wtilde{O}(d)$ times, \mainalg{} ensures that on \emph{most iterations}, a good linear fit of the truncated bonuses $(b\ind{t}_h)$ is achieved, thereby ensuring that \eqref{eq:linreg} holds. However, the caveat is that due to the truncation term $\mathbb{I}\{\|\varphi(\cdot; W_\ell)\|\geq \mu\}$ the bonuses may become less effective at driving exploration. We address how this issue is managed in the following section. Essentially, states \(x \in \cX_\ell\) for which $\|\varphi(x;W_\ell)\| < \mu$ are low-range states (as demonstrated by \cref{lem:subset} and \cref{lem:rangerel}) and can, to a certain extent, be ignored without affecting the algorithm's ability to explore; as discussed in \cref{sec:prelimrange}, low-range states do not significantly impact the process of finding a near-optimal policy.

\begin{remark}[Calls to CSC Oracle within $\texttt{FitOptValue}$] 
    \label{rem:oracle} In the informal version of $\texttt{FitOptValue}$ (\cref{alg:fitoptvalue_p}) it is not clear how the policy optimization steps in \cref{eq:computation_p} and \cref{eq:piargmax_p} reduce to the cost-sensitive classification problem over $\Pi_\bench$ described in \cref{sec:prelims}. However, the reduction can immediately be seen in the full version of $\texttt{FitOptValue}$ in \cref{alg:fitoptvalue}, where the discrepancy $\Delta_{h,\ell}(\pi,v,\tilde \pi)$ (see \cref{line:delta}) is given by 
    \begin{align}
        \Delta_{h, \ell}(\pi,v,\tilde\pi) = \sum_{(x_{1:H}, a_{1:H}, r_{1:H})\in \cDhat_{h,\pi}}  \frac{A \cdot\mathbb{I}\{\phiho(x_{h-1},a_{h-1})^\top v\geq 0\}}{n}\cdot   \left(b_\ell(x_\ell) -   \phih(x_{h},a_{h})^\top \hat\theta^{\bo}_{h,\ell}\right) \cdot \mathbb{I}\{\tilde\pi(x_h)=a_h\}.
    \end{align}
    Thus, the optimization steps in \cref{eq:computation} and \cref{eq:piargmax} of \cref{alg:fitoptvalue} can be reduced to solving 
    \begin{align}
        \argmax_{\tilde\pi\in \Pi'} \ \ \sfrak \cdot \sum_{(x_{1:H}, a_{1:H}, r_{1:H})\in \cDhat_{h,\pi}} c(x_{1:H},a_{1:H},r_{1:H}; v)  \cdot \mathbb{I}\{\tilde\pi(x_h)=a_h\},
    \end{align}
    for all $(\pi,v)\in \Psi_{h-1}$ and $\sfrak \in\{-1,1\}$, where $c(x_{1:H},a_{1:H},r_{1:H};  v)\coloneqq \frac{A \cdot\mathbb{I}\{\phiho(x_{h-1},a_{h-1})^\top v\geq 0\}  \left(b_\ell(x_\ell) -   \phih(x_{h},a_{h})^\top \hat\theta^{\bo}_{h,\ell}\right)}{n}$. This problem is now clearly of the form that our CSC Oracle in \cref{sec:benchmark} solves.
\end{remark}

\begin{remark}
    \label{rem:peculiar}
The reader may have notice the peculiar term $\mathbb{I}\{\phiho(\x_{h-1},\a_{h-1})^\top v \geq 0\}$ in the guarantee we presented in \eqref{eq:linreg} for the optimistic value function fit. As we will see shortly when discussing $\texttt{DesignDir}$, this term is needed for a change of measure argument in the analysis. We would have not needed it if we could get the guarantee in \eqref{eq:linreg} in a least-square sense, which we cannot under \cref{assum:linearqpi} alone.\footnote{The term $\mathbb{I}\{\phiho(\x_{h-1},\a_{h-1})^\top v \geq 0\}$ is reminiscent of a term introduced by \cite{mhammedi2023efficient} in their ``reward-free'' objective to get rid of the requirement for reachability in low-rank MDPs.} 
\end{remark}
   
\subsubsection{\texttt{Evaluate} (\cref{alg:evaluate})}
Before introducing the second main component of \mainalg{}, the \texttt{DesignDir} subroutine, we first describe the \texttt{Evaluate} subroutine in \cref{alg:ops-dp}. The \texttt{Evaluate} subroutine is used to evaluate the performance of the policy \(\pihat\ind{t}_{1:H}\) at iteration \(t\) by calculating the average sum of rewards across \(n_\traj\) trajectories. This subroutine is invoked at the end of each iteration to evaluate the policy's performance. \mainalg{} ultimately returns the best-performing policy after \(T\) rounds. Our analysis of \mainalg{} relies on showing that for \(T = \Omega(d)\), there will be at least one \(O(\veps)\)-optimal policy among \((\pihat\ind{t}_{1:H})_{t \in [T]}\).
 
\begin{algorithm}[H]
\caption{$\texttt{Evaluate}_{h}$: Evaluate a policy.}
\label{alg:evaluate}
\begin{algorithmic}[1]
\setstretch{1.2}
\Statex[0]{\bfseries input:}  \multiline{$\pihat_{1:H}$, $n$.}
\Statex[0] \algcommentbiglight{Gather trajectory data.}
\State Set $\cD \gets \emptyset$.
\For{$n=1,\ldots,n$} 
\State Sample trajectory $(\bx_1,\a_1,\br_1,\dots, \bx_{H},\a_{H},\br_H) \sim \P^{\pihat_{1:H}}$.
\State Update $\cD \gets \cD \cup \{(\bx_{1:H},\a_{1:H},\br_{1:H})\}$.
\EndFor
\State Set $J \gets \frac{1}{n} \sum_{(x_{1:H},a_{1:H}, r_{1:H})\in \cD }\sum_{h\in[H]} r_h$.
\State \textbf{return} $J$.
\end{algorithmic}
\end{algorithm} 
\begin{algorithm}[H]
    \caption{Informal (asymptotic) version of 
	$\texttt{DesignDir}_{h}$ (\cref{alg:fitbonus}); for the formal version, all expectations are replaced by finite sample estimates, where the input $n$ represents the sample size.
	}
	\label{alg:fitbonus_p}
	\begin{algorithmic}[1]
		\setstretch{1.2}
\Statex[0]{\bfseries input:}  \multiline{$h$, $\Psi_{0:h-1}$, $\pihat_{1:h}$, $U_{h}$, $\beta,n$.}
	\State Set $\kappa_h \gets 0$.
	\For{$i\in [d]$}
    \State Set $v_{h,i} = (\beta I + U_h)^{-1/2}e_i \in \reals^d$.
	\State Set $\pi_{h,i,-}(\cdot) = \argmax_{a\in \cA} -\phih(\cdot, a)^\top v_{h,i}$.
	\State Set $\pi_{h,i,+}(\cdot) = \argmax_{a\in \cA}  \phih(\cdot, a)^\top v_{h,i}$.
    \For{$\ell=0,\dots, h-1$}
    \For{$(\pi,v)\in \Psi_\ell$}
	\State Set $u_{h,i,\ell,\pi,-} \gets \E^{\pi \circ_{\ell+1}\pihat_{\ell+1:h}} \left[\phih(\x_h, \pi_{h,i,-}(\x_h))  \cdot \mathbb{I}\{\phih(\x_h, \pi_{h,i,-}(\x_h))^\top v_{h,i}   \leq  0 \}\right]$. \label{line:minus_p}
	\State Set $u_{h,i,\ell,\pi,+} \gets \E^{\pi \circ_{\ell+1}\pihat_{\ell+1:h}}\left[\phih(\x_h, \pi_{h,i,+}(\x_h)) \cdot \mathbb{I}\{\phih(\x_h, \pi_{h,i,+}(\x_h))^\top v_{h,i}  \geq 0 \}\right]$. \label{line:plus_p}
\State Set $(\mathfrak{s},{u}_{h,i,\ell,\pi}) \in \argmax_{(s,u)\in \{(-, u_{h,i,\ell,\pi,-}),\,(+,u_{h,i, \ell,\pi,+})\}}\ |\inner{u}{v_{h,i}}|.$ \label{line:bar_e_p} 	
\If{$|\inner{{u}_{h,i,\ell,\pi}}{v_{h,i}}| \geq \kappa_h$} \label{line:test3_simp_p} 
\State Set $\kappa_h \gets |\inner{{u}_{h,i,\ell, \pi}}{{v}_{h,i}}|$. %
\State Set $u_h\gets {u}_{h,i,\ell,\pi}$ and ${v}_h \gets \mathfrak{s}\cdot {v}_{h,i}$.
\State Set $\tilde\pi_{1:h}\gets \pi' \circ_{\ell+1} \pihat_{\ell+1:h-1}\circ_h \pi_{h,i,\s}$.
\EndIf
\EndFor
\EndFor
\EndFor
		\State \textbf{return} $(u_h,\tilde{\pi}_{1:h},v_h)$.
\end{algorithmic}
\end{algorithm}

\subsubsection{\texttt{DesignDir} (\cref{alg:fitbonus_p})} \mainalg{} uses \texttt{DesignDir} to update the policy sets $(\Psi\ind{t}_{h})$ and the design matrices $(U\ind{t}_h)$. Notice that in \eqref{eq:linreg}, we bound the difference between $Q\ind{t}_h$ and $\Qhat\ind{t}_h$ in expectation under trajectories generated using policies from $\Psi\ind{t}_{h-1}$. Thus, it is desirable for the policies in $\Psi\ind{t}_{h-1}$ to have good coverage over the state space; informally, policies with good coverage would allow us to transfer the guarantee in \eqref{eq:linreg} to any other policy (i.e., perform a change of measure) with minimal cost. Taking inspiration from the classical linear MDP setting, one way we can measure the quality of the coverage of a set of policies $\{\pi\ind{1}, \dots, \pi\ind{m}\}$ is by looking at the ``diversity'' of the expected feature vectors they induce. For example, suppose the policies $\pi\ind{1}, \dots, \pi\ind{m}$ induce a $G$-optimal design in the space of expected features $\{\E^\pi[\phi(\x_h,\a_h)] \mid \pi \in \Pi\}$; that is, suppose that 
\begin{align}
\forall \pi \in \Pi, 
\quad \E^{\pi}\left[\max_{a\in \cA}\|\phi(\x_h,a)\|_{U^\dagger}\right] \leq \sqrt{2 d}, \quad \text{where} \quad U \coloneqq \sum_{i=1}^m \E^{\pi\ind{i}}[\phi(\x_h,\a_h)] \E^{\pi\ind{i}}[\phi(\x_h,\a_h)^\top].
\label{eq:design}
\end{align} 
Such a set is guaranteed to exist for $m= \wtilde{O}(d)$ (see e.g.~\cite{todd2016minimum}).
In this case, for any vector $\theta\in \reals^d$ and any policy $\pi\in \Pi$, we can perform the following change of measure (see proof of \cref{lem:changeofmeasure2}):
\begin{align}
  |\E^{\pi}[\phi(\x_h,\a_h)^\top \theta]| &\leq \sqrt{d}\cdot \E^{\pi}\left[\max_{a\in \cA}\|\phi(\x_h,a)\|_{U^\dagger}\right] \cdot \sum_{i\in[m]}\left|\E^{\pi\ind{i}}\left[\phi(\x_h,\a_h)^\top \theta\right]\right|,\label{eq:change} \\
  & \leq \sqrt{2}d \cdot \sum_{i\in[m]}\left|\E^{\pi\ind{i}}\left[\phi(\x_h,\a_h)^\top \theta\right]\right|. \quad \text{(by \eqref{eq:design})} \label{eq:ideal}
\end{align}
This implies that if $|\E^{\pi\ind{i}}\left[\phi(\x_h,\a_h)^\top \theta\right]|$ is small for all $i\in[m]$, it will also be small for $|\E^{\pi}\left[\phi(\x_h,\a_h)^\top \theta\right]|$, for \emph{any} $\pi \in \Pi$. Thus, one might hope to construct a set $\Psi\ind{t}_{h-1}$ with policies $\pi\ind{1}, \dots, \pi\ind{m}$ satisfying \eqref{eq:design}, which would allow us to transfer the guarantee in \eqref{eq:linreg} from the policies in \(\Psi\ind{t}_{h-1}\) to any other policy with minimal overhead. However, there are two challenges to achieving this:
\begin{enumerate}
   \item Although a set of policies satisfying \eqref{eq:design} always exists with \(m = \wtilde{O}(d)\), there is no straightforward way to compute such a set in our setting. Even in the much simpler linear MDP setting, finding such a set would require solving a non-convex optimization problem.
    \item Even if \(\Psi\ind{t}_{h-1}\) consisted of policies satisfying \eqref{eq:design}, we would not necessarily be able to perform a change of measure in \eqref{eq:linreg} as we did in \eqref{eq:ideal}. This is because \((Q\ind{t}_h - \Qhat\ind{t}_h)(x,a)\) is not necessarily linear in the feature map $\phi(x,a)$, as it would be in the standard linear MDP setting.\label{item:challenge2}
\end{enumerate}
To address the first challenge, we will follow a ``greedy'' approach to construct the set $\Psi\ind{t}_{h-1}$ and the corresponding design matrix $U\ind{t}_{h-1}$ using the $\texttt{DesignDir}_h$ subroutine. The call to $\texttt{DesignDir}_h$ in \mainalg{} at iteration $t$ returns a tuple $(u\ind{t}_h, \tilde\pi\ind{t}_{1:h},v\ind{t}_h)$ that is used to update $\Psi\ind{t}_h$ and $U\ind{t}_h$ as (see \cref{line:updatePsi}):
\begin{align}
U_h\ind{t+1} \gets U_h\ind{t} + u\ind{t}_h (u\ind{t}_h)^\top \quad \text{and} \quad \Psi\ind{t+1}_h \gets \Psi\ind{t}_h \cup \{(\tilde \pi\ind{t}_{1:h},v_h\ind{t})\}. \label{eq:updatetuple}
\end{align}
Furthermore, the tuple $(u\ind{t}_h, \tilde\pi\ind{t}_{1:h},v\ind{t}_h)$ satisfies, with high probability (see \cref{lem:designbound}),
\begin{gather}
    u_{h}\ind{t}\approx  \E^{\tilde\pi_{1:h}\ind{t}}\left[\phih(\x_h,\a_h) \cdot \mathbb{I}\{\phih(\x_h,\a_h)^\top v\ind{t}_{h}\geq 0\}\right], \label{eq:approxexpr}
    \intertext{and for all $\ell\in[0 \ldotst h-1]$ and $(\pi,v)\in \Psi\ind{t}_\ell$\emph{:}}
    \E^{\pi \circ_{\ell+1} \pihat_{\ell+1:H}\ind{t}}\left[\max_{a\in\cA}\|\phih(\x_h,a)\|_{(\beta I + U\ind{t}_h)^{-1}} \right]  \approxleq 2 \sqrt{d}\|u\ind{t}_h\|_{(\beta I + U\ind{t}_h)^{-1}}. \label{eq:mesure}
\end{gather}
Now, thanks to the update rule in \eqref{eq:updatetuple} for $U\ind{t+1}_h$, a standard elliptical potential argument (see proof of \cref{lem:test}) implies that for a large enough iteration $t =\Omega(d)$, we have $\|u\ind{t}_h\|_{(\beta I + U\ind{t}_h)^{-1}}\leq O(1)$. And so, plugging this into \eqref{eq:mesure} implies that for such a $t$, we have that for all $\ell \in[0\ldotst h-1]$ and $(\pi,v)\in \Psi\ind{t}_\ell$,
\begin{align}
    \E^{\pi \circ_{\ell+1} \pihat_{\ell+1:H}\ind{t}}\left[\max_{a\in\cA}\|\phih(\x_h,a)\|_{(\beta I + U\ind{t}_h)^{-1}} \right]  \approxleq 2 \sqrt{d}. \label{eq:bonusbound}
\end{align}
This inequality, which suffices for our analysis, can be seen as a weaker ``on-policy'' version of the $G$-optimal design inequality in \eqref{eq:ideal}, where ``on-policy'' refers to the fact that the expectation in \eqref{eq:bonusbound} is taken under the algorithm's own policies $\pihat_{1:H}$, in addition to the policies in $(\Psi\ind{t}_\ell)$.\footnote{In the RL literature, it is known that on-policy guarantees are sufficient when employing optimism.} 

Before addressing how we tackle the second challenge (\cref{item:challenge2}), there are two additional points worth discussing:

\begin{itemize}
\item First, the reason we include indicators of the form $\mathbb{I}\{\phi(\x_{h-1},\a_{h-1})^\top v \geq 0\}$ in the guarantee in \eqref{eq:linreg} for $\texttt{FitOptValue}_h$ and in the expression of $u\ind{t}_h$ in \eqref{eq:approxexpr} is precisely because we want an inequality like \eqref{eq:mesure} to hold. In general, we cannot find a policy $\pi$ such that
\begin{align}
    \E^{\pi \circ_{\ell+1} \pihat_{\ell+1:H}\ind{t}}\left[\max_{a\in\cA}\|\phih(\x_h,a)\|_{(\beta I + U\ind{t}_h)^{-1}} \right]  \approxleq 2 \sqrt{d}\|\E^{\pi}[\phi(\x_h,\a_h)]\|_{(\beta I + U\ind{t}_h)^{-1}}, \label{eq:eeee}
\end{align}
for all $\ell\in[0 \ldotst h-1]$ and $(\pi,v)\in \Psi\ind{t}_\ell$; Jensen's inequality essentially goes the wrong way (notice the expectation being inside the norm on the right-hand side of \eqref{eq:eeee}). 
\item Second, one may wonder why we require \eqref{eq:mesure} to hold for all $\ell\in[0 \ldotst h-1]$ instead of just $\ell=0$; that is, bounding $\E^{\pihat_{1:H}\ind{t}}\left[\max_{a\in\cA}\|\phih(\x_h,a)\|_{(\beta I + U\ind{t}_h)^{-1}} \right]$. The reason for this is related to the second challenge in \cref{item:challenge2}, which we will discuss next.
\end{itemize}

\subsection{Challenge: Non-Linearity of Optimistic Value Functions}
\label{sec:challenge}
The second challenge described in \cref{item:two}---the non-linearity of $Q\ind{t}_h-\Qhat\ind{t}_h$---is much more serious. We cannot perform a change of measure (as described in \cref{sec:highlevel}) unless the error in \eqref{eq:linreg} is linear in $\phi$; the analysis of \mainalg{} (and essentially any other RL algorithm) requires performing a change of measure at some step. To understand how we can resolve this issue, we need to closely examine the step in the analysis that requires a change of measure. Similar to typical analyses of algorithms that employ local optimism, our approach involves showing via backward induction that for all $\ell = H+1,\dots, 1$, $(x,a)\in \cX_\ell \times \cA$ and $\tilde\pi \in \Pi_\bench$:
\begin{align}
    Q^{\tilde\pi}_\ell(x,a) &  \leq \wtilde Q\ind{t}_\ell(x,a), \label{eq:theQ} \\
    V^{\tilde\pi}_\ell(x) & \leq V\ind{t}_\ell(x),\label{eq:theV}
\end{align}
where $\wtilde{Q}\ind{t}_\ell \coloneqq Q\ind{t}_\ell - b\ind{t}_\ell$ and $V\ind{t}_\ell(\cdot) = Q\ind{t}_\ell(\cdot,\pihat\ind{t}_\ell(\cdot))$. Without going into too much detail, instantiating \eqref{eq:theV} with $\tilde\pi =\pistar$ and $\ell=1$, and using an elliptical potential argument to relate $Q^{\pihat\ind{t}}_h$ to $Q\ind{t}_h$ leads to a suboptimality guarantee for $\pihat\ind{t}_{1:H}$. Now, let's try to perform one step of backward induction to see where the non-linearity of $Q\ind{t}_h$ is problematic. Fix $h\in[H]$ and assume that \eqref{eq:theQ} and \eqref{eq:theV} hold for all $\ell \in[h+1\ldotst H]$, and we want to show that they hold for $\ell = h$. First, \eqref{eq:theQ} follows easily from \eqref{eq:theV} with $\ell=h+1$; in fact, we have that for all $\tilde \pi \in \Pi_\bench$ and $(x,a)\in \cX_h\times \cA$:
\begin{align}
    Q^{\tilde\pi}_h(x,a) & = R(x,a) +\E[V^{\tilde\pi}_{h+1}(\x_{h+1}) \mid \x_h=x,\a_h=a]\leq  R(x,a) +\E[V\ind{t}_{h+1}(\x_{h+1}) \mid \x_h=x,\a_h=a] = \wtilde Q\ind{t}_h(x,a).
\end{align}
Now, let's see how we can use this to show \eqref{eq:theV} for $\ell=h$. Using \eqref{eq:theQ} and assuming that $\Qhat\ind{t}_h$ is such that $\pihat\ind{t}_h(\cdot)= \argmax_{a\in \cA} \Qhat_h\ind{t}(\cdot,a)$, a standard decomposition (see proof of \cref{lem:say}) implies that for all $x\in \cX_h$:
\begin{align}
    V^{\tilde\pi}_h(x) & \leq \wtilde V\ind{t}_h(x) + (Q\ind{t}_h-\Qhat\ind{t}_h)(x,\tilde\pi(x))  + (\Qhat\ind{t}_h-Q\ind{t}_h)(x,\pihath\ind{t}(x)), \label{eq:decomp}
\end{align}
where $\wtilde{V}_h\ind{t} \coloneqq V_h\ind{t} - b\ind{t}_h$. Now, if $Q\ind{t}_h-\Qhat\ind{t}_h$ was linear (which is not the case in our setting); that is, if there exists $\theta\ind{t}_h$ such that $(Q\ind{t}_h-\Qhat\ind{t}_h)(\cdot,\cdot) = \phi(\cdot,\cdot)^\top \theta\ind{t}_h$, then by some standard algebra (similar to the steps in \eqref{eq:change}), we could perform the following change of measure: for all $(x,a)\in \cX_h\times \cA$,
\begin{align}
(Q\ind{t}_h-\Qhat\ind{t}_h)(x,a)&\approxleq \sqrt{d} \|\phi(x,a)\|_{(\beta I + U\ind{t}_h)^{-1}} \sum_{\tau \in [t-1]} |(\theta\ind{t}_h)^\top u\ind{\tau}_h|, \nn \\
&     \approxleq  \sqrt{d} \|\phi(x,a)\|_{(\beta I + U\ind{t}_h)^{-1}} \sum_{(\pi,v)\in \Psi\ind{t}_h} \left|\E^{\pi}\left[ (\theta\ind{t}_h)^\top \phi(\x_h,\a_h) \cdot \mathbb{I}\{\phi(\x_h,\a_h)^\top v  \geq 0 \} \right]\right|, \quad \text{(by \eqref{eq:approxexpr})} \nn \\
\intertext{and by definition of $\theta\ind{t}_h$:}
&  =  \sqrt{d} \|\phi(x,a)\|_{(\beta I + U\ind{t}_h)^{-1}} \sum_{(\pi,v)\in \Psi\ind{t}_h} \left|\E^{\pi}\left[(Q\ind{t}_h(\x_h,\a_h)- \Qhat\ind{t}_h(\x_h,\a_h)) \cdot \mathbb{I}\{\phi(\x_h,\a_h)^\top v  \geq 0 \} \right]\right|, \nn \\
& \approxleq  \frac{\veps}{8H} \|\phi(x,a)\|_{(\beta I + U_h\ind{t})^{-1}},\label{eq:climb}
\end{align}
where the last step follows from \eqref{eq:linreg}. Technically, the right-hand side of \eqref{eq:linreg} involves a sum over elements in $\Psi\ind{t}_{h-1}$ rather than $\Psi\ind{t}_h$; the intention here is not to be overly precise, but rather to illustrate how the point-wise error $(Q\ind{t}_h - \Qhat\ind{t}_h)(x,a)$ can be bounded in terms of expected errors under rollouts when $(Q\ind{t}_h - \Qhat\ind{t}_h)(x,a)$ is linear.
Plugging \eqref{eq:climb} into the decomposition in \eqref{eq:decomp} and setting $b\ind{t}_h(\cdot) \coloneqq  \frac{\veps}{4 H} \max_{a\in \cA}\|\phi(\cdot,a)\|_{(\beta I + U_h\ind{t})^{-1}}$ shows \eqref{eq:theV} for $\ell=h$ and completes the induction. 

Unfortunately, the argument just presented relies on the linearity of $Q\ind{t}_h - \Qhat\ind{t}_h$, which we do not have; the linear-$Q^\pi$ assumption does not imply the linearity of $Q\ind{t}_h - \Qhat\ind{t}_h$ due to the presence of bonus terms in the definition of $Q\ind{t}_h$. We will now explain how we can overcome this issue.

\subsubsection{First Key Idea: Regaining Linearity by Going One Layer Back}
\label{sec:onestepback} The key to resolving this issue is to perform the change of measure in a slightly different way (which ultimately comes at the cost of requiring the CSC Oracle in the call to $\texttt{FitOptValue}$ in \cref{eq:piargmax_p}). First, for the backward induction, we are now going to aim at showing that for all $\ell = H+1,\dots, 1$, $(x,a)\in \cX_\ell \times \cA$, and $\tilde\pi \in \Pi_\bench$:
\begin{align}
    Q^{\tilde\pi}_\ell(x,a) &  \leq  Q\ind{t}_\ell(x,a), \label{eq:theQ2} \\
    V^{\tilde\pi}_\ell(x) & \leq V\ind{t}_\ell(x)+ (Q\ind{t}_\ell-\Qhat\ind{t}_\ell)(x,\tilde\pi(x))  + (\Qhat\ind{t}_\ell-Q\ind{t}_\ell)(x,\pihatell\ind{t}(x)).\label{eq:theV2}
\end{align}  
Suppose that \eqref{eq:theQ2} and \eqref{eq:theV2} hold for all $\ell \in[h+1\ldotst H]$, and we want to show that they hold for $\ell = h$. Once we show \eqref{eq:theQ2}, \eqref{eq:theV2} will follow easily by a standard decomposition as we did in \eqref{eq:decomp} (see also the proof of \cref{lem:say}). Let's examine how we can prove \eqref{eq:theQ2} for $\ell = h$ by using the fact that \eqref{eq:theV2} holds for $\ell = h+1.$ To do this, we will focus on the comparator policy $\tilde\pi = \pihat^\star$ that satisfies:
\begin{align}
\pihat^\star_h(\cdot) = \mathbb{I}\{ \rg(\cdot) \geq \gamma\} \cdot \pistar(\cdot) + \mathbb{I}\{\rg(\cdot) < \gamma\} \cdot \pihat\ind{t}_h(\cdot), \label{eq:pihatstar}
\end{align}
for some small $\gamma > 0$, where $\rg$ is as defined in \cref{par:bench}. First, note that $\tilde\pi \in \Pi_\bench$ by \eqref{eq:talk}. Second, we will see that the policy $\pihat^\star$, which is a mixture of the optimal policy $\pistar$ and $\pihat\ind{t}$, serves as a sufficiently strong benchmark to allow us to derive a good bound on the suboptimality of $\pihat\ind{t}$ relative to $\pistar$ by applying \eqref{eq:theV2} with $\ell = 1$. The strength of $\pihat^\star$ lies in the fact that the definition of $\rg$ ensures that the actions taken by a policy on low-range states—those where $\rg(\cdot) < \gamma$—have minimal impact; see the discussion in \cref{sec:prelimrange} following \cref{def:range}.
 
By definition of $\pihat^\star$ in \eqref{eq:pihatstar}, we have that if $\tilde\pi = \pihat^\star_h$, then for all $x\in \cX_{h+1}$:
 \begin{align}
   \rg(x) <\gamma \implies  
 (Q\ind{t}_{h+1}-\Qhat\ind{t}_{h+1})(x,\tilde\pi(x))  + (\Qhat\ind{t}_{h+1}-Q\ind{t}_{h+1})(x,\pihat_{h+1}\ind{t}(x)) =0. 
 \end{align}
This implies that $x\mapsto (Q\ind{t}_{h+1}-\Qhat\ind{t}_{h+1})(x,\tilde\pi(x))  + (\Qhat\ind{t}_{h+1}-Q\ind{t}_{h+1})(x,\pihat_{h+1}\ind{t}(x))$ is $\alpha$-admissible with $\alpha = \gamma/(2H)$ (see \cref{lem:admissible}). Thus, by \cref{lem:admreal}, there exists $\theta\ind{t}_h$ such that for all $(x,a)\in \cX_h \times \cA$:
\begin{align}
\phi(x,a)^\top \theta\ind{t}_h =   \E\left[(Q\ind{t}_{h+1}-\Qhat\ind{t}_{h+1})(\x_{h+1},\tilde\pi(\x_{h+1}))  + (\Qhat\ind{t}_{h+1}-Q\ind{t}_{h+1})(\x_{h+1},\pihat_{h+1}\ind{t}(\x_{h+1}))\mid \x_h =x, \a_h =a \right].
\end{align}
With this, we can now show \eqref{eq:theQ2} for $\ell=h$ by performing a change of measure as follows: for all $(x,a)\in \cX_h \times \cA$, 
\begin{align}
& Q^{\pihat^\star}_h(x,a)\nn \\
& = R(x,a) + \E[V^{\pihat^\star}_{h+1}(\x_{h+1}) \mid \x_h=x,\a_h=a],\nn \\
& \leq R(x,a) + \E[V\ind{t}_{h+1}(\x_{h+1}) \mid \x_h=x,\a_h=a] \nn \\ 
& \quad +  \E\left[(Q\ind{t}_{h+1}-\Qhat\ind{t}_{h+1})(\x_{h+1},\pihat^\star_{h+1}(\x_{h+1}))  + (\Qhat\ind{t}_{h+1}-Q\ind{t}_{h+1})(\x_{h+1},\pihat_{h+1}\ind{t}(\x_{h+1})) \mid \x_h =x, \a_h =a\right], \quad \text{(by \eqref{eq:theV2})}\\
& \leq R(x,a) + \E[V\ind{t}_{h+1}(\x_{h+1}) \mid \x_h=x,\a_h=a] + \phi(x,a)^\top \theta\ind{t}_h,\nn \\
& = \wtilde{Q}_h\ind{t}(x,a) + \phi(x,a)^\top \theta\ind{t}_h. \label{eq:hall}
\end{align}
Now, by similar steps as in \eqref{eq:climb}, we have that for all $(x,a)\in \cX_h\times \cA$:
\begin{align}
   & \phi(x,a)^\top \theta\ind{t}_h \nn \\
    &\approxleq \sqrt{d} \|\phi(x,a)\|_{(\beta I + U\ind{t}_h)^{-1}} \sum_{\tau \in [t-1]} |(\theta\ind{t}_h)^\top u\ind{\tau}_h|, \nn \\
    &     \approxleq  \sqrt{d} \|\phi(x,a)\|_{(\beta I + U\ind{t}_h)^{-1}} \sum_{(\pi,v)\in \Psi\ind{t}_h} \left|\E^{\pi}\left[ (\theta\ind{t}_h)^\top \phi(\x_h,\a_h) \cdot \mathbb{I}\{\phi(\x_h,\a_h)^\top v  \geq 0 \} \right]\right|, \quad \text{(by \eqref{eq:mesure})}, \nn \\
    \intertext{and by definition of $\theta\ind{t}_h$ and the triangle inequality:}
    &  \leq  \sqrt{d} \|\phi(x,a)\|_{(\beta I + U\ind{t}_h)^{-1}} \sum_{(\pi,v)\in \Psi\ind{t}_h} \left|\E^{\pi\circ_{h+1}\pihat^\star}\left[(Q\ind{t}_{h+1}(\x_{h+1},\a_{h+1})- \Qhat\ind{t}_{h+1}(\x_{h+1},\a_{h+1})) \cdot \mathbb{I}\{\phi(\x_{h},\a_{h})^\top v  \geq 0 \} \right]\right| \nn \\
    &  \  + \sqrt{d} \|\phi(x,a)\|_{(\beta I + U\ind{t}_h)^{-1}} \sum_{(\pi,v)\in \Psi\ind{t}_h} \left|\E^{\pi\circ_{h+1}\pihat\ind{t}}\left[(Q\ind{t}_{h+1}(\x_{h+1},\a_{h+1})- \Qhat\ind{t}_{h+1}(\x_{h+1},\a_{h+1})) \cdot \mathbb{I}\{\phi(\x_{h},\a_{h})^\top v  \geq 0 \} \right]\right|, \nn \\
    & \approxleq  \frac{\veps}{4 H} \|\phi(x,a)\|_{(\beta I + U\ind{t}_h)^{-1}}, \label{eq:climb2}
    \end{align}
    where the last inequality follows from \eqref{eq:linreg}. Thus, by combining \eqref{eq:climb2} and \eqref{eq:hall}, we get that 
$Q^{\pihat^\star}_h(x,a) \leq \wtilde{Q}_h\ind{t}(x,a) +  \frac{\veps}{4H} \max_{a'\in \cA}\|\phi(x,a')\|_{(\beta I + U\ind{t}_h)^{-1}}.$
    On the other hand, we also know that $Q^{\pihat^\star}_h(x,a)\leq H$, and so using that $\min(c+b,H) \leq \min (c,H)+b$ for all $c,b\geq 0$, we get that \begin{align}
        Q^{\pihat^\star}_h(x,a) \leq \wtilde{Q}_h\ind{t}(x,a) +  \bar{b}\ind{t}_h(x),\quad \text{where} \quad \bar{b}\ind{t}_h(x) \coloneqq \min\left(H, \frac{\veps}{4H} \max_{a\in \cA}\|\phi(x,a)\|_{(\beta I + U\ind{t}_h)^{-1}} \right).  \label{eq:doesit}
    \end{align}
    It seems that we are almost done with proving \eqref{eq:theQ2} for $\ell = h$. However, the term $\bar{b}\ind{t}_h$ is not exactly the same as $b\ind{t}_h$ in \eqref{eq:probono}; it is missing the indicator $\mathbb{I}\{\varphi(\cdot; W_h)\geq \mu \}$. Recall that we need this indicator to ensure the linear fit in \eqref{eq:linreg}.

    \subsubsection{Second Key Idea: Skipping Over Low Range States} To deal with the issue that the bonus in \eqref{eq:doesit} is missing the indicator $\mathbb{I}\{\varphi(\cdot; W_h)\geq \mu \}$, we will use a special value decomposition (generalizing \eqref{eq:decomp}) which essentially involves the value functions over different layers and treats low-range states in a special way. Again, we need to slightly modify the target inequalities for our induction; We modify \eqref{eq:theQ2} and \eqref{eq:theV2} so that the goal is to show via backward induction over $\ell =H+1, \dots, 1$ that for all $(x,a)\in \cX_\ell\times \cA$: 
\begin{align}
Q^{\pihat^\star}_\ell(x,a) &\leq  Q\ind{t}_{\ell}(x,a) + \bar{b}\ind{t}_\ell(x) \cdot \mathbb{I}\{\|\varphi(x;W\ind{t}_\ell)\|<\mu\}, \label{eq:ind11} \\
V^{\pihat^\star}_\ell(x) &\leq V\ind{t}_\ell(x) + \xi_\ell\ind{t}(x,\pihatell^\star(x)) - \xi_\ell\ind{t}(x,\pihatell\ind{t}(x)) +  \bar{b}\ind{t}_\ell(x) \cdot \mathbb{I}\{\|\varphi(x;W\ind{t}_\ell)\|<\mu\},  \label{eq:ind22}
\end{align}
where for $(\tilde x,\tilde a)\in \cX_\ell\times \cA$, $\xi_\ell\ind{t}(\tilde x,\tilde a)\coloneqq Q\ind{t}_\ell(\tilde x,\tilde a) - \Qhat\ind{t}_\ell(\tilde x,\tilde a)$. We note that because of the term $\bar{b}\ind{t}_\ell(x) \cdot \mathbb{I}\{\|\varphi(x;W\ind{t}_\ell)\|<\mu\}$, which is not necessarily admissible (\cref{def:admissible}), we cannot easily recover $\eqref{eq:ind11}$ for $\ell=h$ from \eqref{eq:ind22} with $\ell=h+1$. Instead, we will show \eqref{eq:ind11} for $\ell=h$ by leveraging \eqref{eq:ind22} for \emph{all} $\ell \in[h+1\ldotst H]$ and the following ``skip-step'' decomposition (aimed at replace the steps in e.g.~\eqref{eq:hall}) which we prove in \cref{lem:skipping}: for all $(x,a)\in \cX_h\times \cA$:
\begin{align}
&\E[V^{\pihat^\star}_{h+1}(\x_{h+1})- V\ind{t}_{h+1}(\x_{h+1})\mid \x_h=x, \a_h =a]\nn \\
&  \leq  \sum_{\ell=h+1}^H\E^{\pihat\ind{t}}\left[\mathbb{I}\{\rgell(\x_\ell)\geq \gamma\} \prod_{k=h+1}^{\ell-1} \mathbb{I}\{\rgk(\x_k)<\gamma \} \left( \xi\ind{t}_\ell(\x_\ell, \pihatell^\star(\x_\ell)) -\xi\ind{t}_\ell(\x_\ell,\pihatell\ind{t}(\x_\ell))  \right) \mid \x_h=x,\a_h=a \right] \nn \\
& \quad + \sum_{\ell=h+1}^H\E^{\pihat\ind{t}}\left[\mathbb{I}\{\rgell(\x_\ell)\geq \gamma\} \prod_{k=h+1}^{\ell-1} \mathbb{I}\{\rgk(\x_k)<\gamma \}\cdot \mathbb{I}\{\|\varphi(\x_\ell;W_\ell\ind{t})\|<\mu\} \cdot \bar{b}_\ell\ind{t}(\x_\ell)  \mid \x_h=x,\a_h=a \right]. \label{eq:four2}
\end{align}
The advantage of this decomposition, compared to say \eqref{eq:decomp}, is that it eliminates any inadmissible terms on the right-hand side. In fact, by \cref{lem:calm} in \cref{sec:prelimrange} we have that all the terms on the right-hand side of \eqref{eq:four2} are linear in $\phi(x,a)$, allowing us to perform a change of measure as in \eqref{eq:climb2}, which in turn enables us to prove \eqref{eq:ind11} for $\ell=h$. 

Finally, coming back to some earlier points from \cref{sec:highlevel}, the reason we needed \eqref{eq:eeee} for all $\ell\in[0 \ldotst h-1]$ instead of just $\ell=0$ is precisely because we are using the skip-step decomposition in \eqref{eq:four2} (see the proof sketch in \cref{sec:sketch} for more detail). Additionally, the policy optimization step in \texttt{FitOptVal}{} is needed because we are bounding the conditional expectation: 
\begin{align}
    \E\left[(Q\ind{t}_{h+1}-\Qhat\ind{t}_{h+1})(\x_{h+1},\pihat^\star_{h+1}(\x_{h+1}))  + (\Qhat\ind{t}_{h+1}-Q\ind{t}_{h+1})(\x_{h+1},\pihat_{h+1}\ind{t}(\x_{h+1})) \mid \x_h =x, \a_h =a\right];
\end{align} 
that is, we are bounding the regression error ``one layer back'' (as reflected by the title of \cref{sec:onestepback}), and so we need to ensure that we measure the error $(Q\ind{t}_{h+1}-\Qhat\ind{t}_{h+1})(\cdot,\tilde\pi(\cdot))$ for all policies $\tilde{\pi} \in \Pi_\bench$.

\section{Proof Sketch of the Main Theorem (\cref{lem:say})}
   \label{sec:sketch}
   We now provide a proof sketch of \cref{lem:say}, with the full proof deferred to \cref{sec:fullproof}. We recommend that readers first review \cref{sec:algorithm_desc}.

   Let $\tau\in[T]$ be an iteration such that for all $h\in[H]$, $\ell \in[0\ldotst h-1]$, and $(\pi,v)\in \Psi\ind{\tau}_\ell$,
\begin{gather}
    \left|\E^{\pi \circ_h \tilde\pi}\left[\mathbb{I}\{\phiho(\x_{h-1},\a_{h-1})^\top v \geq 0\} \cdot (Q\ind{\tau}_h(\x_h,\a_h) - \Qhat\ind{\tau}_h(\x_h,\a_h)) \right]\right|  \approxleq \frac{\veps}{16 H^2 T^2 d^2},\label{eq:linreg2} \\
    \shortintertext{and}
    \E^{\pi \circ_{\ell+1} \pihat_{\ell+1:H}\ind{\tau}}\left[\max_{a\in\cA}\|\phih(\x_h,a)\|_{(\beta I + U\ind{\tau}_h)^{-1}} \right]  \approxleq 2 \sqrt{d}. \label{eq:bonusbound2}
\end{gather}
We gave a high-level explanation of why such a $\tau$ exists in \cref{sec:highlevel}; see also the formal statements for these bounds in \cref{lem:test} (the bound we display in \eqref{eq:linreg2} is merely a slightly tighter version of the one we presented earlier in \eqref{eq:linreg}). Further, define 
    \begin{gather}\pihath^{\star}(\cdot) \coloneqq \mathbb{I}\{\rgh(\cdot)<\gamma  \} \cdot \pihath\ind{\tau}(\cdot) + \mathbb{I}\{\rgh(\cdot)\geq \gamma \} \cdot \pistarh(\cdot),
    \end{gather}
    for $h\in[H]$, $\gamma = \mu/\sqrt{d_\nu}$ ($\mu, \nu$ are as in \cref{alg:ops-dp}), and $\rgh$ as in \cref{def:designrange}. Note that by \eqref{eq:talk} and the fact that $\pihat\ind{\tau}, \pistar\in \Pi_\base$, where $\Pi_\base=\left\{ x \mapsto \argmax_{a\in \cA} \theta^\top \phih(x,a)  \mid  \theta \in \bbB(H)   \right\}$, we have
    \begin{gather}
    \pihat^\star_h \in \Pi_\bench.
    \label{eq:dona0}
    \end{gather}  
   At a high-level, our strategy will be to show the sequence of inequalities:
    \begin{align}
    J(\pistar)&\leq J(\pihat^\star_{1:H})  + O(\veps),\label{eq:truncate} \\
    J(\pihat^\star_{1:H})&\leq J(\pihat\ind{\tau}_{1:H})  + O(\veps),\label{eq:benchpol_guar} \\
    J(\pihat\ind{\tau}_{1:H})&\leq J(\pihat_{1:H})  + O(\veps). \label{eq:final}
    \end{align}
Summing up these inequalities and telescoping would imply the desired result. The first inequality reflects the fact that it does not matter too much what actions a policy chooses on low range states (see also discussion in \cref{sec:prelimrange} right after \cref{def:range}).
\fakepar{Suboptimality of $\pihat^\star_{1:H}$}
First, by the performance difference lemma (see \cref{lem:pdl}), we have 
\begin{align}
  & J(\pistar) - J(\pihat^\star_{1:H})\nn \\ & = \sum_{h=1}^H \E^{\pihat^\star}\left[ Q_h^{\pi^\star}(\x_h,\pistarh(\x_h))- Q_h^{\pi^\star}(\x_h,\pihath^{\star}(\x_h)) \right], \nn \\
   & = \sum_{h=1}^H \E^{\pihat^\star}\left[ \inner{\phih(\x_h,\pistarh(\x_h))- \phih(\x_h,\pihath^{\star}(\x_h))}{\theta_h^{\pistar}} \right],\nn \\
   & = \sum_{h=1}^H \E^{\pihat^\star}\left[\mathbb{I}\{\rgh(\x_h) <\gamma \} \cdot \inner{\phih(\x_h,\pistarh(\x_h))- \phih(\x_h,\pihath\ind{\tau}(\x_h))}{\theta_h^{\pistar}} \right], \quad \text{(by definition of $\pihat^\star$)}\nn \\
   & \leq \sum_{h=1}^H \E^{\pihat^\star}\left[\mathbb{I}\{\mathrm{Rg}(\x_h) < \sqrt{2d}\gamma \} \cdot \inner{\phih(\x_h,\pistarh(\x_h))- \phih(\x_h,\pihath\ind{\tau}(\x_h))}{\theta_h^{\pistar}} \right],  \quad \text{(by \cref{lem:rangerel})}\nn \\
   & \leq \sum_{h=1}^H \E^{\pihat^\star}\left[\mathbb{I}\{\mathrm{Rg}(\x_h) < \sqrt{2d}\gamma \} \cdot  \mathrm{Rg}(\x_h)\right], \nn \\
   & \leq H\sqrt{2d} \gamma = H \sqrt{2d/d_\nu} \mu.
   \label{eq:perf0}
\end{align}
This implies \eqref{eq:truncate} for the choices of parameters in \cref{alg:ops-dp}.
We now bound the suboptimality of $\pihat\ind{\tau}_{1:H}$ relative to $\pihat^\star_{1:H}$, where we recall that $\tau \in [T]$ is such that \eqref{eq:linreg2} and \eqref{eq:bonusbound2} hold.

\fakepar{Suboptimality of $\pihat\ind{\tau}_{1:H}$} For this part of the proof sketch, we recall some definitions from \cref{sec:algorithm_desc}; for $\ell\in[H]$, $(x,a) \in \cX_\ell\times \cA$, and $t\in[T],$ we define 
\begin{gather}
    \bar{b}\ind{t}_\ell(x) \coloneqq \min\left(H,\frac{\veps}{4 H} \cdot\max_{a'\in \cA} \|\phiell(x, a')\|_{(\beta I + U\ind{t}_\ell)^{-1}}\right),\quad  \quad b\ind{t}_\ell(x) \coloneqq \bar{b}\ind{t}_\ell(x) \cdot \mathbb{I}\{\|\varphi(x;W\ind{t}_\ell)\|\geq \mu\}, \label{eq:bonuses0} \\
 Q\ind{t}_\ell(x,a) = Q^{\pihat\ind{t}}_{\ell}(x,a) + \E^{\pihat\ind{t}}\left[\sum_{k=\ell}^H b\ind{t}_k(\x_k) \mid \x_\ell =x,\a_\ell =a\right],\quad \text{and} \quad V\ind{t}_\ell(x) \coloneqq  Q\ind{t}_\ell(x,\pihat\ind{t}_\ell(x)).\label{eq:optvalue0}
\end{gather} 
Here, $\bar{b}\ind{t}_\ell$ represents the untruncated bonus and $Q\ind{t}_\ell$ represents the optimistic value function.

With this, we proceed via backward induction to show that for all $\ell =H+1, \dots, 1$ and $(x,a)\in \cX_\ell\times \cA$:  
\begin{align}
 Q^{\pihat^\star}_\ell(x,a) &\leq  Q\ind{\tau}_{\ell}(x,a) + \bar{b}\ind{\tau}_\ell(x) \cdot \mathbb{I}\{\|\varphi(x;W\ind{\tau}_\ell)\|<\mu\}, \label{eq:ind10} \\
V^{\pihat^\star}_\ell(x) &\leq V\ind{\tau}_\ell(x) + \xi_\ell\ind{\tau}(x,\pihatell^\star(x)) - \xi_\ell\ind{\tau}(x,\pihatell\ind{\tau}(x)) +  \bar{b}\ind{\tau}_\ell(x) \cdot \mathbb{I}\{\|\varphi(x;W\ind{\tau}_\ell)\|<\mu\},  \label{eq:ind20}
\end{align}
where for $(\tilde x,\tilde a)\in \cX_\ell\times \cA$,
\begin{align}
    V^{\pihat^\star}_\ell(x) \coloneqq Q^{\pihat^\star}_\ell(x,\pihat_\ell^\star(x)), \quad \xi_\ell\ind{\tau}(\tilde x,\tilde a) \coloneqq Q\ind{\tau}_\ell(\tilde x,\tilde a) - \Qhat\ind{\tau}_\ell(\tilde x,\tilde a),\quad \text{and}\quad  \Qhat\ind{t}_\ell(\tilde x,\tilde a) = \phiell(\tilde x,\tilde a)^\top \thetahat\ind{t}_{\ell}+ b\ind{t}_\ell(\tilde x), \label{eq:bhat0}
\end{align}  with the convention that $Q^{\pihat^\star}_{H+1}\equiv Q\ind{\tau}_{H+1}\equiv \Qhat\ind{\tau}_{H+1}\equiv 0$.

    The base case $\ell=H+1$ follows trivially by the convention that $Q^{\pihat^\star}_{H+1}\equiv Q\ind{\tau}_{H+1}\equiv \Qhat\ind{\tau}_{H+1}\equiv 0$. Now, let $h\in[H]$ and suppose that the induction hypothesis holds for all $\ell=[h+1\ldotst H+1]$. We show that it holds for $\ell=h$. 

\paragraphi{We show \eqref{eq:ind10} for $\ell=h$} Fix $(x,a)\in \cX_h \times \cA$.
By the skip-step decomposition in \cref{lem:skipping} (as alluded to in \cref{sec:challenge}) %
we get that 
\begin{align}
\E[V^{\pihat^\star}_{h+1}(\x_{h+1})\mid \x_h= x,\a_h=a] & =  \sum_{\ell=h+1}^H \E^{\pihat\ind{\tau}}\left[\mathbb{I}\{\rgell(\x_\ell) \geq \gamma\}\cdot \prod_{k=h+1}^{\ell-1} \mathbb{I}\{\rgk(\x_k)<\gamma \} \cdot V^{\pihat^\star}_{\ell}(\x_{\ell}) \mid \x_h=x,\a_h= a \right]\nn \\
& \quad + \sum_{\ell=h+1}^H\E^{\pihat\ind{\tau}}\left[   \prod_{k=h+1}^{\ell} \mathbb{I}\{\rgk(\x_k)<\gamma \} \cdot R(\x_\ell,\a_\ell) \mid \x_h=x,\a_h=a \right]. \label{eq:Qstar0}
\end{align}
We instantiate \cref{lem:skipping} again, this time with the optimistic value function, to get that (full details are in \cref{sec:fullproof}) %
\begin{align}
     \E[V\ind{\tau}_{h+1}(\x_{h+1}) \mid \x_h= x,\a_h = a] &  = \sum_{\ell=h+1}^H \E^{\pihat\ind{\tau}}\left[ \mathbb{I}\{\rgell(\x_\ell)\geq \gamma \}\cdot \prod_{k=h+1}^{\ell-1} \mathbb{I}\{\rgk(\x_k)<\gamma \} \cdot V\ind{\tau}_{\ell}(\x_{\ell}) \mid \x_h=x,\a_h= a \right]\nn \\
    & \quad + \sum_{\ell=h+1}^H\E^{\pihat\ind{\tau}}\left[\prod_{k=h+1}^{\ell} \mathbb{I}\{\rgk(\x_k)<\gamma \} \cdot \tilde r_\ell(\x_\ell,\a_\ell) \mid \x_h=x,\a_h=a \right], \label{eq:Qt0}
    \end{align}
where $\tilde r_\ell(\tilde x,\tilde a)\coloneqq R(\tilde x,\tilde a) + {b}\ind{\tau}_\ell(\tilde x)$ and $b\ind{\tau}_\ell$ is as in \eqref{eq:bonuses0}.
Thus, combining \eqref{eq:Qstar0} and \eqref{eq:Qt0} and using that $\tilde{r}_\ell(\cdot)\geq R(\cdot)$, we get 
\begin{align}
    & \E[V^{\pihat^\star}_{h+1}(\x_{h+1})  - V\ind{\tau}_{h+1}(\x_{h+1})\mid \x_h = x,\a_h = a]  \nn \\
    & \leq   \sum_{\ell=h+1}^H \E^{\pihat\ind{\tau}}\left[ \mathbb{I}\{\rgell(\x_\ell)\geq \gamma\}\cdot \prod_{k=h+1}^{\ell-1} \mathbb{I}\{\rgk(\x_k)<\gamma \} \cdot \left(V^{\pihat^\star}_{\ell}(\x_{\ell})- V\ind{\tau}_{\ell}(\x_{\ell}) \right) \mid \x_h=x,\a_h= a  \right], 
\end{align}
and so by the induction hypothesis; in particular \eqref{eq:ind20} for $\ell\in[h+1\ldotst H+1]$, we have
\begin{align}
    & \E[V^{\pihat^\star}_{h+1}(\x_{h+1})  - V\ind{\tau}_{h+1}(\x_{h+1})\mid \x_h = x,\a_h = a] \nn \\
    &  \leq  \sum_{\ell=h+1}^H\E^{\pihat\ind{\tau}}\left[\mathbb{I}\{\rgell(\x_\ell)\geq \gamma\} \prod_{k=h+1}^{\ell-1} \mathbb{I}\{\rgk(\x_k)<\gamma \} \left( \xi\ind{\tau}_\ell(\x_\ell, \pihatell^\star(\x_\ell)) -\xi\ind{\tau}_\ell(\x_\ell,\pihatell\ind{\tau}(\x_\ell))  \right) \mid \x_h=x,\a_h=a \right] \nn \\
    & \quad + \sum_{\ell=h+1}^H\E^{\pihat\ind{\tau}}\left[\mathbb{I}\{\rgell(\x_\ell)\geq \gamma\} \prod_{k=h+1}^{\ell-1} \mathbb{I}\{\rgk(\x_k)<\gamma \}\cdot \mathbb{I}\{\|\varphi(\x_\ell;W_\ell\ind{\tau})\|<\mu\} \cdot \bar{b}_\ell\ind{\tau}(\x_\ell)  \mid \x_h=x,\a_h=a \right]. \label{eq:four0}
\end{align}
Note that this is the inequality we presented in \eqref{eq:four2}. Now, by \cref{lem:subset}, we have that $\rgh(\tilde x)\leq \sqrt{d_\nu} \cdot \|\varphi(\tilde x;W_\ell\ind{\tau})\|$ for all $\tilde x\in \cX_\ell$. Thus, since $\mu = \sqrt{d_\nu}\cdot\gamma$, we have that for all $\tilde x\in \cX_\ell$, $\mathbb{I}\{\|\varphi(\tilde x;W_\ell\ind{\tau})\|<\mu\}=1$ only if $\mathbb{I}\{\rgh(\tilde x)<\gamma\}=1$. This implies that the second sum in \eqref{eq:four0} is zero, and so
\begin{align}
    & \E[V^{\pihat^\star}_{h+1}(\x_{h+1})  - V\ind{\tau}_{h+1}(\x_{h+1})\mid \x_h = x,\a_h = a] \leq   \sum_{\ell=h+1}^H\E^{\pihat\ind{\tau}}\left[g\ind{\tau}_\ell(\x_{h+1:\ell})  \mid \x_h=x,\a_h=a \right], \label{eq:notice0}
\end{align} 
where \begin{align}
    g\ind{\tau}_\ell(\x_{h+1:\ell}) \coloneqq  \prod_{k=h+1}^{\ell-1} \mathbb{I}\{\rgk(\x_k)<\gamma \} \cdot\mathbb{I}\{\rgell(\x_\ell)\geq \gamma\} \cdot\left(\xi\ind{\tau}_\ell(\x_\ell,\pihatell^\star(\x_\ell))  - \xi\ind{\tau}_\ell(\x_\ell,\pihatell\ind{\tau}(\x_\ell)) \right). \label{eq:gdef0}
\end{align}
Now, in \cref{lem:calm}, we show that for any $\pi \in \Pi$, $f: \cX_\ell \rightarrow [-L,L]$ for $L>0$, there exists $\theta\in \reals^d$ such that $\|\theta\|\leq \poly(L,d,H,\gamma^{-1})$ and 
\begin{align}
\E^{\pi}\left[\prod_{k=h+1}^{\ell-1} \mathbb{I}\{\rgk(\x_k)<\gamma \} \cdot\mathbb{I}\{\rgell(\x_\ell)\geq \gamma\} \cdot f(\x_\ell)\mid \x_h =\tilde x ,\a_h =\tilde a\right] = \phi(\tilde x,\tilde a)^\top \theta,
\end{align}
for $(\tilde x,\tilde a)\in \cX_h\times \cA$. Instantiating this with $f(\cdot)= \xi\ind{\tau}_\ell(\cdot,\pihatell^\star(\cdot))  - \xi\ind{\tau}_\ell(\cdot,\pihatell\ind{\tau}(\cdot))$, we get that there exists $\theta\ind{\tau}_{h,\ell} \in \reals^d$ such that $\|\theta\ind{\tau}_{h,\ell}\|\leq \text{poly}(L,d,H,\gamma^{-1},\lambda^{-1})$ and
\begin{align}
  \forall (\tilde x,\tilde a)\in \cX_h\times \cA,\quad  \E^{\pihat\ind{\tau}}\left[g\ind{\tau}_\ell(\x_{h+1:\ell}) \mid \x_h = \tilde x,\a_h = \tilde a\right] = \phih(\tilde x,\tilde a)^\top \theta\ind{\tau}_{h,\ell}.\label{eq:smell0}
\end{align}
Thus, by a change of measure argument (similar to the steps taken in \eqref{eq:climb}; see also formal statement in \cref{lem:changeofmeasure2}), we have that (ignoring some low-order terms to simplify the presentation) 
\begin{align}
& \left|\E^{\pihat\ind{\tau}}\left[g\ind{\tau}_\ell(\x_{h+1:\ell}) \mid \x_h = x,a_h = a\right]\right|\nn \\
&  \approxleq \sqrt{d} \|\phih(x,a)\|_{(\beta I + U\ind{\tau}_h)^{-1}} \sum_{(\pi, v) \in \Psi\ind{\tau}_h}\left| \E^{\pi}[\phih(\x_h,\a_h)^\top \theta\ind{\tau}_{h,\ell} \cdot \mathbb{I}\{\phih(\x_h,\a_h)^\top v \geq 0\}] \right|.\label{eq:talent0}
\end{align}
Now, using \eqref{eq:smell0} again and the law of total expectation, we get that for any $(\pi,v)\in \Psi\ind{\tau}_h$: 
\begin{align}
   & \left| \E^{\pi}[\phih(\x_h,\a_h)^\top \theta\ind{\tau}_{h,\ell} \cdot \mathbb{I}\{\phih(\x_h,\a_h)^\top v \geq 0\}] \right|  = \left| \E^{\pi\circ_{h+1}\pihat\ind{\tau}}[g\ind{\tau}_\ell(\x_{h+1:\ell}) \cdot \mathbb{I}\{\phih(\x_h,\a_h)^\top v \geq 0\}] \right|, 
\end{align}
and so by letting $\mathbf{I}_{h,\ell,v} \coloneqq \mathbb{I}\{\phih(\x_h,\a_h)^\top v \geq 0\} \cdot \prod_{k=h+1}^{\ell-1} \mathbb{I}\{\rgk(\x_k)<\gamma\}$ and using the definition of $g\ind{\tau}_\ell$ in \eqref{eq:gdef0}:
\begin{align}
    & \left| \E^{\pi}[\phih(\x_h,\a_h)^\top \theta\ind{\tau}_{h,\ell} \cdot \mathbb{I}\{\phih(\x_h,\a_h)^\top v \geq 0\}] \right|\nn \\
   & =\left| \E^{\pi \circ_{h+1}\pihat\ind{\tau}}\left[\mathbf{I}_{h,\ell,v} \cdot\E\left[\mathbb{I}\{\rgell(\x_\ell)\geq \gamma\} \cdot \left(\xi\ind{\tau}_\ell(\x_\ell,\pihatell^\star(\x_\ell))-\xi\ind{\tau}_\ell(\x_\ell,\pihatell\ind{\tau}(\x_\ell)) \right)\mid \x_{\ell-1},\a_{\ell-1} \right] \right]\right|. \label{eq:red0}
\end{align}
Now, by \cref{lem:admissible}, the function $x \mapsto \mathbb{I}\{\rgell(x)\geq \gamma\} \cdot \left(\xi\ind{\tau}_\ell(x,\pihatell^\star(x))-\xi\ind{\tau}_\ell(x,\pihatell\ind{\tau}(x)) \right)$ is $\alpha$-admissible with $\alpha = \poly(\lambda, 1/H)$, and so by \cref{lem:admreal}, there exists $\tilde\theta\ind{\tau}_{\ell-1}\in \reals^d$ such that $\|\tilde\theta\ind{\tau}_{\ell-1}\| \leq \poly(d, H, \gamma^{-1}, \lambda^{-1})$ and for all $(\tilde x,\tilde a)\in \cX_{\ell-1}\times \cA$:
\begin{align}
    \phiello(\tilde x,\tilde a)^\top \tilde\theta\ind{\tau}_{\ell-1} & = \E\left[\mathbb{I}\{\rgell(\x_\ell)\geq \gamma\} \cdot \left(\xi\ind{\tau}_\ell(\x_\ell,\pihatell^\star(\x_\ell))-\xi\ind{\tau}_\ell(\x_\ell,\pihatell\ind{\tau}(\x_\ell)) \right)\mid \x_{\ell-1}=\tilde x,\a_{\ell-1}=\tilde a \right], \label{eq:theeta0} \\
    &=\E\left[\xi\ind{\tau}_\ell(\x_\ell,\pihatell^\star(\x_\ell))-\xi\ind{\tau}_\ell(\x_\ell,\pihatell\ind{\tau}(\x_\ell))\mid \x_{\ell-1}=\tilde x,\a_{\ell-1}=\tilde a \right],\label{eq:theetanew0}
\end{align}
where the last equality follows by the fact that $\pihatell^{\star}(\cdot)= \mathbb{I}\{\rgell(\cdot)<\gamma  \} \cdot \pihatell\ind{\tau}(\cdot) + \mathbb{I}\{\rgell(\cdot)\geq \gamma \} \cdot \pistar(\cdot)$, by definition. Using \eqref{eq:theeta0} and a change of measure argument again (see \cref{lem:changeofmeasure2}), we get that for all $(\tilde x,\tilde a)\in \cX_{\ell-1}\times \cA$:
\begin{align}
    & \left|\E\left[\mathbb{I}\{\rgell(\x_\ell)\geq \gamma\} \cdot \left(\xi\ind{\tau}_\ell(\x_\ell,\pihatell^\star(\x_\ell))-\xi\ind{\tau}_\ell(\x_\ell,\pihatell\ind{\tau}(\x_\ell)) \right)\mid \x_{\ell-1}=\tilde x,\a_{\ell-1}=\tilde a \right] \right|\nn \\
    & \approxleq  \sqrt{d}\|\phiello(\tilde x,\tilde a)\|_{(\beta I + U\ind{\tau}_{\ell-1})^{-1}} \sum_{(\pi', v') \in \Psi\ind{\tau}_{\ell-1}}\left|\E^{\pi'}[\phiello(\x_{\ell-1},\a_{\ell-1})^\top \tilde\theta\ind{\tau}_{\ell-1} \cdot \mathbb{I}\{\phiello(\x_{\ell-1},\a_{\ell-1})^\top v' \geq 0\}] \right|,
\intertext{
 and so using the property of $\tilde\theta\ind{\tau}_{\ell-1}$ in \eqref{eq:theetanew0}, the law of total expectation, and the triangle inequality, we get
}
    &  \approxleq \sqrt{d}\|\phiello(\tilde x,\tilde a)\|_{(\beta I + U\ind{\tau}_{\ell-1})^{-1}} \sum_{(\pi', v') \in \Psi\ind{\tau}_{\ell-1}}\left|\E^{\pi'}[\xi\ind{\tau}_\ell(\x_\ell,\pihatell\ind{\tau}(\x_\ell)) \cdot \mathbb{I}\{\phiello(\x_{\ell-1},\a_{\ell-1})^\top v' \geq 0\}]\right|\nn \\
    & \quad + \sqrt{d}\|\phiello(\tilde x,\tilde a)\|_{(\beta I + U\ind{\tau}_{\ell-1})^{-1}} \sum_{(\pi', v') \in \Psi\ind{\tau}_{\ell-1}}\left|\E^{\pi'}[ \xi\ind{\tau}_\ell(\x_\ell,\pihatell^\star(\x_\ell)) \cdot \mathbb{I}\{\phiello(\x_{\ell-1},\a_{\ell-1})^\top v' \geq 0\}]\right|, \nn\\
    & \approxleq \frac{\veps}{8 H^2 T d^{3/2}}\|\phiell(\tilde x,\tilde a)\|_{(\beta I + U\ind{\tau}_{\ell-1})^{-1}} ,\label{eq:see}
\end{align}
where the last inequality follows by \eqref{eq:linreg2} (see also \cref{lem:test} for a formal statement) and the fact that $\pihat^\star \in \Pi_\bench$ (see \eqref{eq:dona0}). Plugging \eqref{eq:see} into \eqref{eq:red0}, we get that for all $(\pi,v)\in \Psi\ind{\tau}_h$:
\begin{align}
    & \left| \E^{\pi}[\phih(\x_h,\a_h)^\top \theta\ind{\tau}_{h,\ell} \cdot \mathbb{I}\{\phih(\x_h,\a_h)^\top v \geq 0\}] \right| \nn \\
  & \approxleq \frac{\veps}{8 H^2 T d^{3/2}}\E^{\pi \circ_{h+1} \pihat\ind{\tau}} \left[\mathbf{I}_{h,\ell,v}\cdot \|\phiello(\x_{\ell-1},\a_{\ell-1})\|_{(\beta I + U_{\ell-1}\ind{\tau})^{-1}}\right],
\end{align}
where we recall that $\mathbf{I}_{h,\ell,v}=\mathbb{I}\{\phih(\x_h,\a_h)^\top v \geq 0\} \cdot \prod_{k=h+1}^{\ell-1} \mathbb{I}\{\rgk(\x_k)<\gamma\}\leq 1$. Thus, we have for all $(\pi,v)\in \Psi\ind{\tau}_h$: 
\begin{align}
   &  \left| \E^{\pi}[\phih(\x_h,\a_h)^\top \theta\ind{\tau}_{h,\ell} \cdot \mathbb{I}\{\phih(\x_h,\a_h)^\top v \geq 0\}] \right|\nn \\
   &  \leq \frac{\veps}{8 H^2 T d^{3/2}}\E^{\pi \circ_{h+1} \pihat\ind{\tau}} \left[\mathbf{I}_{h,\ell,v}\cdot \|\phiello(\x_{\ell-1},\a_{\ell-1})\|_{(\beta I + U_{\ell-1}\ind{\tau})^{-1}}\right], \nn \\
   & \approxleq \frac{\veps}{4 H^2 T d^{1/2}} , 
\end{align} 
where the last inequality follows by \eqref{eq:mesure} and \eqref{eq:bonusbound}; see also \cref{lem:test} for a formal statement.
Combining this with \eqref{eq:talent0} and \eqref{eq:notice0}, we get that 
\begin{align}
&Q^{\pihat^\star}_h(x,a) - \left(R(x,a)+ \E[V\ind{\tau}_{h+1}(\x_{h+1}) \mid \x_h= x,\a_h=a] \right)\nn \\
& =  \E[V^{\pihat^\star}_{h+1}(\x_{h+1})-V\ind{\tau}_{h+1}(\x_{h+1})\mid \x_h=x,\a_h=a], \nn \\
& \leq \frac{\veps}{4 H}\cdot \|\phih(x,a)\|_{(\beta I + U\ind{\tau}_h)^{-1}} \cdot \veps.
\label{eq:Cat0}
\end{align}
On the other hand, we have that 
\begin{align}
Q^{\pihat^\star}_h(x,a)  \leq H \quad \text{and} \quad R(x,a)+ \E[V\ind{\tau}_{h+1}(\x_{h+1})\mid \x_h = x,\a_h = a] \geq 0.
\end{align}
Combining this with \eqref{eq:Cat0} and the fact that $\min(H,c+b)\leq c+\min(H,b)$ for $c,b\geq 0$, we get 
\begin{align}
   & Q^{\pihat^\star}_h(x,a)\nn \\
    & \approxleq  \min\left(H,R(x,a)+ \E[V\ind{\tau}_{h+1}(\x_{h+1}) \mid \x_h= x,\a_h=a]+  \frac{\veps}{4 H} \cdot \|\phih(x,a)\|_{(\beta I + U\ind{\tau}_h)^{-1}} \right), \nn \\
    & \leq  R(x,a)+ \E[V\ind{\tau}_{h+1}(\x_{h+1})\mid \x_h=x,\a_h=a]+ \min\left(H,  \frac{\veps}{4 H}  \cdot \|\phih(x,a)\|_{(\beta I + U\ind{\tau}_h)^{-1}}\right), \nn \\
    & \leq  R(x,a)+ \E[V\ind{\tau}_{h+1}(\x_{h+1})\mid \x_h=x,\a_h=a]+ \min\left(H,  \frac{\veps}{4 H}  \cdot \max_{\tilde a\in \cA}\|\phih(x,\tilde a)\|_{(\beta I + U\ind{\tau}_h)^{-1}} \right), \nn \\
    & = R(x,a)+ \E[V\ind{\tau}_{h+1}(\x_{h+1})\mid \x_h=x,\a_h=a] + b\ind{\tau}_h(x) + \bar b\ind{\tau}_h(x) \cdot \mathbb{I}\{\|\varphi(x;W_h\ind{\tau})\|<\mu\},  \ \text{(see below)}\label{eq:aqua0} \\
    & =   Q_h\ind{\tau}(x,a) + \bar b\ind{\tau}_h(x) \cdot \mathbb{I}\{\|\varphi(x;W_h\ind{\tau})\|<\mu\},
\end{align}
where \eqref{eq:aqua0} follows by the definitions of $b\ind{\tau}_h$ and $\bar{b}\ind{\tau}_h$ in \eqref{eq:bonuses0}, and the last inequality follows by the definition of $Q\ind{\tau}_h$ in \eqref{eq:optvalue0}.
This shows \eqref{eq:ind10} for $\ell=h$.

\paragraphi{We show \eqref{eq:ind20} for $\ell=h$} We have 
\begin{align}
  &  V^{\pihat^\star}_h(x) - V\ind{\tau}_h(x) \nn \\ & = Q^{\pihat^\star}_h(x,\pihath^{\star}(x)) - Q\ind{\tau}_h(x,\pihath\ind{\tau}(x)), \nn \\ 
    & \leq  Q^{\pihat^\star}_h(x,\pihath^{\star}(x)) - Q\ind{\tau}_h(x,\pihath\ind{\tau}(x)) + \Qhat\ind{\tau}_h(x,\pihath\ind{\tau}(x)) - \Qhat\ind{\tau}_h(x,\pihath^{\star}(x)), \quad \text{(by definition of $\pihath\ind{\tau}$)} \nn \\
    & = Q^{\pihat^\star}_h(x,\pihath^{\star}(x)) - Q\ind{\tau}_h(x,\pihath^{\star}(x)) + Q\ind{\tau}_h(x,\pihath^{\star}(x)) - Q\ind{\tau}_h(x,\pihath\ind{\tau}(x)) + \Qhat\ind{\tau}_h(x,\pihath\ind{\tau}(x)) - \Qhat\ind{\tau}_h(x,\pihath^{\star}(x)), \nn \\
    & \leq  \xi\ind{\tau}_h(x,\pihath^{\star}(x))-\xi\ind{\tau}_h(x,\pihath\ind{\tau}(x)) + \bar b\ind{\tau}_h(x) \cdot \mathbb{I}\{\|\varphi(x;W_h\ind{\tau})\|<\mu\},
\end{align}
where the last inequality follows by \eqref{eq:ind10} with $\ell=h$. This shows \eqref{eq:ind20} for $\ell=h$ and completes the induction. Instantiation \eqref{eq:ind20} with $\ell=1$ and using the definition of $V\ind{\tau}_1$ in \eqref{eq:optvalue0}, we get that 
\begin{align}
   & J(\pihat^\star_{1:H}) - J(\pihat\ind{\tau}_{1:H})\nn \\ &
   =\E\left[ V^{\pihat^\star}_1(\x_1)\right] - \E\left[V^{\pihat\ind{\tau}}_1(\x_1)\right],\nn \\ &
   = \E\left[ V^{\pihat^\star}_1(\x_1)\right] - \E\left[V\ind{\tau}_1(\x_1)\right]+ \E\left[\sum_{h=1}^H b\ind{\tau}_h(\x_h)\right], \nn\\
    & \leq  \E\left[\xi\ind{\tau}_1(\x_1,\pihat_1^{\star}(\x_1))-\xi\ind{\tau}_1(\x_1,\pihat_1\ind{\tau}(\x_1)) + \bar b\ind{\tau}_1(\x_1) \cdot \mathbb{I}\{\|\varphi(\x_1;W_1\ind{\tau})\|<\mu\}\right] +  \E\left[\sum_{h=1}^H b\ind{\tau}_h(\x_h)\right],\nn \\
    & \leq 2 \veps + \E\left[\bar b\ind{\tau}_1(\x_1) \cdot \mathbb{I}\{\|\varphi(\x_1;W_1\ind{\tau})\|<\mu\} \right]+ \E\left[\sum_{h=1}^H b\ind{\tau}_h(\x_h)\right], \quad \text{(see below)} \label{eq:intrr0}\\
    & \leq 2  \veps + 2\E\left[\sum_{h=1}^H b\ind{\tau}_h(\x_h)\right], \label{eq:subopte}
\end{align}
where \eqref{eq:intrr0} follows by \eqref{eq:linreg2} (see \cref{lem:test} for a formal statement) and the fact that $\pihat^\star \in \Pi_\bench$ (see \eqref{eq:dona0}). Now, by the definition of the bonuses $({b}_h\ind{\tau})$ in \eqref{eq:bonuses0}, and the bounds in \eqref{eq:mesure} and \eqref{eq:bonusbound}, we get that (see also \cref{lem:test} for a formal statement)
\begin{gather}
    \E\left[\sum_{h=1}^H b\ind{\tau}_h(\x_h)\right] \leq d \veps/2,
    \intertext{and so plugging this into \eqref{eq:subopte}, we get}
    J(\pihat^\star_{1:H}) - J(\pihat\ind{\tau}_{1:H}) \approxleq 2  \veps + d\veps . \label{eq:ter0}
\end{gather}
\paragraph{Suboptimality of $\pihat_{1:H}$} Let $\tau'$ be as in \cref{alg:ops-dp} just before the algorithm returns, and note that the policy $\pihat_{1:H}$ satisfies $\pihat_{1:H}=\pihat\ind{\tau'}_{1:H}$. Now, by \cref{lem:evalmeta} (guarantee of \texttt{Evaluate}), we have that 
\begin{align}
     J(\pihat\ind{\tau}_{1:H}) \approxleq  \max_{t\in [T]}  J(\pihat\ind{t}_{1:H}) \leq   J(\pihat\ind{\tau'}_{1:H}) + H\sqrt{\frac{2 \log T}{n_\traj}} .\label{eq:afg0}
\end{align}
Combining \eqref{eq:perf0}, \eqref{eq:ter0}, and \eqref{eq:afg0}, we get 
\begin{align}
    J(\pistar) - J(\pihat_{1:H}) \leq H\sqrt{\frac{2 \log (T)}{n_\traj}} + 2  \veps+ d \veps + H \sqrt{2d/d_\nu}\cdot \mu. \label{eq:inee0}
\end{align}
Using the choices of $\beta, \mu$, $\nu$, $T$, and $n_\traj$, in \cref{alg:ops-dp}, we get that the right-hand of side of \eqref{eq:inee0} is at most $O(\veps)$, which completes the proof sketch.

        \section{Conclusion, Limitations, and Future Work}
        \label{sec:discussion}
\label{sec:conclusion}
In this paper, we presented an algorithm for the linear $Q^\pi$-realizable setting that is both sample-efficient and makes a polynomial number of calls to a cost-sensitive classification oracle over policies. We also showed that this oracle can be implemented efficiently when the feature dimension is constant. The techniques we used to achieve our results may be of independent interest and could be applicable to other RL settings beyond linear function approximation. 

However, this work leaves open several important questions. First, it remains unclear whether a computationally efficient algorithm can be developed for non-constant feature dimension without relying on a computational oracle. Second, we have not yet determined whether the $\poly(A)$ factor in the sample complexity can be eliminated when using a local optimism-based approach like ours. Addressing these questions could lead to more broadly applicable and computationally feasible algorithms in reinforcement learning. 

Finally, as noted in \cref{foot:layered}, the layered MDP assumption does introduce some loss of generality for linearly $Q^\pi$-realizable MDPs. An interesting direction for future work would be to remove this assumption, which could also pave the way for designing algorithms applicable to the infinite horizon RL setting. 

	\newpage
  \subsection*{Acknowledgements}
  We thank Dylan Foster and Gell\'ert Weisz for several helpful discussions.
	\bibliography{../refs.bib}
	
	\newpage

	\appendix

\clearpage
\addcontentsline{toc}{section}{Appendix} %
\section{Organization of the Appendix}
This appendix is organized as follows:

\begin{itemize}
\item In \cref{sec:additional_related}, we present additional related work in the context of linear function approximation.
\item In \cref{sec:fullversions}, we present the full versions of the \texttt{FitOptValue} and \texttt{DesignDir} algorithms.
\item \cref{sec:params} details the choice of parameters for the \mainalg{} algorithm.
\item \cref{sec:proofsstructural} contains the proofs for structural results related to linearly  $Q^\pi$-realizable MDPs.
\item \cref{sec:linearfit} showcases how a non-zero preconditioning vector can be computed when a linear fit fails (see discussion in \cref{subsec:fitoptv}).
\item \cref{app:designdir} provides the guarantees of the \texttt{DesignDir} algorithm, including both statements and proofs of guarantees.
\item \cref{sec:fitbonusproof} outlines the guarantees of \texttt{FitOptValue}, along with proofs and additional structural results.
\item \cref{app:eval} presents the guarantee for the \texttt{Evaluate} procedure.
\item \cref{sec:fullproof} contains the proof of \cref{lem:say}.
\item \cref{sec:oracle} discusses the implementation of the CSC oracle over $\Pi_\bench$.
\item \cref{sec:otherproofs} provides the upper bound on the growth function of $\Pi_\bench$ and includes the proof of \cref{lem:growth}.
\item \cref{sec:helper} contains a collection of helper lemmas used throughout the analysis.
\end{itemize}

\section{Additonal Related Work}
\label{sec:additional_related}
In the broader context of linear function approximation, the works of \cite{weisz2021exponential} and \cite{wang2021exponential} examine the setting where the optimal state-action value function $Q^\star$ is linear in the feature map, which is a weaker assumption than the linear $Q^\pi$-assumption. \cite{weisz2021exponential} showed that this setting is statistically intractable even with access to a local simulator. Additionally, \cite{wang2021exponential} derived a lower bound for online RL in this setting, showing that even with a gap assumption, the problem remains hard without a simulator.

\cite{du2021bilinear} studied a related assumption where in addition to $Q^\star$, the optimal value function $V^\star$ is also assumed to be linear (i.e.~the $Q^\star/V^\star$ linear setting). \cite{du2021bilinear} showed that this setting is statistically tractable. However, \cite{kane2022computational} later proved a computational-statistical gap in this setting by presenting a computational lower bound.

Another related setting is the linear Bellman complete MDP setting \citep{zanette2020learning}, where Bellman backups of linear functions remain linear in the feature map; this setting is known to generalize the Linear Quadratic Regulator (LQR) problem \cite{wu2024computationally}. \cite{wu2024computationally} presented a sample- and computationally-efficient algorithm for linear Bellman complete MDPs with deterministic dynamics and stochastic rewards. Furthermore, \cite{golowich2024linear} developed an algorithm for the linear Bellman complete setting that is both sample- and computationally-efficient when the number of actions is constant (the complexity depends exponentially on the number of actions).

These works highlight the ongoing challenges in balancing sample efficiency and computational feasibility across various structural assumptions in linear function approximation settings.  
\section{Full Versions of $\texttt{FitOptValue}$ and $\texttt{DesignDir}$}
\label{sec:fullversions}

\begin{algorithm}[H]
    \caption{
	$\texttt{FitOptValue}_{h}$: Fit bonuses and update preconditioning matrices if necessary.
	}
	\label{alg:fitoptvalue}
	\begin{algorithmic}[1]
		\setstretch{1.2}
\Statex[0]{\bfseries input:}  \multiline{$h$, $\Psi_{h-1}$, $\pihat_{h+1:H}$, $U_{h+1:H}$, $W_{h+1:H}$, $\Pi'$, $\mu,\nu, \lambda,\beta,\veps, \delta,n$.}
	\Statex[0]{\bfseries initialize:} For all $\ell\in[h+1\ldotst H]$, $w_\ell\gets 0$. 
	\State Define $\veps'= 2 d c \veps_\reg^{\bo}(\tfrac{\delta}{6HT},\Pi', |\Psi_{h-1}|,n) + \frac{8cd\nu H A}{\mu \lambda}$, with $\veps_\reg^\bo$ be as in \eqref{eq:vepsb} and $c\coloneqq 20 d \log (1+ 16 H^4 \nu^{-4})$.  \label{line:def}
	\Statex[0] \algcommentbiglight{Gather trajectory data.}
 		\For{$(\pi,v)\in \Psi_{h-1}$}\label{line:firstline}
	\State Set $\cDhat_{h,\pi} \gets \emptyset$.
	\For{$n=1,\ldots,n$} 
	\State Sample trajectory $(\bx_1,\a_1,\br_1,\dots, \bx_{H},\a_{H},\br_H) \sim \P^{\pi\circ_h  \pi_\unif \circ_{h+1}\pihat_{h+1:H}}$.
	\State Update $\cDhat_{h,\pi} \gets \cDhat_{h,\pi} \cup \{(\bx_{1:H},\a_{1:H},\br_{1:H})\}$.
	\EndFor
	\EndFor
	\State Define bonuses $b_\ell(\cdot)= \min\left(H,\frac{\veps}{4H}\max_{a\in \cA} \| \phiell(\cdot ,a)\|_{(\beta I + U_\ell)^{-1}}\right) \cdot \mathbb{I}\{\|\varphiell(\cdot;W_\ell)\|\geq \mu\}$. \label{line:bonus}
    \Statex[0] \algcommentbiglight{Fitting the reward and bonuses.}
\State Set $\Sigma_{h} \gets \lambda n |\Psi_{h-1}|  I + \sum_{(\pi,v)\in  \Psi_{h-1}}\sum_{(x_{1:H},a_{1:H},r_{1:H})\in \cDhat_{h,\pi}} \phih(x_{h},a_{h}) \phih(x_{h},a_{h})^\top$. \label{line:sigmah}
\State Set $\thetahat^{\re}_{h} \gets \Sigma_{h}^{-1}\sum_{(\pi,v)\in \Psi_{h-1}}\sum_{(x_{1:H},a_{1:H},r_{1:H})\in \cDhat_{h,\pi}} \phih(x_{h},a_{h}) \cdot \sum_{\ell=h}^H r_{\ell}\in \reals^{d}$. \label{line:thetare}
		\State Set $\thetahat^{\bo}_{h,\ell} \gets \Sigma_{h}^{-1}\sum_{(\pi,v)\in  \Psi_{h-1}}\sum_{(x_{1:H},a_{1:H},r_{1:H})\in \cDhat_{h,\pi}} \phih(x_{h},a_{h}) \cdot b_{\ell}(x_{\ell})\in \reals^{d}$. \label{line:thetahl}
		\State Set $\thetahat_{h}\gets \thetahat_h^{\re} + \sum_{\ell=h+1}^H \thetahat_{h,\ell}^{\bo}$. \label{line:theta}
		\Statex[0] \algcommentbiglight{Check the quality of the linear fit for the bonuses and compute new preconditioning vectors.}
	 \State Define $\Delta_{h, \ell}(\pi,v,\tilde\pi) \gets\sum_{(x_{1:H},a_{1:H},r_{1:H})\in \cDhat_{h,\pi}}  \frac{A \cdot\mathbb{I}\{\phiho(x_{h-1},a_{h-1})^\top v\geq 0\}}{n}\cdot \mathbb{I}\{\tilde\pi(x_h)=a_h\}  \left(b_\ell(x_\ell) -   \phih(x_{h},a_{h})^\top \hat\theta^{\bo}_{h,\ell}\right)$. \label{line:delta}
	\State Define $\cH\gets\{\ell\in[h+1\ldotst H]:  \max_{(\pi,v) \in \Psi_{h-1}} \max_{\tilde\pi\in \Pi'}|\Delta_{h,\ell}(\pi,v,\tilde\pi)|>\veps'\}$. \hfill \algcommentlight{$\veps'$ defined in \cref{line:def}} \label{eq:computation}
	\For{$\ell\in \cH$}
		\State Define $B_\ell(\cdot) = b_\ell(\cdot)\cdot \|\varphiell(\cdot;W_\ell)\|^{-2}\cdot\varphiell(\cdot;W_\ell) \varphiell(\cdot;W_\ell)^\top \in \reals^{d\times d}$.
		\State 
		Set $\hat\vartheta^{\bo}_{h,\ell} \gets \Sigma_{h}^{-1}\sum_{(\pi,v)\in  \Psi_{h-1}}\sum_{(x_{1:H},a_{1:H},r_{1:H})\in \cDhat_{h,\pi}} \phih(x_{h},a_{h}) \otimes B_{\ell}(x_{\ell})\in \reals^{d\times d\times d}$. \label{line:varthetahl}
		\State Compute $((\pi_{h,\ell},v_{h,\ell}), \tilde\pi_{h,\ell})\in  \argmax_{((\pi,v),\tilde\pi)\in \Psi_{h-1}\times \Pi'}|\Delta_{h,\ell}(\pi,v,\tilde\pi)| $. \label{eq:piargmax}
		\State For $(\pi,v,\tilde\pi)= (\pi_{h,\ell},v_{h,\ell}, \tilde\pi_{h,\ell})$, compute \begin{align}
			z_\ell \gets \argmax_{z\in \bbB(1)} \left|\sum_{(x_{1:H},a_{1:H},r_{1:H})\in \cDhat_{h,\pi}} \mathbb{I}\{\phiho(x_{h-1},a_{h-1})^\top v\geq 0\}\cdot \mathbb{I}\{\tilde\pi(x_h)=a_h\}  \cdot \left(z^\top B_\ell(x_\ell) z - \hat\vartheta^{\bo}_{h,\ell}[\phih(x_{h},a_{h}),z,z]\right)\right|.
		\end{align}
		\State Set $\tilde z_{\ell} \gets \mathrm{Proj}_{\cS(\W_{\ell},\nu)}(z_{\ell})$.
		\State $w_\ell \gets W_\ell^{-1} \tilde z_\ell$. \label{eq:dbelus}
		\EndFor
		\State \textbf{return} $(\thetahat_{h},w_{h+1:H})$.
\end{algorithmic}
\end{algorithm}

\begin{algorithm}[H]
    \caption{
	$\texttt{DesignDir}_{h}$: Compute design direction and corresponding partial policy.
	}
	\label{alg:fitbonus}
	\begin{algorithmic}[1]
		\setstretch{1.2}
\Statex[0]{\bfseries input:}  \multiline{$h$, $\Psi_{0:h-1}$, $\pihat_{1:h}$, $U_{h}$, $\beta,n$.}
	\Statex[0] \algcommentbiglight{Gathering trajectories for updating the design matrices.}
	\For{$\ell=0,\dots, h-1$} 
	\For{$(\pi,v)\in \Psi_\ell$}
	\State $\cD_{\ell,\pi} \gets \emptyset$.
	\For{$n=1,\ldots,n$} \label{line:third_simp}
	\State \multiline{Sample trajectory $(\bx_1,\a_1,\br_1,\dots, \bx_{h},\a_{h},\br_{h}) \sim \P^{\pi \circ_{\ell+1}\pihat_{\ell+1:h}}$.\label{eq:phi_sip}
	}
	\State Update $\cD_{\ell,\pi} \gets \cD_{\ell,\pi} \cup \{(\bx_{1:h},\a_{1:h})\}$. \label{line:setend}
	\EndFor
	\EndFor 
	\EndFor
	\Statex[0] \algcommentbiglight{Compute new design direction.}
	\State Set $\kappa_h \gets 0$.
	\For{$i\in [d]$}
    \State Set $v_{h,i} = (\beta I + U_h)^{-1/2}e_i \in \reals^d$.
	\State Set $\pi_{h,i,-}(\cdot) = \argmax_{a\in \cA} -\phih(\cdot, a)^\top v_{h,i}$.
	\State Set $\pi_{h,i,+}(\cdot) = \argmax_{a\in \cA}  \phih(\cdot, a)^\top v_{h,i}$.
    \For{$\ell=0,\dots, h-1$}
    \For{$(\pi,v)\in \Psi_\ell$}
	\State Set $u_{h,i,\ell,\pi,-} \gets \tfrac{1}{n}\sum_{(x_{1:h},a_{1:h})\in \cD_{\ell,\pi}} \phih(x_h, \pi_{h,i,-}(x_h))  \cdot \mathbb{I}\{\phih(x_h, \pi_{h,i,-}(x_h))^\top v_{h,i}   \leq  0 \}$. \label{line:minus}
	\State Set $u_{h,i,\ell,\pi,+} \gets \tfrac{1}{n}\sum_{(x_{1:h},a_{1:h})\in \cD_{\ell,\pi}} \phih(x_h, \pi_{h,i,+}(x_h)) \cdot \mathbb{I}\{\phih(x_h, \pi_{h,i,+}(x_h))^\top v_{h,i}  \geq 0 \}$. \label{line:plus}
\State Set $(\mathfrak{s},{u}_{h,i,\ell,\pi}) \in \argmax_{(s,u)\in \{(-, u_{h,i,\ell,\pi,-}),\,(+,u_{h,i, \ell,\pi,+})\}}\ |\inner{u}{v_{h,i}}|.$ \label{line:bar_e} 	
\If{$|\inner{{u}_{h,i,\ell,\pi}}{v_{h,i}}| \geq \kappa_h$} \label{line:test3_simp} 
\State Set $\kappa_h \gets |\inner{{u}_{h,i,\ell, \pi}}{{v}_{h,i}}|$. %
\State Set $u_h\gets {u}_{h,i,\ell,\pi}$ and ${v}_h \gets \mathfrak{s}\cdot {v}_{h,i}$.
\State Set $\tilde\pi_{1:h}\gets \pi' \circ_{\ell+1} \pihat_{\ell+1:h-1}\circ_h \pi_{h,i,\s}$.
\EndIf
\EndFor
\EndFor
\EndFor
		\State \textbf{return} $(u_h,\tilde{\pi}_{1:h},v_h)$.
\end{algorithmic}
\end{algorithm}

 \clearpage
\section{Choice of Parameters for \mainalg}
\label{sec:params}
For $\frakc = \polylog(d,A,1/\delta, 1/\veps)$ sufficiently large, we set the parameters as:
\begin{align}
    T = 32 d H^2 \cdot \frakc,\ \
    n_\traj  = \frac{A^8 d^{70} H^{80} \cdot \frakc}{\veps^{24}},\ \
    \mu  = \frac{\veps}{H \cdot \frakc}, \ \
    \nu  = \frac{\veps^6}{A^3 H^{20} d^{19} \frakc}, \ \
    \lambda  = \frac{\veps^4}{A^2 d^{14} H^{11}},  \ \
    \beta  = \frac{\veps^{12}}{A^4 d^{24} H^{40}}.\label{eq:params}
\end{align} 
\section{Proofs For Structural Results for Linearly $Q^\pi$-Realizable MDPs}

\label{sec:proofsstructural}
In this section, we present proofs for the structural results for linearly $Q^\pi$-realizable MDPs we presented in \cref{sec:prelimrange}. Many of the results are modification of existing ones in \citep{weisz2024online}.

\begin{proof}[Proof of \cref{lem:rangerel}]
  Fix $h\in[H]$ and $x\in \cX$. Let $\rho_h$ be the approximate optimal design for $\Theta_h$ in \cref{sec:prelimrange}. Further, let $\cC_h \coloneqq \supp \rho_h$ and $\mathrm{G}(\rho_h)\coloneqq \sum_{\pi \in \cC_h} \rho_h(\pi) \theta^\pi_h (\theta_h^\pi)^\top$. Then,
  \begin{align}
    \texttt{Rg}(x) & =  \sup_{a,a'\in\cA}\sup_{\theta_h \in \Theta_h} \inner{\phih(x,a) - \phih(x,a')}{\theta_h},\nn \\
    & = \sup_{a,a'\in\cA}\sup_{\theta_h \in \Theta_h} \inner{\phih(x,a) - \phih(x,a')}{\theta_h}, \nn \\
    & \le  \sup_{a,a'\in\cA}\sup_{\theta_h \in \Theta_h} \|\phih(x,a) - \phih(x,a')\|_{\mathrm{G}(\rho_h)}\cdot \|\theta_h\|_{\mathrm{G}(\rho_h)^\dagger}, \nn \\
    & \leq\sqrt{ \sup_{a,a'\in\cA}\sum_{\pi \in \cC_h} \rho_h(\pi)\cdot \inner{\phih(x,a) - \phih(x,a')}{\theta^{\pi}_h}^2} \cdot \sup_{\theta_h \in \Theta_h} \|\theta_h\|_{\mathrm{G}(\rho_h)^\dagger}, \nn \\
    & \leq\sqrt{ \sup_{a,a'\in\cA}\max_{\pi \in \cC_h}  \inner{\phih(x,a) - \phih(x,a')}{\theta^{\pi}_h}^2} \cdot \sqrt{2d}, \quad\text{(see below)} \label{eq:hire} \\
    & = \sqrt{2d} \cdot \rgh(x),
  \end{align}
  where \eqref{eq:hire} follows by the fact that $\rho_h\in \Delta(\Pi)$ is an approximate optimal design for $\Theta_h = \{\theta_h^\pi \mid \pi \in \Pi\}$; see \cref{def:approxdesign}, and the last equality follows by the definition of the design range. This completes the proof.
    \end{proof}

\begin{proof}[Proof of \cref{lem:calm}]
 We show via induction that for all $i = 1,\dots,\ell - h$, there exists $\theta_i \in (i-1)\bbB(4 \tilde{d} H L/\gamma)$ such that for all $(x,a)\in \cX_h\times \cA$:
 \begin{align}
  &  \E^{\pi}\left[\mathbb{I}\{\rgell(\x_{\ell}) \ge \gamma \} \prod_{k=h+1}^{\ell-1} \mathbb{I}\{\rgk(\x_k) < \gamma\} \cdot f(\x_{\ell}) \mid \x_h =x, \a_h =a \right] \nn \\
  & =  \E^{\pi}\left[\mathbb{I}\{\rgell(\x_{\ell}) \ge \gamma \} \prod_{k=h+i}^{\ell-1} \mathbb{I}\{\rgk(\x_k) < \gamma\} \cdot f(\x_{\ell}) \mid \x_h =x, \a_h =a \right]+ \phih(x,a)^\top \theta_i, \label{eq:weird}
 \end{align}
 where we use the convention that $\prod_{k=\ell}^{\ell-1} \mathbb{I}\{\rgk(\x_k) < \gamma\}=1$.
 The base case $i=1$ follows trivially. Now, suppose that the result holds for $i\in[\ell-h-1]$, and we show that it holds for $i+1$. We have for all $(x,a)\in \cX_h\times \cA$:
 \begin{align}
    &\E^{\pi}\left[\mathbb{I}\{\rgell(\x_{\ell}) \ge \gamma \} \prod_{k=h+i}^{\ell-1} \mathbb{I}\{\rgk(\x_k) < \gamma\} \cdot f(\x_{\ell}) \mid \x_h =x, \a_h =a \right]\nn \\
& = \E^{\pi}\left[\mathbb{I}\{\rgell(\x_{\ell}) \ge \gamma \} \prod_{k=h+i+1}^{\ell-1} \mathbb{I}\{\rgk(\x_k) < \gamma\} \cdot f(\x_{\ell}) \mid \x_h =x, \a_h =a \right]\nn \\
& \quad - \E^{\pi}\left[\mathbb{I}\{\rghi(\x_{h+i}) \ge \gamma \} \cdot\mathbb{I}\{\rgell(\x_{\ell}) \ge \gamma \} \prod_{k=h+i+1}^{\ell-1} \mathbb{I}\{\rgk(\x_k) < \gamma\} \cdot f(\x_{\ell}) \mid \x_h =x, \a_h =a \right]. \label{eq:weird2}
 \end{align}
 Thus, it suffices to show that there exists $\tilde\theta_{i} \in \bbB(4 \tilde{d} HL/\gamma)$ such that $(x,a)\in \cX_h\times \cA$:
 \begin{align}
 \phih(x,a)^\top \tilde\theta_i =  \E^{\pi}\left[\mathbb{I}\{\rghi(\x_{h+i}) \ge \gamma \} \cdot\mathbb{I}\{\rgell(\x_{\ell}) \ge \gamma \} \prod_{k=h+i+1}^{\ell-1} \mathbb{I}\{\rgk(\x_k) < \gamma\} \cdot f(\x_{\ell}) \mid \x_h =x, \a_h =a \right].
 \end{align} 
 Combined with \eqref{eq:weird2}, this would imply \eqref{eq:weird} with $i$ replaced by $i+1$ and $\theta_{i+1}= \theta_i - \tilde \theta_i$. 
 
 Now, define  
 \begin{align}
    g(x') \coloneqq \E^{\pi}\left[ \mathbb{I}\{\rgell(\x_{\ell}) \ge \gamma \} \prod_{k=h+i+1}^{\ell-1} \mathbb{I}\{\rgk(\x_k) < \gamma\} \cdot f(\x_{\ell}) \mid \x_{h+i} = x', \a_{h+i} = \pi(x') \right],
 \end{align}
 for all $x'\in \cX_{h+i}$. By the law of total expectation, we have for all $(x,a)\in \cX_h\times \cA$:
 \begin{align}
   & \E^{\pi}\left[\mathbb{I}\{\rghi(\x_{h+i}) \ge \gamma \} \cdot\mathbb{I}\{\rgell(\x_{\ell}) \ge \gamma \} \prod_{k=h+i+1}^{\ell-1} \mathbb{I}\{\rgk(\x_k) < \gamma\} \cdot f(\x_{\ell}) \mid \x_h =x, \a_h =a \right] \nn \\
& = \E^{\pi}\left[\mathbb{I}\{\rghi(\x_{h+i}) \ge \gamma \} \cdot g(\x_{h+i}) \mid \x_h =x, \a_h =a \right].
 \end{align}  
Now, by \cref{lem:admissible}, $F: x'\mapsto \mathbb{I}\{\rghi(x') \ge \gamma \} \cdot g(x')$ is $\alpha$-admissible (\cref{def:admissible}) with $\alpha \coloneqq L/\gamma$, and so by \cref{lem:admreal}, there exists $\tilde \theta_{i} \in \bbB(4 \tilde{d} HL/\gamma)$ such that for all $(x,a)\in \cX_h\times \cA$:
\begin{align}
& \E^{\pi}\left[\mathbb{I}\{\rghi(\x_{h+i}) \ge \gamma \} \cdot\mathbb{I}\{\rgell(\x_{\ell}) \ge \gamma \} \prod_{k=h+i+1}^{\ell-1} \mathbb{I}\{\rgk(\x_k) < \gamma\} \cdot f(\x_{\ell}) \mid \x_h =x, \a_h =a \right] \nn \\
& = \phih(x,a)^\top \tilde \theta_{i}.
\end{align}
Thus, by \eqref{eq:weird2} and the induction hypothesis (i.e.~\cref{eq:weird}), we have that for all $(x,a)\in \cX_h\times \cA$:
\begin{align}
    & \E^{\pi}\left[\mathbb{I}\{\rgell(\x_{\ell}) \ge \gamma \} \prod_{k=h+i+1}^{\ell-1} \mathbb{I}\{\rgk(\x_k) < \gamma\} \cdot f(\x_{\ell}) \mid \x_h =x, \a_h =a \right]\nn \\
    & = \phih(x,a)^\top \theta_i -\phih(x,a)^\top \tilde\theta_i.
\end{align}
This combined with \eqref{eq:weird2} implies \eqref{eq:weird} with $i$ replaced by $i+1$ and $\theta_{i+1}= \theta_i - \tilde \theta_i$, which completes the induction. Instantiating \eqref{eq:weird} with $i = \ell - h$ (recalling that we are using the convention that $\prod_{k=\ell}^{\ell-1} \mathbb{I}\{\rgk(\x_k) < \gamma\}=1$) implies that there is some $\theta_{h-\ell} \in (h-\ell-1)\bbB(4 \tilde{d} H L/\gamma)$ such that for all $(x,a)\in \cX_h \times \cA$:
\begin{align}
  &   \E^{\pi}\left[\mathbb{I}\{\rgell(\x_{\ell}) \ge \gamma \} \prod_{k=h+1}^{\ell-1} \mathbb{I}\{\rgk(\x_k) < \gamma\} \cdot f(\x_{\ell}) \mid \x_h =x, \a_h =a \right]\nn \\
  & = \E^{\pi}\left[\mathbb{I}\{\rgell(\x_{\ell}) \ge \gamma \} \cdot f(\x_{\ell}) \mid \x_h =x, \a_h =a \right] + \phih(x,a)^\top \theta_{\ell - h}. \label{eq:handle}
\end{align}
By \cref{lem:admissible} and \cref{lem:admreal} again, there exists $\tilde \theta_{\ell-h} \in \bbB(2 \tilde{d} HL/\gamma)$ such that for all $(x,a)\in \cX_h\times \cA$:
\begin{align}
\E^{\pi}\left[\mathbb{I}\{\rgell(\x_{\ell}) \ge \gamma \} \cdot f(\x_{\ell}) \mid \x_h =x, \a_h =a \right] = \phih(x,a)^\top \tilde \theta_{\ell-h}. 
\end{align}
Combining this with \eqref{eq:handle} implies the desired result with $\theta \coloneqq \tilde \theta_{\ell-h} + \theta_{\ell-h} \in (\ell-h)\bbB(4\tilde{d} H L/\gamma) \subseteq \bbB(4 \tilde{d} H^2 L/\gamma)$.
\end{proof}

\begin{proof}[Proof of \cref{lem:fanc}]
  Let $(w_i)_{i\in[k]}$ be the sequence of vectors corresponding to $W_h$ in the definition of a valid preconditioning; see \cref{def:precond}. With this, we have 
    $W_h = \left(H^{-2} I+ \sum_{i \in [k]} w_i w_i^\top \right)^{-1/2}$. Then, for all $\theta \in \Theta_h$:
  \begin{align}
    \|W_h^{-1}\theta\|^2 = \theta^\top \left(H^{-2}I + \sum_{i \in[k]} w_i w_i^\top \right)\theta \leq \|\theta\|^2 H^{-2} + \sum_{i \in[k]} \inner{\theta}{w_i}^2  \stackrel{(a)}{\leq} H^{-2} \|\theta\|^2+ k \leq 1+4 d \log(1 + 16 H^4 \nu^{-4}),
  \end{align}
  where $(a)$ follows by the definition of a valid preconditioning (\cref{def:precond}); i.e.~that $\sup_{\theta \in \Theta_h}|\inner{\theta}{w_i}| \leq 1$, for all $i \in[k]$, and the last inequality follows by \cref{lem:precond} and the fact that $\sup_{\theta \in \Theta_h}\|\theta\|\leq H$ (see \cref{assum:linearqpi}).
\end{proof}

\begin{proof}[Proof of \cref{lem:subset}] Aspects of this proof are inspired by the proof of \cite[Proposition 4.5]{weisz2024online}.

  Fix $x\in\cX_\ell$. Let $\rho_\ell$ be the approximate optimal design for $\Theta_\ell$ in \cref{sec:prelims}. By definition of the design range in \cref{def:designrange}, we have
\begin{align}
\rgell(x) & = \max_{\pi \in  \supp(\rho_\ell)}\sup_{a,a'\in \cA} \inner{\phiell(x,a)- \phiell(x,a')}{\theta^{\pi}_\ell},\nn\\ 
& = \max_{\pi \in  \supp(\rho_\ell)}\sup_{a,a'\in \cA} \inner{W_\ell \phiell(x,a)- W_\ell \phiell(x,a')}{W_\ell^{-1}\theta^{\pi}_\ell},\nn \\
\intertext{and by Cauchy Schwarz and the definition of $\varphi(\cdot;\cdot)$ in \cref{lem:dual-admissibility}, we have}
& \leq \max_{\pi \in  \supp(\rho_\ell)} \|\varphiell(x; W_\ell)\| \cdot \| W_\ell^{-1}\theta^{\pi}_\ell\|,\nn \\
& \leq \sqrt{d_\nu} \cdot \|\varphiell(x; W_\ell)\|,
\end{align}
where the last inequality follows by the fact that $W_\ell$ is a valid $\nu$-preconditioning and \cref{lem:fanc}.
\end{proof}

\section{Fit or Precondition}
\label{sec:linearfit}

\begin{algorithm}[H]
    \caption{
	$\texttt{LinearFit}_{h}$: Given layers $h,\ell\in[H]$ such that $h < \ell$, a function  $f$, and a valid preconditioning matrix $W_{\ell}$, either return $\hat{\theta}^f_h$  such that  $\mathbb{E}^{\pi}[f(\x_\ell) \cdot \mathbb{I}\{\varphi(\x_\ell; W_\ell)\} - \phi(\x_h, \a_h)^\top \hat{\theta}_h^f]$ is small, or compute a new, non-zero vector $w_\ell \in \reals^d$  such that $(W_\ell^{-2} + w_\ell w_\ell^\top)^{-1/2}$ remains a valid preconditioning. For practical implementation, expectations would be replaced with empirical averages; we focus on the asymptotic version of the algorithm for simplicity. 
	}
	\label{alg:fitfunction}
	\begin{algorithmic}[1]
		\setstretch{1.2}
\Statex[0]{\bfseries input:}  \multiline{$h$, $\ell$, $f$, $\Psi$, $\pihat_{h+1:H}$, $W_{h+1:H}$, $\mu,\nu,\lambda$.}
\Statex[0]{\bfseries initialize:} $w_\ell \gets 0$. 
	\State Define $\veps'= \frac{8c\sqrt{\lambda} d\tilde{d}H}{\sqrt{\zeta}\mu}+ \frac{8cd\nu L}{\mu \lambda}$, with $c \coloneqq 20 d \log (1+ 16 H^4 \nu^{-4})$, $\tilde d \coloneqq 5 d \log \log (1 + 16 H^4 \nu^{-4})$, and $\zeta = \frac{1}{8d}$. \label{line:def2_meta}
        \Statex[0] \algcommentbiglight{Fitting the truncated version of the function $f$.}
\State Set $\Sigma_{h} \gets \lambda   I + \sum_{\pi\in  \Psi}\E^{\pi}\left[\phih(\x_{h},\a_{h}) \phih(\x_{h},\a_{h})^\top \right]$. \label{line:sigmah2_meta}
		\State Set $\thetahat^{f}_{h} \gets \Sigma_{h}^{-1}\sum_{\pi\in  \Psi}\E^{\pi \circ_{h+1} \pihat_{h+1:H}}\left[\phih(\x_{h},\a_{h}) \cdot f(\x_{\ell}) \cdot \mathbb{I}\{\|\varphi(\x_\ell; W_\ell)\| \geq \mu \}\right] \in \reals^{d}$. \label{line:thetahl2_meta}
		\Statex[0] \algcommentbiglight{If the quality of the linear fit is poor, compute new non-zero preconditioning vector.}
	 \State Define $\Delta_{h, \ell}(\pi) \gets \E^{\pi \circ_{h+1} \pihat_{h+1:H}}\left[f(\x_{\ell}) \cdot \mathbb{I}\{\|\varphi(\x_\ell; W_\ell)\| \geq \mu \} -   \phih(\x_{h},\a_{h})^\top \hat\theta^{f}_{h} \right]$. \label{line:delta2_meta}
	\For{$\max_{\pi \in \Psi} |\Delta_{h,\ell}(\pi)|>\veps'$} \hfill \algcommentlight{$\veps'$ as in \cref{line:def2_meta}} \label{line:valen}
		\State Define $F(\cdot) = f(\cdot)\cdot  \mathbb{I}\{\|\varphi(\cdot; W_\ell)\| \geq \mu \} \cdot \|\varphiell(\cdot;W_\ell)\|^{-2}\cdot\varphiell(\cdot;W_\ell) \varphiell(\cdot;W_\ell)^\top \in \reals^{d\times d}$.
		\State 
		Set $\hat\vartheta^{f}_{h,\ell} \gets \Sigma_{h}^{-1}\sum_{\pi\in  \Psi}\E^{\pi \circ_{h+1} \pihat_{h+1:H}}\left[\phih(\x_{h},\a_{h}) \otimes F(\x_{\ell}) \right]\in \reals^{d\times d\times d}$. \label{line:varthetahl2_meta}
		\State Compute $\pi_{h,\ell}\in  \argmax_{\pi\in \Psi}|\Delta_{h,\ell}(\pi)| $. \label{eq:piargmax2_meta}
		\State For $\pi= \pi_{h,\ell}$, compute 
        \begin{align}
			z_\ell \gets \argmax_{z\in \bbB(1)} \left|\E^{\pi \circ_{h+1} \pihat_{h+1:H}} \left[ z^\top F(\x_\ell) z - \hat\vartheta^{f}_{h,\ell}[\phih(\x_{h},\a_{h}),z,z]\right] \right|.
		\end{align}
		\State Set $\tilde z_{\ell} \gets \mathrm{Proj}_{\cS(\W_{\ell},\nu)}(z_{\ell})$.
		\State $w_\ell \gets W_\ell^{-1} \tilde z_\ell$. \label{eq:dbelus2_meta}
		\EndFor
		\State \textbf{return} $(\thetahat^f_{h},w_{\ell})$.
\end{algorithmic}
\end{algorithm}

In this section, we show that for any $h\in[H]$, $\ell\in [h+1\ldotst H]$, any function $f:\cX_\ell \rightarrow [-L,L]$ for $L>0$, and any policy $\pi\in\Pi$, if $W_\ell$ is a valid preconditioning for layer $\ell$, then for some small parameters $\mu, \lambda>0$, one of the following holds: 
\begin{enumerate}
    \item The discrepancy $\inf_{\theta \in \reals^d} \left|\E^\pi\left[ \mathbb{I}\{\|\varphi(\x_\ell; W_\ell)\| \geq \mu\} \cdot f(\x_\ell) - \phi(\x_h,\a_h)^\top \theta  \right] \right|$ is small; intuitively, one can think of this as corresponding to case where $\E^{\pi}\left[\mathbb{I}\{\|\varphi(\x_\ell; W_\ell)\| \geq \mu\} \cdot f(\x_\ell)\mid \x_h = x, \a_h=a \right]$ is approximately linear in $\phi(x,a)$ (which would be the case if $\varphi(\cdot;W_\ell) \propto \rg(\cdot)$); or 
\item It is possible to compute a non-zero vector $w_\ell\in \reals^d$ such that $(W_\ell^{-2} + w_\ell w_\ell^\top)^{-1/2}$ remains a valid preconditioning matrix for layer $\ell$; the procedure for finding such a vector is displayed in \cref{alg:fitfunction}; note that many components of \cref{alg:fitfunction} are very similar to those of $\texttt{FitOptValue}$ (\cref{alg:fitoptvalue_p}), which is used in our main algorithm, \mainalg.
\end{enumerate}
Next, we state this result formally and provide a proof.
\begin{lemma}
    \label{lem:decoy-find-new_meta}
    Let $h\in[H]$, $\ell \in[h+1\ldotst H]$, $L>0$, $f : \cX_\ell \rightarrow [-L,L]$, $\Psi, \pihat_{h+1:H}, W_{h+1:H}, \mu, \nu,\lambda$ such that $W_{h+1},\dots, W_H \in \pd$ be given and consider a call to $\texttt{LinearFit}_{h}(\ell, f, \pihat_{h+1:H}, {W}_{h+1:H}, \mu, \nu,\lambda)$. Further, let $d_\nu \coloneqq 5 d \log (1 + 16 H^4 \nu^{-4})$. Then, the variables in \cref{alg:fitfunction} are such that either I) $w_\ell =0$ and
\begin{align}
 \forall \pi \in \Psi, \quad  & \left|\E^{\pi \circ_{h+1} \pihat_{h+1:H}}\left[f(\x_\ell) \cdot \mathbb{I}\{\varphi(\x_\ell;W_\ell) \} - \phih(\x_h,\a_h)^\top \thetahat^f_{h} \right]\right|  \leq   \frac{8c\sqrt{\lambda} d\tilde{d}H}{\sqrt{\zeta}\mu}+ \frac{8cd\nu L}{\mu \lambda}, \label{eq:fitting_meta}
\end{align}
where $c \coloneqq 20 d \log (1+ 16 H^4 \nu^{-4})$, $\tilde d \coloneqq 5 d \log \log (1 + 16 H^4 \nu^{-4})$, and $\zeta = \frac{1}{8d}$;
or II) the following points hold:
    \begin{enumerate}
        \item $\|w_\ell\| \leq \nu^{-1}$; \label{item:one_meta}
        \item $\|W_\ell w_\ell\| \geq 1/2$; and \label{item:two_meta}
        \item $\sup_{\theta \in \Theta_\ell} |\inner{\theta}{w_\ell}|\leq 1$. \label{item:three_meta}
    \end{enumerate}
\end{lemma}
\cref{lem:decoy-find-new_meta} implies that if $W_\ell$ is a valid $\nu$-preconditioning (see \cref{def:precond}), then in the second case of \cref{lem:decoy-find-new_meta}, $(W_\ell^{-2} +w_\ell w_\ell^\top)^{-1/2}$ remains a valid $\nu$-preconditioning.
\begin{proof}[\textbf{Proof of \cref{lem:decoy-find-new_meta}}] 
   Let $\rho_\ell$ be the approximate design for $\Theta_\ell$ in \cref{para:special_paragraph} and define \begin{align}G_\ell \coloneqq W_\ell^{-1}\left( \sum_{\pi\in  \supp \rho_\ell}\rho_\ell(\pi)\theta^{\pi}_\ell (\theta^\pi_\ell)^\top\right)W_\ell^{-1}\in \reals^{d\times d}.\label{eq:Gl_meta} 
    \end{align}
To simplify notation in this proof, we let $c \coloneqq 20 d \log(1 + 16 H^4 \nu^{-4})$ and for any $z\in \reals^d$, we write $z_\pa\coloneqq \Proj_{\cS(G_\ell, \zeta)}(z)$ and $z_\perp \coloneqq \Proj_{\cS(G_\ell, \zeta)^\perp}(z)$, where $\zeta \coloneqq  \frac{1}{8d}$.
    
   \fakepar{Case where $w_\ell\neq 0$} First, assume that $w_\ell$ in \cref{alg:fitfunction} is non-zero. Let 
   \begin{align}
    \pi_{h,\ell}\in \argmax_{\pi \in \Psi} \Delta_{h,\ell}(\pi), 
   \end{align}
   where $\Delta_{h,\ell}$ is as in \cref{alg:fitoptvalue}.
   From \cref{line:valen} of \cref{alg:fitfunction}, we have 
    \begin{align}
|\Delta_{h,\ell}(\pi_{h,\ell})| \geq  \frac{8c\sqrt{\lambda} d\tilde{d}H}{\sqrt{\zeta}\mu}+ \frac{8cd\nu L}{\mu \lambda}. \label{eq:conditio_meta}
    \end{align}
   Moving forward, we let \[M \coloneqq \E^{\pi_{h,\ell} \circ_{h+1} \pihat_{h+1:H}}\left[F(\x_\ell) - \hat\vartheta^{f}_{h,\ell}[\phih(\x_{h},\a_{h}),\cdot,\cdot]\right],\]
    and note that $M\in \reals^{d\times d}$ satisfies 
    \begin{align}
        \tr(M) = \Delta_{h,\ell}(\pi_{h,\ell}). \label{eq:trace_meta}
    \end{align}
    Furthermore, the variables $(z_\ell,\tilde z_\ell,w_\ell)$ in \cref{alg:fitoptvalue} satisfy 
    \[z_\ell \in \argmax_{z\in \bbB(1)} |z^\top M z|,\quad  \tilde z_\ell = \Proj_{\cS(\W_\ell, \nu)}(z_\ell),\quad \text{and}\quad w_\ell = W_\ell^{-1} \tilde z_\ell.\] 
    Now, by definition of the trace, there exist unit vectors (eigenvectors in this case) $c_1,\dots,c_d$ such that $\tr(M) = \sum_{i\in[d]} c_i^\top M c_i$. Combining this with \eqref{eq:trace_meta} and \eqref{eq:conditio_meta}, we get that 
    \begin{align}
    d\cdot |z_\ell^\top M  z_\ell| \geq  \left|\sum_{i \in [d]}  c_i^\top M c_i \right| \geq  \frac{8c\sqrt{\lambda} d\tilde{d}H}{\sqrt{\zeta}\mu} +\frac{8 c d\nu L}{\mu \lambda}. \label{eq:stor_meta}
    \end{align}
Now, let $\wbar{M} \coloneqq \sgn(z_\ell^\top M z_\ell) \cdot M$ and note that $z_\ell \in \argmax_{z\in \bbB(1)} z_\ell^\top \wbar{M}z_\ell$. By \eqref{eq:stor_meta}, we have
    \begin{align}
    z_\ell^\top \wbar{M} z_\ell > \frac{8c\sqrt{\lambda} \tilde{d}H}{\sqrt{\zeta}\mu}+ \frac{8c \nu L}{\mu \lambda}, \label{eq:v-can-be-projd_meta}
    \end{align} and so by \cref{lem:theM} and the definition of $M$, we have 
    \begin{align}
\tilde z_\ell^\top \wbar{M} \tilde z_\ell > \frac{8c\sqrt{\lambda} \tilde{d}H}{\sqrt{\zeta}\mu}+ \frac{8(c-1)\nu L}{\mu \lambda}. \label{eq:fenc_meta}
    \end{align}
    We use this to show the claims of the lemma. 
   \paragraphi{Proving \cref{item:one_meta}} We start by showing $\norm{w_\ell}\le \nu^{-1}$; that is, \cref{item:one_meta} of the lemma. Let $\lambda_1,\dots ,\lambda_d$ be the eigenvalues of $W_\ell$ and let $w_1,\dots,w_d\in \reals^{d}$ be an orthonormal basis such that $W_\ell w_i = \lambda_i w_i$, for all $i\in[d]$ (such a basis exists because $W_\ell \in \pd$). With this, we can write 
   \begin{align}
     W_\ell = \sum_{i=1}^d \lambda_i w_i w_i^\top.
   \end{align}
   Let $P_\ell$ be the matrix whose columns are $w_1,\dots,w_d$, and note that $P_\ell^\top P_\ell=I$ since $w_1,\dots,w_d$ are orthonormal. With this, we have  
   \begin{align}
     W_\ell^{-1} \Proj_{\cS(W_\ell,\nu)}(z_\ell) &=  W_\ell^{-1} \sum_{i\in[d]: \lambda_i \geq \nu}  w_i w_i^\top z_\ell,\nn \\
     & = \left(\sum_{i=1}^d \lambda_i w_i w_i^\top\right)^{-1}\sum_{i\in[d]: \lambda_i \geq \nu} w_i  w_i^\top z_\ell,\nn \\
     & = \left(\sum_{i=1}^d \lambda_i P_\ell e_i e_i^\top P_\ell^\top\right)^{-1}\sum_{i\in[d]: \lambda_i \geq \nu} P_\ell e_i  e_i^\top P_\ell^\top z_\ell,\nn \\
     & = P_\ell \left(\sum_{i=1}^d \lambda_i  e_i e_i^\top \right)^{-1}P_\ell^\top \sum_{i\in[d]: \lambda_i \geq \nu} P_\ell e_i  e_i^\top P_\ell^\top z_\ell,\nn \\
     & = P_\ell \left(\sum_{i=1}^d \lambda_i  e_i e_i^\top \right)^{-1}\sum_{i\in[d]: \lambda_i \geq \nu}  e_i  e_i^\top P_\ell^\top z_\ell,\nn \\
     & = P_\ell \diag (\lambda^{-1}_1,\dots,\lambda^{-1}_d)\sum_{i\in[d]: \lambda_i \geq \nu}  e_i  e_i^\top P_\ell^\top z_\ell,\nn \\
                  & = P_\ell \sum_{i\in[d]: \lambda_i \geq \nu}  \lambda_i^{-1} e_i  e_i^\top P_\ell^\top z_\ell,\nn \\
   \end{align} 
   Thus, taking the norm and using that $\|P_\ell\|_\op =1$, we get 
   \begin{align}
    \|w_\ell\|=\norm{\W_\ell^{-1}\tilde z_\ell}=\norm{\W_\ell^{-1}\Proj_{\cS(\W_\ell,\nu)} (z_\ell)} \le \|P_\ell\|_\op\sqrt{\sum_{i\in[d]: \lambda_i \geq \nu} \lambda_i^{-2} (e_i^\top P_\ell z_\ell)^2} \leq \nu^{-1} \|P_\ell z_\ell\| \leq \nu^{-1}\|P_\ell\|_\op \|z_\ell\|\leq \nu^{-1}. 
   \end{align}
   This shows \cref{item:one_meta}.
    \paragraphi{Proving \cref{item:two_meta}}
    To prove \cref{item:two_meta}, we need to show that $\norm{\W_{\ell} w_\ell} =\norm{\tilde z_\ell} \ge\frac{1}{2}$. Using that $z_\ell \in \argmax_{z\in \bbB(1)} z^\top \wbar{M} z$ and \cref{lem:theM},
    \begin{align}
    \|\tilde z_\ell\|^{-2} \cdot \tilde z_\ell^\top \wbar{M} \tilde z_\ell \le z_\ell^\top \wbar{M} z_\ell & \le \tilde z_\ell^\top \wbar{M} \tilde z_\ell + \frac{8\nu L}{\mu \lambda}, \nn \\
    &  \le \tilde z_\ell^\top \wbar{M} \tilde z_\ell + \frac{1}{c} z_\ell^\top \wbar{M} z_\ell, \label{eq:toarrange_meta}
    \end{align}
    where the last inequality follows by \eqref{eq:v-can-be-projd_meta}. Rearranging \eqref{eq:toarrange_meta} and using that $c \geq 3$, implies that $\norm{\tilde z_\ell}^2 \ge \frac23$. Therefore, get that
    \[
    \norm{\W_\ell w_\ell}^2 = \|\tilde z_\ell\|^2 \ge\frac23,
    \]
    satisfying the inequality in \cref{item:two_meta}.
    
    \paragraphi{Proving \cref{item:three_meta}} It remains to prove \cref{item:three_meta}. We will start by showing that $\norm{(\tilde{z}_\ell)_\pa}^2 \le \frac{1}{c}$ and use this to prove \cref{item:three_meta}. Since $\wbar{M}$ is symmetric, we can decompose $\tilde{z}_\ell^\top \wbar{M} \tilde{z}_\ell$ as
    \begin{align}    \tilde{z}_\ell^\top \wbar{M} \tilde{z}_\ell= (\tilde{z}_\ell)_\pa^\top \wbar{M} \tilde{z}_\ell + (\tilde{z}_\ell)_\pa^\top \wbar{M} (\tilde{z}_\ell)_\perp + (\tilde{z}_\ell)_\perp^\top \wbar{M} (\tilde{z}_\ell)_\perp\,. \label{eq:excu}
    \end{align}
    Now, by \cref{lem:dual-admissibility}, we have that $(\tilde z_\ell)_\pa^\top F(\cdot) z$ is $\alpha$-admissible with $\alpha  \coloneqq \frac{\sqrt{\zeta} \mu }{L}$, for all $z\in \bbB(1)$; by \cref{lem:admreal}, this implies that for all $z\in \bbB(1)$, there exists $\theta^{f,z}_h\in \bbB(4 \tilde d H/\alpha)$ such that 
    \begin{align}
        \E^{\pihat_{h+1:H}}[(\tilde{z}_\ell)_\pa^\top F(\x_\ell) z   \mid \x_h =x,\a_h=a] = \phi(x,a)^\top \theta^{f,z}_{h},\quad \forall (x,a)\in \cX_h\times \cA. \label{eq:admiss_meta}
    \end{align}   
     Thus, by the definitions of $\wbar{M}$ and $\hat\vartheta^f_{h,\ell}$ in \cref{alg:fitfunction}, and \cref{thm:bandits-book-20.5}, we have that for all $z\in \bbB(1)$:
    \begin{align}
    (\tilde{z}_\ell)_\pa^\top \wbar{M} z & =  \frac{4\sqrt{\lambda} \tilde{d}H}{\sqrt{\zeta}\mu}, 
    \end{align}
    where $\tilde d \coloneqq 5 d \log (1+ 16 H^4\nu^{-4})$. Instantiating this with $z\in\{\tilde z_\ell, (z_\ell)_\perp \}$ and using \eqref{eq:excu}, we have
    \begin{align}
    \tilde{z}_\ell^\top \wbar{M} \tilde{z}_\ell \le \frac{8\sqrt{\lambda} \tilde{d}H}{\sqrt{\zeta}\mu}  + (\tilde{z}_\ell)_\perp^\top \wbar{M} (\tilde{z}_\ell)_\perp. \label{eq:light_meta}
    \end{align}
    Combining this with \eqref{eq:fenc_meta}, we get that $\|(\tilde{z}_\ell)_\perp\|\ne 0$.
    Let $\bar{z}_\ell=(\tilde{z}_\ell)_\perp/\norm{(\tilde{z}_\ell)_\perp}$. Since $z_\ell \in \argmax_{z\in \bbB(1)} z^\top \wbar{M} z$, we have
    \begin{align}
    \norm{(\tilde{z}_\ell)_\perp}^{-2}\cdot (\tilde{z}_\ell)_\perp^\top \wbar{M} (\tilde{z}_\ell)_\perp& =(\bar{z}_\ell)^\top \wbar{M} {\bar{z}_\ell} \le z_\ell^\top \wbar{M} z_\ell. \label{eq:mocbd_meta}
    \end{align}
        On the other hand, using \cref{lem:theM} and \eqref{eq:light_meta}, we have  
    \begin{align} 
  z_\ell^\top \wbar{M} z_\ell  & \le \tilde{z}_\ell^\top \wbar{M} \tilde{z}_\ell + \frac{8\nu L}{\mu \lambda},\nn \\
    &  \le  \frac{8\sqrt{\lambda} \tilde{d}H}{\sqrt{\zeta}\mu} + \frac{8\nu L}{\mu \lambda} + (\tilde{z}_\ell)_\perp^\top \wbar{M} (\tilde{z}_\ell)_\perp,\nn \\ 
    & \leq \frac{1}{c} z_\ell^\top \wbar{M} z_\ell  + (\tilde{z}_\ell)_\perp^\top \wbar{M} (\tilde{z}_\ell)_\perp, \label{eq:imme_meta}
\end{align}
where the last inequality follows by \eqref{eq:v-can-be-projd_meta}. Rearranging \eqref{eq:imme_meta} gives
\begin{align}
    z_\ell^\top \wbar{M} z_\ell  & \leq \frac{c}{c-1}  (\tilde{z}_\ell)_\perp^\top \wbar{M} (\tilde{z}_\ell)_\perp.
\end{align}
Combining this with \eqref{eq:mocbd_meta}, dividing by $(\tilde{z}_\ell)_\perp^\top \wbar{M} (\tilde{z}_\ell)_\perp$, and rearranging, we get that $\norm{(\tilde{z}_\ell)_\perp}^2 \ge \frac{c-1}{c}$, and so
    \begin{align}
    \norm{(\tilde{z}_\ell)_\pa}^2 &\le \frac{1}{c}, \label{eq:pitstop_meta}
    \end{align}
    since $1 = \|z_\ell\|^2\geq \|\tilde{z}_\ell\|^2 = \|(\tilde{z}_\ell)_\pa\|^2 + \|(\tilde{z}_\ell)_\perp\|^2$. We use \eqref{eq:pitstop_meta} to prove \cref{item:three_meta}. Note that since \begin{align}G_\ell = W_\ell^{-1}\left( \sum_{\pi\in  \supp \rho_\ell}\rho_\ell(\pi)\theta^{\pi}_\ell (\theta^\pi_\ell)^\top\right)W_\ell^{-1},
    \end{align}
    where $\rho_\ell$ is the approximate design for $\Theta_\ell$ in \cref{para:special_paragraph}, we have 
    \begin{align}
     \forall \theta \in W_\ell^{-1}\Theta_\ell, \quad  \|\theta\|_{G_\ell^\dagger} \leq \sqrt{2 d}.\label{eq:reas_meta}   
    \end{align}
With this, we have 
    \begin{align}
        \sup_{\theta\in\Theta_\ell}\abs{\inner{\theta}{w_\ell}}& = \sup_{\theta\in\Theta_\ell}\abs{\inner{\theta}{\W_\ell^{-1}\tilde{z}_\ell}},\nn \\
    &=\sup_{\theta\in W_\ell^{-1}\Theta_\ell}\abs{\inner{\theta}{\tilde{z}_\ell}},\nn \\
    &\le \sup_{\theta\in W_\ell^{-1}\Theta_\ell}\norm{\theta}\norm{(\tilde{z}_\ell)_\pa} + \sup_{\theta\in W_\ell^{-1}\Theta_\ell}\abs{\inner{\theta}{(\tilde{z}_\ell)_\perp}}, 
    \intertext{and so by \cref{lem:fanc} and $d_\nu \coloneqq 5 d \log(1 + 16 H^4 \nu^{-4})$,}  
    &\le  \sqrt{\frac{d_\nu}{c}} + \sup_{\theta\in W_\ell^{-1}\Theta_\ell}\norm{\theta}_{G_\ell^\dag} \cdot\norm{(\tilde{z}_\ell)_\perp}_{ G_\ell}, \nn \\
    \intertext{and so by \eqref{eq:reas_meta}, we have}
    &\le  \sqrt{\frac{d_\nu}{c}} + \sqrt{2d}\cdot  \norm{(\tilde{z}_\ell)_\perp}_{G_\ell}, \nn \\
    &\le  \sqrt{\frac{d_\nu}{c}} + \sqrt{2d \cdot (\tilde{z}_\ell)_\perp^\top (\zeta I) (\tilde{z}_\ell)_\perp},\ \ \text{(see below)} \label{eq:mobile_meta} \\
    &\le  \sqrt{\frac{d_\nu}{c}} + \sqrt{2d\cdot \zeta},\nn \\
    &  =  \sqrt{\frac{d_\nu}{c}} + \frac12,\nn \quad \text{(since $\zeta= (8d)^{-1}$)} \\
   &  \leq 1, 
    \end{align}
    where in \eqref{eq:mobile_meta} follows from the fact that $(\cdot)_\perp = \Proj_{\cS(G_\ell, \zeta)^\perp}(\cdot)$; and the last inequality uses that $c= 20 d \log (1 + 16 \nu^{-4}H^4)= 4 d_\nu$.
\fakepar{Case where $w_\ell=0$} Note that $w_\ell=0$ only if $\max_{\pi \in \Psi}|\Delta_{h,\ell}(\pi)| \leq\frac{8c\sqrt{\lambda} d\tilde{d}H}{\sqrt{\zeta}\mu}+ \frac{8cd\nu L}{\mu \lambda}$, which implies \eqref{eq:fitting_meta} and completes the proof.  
\end{proof} 
\section{Guarantees of $\texttt{DesignDir}_h$}
\label{app:designdir}
In this section, we present the guarantees for the \texttt{DesignDir} subroutine of \mainalg{} (see \cref{sec:highlevel} for an intuitive explanation of the results). The following subsection outlines these guarantees, with the proofs deferred to \cref{sec:designdirproofs}.
\subsection{Statement of Guarantees}

\begin{lemma}[Concentration result for $\texttt{DesignDir}_h$]
    \label{lem:concentration-new}
    Let $h\in[H]$, $\Psi_{0:h-1}$, $\pihat_{1:h}$, $U_{h}$, $\beta$, $n$ be given and consider a call to 
    $\texttt{DesignDir}_{h}(\Psi_{0:h-1}, \pihat_{1:h} ,U_{h}, \beta, n)$. Then, for any $\delta\in(0,1)$, with probability at least $1-\delta$, the variables in \cref{alg:fitbonus} satisfy for all $h\in[H]$, $i\in[d]$, $\ell\in[0 \ldotst h-1]$, $(\pi,v)\in \Psi\ind{t}_\ell$, and $s\in \{-,+\}$\emph{:}
     \begin{align}
    & \left\|\E^{\pi\circ_{\ell+1}\pihat_{\ell+1:h}}\left[\phih(\x_h,\pi_{h,i,s}(\x_h)) \cdot \mathbb{I}\{s\cdot \phih(\x_h,\pi_{h,i,s}(\x_h))^\top v_{h,i}\geq 0\}\right] - u_{h,i,\ell, \pi, s}\right\|\nn \\
     &  \leq 2 \sqrt{\frac{2\max_{j\in[H]}\log (4 H^2 d|\Psi_j|/\delta)}{n}}. \label{eq:standconcenc-new} 
     \end{align} 
\end{lemma}

\begin{lemma}
\label{lem:designboundmeta}
Consider a call to $\texttt{DesignDir}_{h}(\Psi_{0:h-1}, \pihat_{1:h} ,U_{h}, \beta, n)$ (\cref{alg:fitbonus}) for some given $h\in[H]$, $\Psi_{0:h-1}$, $\pihat_{1:h}$, $U_{h}$, $\beta$, and $n$. Then, for any $\delta'\in(0,1)$, with probability at least $1-\delta'$, the output $(u_h,\tilde{\pi}_{1:h}, v_h)$ of $\texttt{DesignDir}_h$ satisfy for all $h\in[H]$,
\begin{gather}
\left\|\E^{\tilde\pi_{1:h}}\left[\phih(\x_h,\a_h) \cdot \mathbb{I}\{\phih(\x_h,\a_h)^\top v_{h}\geq 0\}\right] - u_{h}\right\| \leq 2 \sqrt{\frac{2\max_{j\in[H]}\log (4 H^2 d|\Psi_j|/\delta')}{n}}, \label{eq:cod}
\intertext{and for all $\ell\in[0 \ldotst h-1]$ and $(\pi,v)\in \Psi_\ell$\emph{:}}
\E^{\pi \circ_{\ell+1} \pihat_{\ell+1:H}}\left[\|\phih(\x_h,\a_h)\|_{(\beta I + U_h)^{-1}} \right]  \leq 2 \sqrt{d}\|u_h\|_{(\beta I + U_h)^{-1}} + \frac{4 d}{\beta} \sqrt{\frac{2\max_{j\in[H]}\log (4 H^2 d|\Psi_j|/\delta')}{ n}}. \label{eq:elliptical}
\end{gather} 
\end{lemma}

\begin{lemma}[Guarantee of $\texttt{DesignDir}$ for $\mainalg$]
    \label{lem:designbound}
    Let $\veps,\delta\in(0,1)$ be given and consider a call $\mainalg(\Pi_\bench,\veps, \delta)$ (\cref{alg:ops-dp}). Then, for any $\delta\in(0,1)$, there is an event $\cE^{\conc}$ of probability at least $1-\delta/2$, under which for all $t\in[T]$ and $h\in[H]$, the output $(u\ind{t}_h, \tilde\pi\ind{t}_{1:h}, v\ind{t}_h)$ of $\texttt{DesignDir}_h$ in \cref{line:designdir} satisfies:
    \begin{gather}
\left\|\E^{\tilde\pi_{1:h}\ind{t}}\left[\phih(\x_h,\a_h) \cdot \mathbb{I}\{\phih(\x_h,\a_h)^\top v\ind{t}_{h}\geq 0\}\right] - u_{h}\ind{t}\right\| \leq \veps_\conc \coloneqq 2  \sqrt{\frac{2\log (4 H^3 d T/\delta)}{n_\traj}}, \label{eq:cod2}
\intertext{and for all $\ell\in[0 \ldotst h-1]$ and $(\pi,v)\in \Psi\ind{t}_\ell$\emph{:}}
\E^{\pi \circ_{\ell+1} \pihat_{\ell+1:H}\ind{t}}\left[\max_{a\in\cA}\|\phih(\x_h,a)\|_{(\beta I + U\ind{t}_h)^{-1}} \right]  \leq 2 \sqrt{d}\|u\ind{t}_h\|_{(\beta I + U\ind{t}_h)^{-1}} + 2 d \beta^{-1} \veps_\conc. \label{eq:elliptical2}
    \end{gather} 
    Furthermore, $\Psi\ind{t}_h$ satisfies $\Psi\ind{t}_h = \{(\tilde\pi_{1:h}\ind{\tau}, v\ind{\tau}_h) \mid \tau \in [h-1]\}$.
\end{lemma}
\begin{proof}
    The result follows from \cref{lem:designboundmeta} with $\delta = \delta'/(HT)$, \cref{lem:unionbound} (essentially the union bound over $t\in[T]$ and $h\in[H]$), and the fact that $|\Psi_h\ind{t}| \leq T$, for all $t\in [T]$ and $h\in[H]$. 
\end{proof}

\subsection{Proofs}
\label{sec:designdirproofs}

\begin{proof}[Proof of \cref{lem:concentration-new}]
    For each $h\in[H]$, $\ell\in [0 \ldotst h-1]$, and $(\pi,v)\in \Psi_\ell$, the dataset $\cD_{\ell,\pi}$ in \cref{alg:fitbonus} consists of i.i.d.~trajectories sampled from $\P^{\pi\circ_{\ell+1}\pihat_{\ell+1:h}}$. What is more, by \cref{line:minus} and \cref{line:plus} in \cref{alg:fitbonus}, $u_{h,i,\ell,\pi,s}$ satisfies: 
    \begin{align}
    u_{h,i,\ell,\pi,s} = \frac{1}{n_\traj}\sum_{(x_{1:H},a_{1:H})\in \cD_{\ell,\pi}} \phih(x_h, \pi _{h,i,s}(x_h))  \cdot \mathbb{I}\{ s \cdot\phih(x_h, \pi _{h,i,s}(x_h))^\top v_{h,i}   \geq  0\}.
    \end{align}
    Thus, by Hoeffding inequality, the fact that $\phih(\cdot,\cdot)\in \bbB(1)$, and the union bound over $h\in[H]$, $i\in[d]$, $\ell\in[0 \ldotst h-1]$, $(\pi,v)\in \Psi_\ell$, and $s\in \{-,+\}$, there is an event of probability at least $1-\delta$ such that the desired inequality in \eqref{eq:standconcenc-new} holds for all $h\in[H]$, $i\in[d]$, $\ell\in[0 \ldotst h-1]$, $(\pi,v)\in \Psi_\ell$, and $s\in \{-,+\}$.
\end{proof}

\begin{proof}[Proof of \cref{lem:designboundmeta}]
Fix $\delta'\in (0,1)$.
In this proof, we condition on the event of \cref{lem:concentration-new} for $\delta = \delta'$; in particular, we assume that \eqref{eq:standconcenc-new} holds for all $h\in[H]$, $i\in[d]$, $\ell\in[0 \ldotst h-1]$, $(\pi,v)\in \Psi_\ell$, $s\in \{-,+\}$, and $\delta = \delta'$.

We note that \eqref{eq:cod} follows immediately from \cref{lem:concentration-new} and the definitions of $u_h$, $\tilde\pi_{1:h}$, and $v_h$ in \cref{alg:fitbonus}. 

We now show \eqref{eq:elliptical}. Fix $h\in[H]$, $\ell\in[0 \ldotst h-1]$, and $(\pi,v)\in \Psi_\ell$. We have
\begin{align}
& \E^{\pi \circ_{\ell+1} \pihat_{\ell+1:H}}\left[\max_{a\in \cA}\|\phih(\x_h,a)\|_{(\beta I + U_h)^{-1}} \right] \nn \\ 
&= \E^{\pi \circ_{\ell+1} \pihat_{\ell+1:H}}\left[\max_{a\in \cA}\| (\beta I + U_h)^{-1/2}\phih(\x_h,a)\| \right]
\nn \\
& \leq \sum_{i\in[d]} \E^{\pi \circ_{\ell+1} \pihat_{\ell+1:H}}\left[\max_{a\in \cA} \left|e_i^\top (\beta I + U_h)^{-1/2} \phih(\x_h,a)\right|\right], \nn \\
& = \sum_{i\in[d]} \E^{\pi \circ_{\ell+1} \pihat_{\ell+1:H}}\left[\max_{a\in \cA}\left|v_{h,i}^\top \phih(\x_h,a)\right|\right],\quad \text{(by definition of $v_{h,i}$ in \cref{alg:fitbonus})} \nn \\
&  = \sum_{i\in[d]} \E^{\pi \circ_{\ell+1} \pihat_{\ell+1:H}}\left[\max_{a\in \cA}v_{h,i}^\top \phih(\x_h,a)\cdot \mathbb{I}\{v_{h,i}^\top \phih(\x_h,a) \geq 0\}\right] \nn \\
& \quad + \sum_{i\in[d]} \E^{\pi \circ_{\ell+1} \pihat_{\ell+1:H}}\left[\max_{a\in \cA}-v_{h,i}^\top \phih(\x_h,a)\cdot \mathbb{I}\{v_{h,i}^\top \phih(\x_h,a) \leq 0\}\right] ,\nn \\
& = \sum_{i\in[d]} \E^{\pi \circ_{\ell+1} \pihat_{\ell+1:H}}\left[v_{h,i}^\top \phih(\x_h,\pi_{h,i,+}(\x_h))\cdot \mathbb{I}\{v_{h,i}^\top \phih(\x_h,\pi_{h,i,+}(\x_h)) \geq 0\}\right] \nn \\
& \quad - \sum_{i\in[d]} \E^{\pi \circ_{\ell+1} \pihat_{\ell+1:H}}\left[v_{h,i}^\top \phih(\x_h,\pi_{h,i,-}(\x_h))\cdot \mathbb{I}\{v_{h,i}^\top \phih(\x_h,\pi_{h,i,-}(\x_h)) \leq 0\}\right],\quad \text{(by def.~of $\pi_{h,i,\pm}$ in Alg.~\ref{alg:fitbonus})} \nn \\
\intertext{by \cref{lem:concentration-new} and $\|v_{h,i}\|\leq \beta^{-1}$,}
& = \sum_{i\in[d]} \left(\inner{v_{h,i}}{u_{h,i,\ell,\pi,+}} - \inner{v_{h,i}}{u_{h,i,\ell,\pi,-}}\right) + \frac{4 d}{\beta} \sqrt{\frac{2\max_{j\in[H]}\log (4 H^2 d|\Psi_j|/\delta')}{n_\traj}},  \nn \\
& \leq 2\sum_{i\in[d]} |v_{h,i}^\top u_h| +  \frac{4d}{\beta} \sqrt{\frac{2\max_{j\in[H]}\log (4 H^2 d|\Psi_j|/\delta')}{ n_\traj}},\quad (\text{by definition of $u_h$ in \cref{alg:fitbonus}}) \nn \\
& \leq 2 \sqrt{d} \|u_h\|_{(\beta I + U_h)^{-1}} + \frac{4 d}{\beta}  \sqrt{\frac{2\max_{j\in[H]}\log (4 H^2 d|\Psi_j|/\delta')}{n_\traj}}, 
\end{align}
where the last inequality follows by Jensen's inequality; in particular,
\begin{align} %
    \sum_{i\in[d]} |v_{h,i}^\top u_h|  &=  \sum_{i\in[d]} |e_i^\top (\beta I + U_h)^{-1/2} u_h|  
   = \|(\beta I + U_h)^{-1/2} u_h\|_1 \leq \sqrt{d} \|(\beta I + U_h)^{-1/2} u_h\|_2 = \|u_h\|_{(\beta I + U_h)^{-1}}.
\end{align} 
This completes the proof. 
\end{proof}

\section{Guarantees of $\texttt{FitOptValue}_h$}
\label{sec:fitbonusproof}

In this section, we present the guarantees for the \texttt{FitOptValue} subroutine of \mainalg{} (see \cref{sec:highlevel} for an intuitive explanation of the results). The following subsection outlines these guarantees, with the proofs deferred to \cref{sec:fitbonusproof}. In \cref{sec:addition-struct}, we present additional structural results that are used in proving the guarantees.

\subsection{Statement of Guarantees}

\begin{lemma}
    \label{lem:littlegen1}
Let $h\in[H]$, $\Psi_{h-1}, \pihat_{h+1:H}, U_{h+1:H}, W_{h+1:H}, \Pi', \mu, \nu, \lambda, \beta, \veps, \delta, n$ be given and consider a call to 
    $\texttt{FitOptValue}_{h}(\Psi_{h-1}, \pihat_{h+1:H} ,U_{h+1:H}, {W}_{h+1:H};\Pi',\mu,\nu,\lambda, \beta, \veps,  \delta, n)$. Then, for any $\delta' \in(0,1)$, with probability at least $1-\delta'$, the variables in \cref{alg:fitoptvalue} satisfy for all $\ell\in[h+1\ldotst H]$, $(\pi,v)\in \Psi_{h-1}$, and $\tilde\pi \in \Pi'$:
    \begin{align}
& \left|\Delta_{h,\ell}(\pi, v, \tilde\pi)- \E^{\pi \circ_h \tilde\pi \circ_{h+1}\pihat_{h+1:H}}\left[\mathbb{I}\{\phiho(\x_{h-1},\a_{h-1})^\top v\geq 0\} \cdot (b_\ell(\x_\ell,\a_\ell) - \phih(\x_h,\a_h)^\top \thetahat^\bo_{h,\ell}) \right]  \right|\nn \\
& \leq \frac{4 A H}{\lambda}\sqrt{\frac{\log (2H^2 |\Psi_{h-1}| \cG(\Pi',n)/\delta')}{n}},\label{eq:sandw}
    \end{align} 
    where $\cG$ is the growth function defined in \cref{sec:benchmark}.
\end{lemma}

\begin{lemma}
    \label{lem:littlegen3}
    Let $h\in[H]$, $\Psi_{h-1}, \pihat_{h+1:H}, U_{h+1:H}, W_{h+1:H}, \Pi', \mu, \nu, \lambda, \beta, \veps, \delta, n$ be given and consider a call to 
    $\texttt{FitOptValue}_{h}(\Psi_{h-1}, \pihat_{h+1:H} ,U_{h+1:H}, {W}_{h+1:H};\Pi',\mu,\nu,\lambda, \beta,\veps, \delta, n)$. Then, for any $\delta' \in(0,1)$, with probability at least $1-\delta'$, the variables in \cref{alg:fitoptvalue} satisfy for all $(\pi,v)\in \Psi_{h-1}$ and $\tilde\pi \in \Pi'$:
    \begin{gather}
          \left|  \E^{\pi \circ_h \tilde\pi}\left[\mathbb{I}\{\phiho(\x_{h-1},\a_{h-1})^\top v\geq 0\} \cdot \left(\phih(\x_h,\a_h)^\top \thetahat^{\re}_h - Q_h^{\pihat}(\x_h,\a_h)\right)  \right] \right| \leq \veps_{\reg}^{\re}(\delta', \Pi', |\Psi_{h-1}|,n), 
        \shortintertext{where}
        \veps_{\reg}^{\re}(\delta', \Pi', |\Psi|,n)   \coloneqq  A H \sqrt{\lambda |\Psi|} + 2A\sqrt{\frac{d \log(\frac{2}{\delta'\lambda})}{n}}+\frac{4 A H}{\lambda}\sqrt{\frac{\log (4|\Psi|\cG(\Pi',n) /\delta')}{n}}, \label{eq:professional-new}
    \end{gather} 
    where $\cG$ is the growth function defined in \cref{par:bench}.
\end{lemma}

\begin{lemma}
    \label{lem:littlegen2}
    Let $h\in[H]$, $\Psi_{h-1}, \pihat_{h+1:H}, U_{h+1:H}, W_{h+1:H}, \Pi', \mu, \nu, \lambda, \beta, \veps, \delta, n$ be given and consider a call to 
    $\texttt{FitOptValue}_{h}(\Psi_{h-1}, \pihat_{h+1:H} ,U_{h+1:H}, {W}_{h+1:H};\Pi',\mu,\nu,\lambda, \beta,\veps,\delta, n)$. Further, for any $\ell\in[h+1 \ldotst H]$, let $\rho_\ell$ be the approximate design for $\Theta_\ell$ in \cref{para:special_paragraph} and define \begin{align}G_\ell \coloneqq W_\ell^{-1}\left( \sum_{\pi\in  \supp \rho_\ell}\rho_\ell(\pi)\theta^{\pi}_\ell (\theta^\pi_\ell)^\top\right)W_\ell^{-1}\in \reals^{d\times d},\quad \text{and} \quad  z_{\pa} \coloneqq \proj_{\cS(G_\ell, \zeta)}(z), \quad \forall z \in \reals^d,\label{eq:Gl} 
    \end{align}
    where $\zeta \coloneqq (8d)^{-1}$. Then, for any $\delta' \in(0,1)$, with probability at least $1-\delta'$, the variables in \cref{alg:fitoptvalue} satisfy for all $\ell\in[h+1\ldotst H]$, $(\pi,v)\in \Psi_{h-1}$, $\tilde\pi \in\Pi'$, and $z,y \in \bbB(1)$:
    \begin{align}
    &\left|\sum_{(x_{1:H},a_{1:H},r_{1:H})\in \cDhat_{h,\pi}} \frac{A \cdot\mathbb{I}\{\phiho(x_{h-1},a_{h-1})^\top v\geq 0, \tilde\pi(x_h)=a_h\}}{n}  \cdot \left(z^\top_{\pa} B_\ell(x_\ell,a_\ell) y - \hat \vartheta^{\bo}_{h,\ell}[\phih(x_{h},a_{h}),z_\pa,y]\right)\right|\nn \\
    &\leq \veps_{\reg}^{\bo}(\delta',\Pi', |\Psi_{h-1}|,n), 
    \end{align}
  where for $\tilde{d}\coloneqq 4 d \log \log d + 16$, $d_\nu \coloneqq 5 d \log (1 + 16 H^4 \nu^{-4})$, and $\cG$ as in \cref{sec:benchmark} (growth function):
  \begin{align}
 \veps_{\reg}^{\bo}(\delta', \Pi', |\Psi|,n) \coloneqq  \frac{\veps}{256 |\Psi| d_\nu d^3 H^5}  + \frac{8A\tilde{d}  H^2}{ \mu} \sqrt{2d\lambda |\Psi|} + 2A\sqrt{ \frac{d \log(\frac{2}{\delta'\lambda})}{n}} \nn \\ \hspace{2cm} + \frac{16 H^2 A \tilde{d} d}{\mu}   \sqrt{ \frac{2\log(2^{12} d^3 d_\nu  H^6  |\Psi|^2\cG(\Pi',n) \veps^{-1}/\delta')}{n}}. \label{eq:vepsb} %
    \end{align} 
\end{lemma}

\begin{lemma}
    \label{lem:decoy-find-new}
    Let $h\in[H]$, $\Psi_{h-1}, \pihat_{h+1:H}, U_{h+1:H}, W_{h+1:H}, \Pi', \mu, \nu, \lambda, \beta, \veps, \delta, n$ such that $W_{h+1},\dots, W_H \in \pd$ be given and consider a call to 
    $\texttt{FitOptValue}_{h}(\Psi_{h-1}, \pihat_{h+1:H} ,U_{h+1:H}, {W}_{h+1:H};\Pi',\mu,\nu,\lambda, \beta,\veps,\delta,  n )$. Then, with probability at least $1-\frac{\delta}{2TH}$, the variables in \cref{alg:fitoptvalue} are such that: if $\cH \neq \emptyset$, then for all $\ell \in \cH$:
    \begin{enumerate}
        \item $\|w_\ell\| \leq \nu^{-1}$; \label{item:one}
        \item $\|W_\ell w_\ell\| \geq 1/2$; and \label{item:two}
        \item $\sup_{\theta \in \Theta_\ell} |\inner{\theta}{w_\ell}|\leq 1$. \label{item:three}
    \end{enumerate}
    On the other hand, if $\cH = \emptyset$, then for all $(\pi,v)\in \Psi_{h-1}$ and $\tilde \pi \in \Pi'$,
\begin{align}
  & \left|\E^{\pi \circ_h \tilde\pi}\left[\mathbb{I}\{\phiho(\x_{h-1},\a_{h-1})^\top v\geq 0\} \cdot (\wtilde{Q}_h(\x_h,\a_h) - \phih(\x_h,\a_h)^\top \thetahat_{h}) \right]\right| \nn \\
  & \leq  \frac{4 A H^2}{\lambda}\sqrt{\frac{\log (4H^3 T |\Psi_{h-1}| \cG(\Pi',n)/\delta)}{n}}  + \frac{32 d_\nu d\nu H^2 A}{\mu \lambda}\nn \\
  &\qquad +8 d_\nu dH\veps_{\reg}^{\bo}(\tfrac{\delta}{6 HT},\Pi', |\Psi_{h-1}|,n)  + \veps_{\reg}^{\re}(\tfrac{\delta}{6 HT},\Pi', |\Psi_{h-1}|,n), \label{eq:fitting}
\end{align}
where $d_\nu \coloneqq 5 d \log (1 + 16 H^4 \nu^{-4})$, $\wtilde{Q}_h(x,a) = Q^{\pihat}_{h}(x,a)+ \E^{\pihat}[\sum_{\ell=h+1}^H b_\ell(\x_h,\a_h) \mid \x_h =a,\a_h =a]$, for $(x,a)\in \cX_h\times \cA$, and $\veps_{\reg}^{\bo}$ [resp.~$\veps_{\reg}^{\re}$] is as in \cref{lem:littlegen2} [resp.~\cref{lem:littlegen3}].
\end{lemma}

    \begin{lemma}[Guarantee of $\texttt{FitOptValue}_h$ for $\mainalg$]
        \label{lem:fitoptvalue}
        Let $\veps,\delta\in(0,1)$ be given and consider a call $\mainalg(\Pi_\bench, \veps, \delta)$ (\cref{alg:ops-dp}) with $\Pi_\bench$ as in \cref{par:bench}. Further, let $\veps_{\reg}^{\bo}$ [resp.~$\veps_{\reg}^{\re}$] be as in \cref{lem:littlegen2} [resp.~\cref{lem:littlegen3}]. Then, there is an event $\cE^{\reg}$ of probability at least $1-\delta/2$, under which for all $t\in[T]$ and $h\in[H]$:
        \begin{itemize}
            \item $W\ind{t}_h$ is a valid $\nu$-preconditioning for layer $h$, where $\nu$ is as in \cref{alg:ops-dp}; and %
            \item If $W\ind{t+1}_{1:H}= W_{1:H}\ind{t}$ at the end of iteration $t$ (i.e., the preconditioning is not updated during the calls to $\texttt{FitOptValue}_{h}$ for $h\in[H]$ at iteration $t$), then for all $(\pi,v)\in \Psi\ind{t}_{h-1}$ and $\tilde\pi \in \Pi_\bench$:
     \begin{align}
                & \left|\E^{\pi \circ_h \tilde\pi}\left[\mathbb{I}\{\phiho(\x_{h-1},\a_{h-1})^\top v \geq 0\} \cdot (Q\ind{t}_h(\x_h,\a_h) - \Qhat\ind{t}_h(\x_h,\a_h)) \right]\right|\\
                & \leq  \veps_{\reg} \coloneqq \frac{4 A H^2}{\lambda}\sqrt{\frac{\log (4H^3 T^2 \cG(\Pi_\bench,n_\traj)/\delta)}{n_\traj}}+ \frac{160 d^2\nu H^2 A \log (1 + 16 H^4 \nu^{-4})}{\mu \lambda} \nn \\
                & \quad  \quad + 50  d^2H\veps_{\reg}^{\bo}(\tfrac{\delta}{6HT},\Pi_\bench, T,n_\traj)\cdot \log (1 + 16 H^4 \nu^{-4})  + \veps_{\reg}^{\re}(\tfrac{\delta}{6 HT},\Pi_\bench, T, n_\traj), 
              \end{align}
              where for all $(x,a)\in \cX_h\times \cA$, ${Q}\ind{t}_h(x,a) = Q^{\pihat\ind{t}}_{h}(x,a) + \E^{\pihat\ind{t}}[\sum_{\ell=h}^H b\ind{t}_\ell(\x_\ell) \mid \x_h =x,\a_h =a]$, $\Qhat\ind{t}_h(x,a) = \phih(x,a)^\top \thetahat\ind{t}_{h} + b\ind{t}_h(x)$, and ${b}\ind{t}_h(x) \coloneqq \min\left(H, \frac{\veps}{4 H} \cdot\max_{a'\in \cA} \|\phiell(x, a')\|_{(\beta I + U\ind{t}_h)^{-1}}\right) \cdot \mathbb{I}\{\|\varphi(x;W\ind{t}_h)\|\geq \mu\}$.
        \end{itemize} 
    \end{lemma}

   \subsection{Proofs} 
    \begin{proof}[Proof of \cref{lem:fitoptvalue}]
        Fix $\delta\in(0,1)$. By the update rule in \cref{line:updateprecond}, we have that for all $t\in[T-1]$ and $h\in[H]$:
        \begin{align}
            W\ind{t+1}_h = \left((W\ind{t}_h)^{-2} +\sum_{\ell \in[h-1]} w\ind{t,\ell}_{h}(w\ind{t,\ell}_{h})^\top \right)^{-1/2}.   \label{eq:hbo}
        \end{align}
        Now, for a given iteration $t\in[T-1]$ and $h\in[H]$, under the success event of \cref{lem:decoy-find-new}, the output $w\ind{t,h}_{h+1:H}$ of the call to $\texttt{FitOptValue}_h$ in \cref{line:fitoptval} is such that
      \begin{itemize} 
        \item For all $\ell\in[h+1\ldotst H]$ either $w\ind{t,h}_\ell =0$; or 
    \begin{align}
        \|w\ind{t,h}_\ell\| \leq \nu^{-1},\qquad  \|W\ind{t}_\ell w_\ell\| \geq 1/2, \qquad \text{and} \qquad  \sup_{\theta \in \Theta_\ell} |\inner{\theta}{w\ind{t,h}_\ell}|\leq 1.
        \end{align}
       \item When $w\ind{t,h}_{\ell}=0$ for all $\ell\in[h+1\ldotst H]$ (corresponding to the case where $\cH$ within the call to $\texttt{FitOptValue}_{h}$ is empty), we have that for all $(\pi,v)\in \Psi\ind{t}_{h-1}$ and $\tilde \pi \in \Pi_\bench$:
       \begin{align}
          \left|\E^{\pi \circ_h \tilde\pi}\left[\mathbb{I}\{\phiho(\x_{h-1},\a_{h-1})^\top v \geq 0\} \cdot (\wtilde{Q}\ind{t}_h(\x_h,\a_h) - \phih(\x_h,\a_h)^\top \thetahat\ind{t}_{h}) \right]\right| \leq \veps_{\reg},
       \end{align}
       where $\veps_{\reg}$ is as in the lemma statement; to get this bound we use \eqref{eq:fitting} and the fact that $|\Psi_{h-1}\ind{t}|\leq T$, for all $t\in[T]$. 
    \end{itemize}
Therefore, by \cref{lem:decoy-find-new} and \cref{lem:unionbound} (essentially the union bound over $t\in[T-1]$ and $h\in[H]$), we get that there is event $\cE^{\reg}$ of probability at least $1-\delta/2$, under which for all $t\in[T-1]$ and $h\in[H]$:
\begin{enumerate} 
    \item \label{item:oneone} For all $\ell\in[h+1\ldotst H]$ either $w\ind{t,h}_\ell =0$; or   
\begin{align}
    \|w\ind{t,h}_\ell\| \leq \nu^{-1},\quad  \|W\ind{t}_\ell w_\ell\| \geq 1/2, \quad \text{and} \quad  \sup_{\theta \in \Theta_\ell} |\inner{\theta}{w\ind{t,h}_\ell}|\leq 1;
    \end{align}
   \item  \label{item:twotwo} When $w\ind{t,h}_{\ell}=0$ for all $\ell\in[h+1\ldotst H]$ (corresponding to the case where $\cH$ within the call to $\texttt{FitOptValue}_{h}$ is empty), we have that for all $(\pi,v)\in \Psi\ind{t}_{h-1}$ and $\tilde \pi \in \Pi_\bench$:
   \begin{align}
      \left|\E^{\pi \circ_h \tilde\pi}\left[\mathbb{I}\{\phiho(\x_{h-1},\a_{h-1})^\top v \geq 0\} \cdot (\wtilde{Q}\ind{t}_h(\x_h,\a_h) - \phih(\x_h,\a_h)^\top \thetahat\ind{t}_{h}) \right]\right| \leq \veps_{\reg}.
   \end{align}
\end{enumerate}
\cref{item:oneone} together with \eqref{eq:hbo} implies that for all $h\in[H]$ and $t\in[T-1]$, $W\ind{t}_h$ is a valid $\nu$-preconditioning for layer $h$ (\cref{def:precond}). Further, by \cref{item:twotwo}, if $W\ind{t+1}_{h} = W\ind{t}_h$ at the end of iteration $t$, or equivalently if $w\ind{t,h}_{\ell}=0$ for all $h\in[H]$ and $\ell\in[h+1\ldotst H]$, then for all $(\pi,v)\in \Psi\ind{t}_{h-1}$ and $\tilde \pi \in \Pi_\bench$:
\begin{align}
    \left|\E^{\pi \circ_h \tilde\pi }\left[\mathbb{I}\{\phiho(\x_{h-1},\a_{h-1})^\top v \geq 0\} \cdot (\wtilde{Q}\ind{t}_h(\x_h,\a_h) - \phih(\x_h,\a_h)^\top \thetahat\ind{t}_{h}) \right]\right| \leq \veps_{\reg}. \label{eq:alley}
 \end{align}
 Using the fact that $Q\ind{t}_h(x,a) = \wtilde{Q}\ind{t}_h(x,a) + b\ind{t}_h(x,a)$ for all $(x,a)\in \cX_h \times \cA$ and $h\in[H]$ completes the proof.
\end{proof}

\label{sec:fitoptvalueproof}
\begin{proof}[Proof of \cref{lem:littlegen1}]
    By definition of $\Delta_{h,\ell}$ in \cref{alg:fitoptvalue}, we have for $(\pi,v)\in \Psi_{h-1}$ and $\tilde\pi \in  \Pi'$:
    \begin{align}
     &\Delta_{h, \ell}(\pi,v,\tilde\pi)\nn \\
     & =\frac{A}{n}\sum_{(x_{1:H},a_{1:H},r_{1:H})\in \cDhat_{h,\pi}} \mathbb{I}\{\tilde\pih(x_h)=a_h\}\cdot  \mathbb{I}\{\phiho(x_{h-1},a_{h-1})^\top v\geq 0\}\cdot   \left(b_\ell(x_\ell) -  \phih(x_{h},a_{h})^\top \hat\theta^{\bo}_{h,\ell}\right),
    \end{align} 
    where the trajectories in $\cDhat_{h,\pi}$ are sampled i.i.d.~according to $\P^{\pi\circ_h  \pi_\unif \circ_{h+1}\pihat_{h+1:H}}$. Note that we have the following: 
    \begin{itemize}
    \item Thanks to the importance weight $A$ and the indicator $\mathbb{I}\{\tilde\pi(\x_h)=\a_h\}$, we have  \begin{align}&\E^{\pi\circ_h \pi_\unif\circ_{h+1} \pihat_{h+1:H}} \left[\mathbb{I}\{\tilde\pih(\x_h)=\a_h\}\cdot  \mathbb{I}\{\phiho(\x_{h-1},\a_{h-1})^\top v\geq 0\}\cdot   \left(b_\ell(\x_\ell) -  \phih(\x_{h},\a_{h})^\top \hat\theta^{\bo}_{h,\ell}\right)\right] \nn\\
        & = \E^{\pi \circ_h \tilde\pi \circ_{h+1}\pihat_{h+1:H}}\left[\mathbb{I}\{\phiho(\x_{h-1},\a_{h-1})^\top v\geq 0\} \cdot (b_\ell(\x_\ell,\a_\ell) - \phih(\x_h,\a_h)^\top \thetahat^\bo_{h,\ell}) \right];
        \end{align}
    \item For all $x'\in \cX_\ell$, we have $|b_\ell(x')| \leq H$;
    \item We have $\|\thetahat^\bo_{h,\ell}\| \leq \frac{H}{\lambda}$, (since $\thetahat^\bo_{h,\ell}=\Sigma_{h}^{-1}\sum_{(\pi,v)\in  \Psi_{h-1}}\sum_{(x_{1:H},a_{1:H},r_{1:H})\in \cDhat_{h,\pi}} \phih(x_{h},a_{h}) b_{\ell}(x_{\ell})$ and $\sigma_{\min}(\Sigma_h)\geq \lambda n |\Psi_{h-1}|$) and so by Cauchy Schwarz inequality, we have for all $(x,a)\in \cX_h\times \cA$, $|\phih(x,a)^\top \thetahat^\bo_{h,\ell}|\leq \frac{H}{\lambda}$.
    \end{itemize}
    Thus, by \cref{lem:rademacher} (generalization bound) and the union bound over $h\in[H]$, $\ell \in[h+1\ldotst H]$ and $(\pi,v)\in \Psi_{h-1}$, we get the desired result.
    \end{proof}

\begin{proof}[Proof of \cref{lem:littlegen3}]
    All the variables in the proof are as in \cref{alg:fitoptvalue}. We start by showing that for all $(\pi,v)\in \Psi_{h-1}$, there is an event $\cE(\pi,v)$ of probability at least $1-\delta'/|\Psi_{h-1}|$ under which \eqref{eq:professional-new} holds. Then, we apply the union bound to obtain the desired result.
    
    Fix $(\pi,v)\in \Psi_{h-1}$. By the linear-$Q^\pi$ assumption (\cref{assum:linearqpi}), there is a $\theta_h^{\pihat}\in \bbB(H)$ such that for all $(x,a)\in \cX_h\times \cA$:
    \begin{align}
       Q^{\pihat}_h(x,a) = \phih(x,a)^\top \theta_h^{\pihat}. \label{eq:loose}
    \end{align}
    Using this and \cref{thm:bandits-book-20.5}, we have that there is an event $\cE$ of probability at least $1-\delta' /2$ such that 
    \begin{gather}
       \|\hat{\theta}^\re_{h}- \theta^{\pihat}_{h}\|_{\Sigma_h} \leq \sqrt{\lambda n |\Psi_{h-1}|} \cdot \|\theta^{\pihat}_{h}\| + \sqrt{2d \log(2  /\delta') + \log \det (\Sigma_h \lambda^{-1} n^{-1}|\Psi_{h-1}|^{-1})}, \label{eq:paramconcenc2}
       \intertext{where}
       \Sigma_h\coloneqq \lambda n |\Psi_{h-1}| I + \sum_{(\pi',v')\in \Psi_{h-1}}\sum_{(x_{1:H},a_{1:H},r_{1:H}) \in \cDhat_{h,\pi'}} \phih(x_{h},a_{h}) \phih(x_{h},a_{h})^\top, 
    \end{gather}
    and $\hat{\theta}^{\re}_{h}$ is as in \cref{alg:fitbonus}.

    For the rest of the proof, we condition on $\cE$. By \eqref{eq:loose} and Jensen's inequality, we have 
    \begin{align}
    &  \left|\frac{A}{n}\sum_{(x_{1:H},a_{1:H},r_{1:H})\in \cDhat_{h,\pi}} \mathbb{I}\{\tilde\pih(x_h)=a_h\}\cdot  \mathbb{I}\{\phiho(x_{h-1},a_{h-1})^\top v\geq 0\} \cdot \left(\phih(x_h,a_h)^\top \thetahat^{\re}_h - Q_h^{\pihat}(x_h,a_h) \right)\right|\nn \\
    & \leq A \sqrt{\frac{1}{n}\sum_{(x_{1:H},a_{1:H},r_{1:H})\in \cDhat_{h,\pi}}  \mathbb{I}\{\tilde\pih(x_h)=a_h,\phiho(x_{h-1},a_{h-1})^\top v\geq 0\}^2 \cdot \inner{\phih(x_h,a_h)}{\thetahat^{\re}_h -  \theta^{\pihat}_h}^2 }, \nn \\
    & \leq A \sqrt{\frac{1}{n}\sum_{(\pi',v')\in \Psi_{h-1}} \sum_{(x_{1:H},a_{1:H},r_{1:H})\in \cDhat_{h,\pi'}}  \inner{\phih(x_h,a_h)}{\thetahat^{\re}_h -  \theta^{\pihat}_h}^2},\nn\\
    & \leq \frac{A}{\sqrt{n}} \|\thetahat^{\re}_h - \theta^{\pihat}_h\|_{\Sigma_h}, \nn \\
    & \leq A\sqrt{\lambda |\Psi_{h-1}|} \|\theta^{\pihat}_h\| + A \sqrt{\frac{2 d \log(2 / \delta')}{n} +  \frac{\log \det (\Sigma_h\lambda^{-1} n^{-1}|\Psi_{h-1}|^{-1})}{n}  },\quad \text{(by \eqref{eq:paramconcenc2})}\nn \\
    & \leq A H \sqrt{\lambda |\Psi_{h-1}|} + A \sqrt{\frac{2d \log(2 / \delta')}{n} + \frac{\log \det (\Sigma_h\lambda^{-1} n^{-1}|\Psi_{h-1}|^{-1})}{n}},\label{eq:milehigh}
    \end{align}
    where the last inequality follows by the fact that $\theta_h^{\pihat} \in \bbB(H)$.
    Now, by Jensen's inequality and the fact that $\|\Sigma_h\|_{\op} \leq (1+\lambda)\cdot|\Psi_{h-1}|\cdot n$, we have \[\log \det (\Sigma_h\lambda^{-1} n^{-1}|\Psi_{h-1}|^{-1}) \leq d \log \tr(\Sigma_h d^{-1} \lambda^{-1} n^{-1}|\Psi_{h-1}|^{-1})\leq d \log (1 + \lambda^{-1}).\] Plugging this into \eqref{eq:milehigh}, we get that 
    \begin{align}
       &   \left|\frac{A}{n}\sum_{(x_{1:H},a_{1:H},r_{1:H})\in \cDhat_{h,\pi}}\mathbb{I}\{\tilde\pih(x_h)=a_h\}\cdot  \mathbb{I}\{\phiho(x_{h-1},a_{h-1})^\top v\geq 0\} \cdot \left(\phih(x_h,a_h)^\top \thetahat^{\re}_h - Q_h^{\pihat}(x_h,a_h) \right)\right|\nn \\
       & \leq A H \sqrt{\lambda |\Psi_{h-1}|} + 2A\sqrt{n^{-1} d \log(2    \lambda^{-1}/\delta')}, \label{eq:eight}
    \end{align}
    Now, by \cref{lem:rademacher} (generalization bound) and the fact that $\thetahat^{\re}_h \in \bbB(H/\lambda)$; since \[\thetahat^{\re}_{h} = \Sigma_{h}^{-1}\sum_{(\pi',v')\in \Psi_{h-1}}\sum_{(x_{1:H},a_{1:H},r_{1:H})\in \cDhat_{h,\pi'}} \phih(x_{h},a_{h}) \cdot \sum_{\ell=h}^H r_{\ell}\] and $r_\ell\in[0,1]$ for all $\ell\in[H]$, there is an event $\cE'(\pi,v)$ of probability at least $1-\delta'/(2|\Psi_{h-1}|)$ under which for all $\tilde\pi\in \Pi'$:
    \begin{align}
    &\left|  \E^{\pi \circ_h \tilde\pi}\left[\mathbb{I}\{\phiho(\x_{h-1},\a_{h-1})^\top v\geq 0\} \cdot \left(\phih(\x_h,\a_h)^\top \thetahat^{\re}_h - Q_h^{\pihat}(\x_h,\a_h)\right)  \right] \right|  \nn \\
    & \leq \left|\frac{A}{n}\sum_{(x_{1:H},a_{1:H},r_{1:H})\in \cDhat_{h,\pi}} \mathbb{I}\{\tilde\pih(x_h)=a_h\}\cdot  \mathbb{I}\{\phiho(x_{h-1},a_{h-1})^\top v\geq 0\}  \cdot \left(\phih(x_h,a_h)^\top \thetahat^{\re}_h - Q_h^{\pihat}(x_h,a_h) \right)\right|\nn \\
    & \quad  +\frac{4A H}{\lambda}\sqrt{\frac{\log (4 |\Psi_{h-1}| \cG_h(\Pi',n)/\delta')}{n}}. \label{eq:nine}
    \end{align}
    Thus, by combining \eqref{eq:eight} and \eqref{eq:nine}, we get that under $\cE \cap \cE'(\pi, v)$: 
    \begin{align}
       &  \left|  \E^{\pi \circ_h \tilde\pi}\left[\mathbb{I}\{\phiho(\x_{h-1},\a_{h-1})^\top v\geq 0\} \cdot \left(\phih(\x_h,\a_h)^\top \thetahat^{\re}_h - Q_h^{\pihat}(\x_h,\a_h)\right)  \right] \right|\nn \\
       & \leq A H \sqrt{\lambda |\Psi_{h-1}|} + 2A\sqrt{ \frac{d \log(\frac{2}{\delta'\lambda})}{n}}+\frac{4 A H}{\lambda}\sqrt{\frac{\log (4|\Psi_{h-1}|\cG_h(\Pi',n) /\delta')}{n}},
    \end{align}
    for all $\tilde\pi\in \Pi'$. Now, the desired result follows by the union bound over $(\pi,v) \in \Psi_{h-1}$.
    \end{proof}
\begin{proof}[Proof of \cref{lem:littlegen2}] Let $\tilde{d}$, $d_\nu$, and $\zeta$ be as in the lemma statement and define \[\vepsd \coloneqq \frac{\veps \zeta \mu}{1024 |\Psi_{h-1}|d_\nu d^3 H^6 A}.\] All other variables in this proof are as in \cref{alg:fitoptvalue}. 

We start by showing that for all $\ell\in[h+1\ldotst H]$, $(\pi,v)\in \Psi_{h-1}$, and $z,y \in \bbB(1)$, there is an event $\cE(\ell, \pi,v,z,y)$ of probability at least $1-\delta' (\vepsd/3)^{2d}/(2|\Psi_{h-1}|H)$ under which the inequality in the lemma's statement holds. Then, we apply the union bound to obtain the desired result.  

Fix $\ell\in[h+1\ldotst H]$, $(\pi,v)\in \Psi_{h-1}$, and $z,y \in \bbB(1)$. Note that $B_\ell$ satisfies
\begin{align}
 \forall x\in\cX , \quad   B_\ell(x) =  b_\ell(x) \cdot \frac{\varphiell(x; W_\ell) \varphiell(x; W_\ell)^\top}{\|\varphiell(x; W_\ell)\|^2},
\end{align}
where $b_\ell$ is as in \cref{line:bonus} of \cref{alg:fitoptvalue}.
Combining this with the fact that $\|b_\ell\|_\infty \leq H$ and \cref{lem:dual-admissibility} implies that $F:x\mapsto z^\top_\parallel B_\ell(x) y$ is $\alpha$-admissible (\cref{def:admissible}) with $\alpha = \sqrt{\zeta} \mu/H$, and so by \cref{lem:admreal} there exists $\theta_{h,\ell} \in \bbB(4 \tilde{d} H/\alpha)$ such that 
\begin{align}
 \forall (x,a)\in \cX_h \times \cA, \quad   \E^{\pihat_{h+1:H}}[z^\top_\parallel B_\ell(\x_\ell) y \mid \x_{h}=x, \a_{h}=a] = \phih(x,a)^\top \theta_{h,\ell}.  \label{eq:conditional2}
\end{align}
Thus, by the law of total expectation and \cref{lem:rademacher} (generalization bound) with the facts that $\|\theta_{h,\ell}\| \leq 4 \tilde{d} H/\alpha$ and $\|B_\ell(x)\|_\op \leq H$, for all $x\in \cX_\ell$, there is an event $\cE(\ell, \pi, v, z, y)$ of probability at least $1-\delta' \cdot (\vepsd/3)^{2d}/(2 H |\Psi_{h-1}|)$ such that for all $\tilde\pi \in\Pi'$:
\begin{align}
  &  \left|\frac{A}{n}\sum_{(x_{1:H},a_{1:H},r_{1:H})\in \cDhat_{h,\pi}} \mathbb{I}\{\tilde\pih(x_h)=a_h\}\cdot  \mathbb{I}\{\phiho(x_{h-1},a_{h-1})^\top v\geq 0\}  \cdot \left(z^\top_{\pa} B_\ell(x_\ell) y - \phih(x_h, a_h)^\top \theta_{h,\ell} \right)\right|\nn\\ & \leq \frac{16 H A \tilde{d}}{\alpha}  \sqrt{ \frac{2 d\log(12 H |\Psi_{h-1}|\cG_h(\Pi',n) \vepsd^{-1}/\delta')}{n}}, \label{eq:hoef}
\end{align}
On the other hand, by \eqref{eq:conditional2} and \cref{thm:bandits-book-20.5}, we have that there is an event $\cE'$ of probability at least $1- \delta'/2$ such that 
\begin{gather}
   \|\hat{\theta}_{h,\ell}- \theta_{h,\ell}\|_{\Sigma_h} \leq \sqrt{\lambda n |\Psi_{h-1}|} \cdot \|\theta_{h,\ell}\| + \sqrt{2d \log(2/\delta') + \log \det (\Sigma_h \lambda^{-1} n^{-1}|\Psi_{h-1}|^{-1} )}, \label{eq:paramconcenc}
   \intertext{where}
   \Sigma_h\coloneqq \lambda n |\Psi_{h-1}| I + \sum_{(\pi',v')\in \Psi_{h-1}}\sum_{(x_{1:H},a_{1:H},r_{1:H}) \in \cDhat_{h,\pi'}} \phih(x_{h},a_{h}) \phih(x_{h},a_{h}), \nn \\
  \text{and}\quad \hat{\theta}_{h,\ell} \coloneqq \Sigma_h^{-1} \sum_{(\pi',v')\in \Psi_{h-1}}\sum_{(x_{1:H},a_{1:H},r_{1:H}) \in \cDhat_{h,\pi'}} \phih(x_{h},a_{h}) \cdot z^\top_\parallel B_\ell(x_{\ell})y.
\end{gather}
We now condition of $\cE(\ell, \pi,v,z,y) \cap \cE'$. Using that $\hat\theta_{h,\ell} = \hat\vartheta^{\bo}_{h,\ell}[\cdot, z_\pa, y]$, we get for all $\tilde\pi\in \Pi'$:
\begin{align}
   & \left|\frac{A}{n}\sum_{(x_{1:H},a_{1:H},r_{1:H})\in \cDhat_{h,\pi}} \mathbb{I}\{\tilde\pih(x_h)=a_h,\phiho(x_{h-1},a_{h-1})^\top v\geq 0\} \cdot \left(z^\top_{\pa} B_\ell(x_\ell) y -\hat  \vartheta^{\bo}_{h,\ell}[\phih(x_{h},a_{h}),z_\pa,y]\right)\right| \nn \\
   & =  \left|\frac{A}{n}\sum_{(x_{1:H},a_{1:H},r_{1:H})\in \cDhat_{h,\pi}} \mathbb{I}\{\tilde\pih(x_h)=a_h\}\cdot  \mathbb{I}\{\phiho(x_{h-1},a_{h-1})^\top v\geq 0\} \cdot \left(z^\top_{\pa} B_\ell(x_\ell) y - \phih(x_h,a_h)^\top \thetahat_{h,\ell} \right)\right|,  \label{eq:target}
   \intertext{and so by the triangle inequality, we get}
& \leq   \left|\frac{A}{n}\sum_{(x_{1:H},a_{1:H},r_{1:H})\in \cDhat_{h,\pi}} \mathbb{I}\{\tilde\pih(x_h)=a_h\}\cdot  \mathbb{I}\{\phiho(x_{h-1},a_{h-1})^\top v\geq 0\} \cdot \left(\phih(x_h,a_h)^\top \thetahat_{h,\ell} - \phih(x_h,a_h)^\top \theta_{h,\ell} \right)\right|\nn \\
& \quad + \left|\frac{A}{n}\sum_{(x_{1:H},a_{1:H},r_{1:H})\in \cDhat_{h,\pi}} \mathbb{I}\{\tilde\pih(x_h)=a_h\}\cdot  \mathbb{I}\{\phiho(x_{h-1},a_{h-1})^\top v\geq 0\} \cdot \left(z^\top_{\pa} B_\ell(x_\ell) y - \phih(x_h,a_h)^\top \theta_{h,\ell} \right)\right|,\nn \\
& \leq  \left|\frac{A}{n}\sum_{(x_{1:H},a_{1:H},r_{1:H})\in \cDhat_{h,\pi}} \mathbb{I}\{\tilde\pih(x_h)=a_h\}\cdot  \mathbb{I}\{\phiho(x_{h-1},a_{h-1})^\top v\geq 0\} \cdot \left(\phih(x_h,a_h)^\top \thetahat_{h,\ell} - \phih(x_h,a_h)^\top \theta_{h,\ell} \right)\right| \nn \\
& \quad +\frac{16 H A \tilde{d}}{\alpha}  \sqrt{ \frac{2 d\log(12 H |\Psi_{h-1}|\cG_h(\Pi',n) \vepsd^{-1}/\delta')}{n}},  \label{eq:hap}
\end{align}
where the last inequality follows by \eqref{eq:hoef}. We now focus on bounding the first term on the right-hand side of \eqref{eq:hap} using \eqref{eq:paramconcenc}. By Jensen's inequality, we have for all $\tilde\pi\in \Pi'$:
\begin{align}
&  \left|\frac{A}{n}\sum_{(x_{1:H},a_{1:H},r_{1:H})\in \cDhat_{h,\pi}} \mathbb{I}\{\tilde\pih(x_h)=a_h\}\cdot  \mathbb{I}\{\phiho(x_{h-1},a_{h-1})^\top v\geq 0\} \cdot \left(\phih(x_h,a_h)^\top \thetahat_{h,\ell} - \phih(x_h,a_h)^\top \theta_{h,\ell} \right)\right|\nn \\
& \leq A \sqrt{\frac{1}{n}\sum_{(x_{1:H},a_{1:H},r_{1:H})\in \cDhat_{h,\pi}} \mathbb{I}\{\tilde\pih(x_h)=a_h\}\cdot  \mathbb{I}\{\phiho(x_{h-1},a_{h-1})^\top v\geq 0\}^2 \cdot \inner{\phih(x_h,a_h)}{\thetahat_{h,\ell} -  \theta_{h,\ell}}^2 }, \nn \\
& \leq A \sqrt{\frac{1}{n}\sum_{(\pi',v')\in \Psi_{h-1}} \sum_{(x_{1:H},a_{1:H},r_{1:H})\in \cDhat_{h,\pi'}}  \inner{\phih(x_h,a_h)}{\thetahat_{h,\ell} -  \theta_{h,\ell}}^2}, \nn\\
& = \frac{A}{\sqrt{n}} \|\thetahat_{h,\ell} - \theta_{h,\ell}\|_{\Sigma_h}, \nn \\
& \leq A\sqrt{\lambda |\Psi_{h-1}|} \|\theta_{h,\ell}\| + A\sqrt{\frac{2 d \log(2 /\delta')}{n} +  \frac{\log \det (\Sigma_h\lambda^{-1} n^{-1}|\Psi_{h-1}|^{-1})}{n} },\nn \\
& \leq \frac{4A\tilde{d}  H}{\alpha} \sqrt{\lambda |\Psi_{h-1}|}+   A\sqrt{\frac{4 d \log(2 /\delta')}{n} + \frac{\log \det (\Sigma_h\lambda^{-1} n^{-1}|\Psi_{h-1}|^{-1})}{n}  },\label{eq:mile}
\end{align}
where the last inequality follows by the fact that $\theta_{h,\ell} \in \bbB(4 H \tilde{d}/\alpha)$ (where we recall that $\alpha = \sqrt{\zeta} \mu/H$).
Now, by Jensen's inequality and the fact that $\|\Sigma_h\|_{\op} \leq (1+\lambda)\cdot|\Psi_{h-1}|\cdot n$, we have \[\log \det (\Sigma_h\lambda^{-1} n^{-1}|\Psi_{h-1}|^{-1}) \leq d \log \tr(\Sigma_h d^{-1} \lambda^{-1} n^{-1}|\Psi_{h-1}|^{-1})\leq d \log (1 + \lambda^{-1}).\] Plugging this into \eqref{eq:mile}, we get that for all $\tilde\pi\in \Pi'$:
\begin{align}
   &  \left|\frac{A}{n}\sum_{(x_{1:H},a_{1:H},r_{1:H})\in \cDhat_{h,\pi}} \mathbb{I}\{\tilde\pih(x_h)=a_h\}\cdot  \mathbb{I}\{\phiho(x_{h-1},a_{h-1})^\top v\geq 0\} \cdot \left(\phih(x_h,a_h)^\top \thetahat_{h,\ell} - \phih(x_h,a_h)^\top \theta_{h,\ell} \right)\right|\nn \\
   & \leq\frac{4 A\tilde{d}  H}{\alpha} \sqrt{\lambda |\Psi_{h-1}|} + 2A\sqrt{ \frac{d \log(\frac{2}{\lambda \delta'})}{n}},
\end{align}
Combining this with \eqref{eq:hap}, we get that
\begin{align}
   & \left|\frac{A}{n}\sum_{(x_{1:H},a_{1:H},r_{1:H})\in \cDhat_{h,\pi}} \mathbb{I}\{\tilde\pih(x_h)=a_h\}\cdot  \mathbb{I}\{\phiho(x_{h-1},a_{h-1})^\top v\geq 0\} \cdot \left(z^\top_{\pa} B_\ell(x_\ell) y - \hat\vartheta^{\bo}_{h,\ell}[\phih(x_{h},a_{h}),z_\pa,y]\right)\right| \nn \\
    & \leq \frac{4A\tilde{d}  H}{\alpha} \sqrt{\lambda |\Psi_{h-1}|} + 2A\sqrt{ \frac{d \log(\frac{2} {\lambda \delta'})}{n}} + \frac{16 H A \tilde{d}}{\alpha}  \sqrt{ \frac{2 d\log(12 H |\Psi_{h-1}|\cG_h(\Pi',n) \vepsd^{-1}/\delta')}{n}},
\end{align}
for all $\tilde\pi\in \Pi'$. Now, we apply \cref{lem:unionbound} instantiated with 
\begin{itemize}
    \item $\cK \coloneqq \big\{ \Proj_{\cS(G_\ell, \zeta)}(z) : z \in \bbB(1)\big\}$;
    \item $\veps'=\vepsd =\frac{\veps \zeta \mu}{1024 |\Psi_{h-1}|d_\nu d^3 H^6 A}$; 
    \item $\cZ = \Pi'$;
    \item $L = 2 AH/\lambda$;
    \item and 
    \[\bm{M}^{\tilde\pi} = \frac{A}{n}\sum_{(x_{1:H},a_{1:H},r_{1:H})\in \cDhat_{h,\pi}} \mathbb{I}\{\tilde\pih(x_h)=a_h\}\cdot  \mathbb{I}\{\phiho(x_{h-1},a_{h-1})^\top v\geq 0\} \cdot \left(B_\ell(x_\ell) - \hat\vartheta^{\bo}_{h,\ell}[\phih(x_{h},a_{h}),\cdot,\cdot]\right),\]
\end{itemize} to get that there exists an event $\cE(\ell, \pi,v)$ of probability at least $1 - \delta'/(2H |\Psi_{h-1}|)$ under which we have for all $z,y\in \bbB(1)$ and $\tilde\pi\in \Pi'$:
\begin{align}
    & \left|\frac{A}{n}\sum_{(x_{1:H},a_{1:H},r_{1:H})\in \cDhat_{h,\pi}} \mathbb{I}\{\tilde\pih(x_h)=a_h\}\cdot  \mathbb{I}\{\phiho(x_{h-1},a_{h-1})^\top v\geq 0\} \cdot \left(z^\top_{\pa} B_\ell(x_\ell) y - \hat\vartheta^{\bo}_{h,\ell}[\phih(x_{h},a_{h}),z_\pa,y]\right)\right| \nn \\
    & \leq \frac{4A H \vepsd}{\lambda}  + \frac{4A\tilde{d}  H}{\alpha} \sqrt{\lambda |\Psi_{h-1}|} + 2A\sqrt{ \frac{d \log(\frac{2}{\delta'\lambda})}{n}} +\frac{16 H A \tilde{d}}{\alpha}  \sqrt{ \frac{2 d\log(12 H |\Psi_{h-1}|\cG_h(\Pi',n) \vepsd^{-1}/\delta')}{n}}.
\end{align}
Note that the condition of \cref{lem:unionbound} that requires $\|\bm{M}^{\tilde\pi}\|_{\op} \leq L = 2 AH/\lambda$, for all $\tilde \pi \in \Pi'$, is indeed satisfied by definition of $\hat\vartheta^{\bo}_{h,\ell}$ in \cref{alg:fitoptvalue} and the fact that $\sigma_{\min}(\Sigma_h)\geq \lambda |\Psi_{h-1}|n$.
Now, the desired result follows by the union bound over $\ell\in[h+1\ldotst H]$ and $(\pi,v) \in \Psi_{h-1}$ and the facts that $\vepsd = \frac{\veps \zeta \mu}{1024 |\Psi_{h-1}|d_\nu d^3 H^6 A}$ and $\zeta = \frac{1}{8d}$. 
\end{proof}
    
        \begin{proof}[\textbf{Proof of \cref{lem:decoy-find-new}}] 
            In this proof, we condition on the intersection of the events of \cref{lem:littlegen1}, \cref{lem:littlegen3}, and \cref{lem:littlegen2} for $\delta' = \delta/(6HT)$; by the union bound, the probability of this event intersection is at least $1-\frac{\delta}{2 HT}$. To simplify notation in this proof, we let $c \coloneqq 20 d \log(1 + 16 H^4 \nu^{-4})$ and for any $z\in \reals^d$, we write $z_\pa\coloneqq \Proj_{\cS(G_\ell, \zeta)}(z)$ and $z_\perp \coloneqq \Proj_{\cS(G_\ell, \zeta)^\perp}(z)$, where $G_\ell$ is as in \eqref{eq:Gl} and $\zeta \coloneqq  \frac{1}{8d}$.
            
           \fakepar{Case where $\cH\neq \emptyset$} First, assume that $\cH \neq \emptyset$. Fix $\ell\in\cH$ and let 
           \begin{align}
            ((\pi_{h,\ell},v_{h,\ell}), \tilde\pi_{h,\ell}) \in \argmax_{((\pi,v),\tilde\pi)\in \Psi_{h-1}\times  \Pi'} \Delta_{h,\ell}(\pi, v,\tilde\pi), 
           \end{align}
           where $\Delta_{h,\ell}$ is as in \cref{alg:fitoptvalue}.
           By the definition of $\cH$, we have 
            \begin{align}
|\Delta_{h,\ell}(\pi_{h,\ell}, v_{h,\ell}, \tilde\pi_{h,\ell})| \geq 2 d c \veps_{\reg}^{\bo}(\tfrac{\delta}{6HT},\Pi', |\Psi_{h-1}|,n) + \frac{8cd\nu H A}{\mu \lambda}, \label{eq:conditio}
            \end{align}
where $\veps^\bo_\reg$ is as in \cref{lem:littlegen2}. Moving forward, we let \[M \coloneqq \frac{A}{n}\sum_{(x_{1:H},a_{1:H},r_{1:H})\in \cDhat_{h,\pi_{h,\ell}}} \mathbb{I}\{\tilde\pi_{h,\ell}(x_h)=a_h\}\cdot  \mathbb{I}\{\phiho(x_{h-1},a_{h-1})^\top v_{h,\ell}\geq 0\} \cdot \left(B_\ell(x_\ell) - \hat\vartheta^{\bo}_{h,\ell}[\phih(x_{h},a_{h}),\cdot,\cdot]\right),\]
            and note that $M\in \reals^{d\times d}$ satisfies 
            \begin{align}
                \tr(M) = \Delta_{h,\ell}(\pi_{h,\ell}, v_{h,\ell}, \tilde\pi_{h,\ell}). \label{eq:trace}
            \end{align}
            Furthermore, the variables $(z_\ell,\tilde z_\ell,w_\ell)$ in \cref{alg:fitoptvalue} satisfy 
            \[z_\ell \in \argmax_{z\in \bbB(1)} |z^\top M z|,\quad  \tilde z_\ell = \Proj_{\cS(\W_\ell, \nu)}(z_\ell),\quad \text{and}\quad w_\ell = W_\ell^{-1} \tilde z_\ell.\] 
            Now, by definition of the trace, there exist unit vectors (eigenvectors in this case) $c_1,\dots,c_d$ such that $\tr(M) = \sum_{i\in[d]} c_i^\top M c_i$. Combining this with \eqref{eq:trace} and \eqref{eq:conditio}, we get that
            \begin{align}
            d\cdot |z_\ell^\top M  z_\ell| \geq  \left|\sum_{i \in [d]}  c_i^\top M c_i \right| \geq 2 d c \veps_{\reg}^{\bo}(\tfrac{\delta}{6 HT},\Pi', |\Psi_{h-1}|,n) + \frac{8 c d\nu H A}{\mu \lambda}. \label{eq:stor}
            \end{align}
            Now, let $\wbar{M} \coloneqq \sgn(z_\ell^\top M z_\ell) \cdot M$ and note that $z_\ell \in \argmax_{z\in \bbB(1)} z_\ell^\top \wbar{M}z_\ell$. By \eqref{eq:stor}, we have
            \begin{align}
            z_\ell^\top \wbar{M} z_\ell > 2c \veps_{\reg}^{\bo}(\tfrac{\delta}{6 HT},\Pi', |\Psi_{h-1}|,n) + \frac{8c \nu H A}{\mu \lambda}, \label{eq:v-can-be-projd}
            \end{align} and so by \cref{lem:theta-precond-norm-bound} and the definition of $M$, we have 
            \begin{align}
        \tilde z_\ell^\top \wbar{M} \tilde z_\ell > 2 c \veps_{h}^{\bo}(\tfrac{\delta}{6 HT},\Pi', |\Psi_{h-1}|,n) + \frac{8(c-1)\nu H A}{\mu \lambda}. \label{eq:fenc}
            \end{align}
            We use this to show the claims of the lemma. 
           \paragraphi{Proving \cref{item:one}} We start by showing $\norm{w_\ell}\le \nu^{-1}$; that is, \cref{item:one} of the lemma. Let $\lambda_1,\dots ,\lambda_d$ be the eigenvalues of $W_\ell$ and let $w_1,\dots,w_d\in \reals^{d}$ be an orthonormal basis such that $W_\ell w_i = \lambda_i w_i$, for all $i\in[d]$ (such a basis exists because $W_\ell \in \pd$). With this, we can write 
           \begin{align}
             W_\ell = \sum_{i=1}^d \lambda_i w_i w_i^\top.
           \end{align}
           Let $P_\ell$ be the matrix whose columns are $w_1,\dots,w_d$, and note that $P_\ell^\top P_\ell=I$ since $w_1,\dots,w_d$ are orthonormal. With this, we have  
           \begin{align}
             W_\ell^{-1} \Proj_{\cS(W_\ell,\nu)}(z_\ell) &=  W_\ell^{-1} \sum_{i\in[d]: \lambda_i \geq \nu}  w_i w_i^\top z_\ell,\nn \\
             & = \left(\sum_{i=1}^d \lambda_i w_i w_i^\top\right)^{-1}\sum_{i\in[d]: \lambda_i \geq \nu} w_i  w_i^\top z_\ell,\nn \\
             & = \left(\sum_{i=1}^d \lambda_i P_\ell e_i e_i^\top P_\ell^\top\right)^{-1}\sum_{i\in[d]: \lambda_i \geq \nu} P_\ell e_i  e_i^\top P_\ell^\top z_\ell,\nn \\
             & = P_\ell \left(\sum_{i=1}^d \lambda_i  e_i e_i^\top \right)^{-1}P_\ell^\top \sum_{i\in[d]: \lambda_i \geq \nu} P_\ell e_i  e_i^\top P_\ell^\top z_\ell,\nn \\
             & = P_\ell \left(\sum_{i=1}^d \lambda_i  e_i e_i^\top \right)^{-1}\sum_{i\in[d]: \lambda_i \geq \nu}  e_i  e_i^\top P_\ell^\top z_\ell,\nn \\
             & = P_\ell \diag (\lambda^{-1}_1,\dots,\lambda^{-1}_d)\sum_{i\in[d]: \lambda_i \geq \nu}  e_i  e_i^\top P_\ell^\top z_\ell,\nn \\
                          & = P_\ell \sum_{i\in[d]: \lambda_i \geq \nu}  \lambda_i^{-1} e_i  e_i^\top P_\ell^\top z_\ell,\nn \\
           \end{align} 
           Thus, taking the norm and using that $\|P_\ell\|_\op =1$, we get 
           \begin{align}
            \|w_\ell\|=\norm{\W_\ell^{-1}\tilde z_\ell}=\norm{\W_\ell^{-1}\Proj_{\cS(\W_\ell,\nu)} (z_\ell)} \le \|P_\ell\|_\op\sqrt{\sum_{i\in[d]: \lambda_i \geq \nu} \lambda_i^{-2} (e_i^\top P_\ell z_\ell)^2} \leq \nu^{-1} \|P_\ell z_\ell\| \leq \nu^{-1}\|P_\ell\|_\op \|z_\ell\|\leq \nu^{-1}. 
           \end{align}
           This shows \cref{item:one}.
            \paragraphi{Proving \cref{item:two}}
            To prove \cref{item:two}, we need to show that $\norm{\W_{\ell} w_\ell} =\norm{\tilde z_\ell} \ge\frac{1}{2}$. Using that $z_\ell \in \argmax_{z\in \bbB(1)} z^\top \wbar{M} z$ and \cref{lem:theta-precond-norm-bound},
            \begin{align}
            \|\tilde z_\ell\|^{-2} \cdot \tilde z_\ell^\top \wbar{M} \tilde z_\ell \le z_\ell^\top \wbar{M} z_\ell & \le \tilde z_\ell^\top \wbar{M} \tilde z_\ell + \frac{8\nu H A}{\mu \lambda}, \nn \\
            &  \le \tilde z_\ell^\top \wbar{M} \tilde z_\ell + \frac{1}{c} z_\ell^\top \wbar{M} z_\ell, \label{eq:toarrange}
            \end{align}
            where the last inequality follows by \eqref{eq:v-can-be-projd}. Rearranging \eqref{eq:toarrange} and using that $c \geq 3$, implies that $\norm{\tilde z_\ell}^2 \ge \frac23$. Therefore, we get that
            \[
            \norm{\W_\ell w_\ell}^2 = \|\tilde z_\ell\|^2 \ge\frac23,
            \]
            satisfying the inequality in \cref{item:two}.
            
            \paragraphi{Proving \cref{item:three}} It remains to prove \cref{item:three}. We will start by showing that $\norm{(\tilde{z}_\ell)_\pa}^2 \le \frac{1}{c}$ and use this to prove \cref{item:three}. Since $\wbar{M}$ is symmetric, we can decompose $\tilde{z}_\ell^\top \wbar{M} \tilde{z}_\ell$ as
            \[
            \tilde{z}_\ell^\top \wbar{M} \tilde{z}_\ell= (\tilde{z}_\ell)_\pa^\top \wbar{M} \tilde{z}_\ell + (\tilde{z}_\ell)_\pa^\top \wbar{M} (\tilde{z}_\ell)_\perp + (\tilde{z}_\ell)_\perp^\top \wbar{M} (\tilde{z}_\ell)_\perp\,.
            \]
            By the conditioning on the event of \cref{lem:littlegen2} with $\delta' = \delta/(6HT)$, we have $(\tilde{z}_\ell)_\pa^\top \wbar{M} \tilde{z}_\ell \leq \veps_{\reg}^{\bo}(\tfrac{\delta}{6 HT},\Pi', |\Psi_{h-1}|,n)$ and $ (\tilde{z}_\ell)_\pa^\top \wbar{M} (\tilde{z}_\ell)_\perp \leq \veps_{\reg}^{\bo}(\tfrac{\delta}{6 HT},\Pi', |\Psi_{h-1}|,n)$. Therefore, we have
            \begin{align}
            \tilde{z}_\ell^\top \wbar{M} \tilde{z}_\ell \le 2\veps_{\reg}^{\bo}(\tfrac{\delta}{6 HT},\Pi', |\Psi_{h-1}|,n)  + (\tilde{z}_\ell)_\perp^\top \wbar{M} (\tilde{z}_\ell)_\perp. \label{eq:light}
            \end{align}
            Combining this with \eqref{eq:fenc}, we get that $\|(\tilde{z}_\ell)_\perp\|\ne 0$.
            Let $\bar{z}_\ell=(\tilde{z}_\ell)_\perp/\norm{(\tilde{z}_\ell)_\perp}$. Since $z_\ell \in \argmax_{z\in \bbB(1)} z^\top \wbar{M} z$, we have
            \begin{align}
            \norm{(\tilde{z}_\ell)_\perp}^{-2}\cdot (\tilde{z}_\ell)_\perp^\top \wbar{M} (\tilde{z}_\ell)_\perp& =(\bar{z}_\ell)^\top \wbar{M} {\bar{z}_\ell} \le z_\ell^\top \wbar{M} z_\ell. \label{eq:mocbd}
            \end{align}
                On the other hand, using \cref{lem:theta-precond-norm-bound} and \eqref{eq:light}, we have  
            \begin{align} 
          z_\ell^\top \wbar{M} z_\ell  & \le \tilde{z}_\ell^\top \wbar{M} \tilde{z}_\ell + \frac{8\nu H A}{\mu \lambda},\nn \\
            &  \le 2 \veps_{\reg}^{\bo}(\tfrac{\delta}{6 HT},\Pi', |\Psi_{h-1}|,n) + \frac{8\nu H A}{\mu \lambda} + (\tilde{z}_\ell)_\perp^\top \wbar{M} (\tilde{z}_\ell)_\perp,\nn \\ 
            & \leq \frac{1}{c} z_\ell^\top \wbar{M} z_\ell  + (\tilde{z}_\ell)_\perp^\top \wbar{M} (\tilde{z}_\ell)_\perp, \label{eq:imme}
        \end{align}
        where the last inequality follows by \eqref{eq:v-can-be-projd}. Rearranging \eqref{eq:imme} gives
        \begin{align}
            z_\ell^\top \wbar{M} z_\ell  & \leq \frac{c}{c-1}  (\tilde{z}_\ell)_\perp^\top \wbar{M} (\tilde{z}_\ell)_\perp.
        \end{align}
    Combining this with \eqref{eq:mocbd}, dividing by $(\tilde{z}_\ell)_\perp^\top \wbar{M} (\tilde{z}_\ell)_\perp$, and rearranging, we get that $\norm{(\tilde{z}_\ell)_\perp}^2 \ge \frac{c-1}{c}$, and so
            \begin{align}
            \norm{(\tilde{z}_\ell)_\pa}^2 &\le \frac{1}{c}, \label{eq:pitstop}
            \end{align}
            since $1 = \|z_\ell\|^2\geq \|\tilde{z}_\ell\|^2 = \|(\tilde{z}_\ell)_\pa\|^2 + \|(\tilde{z}_\ell)_\perp\|^2$. We use \eqref{eq:pitstop} to prove \cref{item:three}. Note that since \begin{align}G_\ell = W_\ell^{-1}\left( \sum_{\pi\in  \supp \rho_\ell}\rho_\ell(\pi)\theta^{\pi}_\ell (\theta^\pi_\ell)^\top\right)W_\ell^{-1},
            \end{align}
            where $\rho_\ell$ is the approximate design for $\Theta_\ell$ in \cref{para:special_paragraph}, we have 
            \begin{align}
             \forall \theta \in W_\ell^{-1}\Theta_\ell, \quad  \|\theta\|_{G_\ell^\dagger} \leq \sqrt{2 d}.\label{eq:reas}   
            \end{align}
    With this, we have 
            \begin{align}
                \sup_{\theta\in\Theta_\ell}\abs{\inner{\theta}{w_\ell}}& = \sup_{\theta\in\Theta_\ell}\abs{\inner{\theta}{\W_\ell^{-1}\tilde{z}_\ell}},\nn \\
            &=\sup_{\theta\in W_\ell^{-1}\Theta_\ell}\abs{\inner{\theta}{\tilde{z}_\ell}},\nn \\
            &\le \sup_{\theta\in W_\ell^{-1}\Theta_\ell}\norm{\theta}\norm{(\tilde{z}_\ell)_\pa} + \sup_{\theta\in W_\ell^{-1}\Theta_\ell}\abs{\inner{\theta}{(\tilde{z}_\ell)_\perp}}, 
            \intertext{and so by \cref{lem:fanc} and $d_\nu \coloneqq 5 d \log(1 + 16 H^4 \nu^{-4})$,}  
            &\le  \sqrt{\frac{d_\nu}{c}} + \sup_{\theta\in W_\ell^{-1}\Theta_\ell}\norm{\theta}_{G_\ell^\dag} \cdot\norm{(\tilde{z}_\ell)_\perp}_{ G_\ell}, \nn \\
            \intertext{and so by \eqref{eq:reas}, we have}
            &\le  \sqrt{\frac{d_\nu}{c}} + \sqrt{2d}\cdot  \norm{(\tilde{z}_\ell)_\perp}_{G_\ell},  \\
            &\le  \sqrt{\frac{d_\nu}{c}} + \sqrt{2d \cdot (\tilde{z}_\ell)_\perp^\top (\zeta I) (\tilde{z}_\ell)_\perp},\ \ \text{(see below)} \label{eq:mobile} \\
            &\le  \sqrt{\frac{d_\nu}{c}} + \sqrt{2d\cdot \zeta},\nn \\
            &  =  \sqrt{\frac{d_\nu}{c}} + \frac12,\nn \quad \text{(since $\zeta= (8d)^{-1}$)} \\
           &  \leq 1, 
            \end{align}
            where in \eqref{eq:mobile} follows from the fact that $(\cdot)_\perp = \Proj_{\cS(G_\ell, \zeta)^\perp}(\cdot)$; and the last inequality uses that $c= 20 d \log (1 + 16 \nu^{-4}H^4)= 4 d_\nu$.
\fakepar{Case where $\cH = \emptyset$} We now consider the case where $\cH = \emptyset$. By definition of $\cH$ in \cref{alg:fitoptvalue}, we have that for all $\ell\in[h+1\ldotst H]$, $(\pi, v)\in \Psi_{h-1}$, and $\tilde \pi \in  \Pi'$:
\begin{align}
   | \Delta_{h,\ell}(\pi, v,\tilde\pi)| \leq 2c d\veps_{\reg}^{\bo}(\tfrac{\delta}{6 HT},\Pi', |\Psi_{h-1}|,n) + \frac{8cd \nu H A}{\mu \lambda}. \label{eq:waho}
\end{align}
On the other hand, by the conditioning on the event of \cref{lem:littlegen1} with $\delta' = \delta/(6HT)$, we have for all $\ell\in[h+1\ldotst H]$, $(\pi, v)\in \Psi_{h-1}$, and $\tilde \pi \in  \Pi'$:
\begin{align}
    & \left|\Delta_{h,\ell}(\pi, v, \tilde\pi)- \E^{\pi \circ_h \tilde\pi \circ_{h+1}\pihat_{h+1:H}}\left[\mathbb{I}\{\phiho(\x_{h-1},\a_{h-1})^\top v\geq 0\} \cdot (b_\ell(\x_\ell) - \phih(\x_h,\a_h)^\top \thetahat^\bo_{h,\ell}) \right]  \right|\nn \\
    & \leq \frac{4 A H}{\lambda}\sqrt{\frac{\log (2H^2 |\Psi_{h-1}| \cG_h(\Pi',n)/\delta)}{n}}.
\end{align}
Thus, by the triangle inequality and \eqref{eq:waho}, we have that for all $\ell\in[h+1\ldotst H]$, $(\pi, v)\in \Psi_{h-1}$, and $\tilde \pi \in  \Pi'$:
\begin{align}
   & \left| \E^{\pi \circ_h \tilde\pi \circ_{h+1}\pihat_{h+1:H}}\left[\mathbb{I}\{\phiho(\x_{h-1},\a_{h-1})^\top v\geq 0\} \cdot (b_\ell(\x_\ell) - \phih(\x_h,\a_h)^\top \thetahat^\bo_{h,\ell}) \right]\right| \nn \\
  & \leq \frac{4 A H}{\lambda}\sqrt{\frac{\log (2H^2 |\Psi_{h-1}| \cG_h(\Pi',n)/\delta)}{n}} + 2c d\veps_{\reg}^{\bo}(\tfrac{\delta}{6 HT},\Pi', |\Psi_{h-1}|,n) + \frac{8c d\nu H A}{\mu \lambda}. 
\end{align}
Thus, summing over $\ell\in[h+1\ldotst H]$ and using the triangle inequality, we get that for all $(\pi,v)\in \Psi_{h-1}$ and $\tilde \pi \in  \Pi'$:
\begin{align}
    & \left| \E^{\pi \circ_h \tilde\pi \circ_{h+1}\pihat_{h+1:H}}\left[\mathbb{I}\{\phiho(\x_{h-1},\a_{h-1})^\top v\geq 0\} \cdot \left(\sum_{\ell=h+1}^H b_\ell(\x_\ell) - \sum_{\ell=h+1}^H \phih(\x_h,\a_h)^\top \thetahat^\bo_{h,\ell} \right) \right]\right| \nn \\
    & \leq  \frac{4 A H^2}{\lambda}\sqrt{\frac{\log (2H^2 |\Psi_{h-1}| \cG_h(\Pi',n)/\delta)}{n}} + 2c d H\veps_{\reg}^{\bo}(\tfrac{\delta}{6 HT},\Pi', |\Psi_{h-1}|,n) + \frac{8cd \nu H^2 A}{\mu \lambda}.
\end{align}
Combining this with the conditioning on the event of \cref{lem:littlegen3} with $\delta' = \delta/(6HT)$ and the triangle inequality, we get that for all $(\pi,v)\in \Psi_{h-1}$ and $\tilde \pi \in  \Pi'$:
\begin{align}
        & \left| \E^{\pi \circ_h \tilde\pi \circ_{h+1}\pihat_{h+1:H}}\left[\mathbb{I}\{\phiho(\x_{h-1},\a_{h-1})^\top v\geq 0\} \cdot \left(\sum_{\ell=h}^H \br_h+ \sum_{\ell=h+1}^H b_\ell(\x_\ell) - \phih(\x_h,\a_h)^\top \left(\thetahat^{\re}_h +\sum_{\ell=h+1}^H  \thetahat^\bo_{h,\ell} \right) \right)\right]\right| \nn \\
        & \leq  \frac{4 A H^2}{\lambda}\sqrt{\frac{\log (2H^2 |\Psi_{h-1}| \cG_h(\Pi',n)/\delta)}{n}} + 2c dH\veps_{\reg}^{\bo}(\tfrac{\delta}{6 HT},\Pi', |\Psi_{h-1}|,n) + \frac{8c d\nu H^2 A}{\mu \lambda} + \veps_{\reg}^{\re}(\tfrac{\delta}{6 HT},\Pi', |\Psi_{h-1}|,n),
\end{align}
where $\veps_{\reg}^{\re}$ is as in \cref{lem:littlegen3}. Now, \eqref{eq:fitting} follows from the definition of $\wtilde{Q}_h$ in the lemma's statement and the fact that $\thetahat_h = \thetahat^{\re}_h +\sum_{\ell=h+1}^H  \thetahat^\bo_{h,\ell}$ and $c= 4d_\nu$. 
    \end{proof}

    \subsection{Additional Structural Results for the Proofs of $\texttt{FitOptValue}$}
    \label{sec:addition-struct}
    In this subsection, we present additional structural results we require for proving the guarantees of \texttt{FitOptValue}. These results are closely related to the structural results in \citep{weisz2024online}.

\begin{lemma}
    \label{lem:dual-admissibility}
    Let $\ell\in[H]$, $L>0$, $\mu, \zeta>0$, $W_\ell \in \pd$, and $f: \cX_\ell\rightarrow [-L,L]$ be given, and let $F: \cX_\ell  \rightarrow [-L,L]^{d\times d}$ be the function defined as \begin{gather}F(x) = \mathbb{I}\{\|\varphiell(x; W_\ell)\|\geq \mu\}\cdot \frac{\varphiell(x; W_\ell)\varphiell(x; W_\ell)^\top}{\|\varphiell(x; W_\ell)\|^2} \cdot f(x), 
       \intertext{where} 
        \varphiell(x; W_\ell)   = \W_\ell \phiell(x,a_x) - \W_\ell \phiell(x,a'_x), \quad  \text{and}
 \quad         (a_x,a_x')   \in \argmax_{(a,a')\in \cA^2} \|\W_\ell \phiell(x,a) - \W_\ell\phiell(x,a')\|. \label{eq:varphi}
    \end{gather}
    Further, let $\rho_\ell$ denote the approximate optimal design for $\Theta_\ell$ in \cref{par:bench}, and define \[G_\ell \coloneqq W_\ell^{-1}\left( \sum_{\pi\in  \supp \rho_\ell}\rho_\ell(\pi)\theta^{\pi}_\ell (\theta^\pi_\ell)^\top\right)W_\ell^{-1}\in \reals^{d\times d},\quad \text{and} \quad z_\pa\coloneqq \Proj_{\cS(G_\ell, \zeta)}(z), \quad \forall z\in \reals^d.  \]
   Then, for all $z,y\in\mathbb{B}(1)$, the function $z_\pa^\top F(\cdot) y$ is $\alpha$-admissible with $\alpha \coloneqq  \frac{\sqrt{\zeta} \mu}{L}$; that is, $z_\pa^\top F(x) y\leq \rgell(x)/\alpha$ for all $x\in \cX_\ell$. 
\end{lemma}
\begin{proof}[\textbf{Proof of \cref{lem:dual-admissibility}}]\label{proof:dual-admissibility}
Aspects of this proof are inspired by the proof of \cite[Lemma 4.9]{weisz2024online}.

Fix $x\in\cX_\ell$ and let $\vphibell(x; W_\ell) \coloneqq \varphiell(x; W_\ell)/\|\varphiell(x; W_\ell)\|$. Using that $|f(x)|\le L$, $z,y\in\mathbb{B}(1)$, and $\|\vphibell(x; W_\ell)\|\le 1$, we obtain
    \begin{align}
    |z_\pa^\top F(x) y|
    &\le |{\vphibell(x; W_\ell)^\top z_\pa}\cdot {\vphibell(x; W_\ell)^\top y}|\cdot \mathbb{I}\{\|\varphiell(x; W_\ell)\|\geq \mu\} \cdot  L,  \\
    & \leq  |{\varphiell(x; W_\ell)^\top z_\pa}|\cdot \mu^{-1} \cdot  L, \\
    &\le \|\varphiell(x; W_\ell)_\pa\| \cdot \mu^{-1} \cdot L. \label{eq:previous}
    \end{align}
On the other hand, we can write
\begin{align}
\rgell(x)^2 & = \max_{\pi \in  \supp(\rho_\ell)}\sup_{a,a'\in \cA} \inner{\phiell(x,a)- \phiell(x,a')}{\theta^{\pi}_\ell}^2  ,\nn\\ 
& = \max_{\pi \in  \supp(\rho_\ell)}\sup_{a,a'\in \cA} \inner{W_\ell \phiell(x,a)- W_\ell \phiell(x,a')}{W_\ell^{-1}\theta^{\pi}_\ell}^2,\nn \\
&= \max_{\pi \in  \supp(\rho_\ell)}\inner{\varphiell(x; W_\ell)}{W_\ell^{-1}\theta^{\pi}_\ell}^2,\nn \\
& = \max_{\pi \in  \supp(\rho_\ell)}\varphiell(x; W_\ell)^\top W_\ell^{-1}\theta^{\pi}_\ell (\theta^{\pi}_\ell)^\top  W_\ell^{-1}  \varphiell(x; W_\ell) ,\nn \\
& \ge \varphiell(x; W_\ell)^\top G_\ell  \varphiell(x; W_\ell) ,\nn \\
& \ge \varphiell(x; W_\ell)_{\pa}^\top G_\ell \varphiell(x; W_\ell)_{\pa}, \nn \\
& \ge \|\varphiell(x; W_\ell)_{\pa}\|^2 \zeta,\label{eq:gd}
\end{align}
where the last inequality follows from the fact that the eigenvalues of $G_\ell$ corresponding to the subspace in which $\varphiell(x; W_\ell)_\pa$ lies are by definition at least $\zeta$. Combining \eqref{eq:gd} with \eqref{eq:previous}, we get that 
    \begin{align}
    \rgell(x)\ge \sqrt\zeta \cdot \|\varphiell(x; W_\ell)_{\pa}\| \ge \frac{\sqrt\zeta \mu}{L}\cdot  |z_\pa^\top F(x) y|,
    \end{align}
    finishing the proof. 
    \end{proof}

\begin{lemma}
    \label{lem:theM}
Let $\ell\in[H]$, $\mu, \nu>0$, $W_\ell \in \pd$ be given. Further, let $\varphiell(\cdot; W_\ell)$ and $\vphibell(\cdot;W_\ell)$ be as in \cref{lem:dual-admissibility} and define
\begin{align}
    M_\ell(x)  \coloneqq \mathbb{I}\{\varphiell(x; W_\ell) \geq\mu \} \cdot \vphibell(x;W_\ell) \vphibell(x; W_\ell)^\top, \quad \text{where} \quad \vphibell(x; W_\ell)\coloneqq  \frac{\varphiell(x;W_\ell)}{\|\varphiell(x;W_\ell)\|}. \label{eq:theM}
\end{align}
Then, for any $z\in \bbB(1)$, $\tilde z \coloneqq \Proj_{\cS(W_\ell,\nu)}(z)$, and $x \in \cX_\ell$:
\begin{align}
|\tilde z ^\top M_\ell(x) \tilde z - z^\top M_\ell(x) z|\leq  \frac{4\nu}{\mu}.
\end{align}
\end{lemma}
\begin{proof}
  Aspects of this proof are inspired by the proof of \cite[Lemma E.1]{weisz2024online}.

Fix $z\in \bbB(1)$ and let $y = \Proj_{\cS(W_\ell,\nu)^\perp}(z)$ and note that $z = \tilde z + y$. By symmetry of $M_\ell(x)$, we have 
\begin{align}
z^\top M_\ell(x) z & = (\tilde z + y)^\top M_\ell(x) (\tilde z + y),\nn \\
& = \tilde z^\top M_\ell(x) (\tilde z + y) + y^\top M_\ell(x) (\tilde z + y),\nn \\
& =  \tilde z^\top M_\ell(x) \tilde z  + y^\top M_\ell(x) (2\tilde z + y), \nn \\
& =  \tilde z^\top M_\ell(x) \tilde z  + y^\top M_\ell(x) (\tilde z + z),
\end{align}
where the last step follows by the fact that $z= \tilde z + y$.
Thus, it suffices to bound $y^\top M_\ell(x) (\tilde z+z)$. First, note that $\| z\|\leq 1$ and so $\|\tilde z+ z\|\leq 2$. Therefore, we have 
\begin{align}
|y^\top M_\ell(x) (\tilde z + z)| 
& = \mathbb{I}\{\varphiell(x;W_\ell)\geq \mu\}  \cdot |y^\top \vphibell(x; W_\ell)| \cdot |(\tilde z + z)^\top \vphibell(x; W_\ell)|,\nn \\
& \leq 2\mathbb{I}\{\varphiell(x;W_\ell)\geq \mu\} \cdot |y^\top \vphibell(x; W_\ell)|,\nn \\
& =  2\mathbb{I}\{\varphiell(x;W_\ell)\geq \mu\} \cdot \left|y^\top \frac{\varphiell(x; W_\ell)}{\|\varphiell(x; W_\ell)\|}\right|,\nn \\
& \leq 2\mu^{-1} \cdot |y^\top \varphiell(x; W_\ell)|,\nn \\
& \leq 2\mu^{-1} \cdot \|\Proj_{\cS(W_\ell,\nu)^\perp}(\varphiell(x;W_\ell))\|. \label{eq:thisone}
\end{align}
Let $\lambda_1,\dots ,\lambda_d$ be the eigenvalues of $W_\ell$ and and let $w_1,\dots,w_d\in \reals^{d}$ be an orthonormal basis such that $W_\ell w_i = \lambda_i w_i$, for all $i\in[d]$ (such a basis exists because $W_\ell \in \pd$). With this, we can write 
\begin{align}
  W_\ell = \sum_{i=1}^d \lambda_i w_i w_i^\top.
\end{align}
Therefore, since $\varphi(\tilde x; W_\ell)= W_\ell \varphi(x',a_{\tilde x},a'_{\tilde x})$ with $(a_{\tilde x},a_{\tilde x}')$ as in \eqref{eq:varphi}. we have  
\begin{align}
  \Proj_{\cS(W_\ell,\nu)^\perp}(\varphiell(x;W_\ell)) &=  \sum_{i=1}^d \lambda_i \cdot w_i^\top \varphi(x,a_x,a_x') \cdot \Proj_{\cS(W_\ell,\nu)^\perp}(w_i),\nn \\
  \intertext{and by definition of $\Proj_{\cS(W_\ell,\nu)^\perp}$:}
  & =   \sum_{i\in[d]: \lambda_i \leq \nu} \lambda_i \cdot w_i^\top \varphi(x,a_x,a_x'),\nn \\
  & \leq \nu \|\varphi(x,a_x,a_{x'})\|,\nn \\
  & \leq 2 \nu.
\end{align}  
Plugging this bound into \eqref{eq:thisone} completes the proof.
\end{proof}

\begin{lemma}
    \label{lem:theta-precond-norm-bound}
   For $h\in[H]$ and $\ell\in[h+1\ldotst H]$, let $B_\ell$ and $\hat\vartheta^{\bo}_{h,\ell}$ be as in \cref{alg:fitbonus}. Then, for any $z\in\mathbb{B}(1)$ and $\tilde z \coloneqq \Proj_{\cS(W_\ell,\nu)}(z)$, we have:
    \begin{align}
  \forall x\in \cX_\ell, \quad    |\tilde z^\top B_\ell(x) \tilde z- z^\top  B_\ell(x) z |  & \leq  \frac{4\nu H}{\mu},\label{eq:mag} \\
  \forall (x,a)\in \cX_h\times \cA, \quad   |\hat\vartheta^{\bo}_{h,\ell}[\phih(x,a), \tilde z, \tilde z] - \hat\vartheta^{\bo}_{h,\ell}[\phih(x,a),z, z] |& \leq  \frac{4\nu H}{\mu \lambda}. \label{eq:pic}
    \end{align}
\end{lemma}
\begin{proof}
Fix $x'\in \cX_\ell$ and $z\in \bbB(1)$. Note that $B_\ell$ is of the form 
\begin{align}
   B_\ell(x') \coloneqq \bar{b}_\ell(x') \cdot \mathbb{I}\{\|\varphiell(x';W_\ell)\| \geq \mu \} \cdot \frac{\varphiell(x';W_\ell) \varphiell(x';W_\ell)^\top}{\|\varphiell(x';W_\ell)\|^2}.
\end{align}
where $\bar{b}_\ell(x') \coloneqq \min\left(H,\frac{\veps}{4 H} \max_{a'\in\cA}\|\phiell(x',a')\|_{(\lambda I +U_\ell)^{-1}}\right)$. Using the notation of \cref{lem:theM}; specifically using the matrix $M_\ell(x')$ in \eqref{eq:theM}, we can write $B_\ell$ as 
\begin{align}
    B_\ell(x') =  \bar b_\ell(x') \cdot  M_\ell(x').
\end{align}
Thus, applying \cref{lem:theM} with the fact that $\bar{b}_\ell(x')\leq H$ implies that 
\begin{align}
    |\tilde z^\top B_\ell(x') \tilde z - z^\top B_\ell(x') z| \leq  \frac{4\nu H}{\mu}, \label{eq:theBbound}
\end{align}
where $\tilde z\coloneqq \proj_{\cS(W_\ell,\nu)}(z)$.
This shows \eqref{eq:mag}. We now show \eqref{eq:pic}. Fix $(x,a)\in \cX_h\times \cA$. Using the definition of $\hat\vartheta^{\bo}_{h,\ell}$ in \cref{line:varthetahl} of \cref{alg:fitoptvalue}, we can write 
\begin{align}
   \hat\vartheta^{\bo}_{h,\ell}[\phih(x,a),\cdot,\cdot] = \phih(x,a)^\top \Sigma_{h}^{-1}\sum_{(\pi,v)\in  \Psi_{h-1}}\sum_{(x_{1:H},a_{1:H},r_{1:H})\in \cDhat_{h,\pi}} \phih(x_{h},a_{h}) \cdot  B_\ell(x_\ell) \in \reals^{d\times d},  
\end{align}
where $\lambda n_\traj |\Psi_{h-1}|  I + \sum_{(\pi,v)\in  \Psi_{h-1}}\sum_{(x_{1:H},a_{1:H},r_{1:H})\in \cDhat_{h,\pi}} \phih(x_{h},a_{h}) \phih(x_{h},a_{h})^\top$. Thus, using that 
\[
\left\|  \Sigma_{h}^{-1}\sum_{(\pi,v)\in  \Psi_{h-1}}\sum_{(x_{1:H},a_{1:H},r_{1:H})\in \cDhat_{h,\pi}} \phih(x_{h},a_{h})\right\| \leq \frac 1 \lambda,
\]
together with \eqref{eq:theBbound} and the fact that $\|\phi(x,a)\|\leq 1$, we get that
\begin{align}
    |\hat\vartheta^{\bo}_{h,\ell}[\phi(x,a),\tilde z,\tilde z] -  \hat\vartheta^{\bo}_{h,\ell}[\phi(x,a),z,z]| \leq  \frac{4\nu H}{\mu \lambda}.
\end{align}
This completes the proof.
\end{proof}

\section{Guarantee of $\texttt{Evaluate}$}
\label{app:eval}
\begin{lemma}[Guarantee of $\texttt{Evaluate}$]
    \label{lem:eval}
    Let $\pihat_{h+1:H}$ and $n$ be given and consider a call to 
    $\texttt{Evaluate}_{h}(\pihat_{1:H}, n)$ (\cref{alg:evaluate}). Then, for any $\delta' \in(0,1)$, with probability at least $1-\delta'$, the output $J$ of \cref{alg:evaluate} is such that 
    \begin{align}
\left|J - \E^{\pihat}\left[\sum_{h\in[H]}\br_h\right]\right| \leq H \sqrt{\frac{2\log (1/\delta')}{n}}.
    \end{align}
\end{lemma}
\begin{proof}
The result follows from Hoeffding's inequality and the fact that the rewards take values in $[0,1]$.
\end{proof} 
\section{Analysis: Proof of \cref{lem:say}}
\label{sec:fullproof}

Central to the proof of \cref{lem:say} is a change of measure argument (as described in \cref{sec:highlevel}), which we now state.
\begin{lemma}[Change of measure]
    \label{lem:changeofmeasure2}
    Let $\veps,\delta\in(0,1)$ be given and consider a call to $\mainalg(\Pi_\bench,\veps,\delta)$ with $\Pi_\bench$ as in \cref{par:bench}. Further, let $(\veps_\conc, \cE^\conc)$ be as in \cref{lem:designbound}. Then, under the event $\cE^\conc$, we have for all $\theta \in \reals^d$, $t\in[T]$, $h \in[H]$ and $(x,a)\in \cX_h\times \cA$:
   \begin{align}
    & |\theta^\top\phih(x, a)|\nn\\
     &\leq \sqrt{d}\|\phih(x_h, a)\|_{(\beta I + U\ind{t}_h)^{-1}} \cdot \left(\sqrt{\frac{\beta}{d}}\|\theta\| + t\veps_\conc +  \sum_{(\pi, v) \in \Psi\ind{t}_h} \left| \E^{\pi}[\theta^\top \phih(\x_h,\a_h)\cdot \mathbb{I}\{\phih(\x_h,\a_h)^\top v \geq 0\}]\right|  \right).
   \end{align} 
\end{lemma}
\begin{proof}
    In this proof, we condition on the event $\cE^{\conc}$. First, we note that $U\ind{t}_h = \sum_{k=1}^{t-1} u_{h}\ind{k} (u_h\ind{k})^\top$, where $(u_h\ind{k})$ are as in \cref{alg:ops-dp}. Thus, by letting $A_h = \beta I + U\ind{t}_h$, we have that for all $\theta \in \reals^d$, $t\in[T]$, $h \in[H]$, and $(x,a)\in \cX_h\times \cA$:
\begin{align}
    |\theta^\top\phih(x, a)| & = |\theta^\top A_h A^{-1}_h \phih(x,a)|, \nn \\
    & \leq \left|\left(\beta \theta^\top+ \sum_{k=1}^{t-1} \theta^\top  u\ind{k}_h (u\ind{k}_h)^\top\right) A_h^{-1/2} A_h^{-1/2} \phih(x,a)\right|,\nn \\
    & \leq \beta \|\theta\|\cdot \|A_h^{-1/2}\|_\op \cdot \|\phih(x,a)\|_{A_h^{-1}} + \sum_{k=1}^{t-1} |\theta^\top u\ind{k}_h|\cdot \|u\ind{k}_h \|_{A_h^{-1}} \cdot   \|\phih(x,a)\|_{A_h^{-1}}, \nn \\
    &\leq  \sqrt{\beta}\cdot \|\theta\| \cdot \|\phih(x,a)\|_{A_h^{-1}} + \sqrt{d} \sum_{k=1}^{t-1} |\theta^\top u\ind{k}_h|\cdot \|\phih(x,a)\|_{A_h^{-1}}, \label{eq:plug}
\end{align}
where the last inequality follows by the facts that $\|A_h^{-1/2}\|_\op \leq \beta^{-1/2}$ and $\|u\ind{k}_h \|^2_{A_h^{-1}} \leq d$. Now, by \cref{lem:designbound}, the conditioning on $\cE^\conc$, and the triangle inequality, we have that for all $k\in[t-1]$:
\begin{align}
    |\theta^\top u\ind{k}_h| \leq \left| \E^{\tilde\pi_{1:h}\ind{k}}[\theta^\top \phih(\x_h,\a_h)\cdot \mathbb{I}\{\phih(\x_h,\a_h)^\top v\ind{k}_h \geq 0\}]\right| + \veps_\conc.
\end{align} 
Plugging this into \eqref{eq:plug} and using that $\Psi\ind{t}_h = \bigcup_{k=1}^{t-1} \{(\tilde\pi_{1:h}\ind{k}, v\ind{k}_h)\}$ (see the update rule in \cref{line:updatePsi} of \cref{alg:ops-dp}), we get the desired result.
\end{proof}
 
The following key lemma consolidates the guarantees of \texttt{DesignDir} and \texttt{FitOptValue}, presented in \cref{app:designdir} and \cref{sec:fitbonusproof}, respectively, into a single statement. It also employs an elliptical potential argument to bound the expected bonuses under rollouts (see \cref{sec:highlevel} for a high-level explanation of this argument)

\begin{lemma}
    \label{lem:test}
    Let $\veps,\delta\in(0,1)$ be given and consider a call to $\mainalg(\Pi_\bench,\veps, \delta)$ (\cref{alg:ops-dp}) with $\Pi_\bench$ as in \cref{par:bench}. Let $(\veps_\conc, \cE^{\conc})$ and $(\veps_\reg,\cE^\reg)$ be as in \cref{lem:designbound} and \cref{lem:fitoptvalue}, respectively. Then, under the event $\cE^\conc\cap \cE^\reg$, there exists $\tau \in [T]$ such that for all $h\in[H]$ and $\ell\in[0 \ldotst h-1]$:
    \begin{gather}
        \E^{\pi \circ_{\ell+1} \pihat\ind{\tau}_{\ell+1:H}} \left[\max_{a\in \cA}\|\phih(\x_h,a)\|_{(\beta I + U\ind{\tau}_{h})^{-1}} \right] \leq 1/2+ 2 d\beta^{-1} \veps_\conc , 
        \intertext{and for all $\tilde \pi \in \Pi_\bench$ and $(\pi, v)\in \Psi\ind{\tau}_{h-1}$:}
        \E^{\pi\circ_h \tilde\pi\circ_{h+1}\pihat\ind{\tau}_{h+1:H}} \left[ \mathbb{I}\{\phiho(\x_{h-1},\a_{h-1})^\top v\geq 0\} \cdot (Q_h\ind{\tau}(\x_h,\a_h)- \Qhat\ind{\tau}_h(\x_h,\a_h)) \right] \leq \veps_\reg,  
    \end{gather}
    where for all $(x,a)\in \cX_h\times \cA$, $Q\ind{\tau}_h(x,a)=Q^{\pihat\ind{\tau}}_h(x,a)+ \E^{\pihat\ind{\tau}}[\sum_{\ell=h}^H b\ind{\tau}_\ell(\x_h)\mid \x_h = x,\a_h=a]$, $\Qhat\ind{\tau}_h(x,a) \coloneqq \phih(x,a)^\top \thetahat_h\ind{\tau} + b\ind{\tau}_h(x)$, and ${b}\ind{\tau}_h(x) \coloneqq \min\left(H,\frac{\veps}{4 H } \cdot\max_{a'\in \cA} \|\phiell(x, a')\|_{(\beta I + U\ind{\tau}_h)^{-1}}\right) \cdot \mathbb{I}\{\|\varphi(x;W\ind{\tau}_h)\|\geq \mu\}$.
    \end{lemma}
\begin{proof}
    Throughout the proof, we condition on the event $\cE^\conc\cap \cE^\reg$. For $h\in[H]$, define 
   \begin{align}
    \cT_h  \coloneqq  \{ (t,\ell) \in [T]\times [h-1]\mid  w\ind{t,\ell}_h \neq 0\},
   \end{align}
   where $(w\ind{t,\ell}_h)$ are as in \cref{alg:ops-dp}. By \cref{line:updateprecond} of \cref{alg:ops-dp}, we have that 
   \begin{align}
    W_h\ind{T+1}  = \left(H^{-2} I +\sum_{(t,\ell)\in \cT_h} w\ind{t,\ell}_h (w\ind{t,\ell}_h)^\top \right)^{-1/2}.
   \end{align}
   By \cref{lem:fitoptvalue} (and the conditioning on $\cE^\reg$), $W_h\ind{T+1}$ is a valid $\nu$-preconditioning, and so by \cref{lem:precond}, we must have that \begin{align}|\cT_h| \leq 4 d \log(1 + 16 \nu^{-4}H^{4}). \label{eq:dist}
   \end{align} 
   Now, let 
   \begin{align}
    \cT \coloneqq \{t\in[T] \mid (t,\ell) \not\in \cT_h, \forall h \in[H], \forall \ell \in[h-1]\}.
   \end{align}
   By \eqref{eq:dist}, we have that 
   \begin{align}
    |\cT|& \geq T - 4 H d \log(1 + 16 \nu^{-4}H^4),\nn \\
    & \geq T/2, \label{eq:inhere}
   \end{align}
   where the last inequality follows by the choice of $T$ in \cref{alg:ops-dp}. 
   On the other hand, we have that for all $h\in[H]$: 
   \begin{align}
    \sum_{t\in\cT}  \sum_{h\in[H]} \|u\ind{t}_h\|_{(\beta I + U\ind{t}_h)^{-1}}  \leq    \sum_{h\in[H]} \sum_{t\in[T]}  \|u\ind{t}_h\|_{(\beta I + U\ind{t}_h)^{-1}}  \leq  H \sqrt{T d \log (1+ T/\beta)},  \label{eq:bodd}
   \end{align}
   where the last inequality follows by \cref{lem:elliptical}. Therefore, there is a $\tau \in\cT$ such that for all $h\in[H]$:
   \begin{align}
    \|u\ind{\tau}_h\|_{(\beta I + U\ind{\tau}_h)^{-1}}  \leq  \frac{1}{|\cT|}\sum_{t\in\cT}  \sum_{h\in[H]} \|u\ind{t}_h\|_{(\beta I + U\ind{t}_h)^{-1}} &\leq  \frac{H \sqrt{T d \log (1+ T/\beta)}}{|\cT|}, \nn \\
    & \leq 2H \sqrt{\frac{d \log (1+ T/\beta)}{T}}, \quad \text{(by \eqref{eq:inhere})}\nn \\
    &  \leq  1/2, \label{eq:look}
   \end{align}
   where the last inequality follows from the fact that $T\geq 16 d H^2 \log (1+T/\beta)$. Combining \eqref{eq:look} with \cref{lem:designbound} (and the conditioning on $\cE^{\conc}$), we have that for all $h\in[H]$ and $\ell \in [0 \ldotst h-1]$:
   \begin{align}
    \E^{\pi \circ_{\ell+1} \pihat\ind{\tau}_{\ell+1:H}} \left[\|\phih(\x_h,\a_h)\|_{(\beta I + U\ind{\tau}_h)^{-1}} \right]\leq 2 \sqrt{d}  \|u\ind{\tau}_h\|_{(\beta I + U\ind{\tau}_h)^{-1}}   + 2 d\beta^{-1} \veps_\conc  \leq 1/2+ 2 d\beta^{-1} \veps_\conc .
   \end{align}
   Now, by definition of $\cT$, we have that $w_h\ind{t,\ell}=0$ for all $h \in [H]$, and $\ell\in [h-1]$ at the end of each iteration $t\in \cT$. This implies that $W\ind{t+1}_h = W_h\ind{t}$ at the end of any iteration $t\in \cT$, and so by \cref{lem:fitoptvalue} (and the conditioning on $\cE^{\reg}$), we have that for all $\tilde \pi \in \Pi_\bench$ and $(\pi, v)\in \Psi\ind{\tau}_{h-1}$:
   \begin{align}
    \E^{\pi\circ_h \tilde\pi\circ_{h+1}\pihat\ind{t}_{h+1:H}} \left[ \mathbb{I}\{\phiho(\x_{h-1},\a_{h-1})^\top v\geq 0\} \cdot (Q_h\ind{t}(\x_h,\a_h)- \Qhat\ind{t}_h(\x_h,\a_h)) \right] \leq \veps_\reg.
   \end{align}
   Instantiating this with $t = \tau \in \cT$ completes the proof. 
    \end{proof}

We still need a guarantee for the $\texttt{Evaluate}$ subroutine within $\mainalg$.
    \begin{lemma}
        [Guarantee of $\texttt{Evaluate}$ for $\mainalg$]
            \label{lem:evalmeta}
            Let $\veps,\delta\in(0,1)$ be given and consider a call $\mainalg(\Pi_\bench,\veps, \delta)$ (\cref{alg:ops-dp}) with $\Pi_\bench$ as in \cref{par:bench}. Then, for any $\delta\in(0,1)$, there is an event $\cE^{\mathrm{eval}}$ of probability at least $1-\delta$, under which for all $t\in[T]$, the variable $\tau'$ in \cref{alg:ops-dp} satisfies:
            \begin{align}
        \max_{t\in[T]}J(\pihat\ind{t}_{1:H}) - J(\pihat\ind{\tau'}_{1:H})   \leq \veps_\mathrm{eval} \coloneqq H \sqrt{\frac{2\log (T/\delta)}{n_\traj}}.
            \end{align} 
        \end{lemma}
    \begin{proof}[Proof of \cref{lem:evalmeta}]
    The result follows from \cref{lem:eval} with $\delta'=\delta/T$ and \cref{lem:unionbound} (essentially the union bound over $t\in[T]$).
    \end{proof}

We now have all the ingredients to prove \cref{lem:say}.

\begin{proof}[Proof of \cref{lem:say}]
    Let $(\veps_\conc, \cE^{\conc})$, $(\veps_\reg,\cE^\reg)$, and $(\veps_\evale, \cE^{\evale})$ be as in \cref{lem:designbound}, \cref{lem:fitoptvalue}, and \cref{lem:evalmeta}, respectively. By the union bound, we have that $\P[\cE^{\conc} \cap \cE^\reg \cap \cE^\evale]\geq 1 - 2\delta$. Throughout this proof, we condition on $\cE^{\conc} \cap \cE^\reg \cap \cE^\evale$.

    For $\mu, \nu$ as in \cref{alg:ops-dp}, let $\gamma \coloneqq  \mu/\sqrt{d_\nu}$, where $d_\nu \coloneqq 5 d \log(1+16 H^4 \nu^{-4})$. Further, let $\tau \in [T]$ be as in \cref{lem:test} and define 
    \begin{gather}\pihath^{\star}(\cdot) \coloneqq \mathbb{I}\{\rgh(\cdot)<\gamma  \} \cdot \pihath\ind{\tau}(\cdot) + \mathbb{I}\{\rgh(\cdot)\geq \gamma \} \cdot \pistarh(\cdot),
    \end{gather}
    for $h\in[H]$ and $\rgh$ as in \cref{def:designrange}; note that by \eqref{eq:talk} and the fact that $\pihat\ind{\tau}_h, \pistar\in \Pi_\base$, where $\Pi_\base=\left\{ x \mapsto \argmax_{a\in \cA} \theta^\top \phih(x,a)  \mid  \theta \in \bbB(H)   \right\}$, we have
    \begin{gather}
    \pihat^\star_h \in \Pi_\bench.
    \label{eq:dona}
    \end{gather}  
   At a high-level, our strategy will be to show the sequence of inequalities:
    \begin{align}
    J(\pistar)&\leq J(\pihat^\star_{1:H})  + O(\veps),\nn \\
    J(\pihat^\star_{1:H})&\leq J(\pihat\ind{\tau}_{1:H})  + O(\veps),\nn \\
    J(\pihat\ind{\tau}_{1:H})&\leq J(\pihat_{1:H})  + O(\veps).
    \end{align}
Summing up these inequalities and telescoping would imply the desired result.
\fakepar{Suboptimality of $\pihat^\star_{1:H}$}
First, by the performance difference lemma (see \cref{lem:pdl}), we have 
\begin{align}
  & J(\pistar) - J(\pihat^\star_{1:H})\nn \\ & = \sum_{h=1}^H \E^{\pihat^\star}\left[ Q_h^{\pi^\star}(\x_h,\pistarh(\x_h))- Q_h^{\pi^\star}(\x_h,\pihath^{\star}(\x_h)) \right], \nn \\
   & = \sum_{h=1}^H \E^{\pihat^\star}\left[ \inner{\phih(\x_h,\pistarh(\x_h))- \phih(\x_h,\pihath^{\star}(\x_h))}{\theta_h^{\pistar}} \right],\nn \\
   & = \sum_{h=1}^H \E^{\pihat^\star}\left[\mathbb{I}\{\rgh(\x_h) <\gamma \} \cdot \inner{\phih(\x_h,\pistarh(\x_h))- \phih(\x_h,\pihath\ind{\tau}(\x_h))}{\theta_h^{\pistar}} \right], \quad \text{(by definition of $\pihat^\star$)}\nn \\
   & \leq \sum_{h=1}^H \E^{\pihat^\star}\left[\mathbb{I}\{\mathrm{Rg}(\x_h) < \sqrt{2d}\gamma \} \cdot \inner{\phih(\x_h,\pistarh(\x_h))- \phih(\x_h,\pihath\ind{\tau}(\x_h))}{\theta_h^{\pistar}} \right],  \quad \text{(by \cref{lem:rangerel})}\nn \\
   & \leq \sum_{h=1}^H \E^{\pihat^\star}\left[\mathbb{I}\{\mathrm{Rg}(\x_h) < \sqrt{2d}\gamma \} \cdot  \mathrm{Rg}(\x_h)\right], \nn \\
   & \leq H\sqrt{2d} \gamma = H \sqrt{2d/d_\nu} \mu.
   \label{eq:perf}
\end{align}
We now bound the suboptimality of $\pihat\ind{\tau}_{1:H}$ relative to $\pihat^\star_{1:H}$, where we recall that $\tau \in [T]$ is as in \cref{lem:test}.

\fakepar{Suboptimality of $\pihat\ind{\tau}_{1:H}$} For this part of the proof, we need some additional notation for the bonuses and optimistic value functions; for $\ell\in[H]$, $(x,a) \in \cX_\ell\times \cA$, and $t\in[T],$ we define 
\begin{gather}
    \bar{b}\ind{t}_\ell(x) \coloneqq \min\left(H,\frac{\veps}{4 H} \cdot\max_{a'\in \cA} \|\phiell(x, a')\|_{(\beta I + U\ind{t}_\ell)^{-1}}\right),\quad  \quad b\ind{t}_\ell(x) \coloneqq \bar{b}\ind{t}_\ell(x) \cdot \mathbb{I}\{\|\varphi(x;W\ind{t}_\ell)\|\geq \mu\}, \label{eq:bonuses} \\
 Q\ind{t}_\ell(x,a) = Q^{\pihat\ind{t}}_{\ell}(x,a) + \E^{\pihat\ind{t}}\left[\sum_{k=\ell}^H b\ind{t}_k(\x_k) \mid \x_\ell =x,\a_\ell =a\right],\quad \text{and} \quad V\ind{t}_\ell(x) \coloneqq  Q\ind{t}_\ell(x,\pihat\ind{t}_\ell(x)).\label{eq:optvalue}
\end{gather} 
With this, we proceed via backward induction to show that for all $\ell =H+1, \dots, 1$ and $(x,a)\in \cX_\ell\times \cA$:  
\begin{align}
 Q^{\pihat^\star}_\ell(x,a) &\leq  Q\ind{\tau}_{\ell}(x,a) + \bar{b}\ind{\tau}_\ell(x) \cdot \mathbb{I}\{\|\varphi(x;W\ind{\tau}_\ell)\|<\mu\}, \label{eq:ind1} \\
V^{\pihat^\star}_\ell(x) &\leq V\ind{\tau}_\ell(x) + \xi_\ell\ind{\tau}(x,\pihatell^\star(x)) - \xi_\ell\ind{\tau}(x,\pihatell\ind{\tau}(x)) +  \bar{b}\ind{\tau}_\ell(x) \cdot \mathbb{I}\{\|\varphi(x;W\ind{\tau}_\ell)\|<\mu\},  \label{eq:ind2}
\end{align}
where for $(\tilde x,\tilde a)\in \cX_\ell\times \cA$,
\begin{align}
 \xi_\ell\ind{\tau}(\tilde x,\tilde a) \coloneqq Q\ind{\tau}_\ell(\tilde x,\tilde a) - \Qhat\ind{\tau}_\ell(\tilde x,\tilde a)\quad \text{and}\quad  \Qhat\ind{t}_\ell(\tilde x,\tilde a) = \phiell(\tilde x,\tilde a)^\top \thetahat\ind{t}_{\ell}+ b\ind{t}_\ell(\tilde x), \label{eq:bhat}
\end{align}  with the convention that $Q^{\pihat^\star}_{H+1}\equiv Q\ind{\tau}_{H+1}\equiv \Qhat\ind{\tau}_{H+1}\equiv 0$.

    The base case $\ell=H+1$ follows trivially by the convention that $Q^{\pihat^\star}_{H+1}\equiv Q\ind{\tau}_{H+1}\equiv \Qhat\ind{\tau}_{H+1}\equiv 0$. Now, let $h\in[H]$ and suppose that the induction hypothesis holds for all $\ell=[h+1\ldotst H+1]$. We show that it holds for $\ell=h$. 

\paragraphi{We show \eqref{eq:ind1} for $\ell=h$} Fix $(x,a)\in \cX_h \times \cA$.
By \cref{lem:skipping} instantiated with 
\begin{itemize}
    \item $(\pi'_k,\pihat_{k},\pihat'_k)= (\pi^\star,\pihat\ind{\tau}_k, \pihat^\star_k)$;
    \item $\cK_k = \{\tilde x \in \cX_k\mid \rgell(\tilde x)<\gamma\}$;
    \item $\tilde r_k\equiv R$ ($R$ is the reward function); and 
    \item $V_k = V_k^{\pihat^\star}$, 
\end{itemize}
for all $k\in [h+1\ldotst H]$, we get that 
\begin{align}
\E[V^{\pihat^\star}_{h+1}(\x_{h+1})\mid \x_h =x,\a_h =a] & =  \sum_{\ell=h+1}^H \E^{\pihat\ind{\tau}}\left[\mathbb{I}\{\rgell(\x_\ell) \geq \gamma\}\cdot \prod_{k=h+1}^{\ell-1} \mathbb{I}\{\rgk(\x_k)<\gamma \} \cdot V^{\pihat^\star}_{\ell}(\x_{\ell}) \mid \x_h=x,\a_h= a \right]\nn \\
& \quad + \sum_{\ell=h+1}^H\E^{\pihat\ind{\tau}}\left[   \prod_{k=h+1}^{\ell} \mathbb{I}\{\rgk(\x_k)<\gamma \} \cdot r_\ell(\x_\ell,\a_\ell) \mid \x_h=x,\a_h=a \right]. \label{eq:Qstar}
\end{align}
We instantiate \cref{lem:skipping} again, this time with 
\begin{itemize}
    \item $(\pi'_k,\pihat_k,\pihat'_k) =  (\pihat\ind{\tau}_k,\pihat\ind{\tau}_k,\pihat\ind{\tau}_k)$;
    \item $\cK_k = \{\tilde x \in \cX_k\mid \rgell(\tilde x)<\gamma\}$;
    \item $\tilde r_k(\tilde x,\tilde a)\coloneqq R(\tilde x,\tilde a) + {b}\ind{\tau}_k(\tilde x)$ (with ${b}\ind{\tau}_k$ as in \eqref{eq:bonuses}); and 
    \item $V_k = V_k\ind{\tau}$,
\end{itemize}
for all $k\in [h+1\ldotst H]$, to get that 
\begin{align}
     \E[V\ind{\tau}_{h+1}(\x_{h+1})\mid \x_h=x,\a_h =a] &  = \sum_{\ell=h+1}^H \E^{\pihat\ind{\tau}}\left[ \mathbb{I}\{\rgell(\x_\ell)\geq \gamma \}\cdot \prod_{k=h+1}^{\ell-1} \mathbb{I}\{\rgk(\x_k)<\gamma \} \cdot V\ind{\tau}_{\ell}(\x_{\ell}) \mid \x_h=x,\a_h= a \right]\nn \\
    & \quad + \sum_{\ell=h+1}^H\E^{\pihat\ind{\tau}}\left[\prod_{k=h+1}^{\ell} \mathbb{I}\{\rgk(\x_k)<\gamma \} \cdot \tilde r_\ell(\x_\ell,\a_\ell) \mid \x_h=x,\a_h=a \right]. \label{eq:Qt}
    \end{align}
Thus, combining \eqref{eq:Qstar} and \eqref{eq:Qt} and using that $\tilde{r}_\ell(\cdot)\geq r_\ell(\cdot)$, we get 
\begin{align}
    & \E[V^{\pihat^\star}_{h+1}(\x_{h+1})  - V\ind{\tau}_{h+1}(\x_{h+1})\mid \x_h = x,\a_h = a]  \nn \\
    & \leq   \sum_{\ell=h+1}^H \E^{\pihat\ind{\tau}}\left[ \mathbb{I}\{\rgell(\x_\ell)\geq \gamma\}\cdot \prod_{k=h+1}^{\ell-1} \mathbb{I}\{\rgk(\x_k)<\gamma \} \cdot \left(V^{\pihat^\star}_{\ell}(\x_{\ell})- V\ind{\tau}_{\ell}(\x_{\ell}) \right) \mid \x_h=x,\a_h= a  \right], 
\end{align}
and so by the induction hypothesis; in particular \eqref{eq:ind2} for $\ell\in[h+1\ldotst H+1]$, we have
\begin{align}
    & \E[V^{\pihat^\star}_{h+1}(\x_{h+1})  - V\ind{\tau}_{h+1}(\x_{h+1})\mid \x_h = x,\a_h = a]\nn \\
    &  \leq  \sum_{\ell=h+1}^H\E^{\pihat\ind{\tau}}\left[\mathbb{I}\{\rgell(\x_\ell)\geq \gamma\} \prod_{k=h+1}^{\ell-1} \mathbb{I}\{\rgk(\x_k)<\gamma \} \left( \xi\ind{\tau}_\ell(\x_\ell, \pihatell^\star(\x_\ell)) -\xi\ind{\tau}_\ell(\x_\ell,\pihatell\ind{\tau}(\x_\ell))  \right) \mid \x_h=x,\a_h=a \right] \nn \\
    & \quad + \sum_{\ell=h+1}^H\E^{\pihat\ind{\tau}}\left[\mathbb{I}\{\rgell(\x_\ell)\geq \gamma\} \prod_{k=h+1}^{\ell-1} \mathbb{I}\{\rgk(\x_k)<\gamma \}\cdot \mathbb{I}\{\|\varphi(\x_\ell;W_\ell\ind{\tau})\|<\mu\} \cdot \bar{b}_\ell\ind{\tau}(\x_\ell)  \mid \x_h=x,\a_h=a \right]. \label{eq:four}
\end{align}
Now, by \cref{lem:subset}, we have that $\rgh(\tilde x)\leq \sqrt{d_\nu} \cdot \|\varphi(\tilde x;W_\ell\ind{\tau})\|$ for all $\tilde x\in \cX_\ell$. Thus, since $\mu = \sqrt{d_\nu}\cdot\gamma$, we have that for all $\tilde x\in \cX_\ell$, $\mathbb{I}\{\|\varphi(\tilde x;W_\ell\ind{\tau})\|<\mu\}=1$ only if $\mathbb{I}\{\rgh(\tilde x)<\gamma\}=1$. This implies that the second sum in \eqref{eq:four} is zero, and so
\begin{align}
    & \E[V^{\pihat^\star}_{h+1}(\x_{h+1})  - V\ind{\tau}_{h+1}(\x_{h+1})\mid \x_h = x,\a_h = a] \leq   \sum_{\ell=h+1}^H\E^{\pihat\ind{\tau}}\left[g\ind{\tau}_\ell(\x_{h+1:\ell})  \mid \x_h=x,\a_h=a \right], \label{eq:notice}
\end{align} 
where \begin{align}
    g\ind{\tau}_\ell(\x_{h+1:\ell}) \coloneqq  \prod_{k=h+1}^{\ell-1} \mathbb{I}\{\rgk(\x_k)<\gamma \} \cdot\mathbb{I}\{\rgell(\x_\ell)\geq \gamma\} \cdot\left(\xi\ind{\tau}_\ell(\x_\ell,\pihatell^\star(\x_\ell))  - \xi\ind{\tau}_\ell(\x_\ell,\pihatell\ind{\tau}(\x_\ell)) \right). \label{eq:gdef}
\end{align}
Now, by definition of the bonuses and $Q\ind{\tau}_\ell$ in \eqref{eq:bonuses} and \eqref{eq:optvalue}, we have $Q\ind{\tau}_\ell \in [0, H + H^2]$. On the other hand, since $\thetahat\ind{\tau}_\ell$ in the definition of $\Qhat\ind{\tau}_\ell$ in \eqref{eq:bhat} satisfies $\thetahat_\ell\ind{\tau}\in \bbB(H + H^2/\lambda)$; see \cref{line:sigmah}-\cref{line:theta} of \cref{alg:fitoptvalue}, we have $\Qhat\ind{\tau}_\ell(\tilde x,\tilde a) \in[-2H^2/\lambda, 2 H^2/ \lambda]$, for all $(\tilde x,\tilde a)\in \cX_\ell\times \cA$. Therefore, \begin{align}\xi\ind{\tau}_\ell(\tilde x,\tilde a) = Q\ind{\tau}_\ell(\tilde x,\tilde a) - \Qhat\ind{\tau}_\ell(\tilde x,\tilde a) \in [- 3 H^2/\lambda,  3H^2/\lambda],\label{eq:boundedxi}\end{align}for all $(\tilde x,\tilde a)\in \cX_\ell\times \cA$ and so by \cref{lem:calm} instantiated with $\pi = \pihat\ind{\tau}$, $f(\cdot)= \xi\ind{\tau}_\ell(\cdot,\pihatell^\star(\cdot))  - \xi\ind{\tau}_\ell(\cdot,\pihatell\ind{\tau}(\cdot))$, and $L= 6 H^2/\lambda$, we get that there exists \begin{align}\theta\ind{\tau}_{h,\ell}\in \bbB(24 \tilde{d} H^4/(\gamma \lambda)),  \label{eq:firsttheta}
\end{align} where $\tilde{d}= 4 d \log \log d + 16$, such that 
\begin{align}
   \E^{\pihat\ind{\tau}}\left[g\ind{\tau}_\ell(\x_{h+1:\ell}) \mid \x_h = x,\a_h = a\right] = \phih(x,a)^\top \theta\ind{\tau}_{h,\ell}. \label{eq:smell}
\end{align}
Thus, by \cref{lem:changeofmeasure2} and $\veps_\conc$ as in \cref{lem:designbound}, we have that 
\begin{align}
& \left|\E^{\pihat\ind{\tau}}\left[g\ind{\tau}_\ell(\x_{h+1:\ell}) \mid \x_h = x,a_h = a\right]\right|\nn \\
&  \leq \sqrt{d} \|\phih(x,a)\|_{(\beta I + U\ind{\tau}_h)^{-1}} \cdot \left(\sqrt{\frac{\beta}{d}}\|\theta\ind{\tau}_{h,\ell}\| +T\veps_\conc + \sum_{(\pi, v) \in \Psi\ind{\tau}_h}\left| \E^{\pi}[\phih(\x_h,\a_h)^\top \theta\ind{\tau}_{h,\ell} \cdot \mathbb{I}\{\phih(\x_h,\a_h)^\top v \geq 0\}] \right| \right).\label{eq:talent}
\end{align}
Now, using \eqref{eq:smell} again and the law of total expectation, we get that for any $(\pi,v)\in \Psi\ind{\tau}_h$: 
\begin{align}
   & \left| \E^{\pi}[\phih(\x_h,\a_h)^\top \theta\ind{\tau}_{h,\ell} \cdot \mathbb{I}\{\phih(\x_h,\a_h)^\top v \geq 0\}] \right|  = \left| \E^{\pi\circ_{h+1}\pihat\ind{\tau}}[g\ind{\tau}_\ell(\x_{h+1:\ell}) \cdot \mathbb{I}\{\phih(\x_h,\a_h)^\top v \geq 0\}] \right|, 
\end{align}
and so by letting $\mathbf{I}_{h,\ell,v} \coloneqq \mathbb{I}\{\phih(\x_h,\a_h)^\top v \geq 0\} \cdot \prod_{k=h+1}^{\ell-1} \mathbb{I}\{\rgk(\x_k)<\gamma\}$ and using the definition of $g\ind{\tau}_\ell$ in \eqref{eq:gdef}:
\begin{align}
    & \left| \E^{\pi}[\phih(\x_h,\a_h)^\top \theta\ind{\tau}_{h,\ell} \cdot \mathbb{I}\{\phih(\x_h,\a_h)^\top v \geq 0\}] \right|\nn \\
   & =\left| \E^{\pi \circ_{h+1}\pihat\ind{\tau}}\left[\mathbf{I}_{h,\ell,v} \cdot\E\left[\mathbb{I}\{\rgell(\x_\ell)\geq \gamma\} \cdot \left(\xi\ind{\tau}_\ell(\x_\ell,\pihatell^\star(\x_\ell))-\xi\ind{\tau}_\ell(\x_\ell,\pihatell\ind{\tau}(\x_\ell)) \right)\mid \x_{\ell-1},\a_{\ell-1} \right] \right]\right|. \label{eq:red}
\end{align}
Now, by \eqref{eq:boundedxi}, \cref{lem:admissible}, and \cref{lem:admreal}, there exists  \begin{align}\tilde\theta\ind{\tau}_{\ell-1} \in \bbB(24 \tilde{d} H^3/(\gamma \lambda)) \label{eq:secondtheta}
\end{align} such that for all $(\tilde x,\tilde a)\in \cX_{\ell-1}\times \cA$:
\begin{align}
    \phiello(\tilde x,\tilde a)^\top \tilde\theta\ind{\tau}_{\ell-1} & = \E\left[\mathbb{I}\{\rgell(\x_\ell)\geq \gamma\} \cdot \left(\xi\ind{\tau}_\ell(\x_\ell,\pihatell^\star(\x_\ell))-\xi\ind{\tau}_\ell(\x_\ell,\pihatell\ind{\tau}(\x_\ell)) \right)\mid \x_{\ell-1}=\tilde x,\a_{\ell-1}=\tilde a \right], \label{eq:theeta} \\
    &=\E\left[\xi\ind{\tau}_\ell(\x_\ell,\pihatell^\star(\x_\ell))-\xi\ind{\tau}_\ell(\x_\ell,\pihatell\ind{\tau}(\x_\ell))\mid \x_{\ell-1}=\tilde x,\a_{\ell-1}=\tilde a \right],\label{eq:theetanew}
\end{align}
where the last equality follows by the fact that $\pihatell^{\star}(\cdot)= \mathbb{I}\{\rgell(\cdot)<\gamma  \} \cdot \pihatell\ind{\tau}(\cdot) + \mathbb{I}\{\rgell(\cdot)\geq \gamma \} \cdot \pistar(\cdot)$, by definition. Using \eqref{eq:theeta} and \cref{lem:changeofmeasure2}, we get that for all $(\tilde x,\tilde a)\in \cX_{\ell-1}\times \cA$:
\begin{align}
    & \left|\E\left[\mathbb{I}\{\rgell(\x_\ell)\geq \gamma\} \cdot \left(\xi\ind{\tau}_\ell(\x_\ell,\pihatell^\star(\x_\ell))-\xi\ind{\tau}_\ell(\x_\ell,\pihatell\ind{\tau}(\x_\ell)) \right)\mid \x_{\ell-1}=\tilde x,\a_{\ell-1}=\tilde a \right] \right|\nn \\
    & \leq \sqrt{d} \|\phiello(\tilde x,\tilde a)\|_{(\beta I + U\ind{\tau}_{\ell-1})^{-1}}\cdot \left(\sqrt{\beta /d}\cdot \|\tilde\theta\ind{\tau}_{\ell-1}\|+ T \veps_\conc\right) \nn \\
    & \quad + \sqrt{d}\|\phiello(\tilde x,\tilde a)\|_{(\beta I + U\ind{\tau}_{\ell-1})^{-1}} \sum_{(\pi', v') \in \Psi\ind{\tau}_{\ell-1}}\left|\E^{\pi'}[\phiello(\x_{\ell-1},\a_{\ell-1})^\top \tilde\theta\ind{\tau}_{\ell-1} \cdot \mathbb{I}\{\phiello(\x_{\ell-1},\a_{\ell-1})^\top v' \geq 0\}] \right|,
\intertext{
 and so using \eqref{eq:theetanew}, the law of total expectation, and the triangle inequality, we get
}
    &  \leq \sqrt{d} \|\phiello(\tilde x,\tilde a)\|_{(\beta I + U\ind{\tau}_{\ell-1})^{-1}}\cdot \left(\sqrt{\beta /d}\cdot \|\tilde\theta\ind{\tau}_{\ell-1}\|+ T \veps_\conc\right) \nn \\
    & \quad + \sqrt{d}\|\phiello(\tilde x,\tilde a)\|_{(\beta I + U\ind{\tau}_{\ell-1})^{-1}} \sum_{(\pi', v') \in \Psi\ind{\tau}_{\ell-1}}\left|\E^{\pi'}[\xi\ind{\tau}_\ell(\x_\ell,\pihatell\ind{\tau}(\x_\ell)) \cdot \mathbb{I}\{\phiello(\x_{\ell-1},\a_{\ell-1})^\top v' \geq 0\}]\right|\nn \\
    & \quad + \sqrt{d}\|\phiello(\tilde x,\tilde a)\|_{(\beta I + U\ind{\tau}_{\ell-1})^{-1}} \sum_{(\pi', v') \in \Psi\ind{\tau}_{\ell-1}}\left|\E^{\pi'}[ \xi\ind{\tau}_\ell(\x_\ell,\pihatell^\star(\x_\ell)) \cdot \mathbb{I}\{\phiello(\x_{\ell-1},\a_{\ell-1})^\top v' \geq 0\}]\right|, \nn\\
    & \leq \sqrt{d}\|\phiell(\tilde x,\tilde a)\|_{(\beta I + U\ind{\tau}_{\ell-1})^{-1}} \cdot \left(\sqrt{\beta/d} \cdot \|\tilde\theta\ind{\tau}_{\ell-1}\| +T \veps_\conc+ 2  T \veps_\reg \right),
\end{align}
where the last inequality follows by  \cref{lem:test}, the conditioning on $\cE^\reg \cap \cE^\conc$, and the fact that $\pihat^\star \in \Pi_\bench$ (see \eqref{eq:dona}). Plugging this into \eqref{eq:red}, we get that for all $(\pi,v)\in \Psi\ind{\tau}_h$:
\begin{align}
    & \left| \E^{\pi}[\phih(\x_h,\a_h)^\top \theta\ind{\tau}_{h,\ell} \cdot \mathbb{I}\{\phih(\x_h,\a_h)^\top v \geq 0\}] \right| \nn \\
  & \leq \sqrt{d}\E^{\pi \circ_{h+1} \pihat\ind{\tau}} \left[\mathbf{I}_{h,\ell,v}\cdot \|\phiello(\x_{\ell-1},\a_{\ell-1})\|_{(\beta I + U_{\ell-1}\ind{\tau})^{-1}}\right] \cdot \left(\sqrt{\beta/d}\cdot \| \tilde\theta\ind{\tau}_{\ell-1}\| +T \veps_\conc+ 2 T \veps_\reg \right),
\end{align}
where we recall that $\mathbf{I}_{h,\ell,v}=\mathbb{I}\{\phih(\x_h,\a_h)^\top v \geq 0\} \cdot \prod_{k=h+1}^{\ell-1} \mathbb{I}\{\rgk(\x_k)<\gamma\}\leq 1$. Thus, we have for all $(\pi,v)\in \Psi\ind{\tau}_h$: 
\begin{align}
   &  \left| \E^{\pi}[\phih(\x_h,\a_h)^\top \theta\ind{\tau}_{h,\ell} \cdot \mathbb{I}\{\phih(\x_h,\a_h)^\top v \geq 0\}] \right|\nn \\
   &  \leq \sqrt{d}\E^{\pi \circ_{h+1} \pihat\ind{\tau}} \left[\|\phiello(\x_{\ell-1},\a_{\ell-1})\|_{(\beta I + U_{\ell-1}\ind{\tau})^{-1}}\right]\cdot  \left(\sqrt{\beta/d}\cdot \|\tilde\theta\ind{\tau}_{\ell-1}\| +T \veps_\conc+ 2  T \veps_\reg \right), \nn \\
   & \leq \left(\sqrt{d}/2 + 2 d^{3/2}\beta^{-1} \veps_\conc\right) \cdot \left(\sqrt{\beta/d}\cdot\|\tilde\theta\ind{\tau}_{\ell-1}\| +T \veps_\conc+ 2 T \veps_\reg \right), 
\end{align} 
where the last inequality follows by \cref{lem:test} and the conditioning on $\cE^\reg \cap \cE^\conc$. 
Combining this with \eqref{eq:talent} and \eqref{eq:notice}, we get that 
\begin{align}
&Q^{\pihat^\star}_h(x,a) - \left(R(x,a)+ \E[V\ind{\tau}_{h+1}(\x_{h+1})\mid \x_h = x,\a_h = a] \right)\nn \\
& =  \E[V^{\pihat^\star}_{h+1}(\x_{h+1})  - V\ind{\tau}_{h+1}(\x_{h+1})\mid \x_h = x,\a_h = a], \nn \\
& \leq \sqrt{d}\cdot \|\phih(x,a)\|_{(\beta I + U\ind{\tau}_h)^{-1}} \cdot \left(\sqrt{\frac \beta d}\cdot \sum_{\ell=h+1}^H\|\theta\ind{\tau}_{h,\ell}\| +T(H-h)\veps_\conc \right)\nn \\
& \quad + T\cdot \|\phih(x,a)\|_{(\beta I + U\ind{\tau}_h)^{-1}} \cdot \left(\left(\frac d 2 + \frac{2 d^{2} \veps_\conc}{\beta}\right) \cdot \left(\sqrt{\frac \beta d}\cdot\sum_{\ell=h+1}^H\|\tilde\theta\ind{\tau}_{\ell-1}\| +T (H-h) \veps_\conc+ 2 T(H-h) \veps_\reg \right) \right),\nn \\
& \leq  \|\phih(x,a)\|_{(\beta I + U\ind{\tau}_h)^{-1}} \cdot \left( \left(\frac{24\sqrt{\beta}\tilde{d} H^4}{\gamma \lambda \sqrt{d}} + T \veps_\conc+   T \veps_\reg \right) \cdot \left( HTd + \frac{4 d^2 H T  \veps_\conc}{\beta}\right) \right),  \label{eq:lighter}
\\
& \leq \|\phih(x,a)\|_{(\beta I + U\ind{\tau}_h)^{-1}} \cdot \frac{\veps}{4H},
\label{eq:Cat}
\end{align}
where \eqref{eq:lighter} follows by the bounds on $\|\theta\ind{\tau}_{\ell-1}\|$ and $\|\tilde\theta\ind{\tau}_{\ell-1}\|$ from \eqref{eq:firsttheta} and \eqref{eq:secondtheta}, respectively; and \eqref{eq:Cat} follows from the choices of parameters $n_\traj, \mu,\nu, \lambda$, and $\beta$ in \cref{alg:fitbonus}.
On the other hand, we have that 
\begin{align}
Q^{\pihat^\star}_h(x,a)  \leq H \quad \text{and} \quad R(x,a)+ \E[V\ind{\tau}_{h+1}(\x_{h+1})\mid \x_h = x,\a_h = a] \geq 0.
\end{align}
Combining this with \eqref{eq:Cat} and the fact that $\min(H,c+b)\leq c+\min(H,b)$ for $c,b\geq 0$, we get 
\begin{align}
    Q^{\pihat^\star}_h(x,a)
    & \leq  \min\left(H,R(x,a)+ \E[V\ind{\tau}_{h+1}(\x_{h+1})\mid \x_h = x,\a_h = a]+  \|\phih(x,a)\|_{(\beta I + U_h\ind{\tau})^{-1}} \cdot \frac{\veps}{4 H}\right), \nn \\
    & \leq  R(x,a)+ \E[V\ind{\tau}_{h+1}(\x_{h+1})\mid \x_h = x,\a_h = a]+ \min\left(H, \|\phih(x,a)\|_{(\beta I + U_h\ind{\tau})^{-1}} \cdot \frac{\veps}{4 H}\right), \nn \\
    & \leq R(x,a)+ \E[V\ind{\tau}_{h+1}(\x_{h+1})\mid \x_h = x,\a_h = a] + \min\left(H,\max_{\tilde a\in \cA} \|\phih(x,\tilde a)\|_{(\beta I + U_h\ind{\tau})^{-1}} \cdot \frac{\veps}{4 H}\right), \nn \\
    & = R(x,a)+ \E[V\ind{\tau}_{h+1}(\x_{h+1})\mid \x_h = x,\a_h = a] + b\ind{\tau}_h(x) + \bar b\ind{\tau}_h(x) \cdot \mathbb{I}\{\|\varphi(x;W_h\ind{\tau})\|<\mu\}, \label{eq:aqua} \\
    & =   Q_h\ind{\tau}(x,a) + \bar b\ind{\tau}_h(x) \cdot \mathbb{I}\{\|\varphi(x;W_h\ind{\tau})\|<\mu\},
\end{align}
where \eqref{eq:aqua} follows by the definitions of $b\ind{\tau}_h$ and $\bar{b}\ind{\tau}_h$ in \eqref{eq:bonuses}, and the last inequality follows by the definition of $Q\ind{\tau}_h$ in \eqref{eq:optvalue}.
This shows \eqref{eq:ind1} for $\ell=h$.

\paragraphi{We show \eqref{eq:ind2} for $\ell=h$} We have 
\begin{align}
  &  V^{\pihat^\star}_h(x) - V\ind{\tau}_h(x) \nn \\ & = Q^{\pihat^\star}_h(x,\pihath^{\star}(x)) - Q\ind{\tau}_h(x,\pihath\ind{\tau}(x)), \nn \\ 
    & \leq  Q^{\pihat^\star}_h(x,\pihath^{\star}(x)) - Q\ind{\tau}_h(x,\pihath\ind{\tau}(x)) + \Qhat\ind{\tau}_h(x,\pihath\ind{\tau}(x)) - \Qhat\ind{\tau}_h(x,\pihath^{\star}(x)), \quad \text{(by definition of $\pihath\ind{\tau}$)} \nn \\
    & = Q^{\pihat^\star}_h(x,\pihath^{\star}(x)) - Q\ind{\tau}_h(x,\pihath^{\star}(x)) + Q\ind{\tau}_h(x,\pihath^{\star}(x)) - Q\ind{\tau}_h(x,\pihath\ind{\tau}(x)) + \Qhat\ind{\tau}_h(x,\pihath\ind{\tau}(x)) - \Qhat\ind{\tau}_h(x,\pihath^{\star}(x)), \nn \\
    & \leq  \xi\ind{\tau}_h(x,\pihath^{\star}(x))-\xi\ind{\tau}_h(x,\pihath\ind{\tau}(x)) + \bar b\ind{\tau}_h(x) \cdot \mathbb{I}\{\|\varphi(x;W_h\ind{\tau})\|<\mu\},
\end{align}
where the last inequality follows by \eqref{eq:ind1} with $\ell=h$. This shows \eqref{eq:ind2} for $\ell=h$ and completes the induction. Instantiation \eqref{eq:ind2} with $\ell=1$ and using the definition of $V\ind{\tau}_1$ in \eqref{eq:optvalue}, we get that 
\begin{align}
   & J(\pihat^\star_{1:H}) - J(\pihat\ind{\tau}_{1:H})\nn \\ &
   =\E\left[ V^{\pihat^\star}_1(\x_1)\right] - \E\left[V^{\pihat\ind{\tau}}_1(\x_1)\right],\nn \\ &
   = \E\left[ V^{\pihat^\star}_1(\x_1)\right] - \E\left[V\ind{\tau}_1(\x_1)\right]+ \E\left[\sum_{h=1}^H b\ind{\tau}_h(\x_h)\right], \nn\\
    & \leq  \E\left[\xi\ind{\tau}_1(\x_1,\pihat_1^{\star}(\x_1))-\xi\ind{\tau}_1(\x_1,\pihat_1\ind{\tau}(\x_1)) + \bar b\ind{\tau}_1(\x_1) \cdot \mathbb{I}\{\|\varphi(\x_1;W_1\ind{\tau})\|<\mu\}\right] +  \E\left[\sum_{h=1}^H b\ind{\tau}_h(\x_h)\right],\nn \\
    & \leq 2  \veps_\reg + \E\left[\bar b\ind{\tau}_1(\x_1) \cdot \mathbb{I}\{\|\varphi(\x_1;W_1\ind{\tau})\|<\mu\} \right]+ \E\left[\sum_{h=1}^H b\ind{\tau}_h(\x_h)\right], \quad \text{(see below)} \label{eq:intrr}\\
    & \leq 2  \veps_\reg + 2\E\left[\sum_{h=1}^H b\ind{\tau}_h(\x_h)\right],
\end{align}
where \eqref{eq:intrr} follows by \cref{lem:test}, the conditioning on $\cE^\reg \cap \cE^\conc$, and the fact that $\pihat^\star \in \Pi_\bench$ (see \eqref{eq:dona}). Now, by the definition of $({b}_h\ind{\tau})$ in \eqref{eq:bonuses}, \cref{lem:test}, and the conditioning on $\cE^\conc \cap \cE^{\reg}$, we get that
\begin{align}
    J(\pihat^\star_{1:H}) - J(\pihat\ind{\tau}_{1:H}) & \leq 2  \veps_\reg + \frac{\veps}{4} \cdot \left(1 + 4 d \beta^{-1}\veps_\conc\right). \label{eq:ter}
\end{align}
\paragraph{Suboptimality of $\pihat_{1:H}$} Let $\tau'$ be as in \cref{alg:ops-dp} just before the algorithm returns, and note that the policy $\pihat_{1:H}$ satisfies $\pihat_{1:H}=\pihat\ind{\tau'}_{1:H}$. Now, by \cref{lem:evalmeta} and the conditioning on $\cE^\evale$, we have that 
\begin{align}
     J(\pihat\ind{\tau}_{1:H}) \leq  \max_{t\in [T]}  J(\pihat\ind{t}_{1:H}) \leq   J(\pihat\ind{\tau'}_{1:H}) + H\sqrt{\frac{2 \log (T/\delta)}{n_\traj}} .\label{eq:afg}
\end{align}
Combining \eqref{eq:perf}, \eqref{eq:ter}, and \eqref{eq:afg}, we get 
\begin{align}
    J(\pistar) - J(\pihat_{1:H}) \leq H\sqrt{\frac{2 \log (T/\delta)}{n_\traj}} + 2  \veps_\reg + \veps \cdot \left({H} + 2 d H \beta^{-1}\veps_\conc\right) + H \sqrt{2d/d_\nu}\cdot \mu. \label{eq:inee}
\end{align}
Using the expressions of $\veps_\conc$ and $\veps_\reg$ in \cref{lem:designbound} and \cref{lem:fitoptvalue}, respectively, and the choices of $\beta, \mu$, $\nu$, $T$, and $n_\traj$ in \cref{alg:ops-dp}, we get that the right-hand of side of \eqref{eq:inee} is at most $\veps$, which completes the proof.

\paragraph{Number of Oracle calls}
A single call to the subroutine $\texttt{FitOptVal}$ at iteration $t$ makes $H t$ calls to the policy optimization Oracle; this is because the problem in \cref{eq:computation} of \cref{alg:fitoptvalue} (i.e.~$\min_{\tilde\pi\in \Pi_\bench}\Delta_{h,\ell}(\pi,v,\tilde \pi)$) has to be solved for all $\ell\in[h+1\ldotst H]$ and $(\pi,v)\in \Psi\ind{t}_{h-1}$, and we know that $|\Psi\ind{t}_{h-1}|\leq t$. Thus, since \cref{alg:ops-dp} calls $\texttt{FitOptVal}$ $T H$ times, the total number of calls to the policy optimization Oracle is at most $H^2 T^2$. Now the desired number of Oracle calls follows by the choice of $T$ in \cref{alg:ops-dp}.
\end{proof}

\section{Implementation of the CSC Oracle over $\Pi_\bench$}
\label{sec:oracle}

\begin{lemma}
    \label{lem:comp}
        Let $n\in \mathbb{N}$, $h\in [H]$, and $(c\ind{1},x\ind{1},a\ind{1}),\dots, (c\ind{n},x\ind{n}, a\ind{n})\in \reals \times \cX_h \times \cA$ be given. Then, for the benchmark policy class $\Pi_\bench$ in \cref{par:bench}, it is possible to find 
        \begin{align}
          \pi' \in   \argmin_{\pi \in \Pi_\bench} \sum_{i=1}^n c\ind{i} \cdot \mathbb{I}\{\pi(x\ind{i}) =a\ind{i}\}, \label{eq:tosolve}
        \end{align}
        in $ O(\poly(n,d,A) \cdot(9n^2 A^2/d)^{(d+1)^2})$ time.  
    \end{lemma}
   
\begin{proof}
    Note that to solve \eqref{eq:tosolve}, it suffices to enumerate points in the set 
    \begin{align}
    \text{Val}_{\bench}  \coloneqq   \{ (\pi(x\ind{1}),\dots, \pi(x\ind{n})) \mid \pi \in \Pi_\bench \} \subseteq \cA^n, \label{eq:cadt}
    \end{align} 
    and take $\pi'$ to be the policy corresponding to the point in $\text{Val}_{\bench}$ that minimizes the objective in \eqref{eq:tosolve}. We know by \cref{lem:growth} that the cardinality of $\text{Val}_{\bench}$ is at most $(9^2 n A^2/d)^{(d+1)^2}$, but we still need a way of enumerating this set efficiently (in the dimension is constant). To do this, we will use the characterization of $\Pi_\bench$ in the proof of \cref{lem:growth} to reduce our problem to enumerating regions of a Euclidean space where certain polynomials change sign configurations. 

   As argued in the proof of \cref{lem:growth}, we have that for any $\pi \in \Pi_\bench$, there exist $\theta_1,\dots,\theta_d$, $\tilde\theta_1, \tilde\theta_2\in \bbB(H)$, and $\gamma>0$ such that:
    \begin{gather}
    \pi(\cdot)= \bar\pi(\cdot;\theta_{1:d},\tilde\theta_{1:2},\gamma),
    \intertext{where}
    \bar\pi(\cdot;\theta_{1:d},\tilde\theta_{1:2},\gamma) \coloneqq \left(1 - \prod_{a,a'\in \cA, i\in[d]}\mathbb{I}\left\{ \varphih(\cdot,a,a')^\top\theta_i \ge \gamma \right\}\right) \cdot \tilde\pi(\cdot;\tilde\theta_1) + \prod_{a,a'\in \cA, i\in[d]}\mathbb{I}\left\{ \varphih(\cdot,a,a')^\top\theta_i < \gamma \right\}  \cdot \tilde\pi(\cdot;\tilde\theta_2), \label{eq:polcc}
    \end{gather}
    and $\tilde \pi(\cdot; \theta) = \argmax_{a\in \cA}\phi(\cdot,a)^\top \theta$. Furthermore, if we let $y\ind{j}_a \coloneqq \phi(x\ind{j},a)$ for $j\in[n]$ and $a\in \cA$, then for any $\theta_{1:d},\tilde\theta_{1:2}\in \bbB(H)$ and $\gamma\in \reals$, the value of the tuple \[(\pibar(x\ind{1}_1;\theta_{1:d},\tilde\theta_{1:2},\gamma), \dots,\pibar(x\ind{n};\theta_{1:d},\tilde\theta_{1:2},\gamma))\in \cA^n\] is completely determined by the signs of the linear functions (see proof of \cref{lem:growth}):  
    \begin{align}
    P_{i,j,a,a'}(\theta_{1:d},\tilde\theta_{1:2},\gamma) & \coloneqq \theta_i^\top y\ind{j}_{a} - \theta_i^\top y\ind{j}_{a'}  -\gamma, \nn\\
      \wtilde P_{1,j,a,a'}( \theta_{1:d},\tilde\theta_{1:2},\gamma) &\coloneqq \tilde \theta_1^\top  y\ind{j}_{a} -  \tilde \theta_1^\top  y\ind{j}_{a'}, \nn\\
      \wtilde P_{2,j, a,a'}(\theta_{1:d},\tilde\theta_{1:2},\gamma) &\coloneqq \tilde \theta_2^\top  y\ind{j}_{a} -  \tilde \theta_2^\top  y\ind{j}_{a'}.
    \end{align}
    In other words, if we let $(S_{i,j,a,a'},\wtilde S_{1,j,a,a'},\wtilde S_{2,j,a,a'}) \coloneqq  (\sgn(P_{i,j,a,a'}),\sgn(\wtilde P_{1,j,a,a'}),\sgn(\wtilde P_{2,j,a,a'}))$, then there is a surjective mapping from 
    \begin{align}
     &\text{Sign}_{\bench} \nn \\
      &\coloneqq  \{ (S_{i,j,a,a'}(\theta_{1:d},\tilde\theta_{1:2}, \gamma), \wtilde{S}_{1,j,a,a'}(\theta_{1:d},\tilde\theta_{1:2},\gamma), \wtilde{S}_{2,j,a,a'}(\theta_{1:d},\tilde\theta_{1:2},\gamma))_{i\in[d], j\in[n],a,a'\in \cA} \mid \theta_{1:d},\tilde\theta_{1:2}\in \bbB(H), \gamma \in \reals  \}
    \end{align}
   to the set $\text{Val}_{\bench}$ in \eqref{eq:cadt}. Note that the elements of the set $\text{Sign}_{\bench}$ take values in the finite set \[(\{-1,1\}\times \{-1,1\}\times \{-1,1\})^{n d A^2},\] and thus $\text{Sign}_{\bench}$ induces a partition over $\bbB(H)^{d+2} \times \reals$, where each region of the partition corresponds to values of $(\theta_{1:d},\tilde\theta_{1:2}, \gamma)$ mapping to a fixed sign tuple:
   \begin{align}
    (S_{i,j,a,a'}(\theta_{1:d},\tilde\theta_{1:2}, \gamma), \wtilde{S}_{1,j,a,a'}(\theta_{1:d},\tilde\theta_{1:2},\gamma), \wtilde{S}_{2,j,a,a'}(\theta_{1:d},\tilde\theta_{1:2},\gamma))_{i\in[d], j\in[n],a,a'\in \cA}.
   \end{align}  
   Thus, enumerating the elements of $\text{Val}_{\bench}$ reduces to enumerating the regions of this partition. Since $P_{i,j,a,a'}, \wtilde{P}_{1,j,a,a'}, \wtilde{P}_{2,j,a,a'}$ are linear functions, \cref{lem:feasibilitypol} implies that enumerating the elements of the partition induced by $\text{Sign}_{\bench}$ can be done using $O(3 n d A^2 \cdot (9 n^2 A^2/d)^{(d+1)^2})$ calls to an Oracle for checking the feasibility of a linear program over $\reals^{(d+1)^2}$ with at most $3d n A^2$ constraints; this Oracle can be implemented in $\poly(d,A, n)$ time. We now explain what exactly it means to enumerate elements of the partition induced by $\text{Sign}_{\bench}$.

  \paragraphi{Enumerating partition elements} By calling \cref{alg:BFS} with the input vectors corresponding to the linear functions $(P_{i,j,a,a'}, \wtilde{P}_{1,j,a,a'}, \wtilde{P}_{2,j,a,a'})_{i\in[d],j\in[n], a,a'\in \cA}$, we get a set $\cG$ of size at most $(9 n^2 A^2/d)^{(d+1)^2}$ such that for any tuple of signs $\text{tup} \in \text{Sign}_{\bench}$, there is a set $\cV= \{v_1,\dots, v_N\}\in \cG$ of vectors in $\reals^K$, where $N \coloneqq 3 dn A^2$ and $K \coloneqq  (d+1)^2$ such that $\bigcap_{i\in [N]}\{\theta \mid \theta^\top v_i \geq 0\} \neq \emptyset$ and 
  \begin{align}
    \text{tup} = (\sgn(\theta^\top v_1), \dots, \sgn(\theta^\top v_N)), \quad \forall \theta \in \bigcap_{i\in [N]}\{\tilde\theta \mid \tilde\theta^\top v_i \geq 0\}.
  \end{align} 
  This means that we can enumerate and evaluate the elements of $\text{Sign}_{\bench}$ (and thus also $\text{Val}_\bench$) by enumerating the elements of $\cG$. 
\end{proof}

\begin{algorithm}[H]
    \caption{Breath-First-Search for enumerating the state space regions corresponding to different combination of signs of a set of linear functions.}
	\label{alg:BFS}
	\begin{algorithmic}[1]
\setstretch{1.2}
\Statex[0]{\bfseries input:} Vectors $v_1,\dots, v_N\in \reals^K$. \hfill \algcommentlight{These correspond to linear functions $\theta \mapsto v_i^\top \theta$.}
\State Set $\cG_1\gets \emptyset$. \label{line:initi}
\If{$\{\theta \in \reals^K \mid \theta^\top v_1\geq 0  \} \neq \emptyset $} \label{eq:checkfeasibility_pre}
\State $\cG_{1} \gets \cG_{1}\cup\{ \{v_1\}\}$.
\EndIf
\If{$\{\theta \in \reals^N\mid \theta^\top v_1 <0 \} \neq \emptyset$} \label{eq:checkfeasibility2_pre}
\State $\cG_{1} \gets \cG_{1}\cup \{  \{-v_1\}\}$. \label{line:endi}
\EndIf
\For{$i=2,\dots, N$} 
\State Set $\cG_{i}\gets \emptyset$. \label{line:initii}
\For{$\cV \in \cG_{i-1}$}
\If{$\bigcap_{v\in \cV \cup \{v_i\}}\{\theta \in \reals^N \mid \theta^\top v \geq 0 \}  \neq \emptyset $} \label{eq:checkfeasibility}
\State $\cG_{i} \gets \cG_{i}\cup\{\cV\cup \{v_i\}\}$.
\EndIf
\If{$\bigcap_{v\in \cV \cup\{-v_i\}}\{\theta \in \reals^N\mid \theta^\top v \geq 0 \}  \neq \emptyset$} \label{eq:checkfeasibility2}
\State $\cG_{i} \gets \cG_{i}\cup \{ \cV\cup \{-v_i\}\}$. \label{line:endii}
\EndIf
\EndFor
\EndFor
\State \textbf{return} $\cG_N$.
\end{algorithmic}
\end{algorithm}

 \begin{lemma}
\label{lem:feasibilitypol}
Let $N,K\in \mathbb{N}$ be given. Then, for any input vectors $v_1,\dots, v_N\in \reals^K$, \cref{alg:BFS} returns a set $\cG_N$ such that for any tuple of signs 
\begin{align}
\text{tup} \in \{(\sgn(v_1^\top \theta),\dots,\sgn(v_N^\top \theta))\mid \theta \in \reals^K\},
\end{align} 
there exists an element $\cV \in \cG_N$ such that $\bigcap_{v\in \cV}\{\tilde\theta\in \reals^K \mid v^\top \tilde\theta \geq 0 \} \neq \emptyset$ and
\begin{align}
    \text{tup} = (\sgn(v_1^\top \theta),\dots,\sgn(v_N^\top \theta)),\quad  \forall \theta \in \bigcap_{v\in \cV}\{\tilde\theta\in \reals^K \mid v^\top \tilde\theta \geq 0 \}.
\end{align}
Furthermore, the algorithm makes at most $N \cdot (8e N/K)^K$ calls to an Oracle for checking the feasibility of a linear program over $\reals^K$ with at most $N$ constraints.
\end{lemma}
\cref{lem:feasibilitypol} implies that the elements of set $\cG_N$ returned by \cref{alg:BFS} characterizes the partition over $\reals^K$ that the set 
\begin{align}
    \{(\sgn(v_1^\top \theta),\dots,\sgn(v_N^\top \theta))\mid \theta \in \reals^K\}
\end{align} 
induces.
\begin{proof}[Proof of \cref{lem:feasibilitypol}]
We will show by induction over $i=1,\dots, N$ that the set $\cG_i$ at the end of the $i$-th iteration of the outer loop of \cref{alg:BFS} satisfies the property that for any tuple of signs 
\begin{align}
\text{tup} \in \{(\sgn(v_1^\top \theta),\dots,\sgn(v_i^\top \theta))\mid \theta \in \reals^K\},
\end{align} 
there exists an element $\cV \in \cG_i$ such that $\bigcap_{v\in \cV}\{\tilde\theta\in \reals^K \mid  v^\top \tilde\theta \geq 0 \} \neq \emptyset$ and
\begin{align}
    \text{tup} = (\sgn(v_1^\top \theta),\dots,\sgn(v_i^\top \theta)),\quad  \forall \theta \in \bigcap_{v\in \cV}\{\tilde\theta\in \reals^K \mid  v^\top \tilde\theta \geq 0 \}.
\end{align}
Instantiating this with $i = n$ implies the first claim of the lemma. 
\paragraphi{Base case $i=1$} From \cref{line:initi}-\cref{line:endi}, the set $\cG_1$ consists of:
\begin{itemize}
\item The elements $\{-v_i\}$ and $\{v_i\}$ if $\{\theta\in \reals^K\mid \theta^\top v_i\geq 0\} \neq \emptyset$ and $\{\theta\in \reals^K\mid \theta^\top v_i< 0\} \neq \emptyset$;
\item The element $\{-v_i\}$ if $\{\theta\in \reals^K\mid \theta^\top v_i\geq 0\}= \emptyset$ and $\{\theta\in \reals^K\mid \theta^\top v_i< 0\} \neq \emptyset$;
\item The element $\{v_i\}$ if $\{\theta\in \reals^K\mid \theta^\top v_i\geq 0\} \neq \emptyset$ and $\{\theta\in \reals^K\mid \theta^\top v_i< 0\} = \emptyset$.
\end{itemize}
In all cases, we have for any \begin{align}
    \text{tup} \in \{ \sgn(v_1^\top \theta)\mid \theta \in \reals^K\},
    \end{align} 
    there exists an element $\cV \in \cG_i$ such that $\bigcap_{v\in \cV}\left\{\tilde\theta\in \reals^K \mid v^\top \tilde\theta \geq 0 \right\}\neq \emptyset$ and
    \begin{align}
        \text{tup} = \sgn(v_1^\top \theta) ,\quad  \forall \theta \in \bigcap_{v\in \cV}\left\{\tilde\theta\in \reals^K \mid v^\top \tilde\theta \geq 0 \right\}.
    \end{align}
\paragraphi{General case} Now, let $i\in [N-1]$ and suppose that we have the property that for any \[\text{tup} \in \{(\sgn(v_1^\top \theta),\dots,\sgn(v_i^\top \theta))\mid \theta \in \reals^K\},\] there exists an element $\cV \in \cG_i$ such that $\bigcap_{v\in \cV}\left\{\tilde\theta\in \reals^K \mid v^\top \tilde\theta \geq 0 \right\}\neq \emptyset$ and
\begin{align}
    \text{tup} = (\sgn(v_1^\top \theta),\dots,\sgn(v_i^\top \theta)),\quad  \forall \theta \in \bigcap_{v\in \cV}\left\{\tilde\theta\in \reals^K \mid v^\top \tilde\theta \geq 0 \right\}.
\end{align}
We show that this property holds for $i$ replaced by $i+1$. Let
\begin{align}
    \text{tup}' \in  \{(\sgn(v_1^\top \theta),\dots,\sgn(v_{i+1}^\top \theta))\mid \theta \in \reals^K\}.
\end{align}
Further, let $\theta'\in \reals^K$ be such that $\text{tup}' \in (\sgn(v_1^\top \theta'),\dots,\sgn(v_{i+1}^\top \theta'))$ and define 
\begin{align}
   \widebar{\text{tup}}' \coloneqq  (\sgn(v_1^\top \theta'),\dots,\sgn(v_{i}^\top \theta'));
   \intertext{Note that}
   \text{tup}' = (\widebar{\text{tup}}'_1,\dots,\widebar{\text{tup}}'_i,\sgn(v_{i+1}^\top \theta')). \label{eq:note}
\end{align}
By the induction hypothesis, there exists an element $\cV' \in \cG_i$ such that 
\begin{align}
    \widebar{\text{tup}}' = (\sgn(v_1^\top \theta),\dots,\sgn(v_i^\top \theta)),\quad  \forall \theta \in \bigcap_{v\in \cV'}\left\{\tilde\theta\in \reals^K \mid v^\top \tilde\theta \geq 0 \right\}\neq \emptyset. \label{eq:forall}
\end{align}
On the other hand, by \cref{line:initii}-\cref{line:endii} of \cref{alg:BFS}, the set $\cG_{i+1}$ at the end of iteration $i+1$ contains:
\begin{itemize}
    \item The elements $\cV' \cup \{-v_{i+1}\}$ and $\cV' \cup \{v_{i+1}\}$ if 
    \begin{align}
    \bigcap_{v\in \cV' \cup \{v_{i+1}\}} \{\theta \in \reals^K \mid \theta^\top v \geq 0\}\neq\emptyset, \quad \text{and} \quad 
    \bigcap_{v\in \cV'\cup \{-v_{i+1}\}} \{\theta \in \reals^K \mid \theta^\top v \geq 0\} \neq \emptyset; \label{eq:notempty0}
    \end{align} 
    In this case, by \eqref{eq:note} and the property of $\cV'$ in \eqref{eq:forall}, there exists $s\in\{-,+\}$ such that
    \begin{align}
        \text{tup}' = (\sgn(v_1^\top \theta),\dots,\sgn(v_{i+1}^\top \theta)),\quad  \forall \theta \in \bigcap_{v\in \cV' \cup \{s v_{i+1}\}}\left\{\tilde\theta\in \reals^K \mid v^\top \tilde\theta \geq 0 \right\} \stackrel{\eqref{eq:notempty0}}{\neq} \emptyset.
    \end{align} 
    \item The element $\cV'\cup \{-v_{i+1}\}$ if 
    \begin{align}
        \bigcap_{v\in \cV' \cup \{v_{i+1}\}} \{\theta \in \reals^K \mid \theta^\top v \geq 0\}=\emptyset,  \quad 
    \text{and} \quad 
    \bigcap_{v\in \cV' \cup \{-v_{i+1}\}} \{\theta \in \reals^K \mid \theta^\top v \geq 0\} \neq \emptyset; \label{eq:notempty}
    \end{align}
    In this case, by \eqref{eq:note} and the property of $\cV'$ in \eqref{eq:forall}, we have $\sgn(v_{i+1}^\top \theta')=-1$ and 
    \begin{align}
        \text{tup}' = (\sgn(v_1^\top \theta),\dots,\sgn(v_{i+1}^\top \theta)),\quad  \forall \theta \in \bigcap_{v\in \cV' \cup \{-v_{i+1}\}}\left\{\tilde\theta\in \reals^K \mid v^\top \tilde\theta \geq 0 \right\}\stackrel{\eqref{eq:notempty}}{\neq} \emptyset.
    \end{align} 
    \item The element $\cV' \cup \{v_{i+1}\}$ if 
    \begin{align}
        \bigcap_{v\in \cV' \cup \{v_{i+1}\}} \{\theta \in \reals^K \mid \theta^\top v \geq 0\}\neq \emptyset, \quad \text{and} \quad
        \bigcap_{v\in \cV' \cup \{-v_{i+1}\}} \{\theta \in \reals^K \mid \theta^\top v \geq 0\}= \emptyset. \label{eq:notempty1}
    \end{align}
    In this case, by \eqref{eq:note} and the property of $\cV'$ in \eqref{eq:forall}, we have $\sgn(v_{i+1}^\top \theta')=+1$ and 
    \begin{align}
        \text{tup}' = (\sgn(v_1^\top \theta),\dots,\sgn(v_{i+1}^\top \theta)),\quad  \forall \theta \in \bigcap_{v\in \cV' \cup \{v_{i+1}\}}\left\{\tilde\theta\in \reals^K \mid v^\top \tilde\theta \geq 0 \right\}\stackrel{\eqref{eq:notempty1}}{\neq} \emptyset.
    \end{align} 
    \end{itemize}
    In all cases, the set $\cG_{i+1}$ at the end of iteration $i+1$ will contain an element $\cV$ such that $\bigcap_{v\in \cV}\left\{\tilde\theta\in \reals^K \mid v^\top \tilde\theta \geq 0 \right\}\neq \emptyset$ and
    \begin{align}
        \text{tup}' = (\sgn(v_1^\top \theta),\dots,\sgn(v_{i+1}^\top \theta)),\quad  \forall \theta \in \bigcap_{v\in \cV}\left\{\tilde\theta\in \reals^K \mid v^\top \tilde\theta \geq 0 \right\}.
    \end{align}
    This completes the induction step. We now bound the computational cost of running the algorithm.
    \paragraph{Computational cost}
    The computational cost of \cref{alg:BFS} is bounded by $N \cdot |\cG_N|$ (due to the two for loops in the algorithm) times the cost of checking the feasibility of a linear program. Thus, it suffices to bound the number of elements in $\cG_N$. The number of elements in $\cG_N$ corresponds to the number of configurations of signs of the linear functions (polynomials of degree 1) $(\theta \mapsto v_i^\top \theta)_{i\in [N]}$. And so, by \cref{lem:poly}, $|\cG_N|$ is at most $(8 e N/K)^K$. This completes the proof.
\end{proof}

    \section{Upper Bound on the Growth Function of $\Pi_\bench$ (Proof of \cref{lem:growth})}
    \label{sec:otherproofs}

  \begin{proof}[Proof of \cref{lem:growth}]
    Some aspects of this proof are inspired by proofs in \citep{jin2023upper}.
    
    Fix $n\in \mathbb{N}$ and $h\in[H]$. Note that by definition of $\Pi_\bench$, we have that for any $\pi \in \Pi_\bench$, there exist $\theta_1,\dots,\theta_d$, $\tilde\theta_1, \tilde\theta_2\in \bbB(H)$, and $\gamma>0$ such that:
    \begin{gather}
    \pi(\cdot)= \bar\pi(\cdot;\theta_{1:d},\tilde\theta_{1:2},\gamma),
    \intertext{where}
    \bar\pi(\cdot;\theta_{1:d},\tilde\theta_{1:2},\gamma) \coloneqq \left(1 - \prod_{a,a'\in \cA, i\in[d]}\mathbb{I}\left\{ \varphih(\cdot,a,a')^\top\theta_i \ge \gamma \right\}\right) \cdot \tilde\pi(\cdot;\tilde\theta_1) + \prod_{a,a'\in \cA, i\in[d]}\mathbb{I}\left\{ \varphih(\cdot,a,a')^\top\theta_i < \gamma \right\}  \cdot \tilde\pi(\cdot;\tilde\theta_2),
    \end{gather}
    and $\tilde \pi(\cdot; \theta) = \argmax_{a\in \cA}\phi(\cdot,a)^\top \theta$; note that the expression of $\bar\pi(\cdot;\theta_{1:d},\tilde\theta_{1:2},\gamma)$ is equivalent to the right-hand side of \eqref{eq:benchpol} with $\pi'(\cdot)\equiv \tilde\pi(\cdot; \tilde\theta_1)$ and $\pi''(\cdot) \equiv \tilde \pi(\cdot ; \tilde\theta_2)$. Thus, we may write the growth function $\cG_h(\Pi_\bench, n)$ as 
    \begin{align}
       \cG_{h}(\Pi_\bench,n) = \max_{(x_1,\dots,x_n)\in \cX_h^n} \left| \left\{(\pibar(x_1;\theta_{1:d},\tilde\theta_{1:2},\gamma), \dots,\pibar(x_d;\theta_{1:d},\tilde\theta_{1:2},\gamma)) \mid \theta_{1:d},\tilde\theta_{1:2}\in \bbB(H), \gamma \in \reals \right\}\right|  .
    \end{align}
    Moving forward, we let 
    \begin{align}
        (x'_1,\dots,x'_n) \in \argmax_{(x_1,\dots,x_n)  \in \cX_h^n} \left| \left\{(\pibar(x_1;\theta_{1:d},\tilde\theta_{1:2}, \gamma), \dots,\pibar(x_d;\theta_{1:d},\tilde\theta_{1:2},\gamma)) \mid \theta_{1:d},\tilde\theta_{1:2}\in \bbB(H), \gamma \in \reals \right\}\right|,
        \intertext{and note that by definition of $\cG_h(\Pi_\bench,n)$, we have}
        \cG_h(\Pi_\bench,n) =  \left|\left\{(\pibar(x'_1;\theta_{1:d},\tilde\theta_{1:2},\gamma), \dots,\pibar(x'_d;\theta_{1:d},\tilde\theta_{1:2},\gamma)) \mid \theta_{1:d},\tilde\theta_{1:2} \in \bbB(H) \right\}\right|.
    \end{align}
    Finally, define the vectors $y_{j,a}\coloneqq \phi(x_j',a)\in \reals^d$ for $j\in[n]$ and $a\in \cA$. With this, observe that for any $\theta_{1:d},\tilde\theta_{1:2}\in \bbB(H)$ and $\gamma\in \reals$, the value of $(\pibar(x'_1;\theta_{1:d},\tilde\theta_{1:2},\gamma), \dots,\pibar(x'_d;\theta_{1:d},\tilde\theta_{1:2},\gamma))$ is fully determined by the signs of the following functions:
    \begin{align}
    P_{i,j,a,a'}(\theta_{1:d},\tilde\theta_{1:2},\gamma) & \coloneqq \theta_i^\top y_{j,a} - \theta_i^\top y_{j,a'}  -\gamma, \nn\\
      \wtilde P_{1,j,a,a'}( \theta_{1:d},\tilde\theta_{1:2},\gamma) &\coloneqq \tilde \theta_1^\top  y_{j,a} -  \tilde \theta_1^\top  y_{j,a'}, \nn\\
      \wtilde P_{2,j, a,a'}(\theta_{1:d},\tilde\theta_{1:2},\gamma) &\coloneqq \tilde \theta_2^\top  y_{j,a} -  \tilde \theta_2^\top  y_{j,a'}.
    \end{align}
    In fact, we have that 
    \begin{itemize}
        \item The signs of $P_{i,j,a,a'}(\theta_{1:d},\tilde\theta_{1:2},\gamma)$ for $i\in[d]$, $j\in [n]$, and $a,a'\in\cA$ determine the value of \[\prod_{a,a'\in \cA, i\in[d]}\mathbb{I}\left\{ \varphih(x_j',a,a')^\top\theta_i < \gamma \right\}\] in the definition of $\pibar(x_j'; \theta_{1:d},\tilde\theta_{1:2},\gamma)$;
        \item The signs of $\wtilde P_{1,j,a,a'}( \theta_{1:d},\tilde\theta_{1:2},\gamma)$ for $j\in[n]$ and $a,a'\in\cA$  determine the value of $\tilde\pi(\cdot;\tilde\theta_1)$ in the definition of $\pibar(\cdot; \theta_{1:d},\tilde\theta_{1:2},\gamma)$; and
        \item The signs of $\wtilde P_{2,j,a,a'}( \theta_{1:d},\tilde\theta_{1:2},\gamma)$ for $j\in [n]$, and $a,a'\in\cA$  determine the value of $\tilde\pi(\cdot;\tilde\theta_2)$ in the definition of $\pibar(\cdot; \theta$$_{1:d},\tilde\theta_{1:2},\gamma)$.
    \end{itemize} 
    Therefore, $\cG_h(\Pi_\bench,n)$ is bounded by the number of configuration of signs of the tuples 
    \begin{align}
        (P_{i,j,a,a'}, \wtilde{P}_{1,j,a,a'}, \wtilde{P}_{2,j,a,a'})_{i\in[d], j\in[n],a,a'\in \cA}.\label{eq:same}
    \end{align} Since for each $i,j\in[d]$ and $a,a'\in \cA$, $P_{i,j,a,a'}$, $\wtilde{P}_{1,j,a,a'}$, and $\wtilde P_{2,j,a,a'}$ are polynomials in $(\theta_{1:d},\tilde\theta_{1:2},\gamma)$ of degree 1, \cref{lem:poly} instantiated with $(p, N, K)=(1,3 d n A^2, d^2 +2d+1)$ implies that the number of configuration of signs of the functions in \eqref{eq:same} is at most $(24 e nA^2/d)^{d^2 + 2d+1}$, and so we have 
    \begin{align}
        \cG_h(\Pi_\bench,n)\leq (24 e n A^2/d)^{d^2 + 2d} \leq (9^2 n A^2/d)^{(d+1)^2}.
    \end{align}  
    Combining this with the fact that $\cG(\Pi_\bench, n) = \max_{h\in[H]} \cG_h(\Pi_\bench, n)$ completes the proof.
    \end{proof}

    \section{Helper Lemmas}
 	\label{sec:helper}
\begin{lemma}[\cite{goldberg1993bounding}]
    \label{lem:poly}
    Let $\{P_1, \dots, P_N\}$
 be $N$ polynomials of degree at most $p$ in $K$ reals variables with $N\geq K$, then the number of different configurations of signs of $\{P_1,\dots, P_N\}$ is at most $(8 e p N/K)^K$.
\end{lemma}

\begin{lemma}
	\label{lem:rademacher}
	Let $\Pi'\subseteq \Pi$, $\pi \in \Pi$, $B>0$, and $n\in \mathbb{N}$ be given. Further, let $(\x_{1:H}\ind{i},\a_{1:H}\ind{i})_{i\in[n]}$ be $n$ i.i.d.~trajectories generated according to $\P^{\pi}$. Then, for any function $f:(\bigtimes_{h=1}^H \cX_h) \times \cA^H\rightarrow [-B,B]$, $h\in[H]$, and $\delta \in(0,1)$, there is an event of probability at least $1-\delta$ such that for all $\pi'\in \Pi'$:
	\begin{align}
		& \left|\E^{\pi}[\mathbb{I}\{\pi'(\x_h)=\a_h\} \cdot f(\x_{1:H},\a_{1:H})] - \frac{1}{n}\sum_{i\in[n]}\mathbb{I}\{\pi'(\x\ind{i}_h)=\a\ind{i}_h\}\cdot  f(\x_{1:H}\ind{i},\a_{1:H}\ind{i}) \right|\leq 4B\sqrt{\frac{\log(2\cG_h(\Pi',n)/\delta)}{n}}. %
	\end{align}
\end{lemma}
\begin{proof}
	Fix $f:(\bigtimes_{h=1}^H \cX_h) \times \cA^H\rightarrow [-B,B]$, $h\in[H]$, and $\delta \in(0,1)$.
By a standard Rademacher complexity argument (see e.g.~\cite[Section 11]{mohri2012}), there is an event of probability at least $1-\delta$ such that for all $\pi'\in \Pi'$:
\begin{align}
&\left|	\E^{\pi}[\mathbb{I}\{\pi'(\x_h)=\a_h\} \cdot f(\x_{1:H},\a_{1:H})]-  \frac{1}{n}\sum_{i\in[n]}\mathbb{I}\{\pi'(\x\ind{i}_h)=\a\ind{i}_h\}\cdot  f(\x_{1:H}\ind{i},\a_{1:H}\ind{i}) \right| \\
& \leq 2\mathfrak{R}_n + B\sqrt{\frac{\log(2/\delta)}{2n}}, \label{eq:al}
\end{align}
where 
\begin{align}
	\mathfrak{R}_n \coloneqq \E_{(\x\ind{i}_{1:H},\a_{1:H}\ind{i})_{i\in[d]}}\E_{\bm{\sigma}_{1:n}} \left[\sup_{\pi'\in\Pi'} \frac{1}{n}\sum_{i\in[n]}\bm{\sigma}_i\cdot\mathbb{I}\{\pi'(\x\ind{i}_h)=\a\ind{i}_h\}\cdot  f(\x_{1:H}\ind{i},\a_{1:H}\ind{i})\right], \label{eq:sed}
\end{align}
and $\bm{\sigma}_1,\dots, \bm{\sigma}_n$ are i.i.d.~Rademacher random variables. Now, by the weighted version of Massart's lemma (\cref{lem:massart}), we have for any $(x_{1:H}\ind{i},a_{1:H}\ind{i})_{i\in[n]}$: \begin{align}
	\E_{\bm{\sigma}_{1:n}} \left[\sup_{\pi'\in\Pi'} \frac{1}{n}\sum_{i\in[n]}\bm{\sigma}_i\cdot\mathbb{I}\{\pi'(x\ind{i}_h)=a\ind{i}_h\}\cdot  f(x_{1:H}\ind{i},a_{1:H}\ind{i})\right]\leq B \sqrt{2n \log |\cJ|}, \label{eq:le}
\end{align}
where \[\cJ \coloneqq  \left\{ \left(\mathbb{I}\{\pi'(x\ind{1}_h)=a\ind{1}_h\},\dots, \mathbb{I}\{\pi'(x\ind{n}_h)=a\ind{n}_h\} \right) \mid (x\ind{i}_{1:H},a_{1:H}\ind{i})\in \left(\bigtimes_{h=1}^H \cX_h\right) \times \cA  \right\}. \]
Note that $|\cJ| \leq \cG_h(\Pi',n)$. Plugging this in \eqref{eq:le} and using \eqref{eq:al} and \eqref{eq:sed}, we get the desired result.
\end{proof}

We now state a weighted version of Massart's lemma, which we use in the proof of \cref{lem:rademacher}.
\begin{lemma}
	\label{lem:massart}
	Let $n\in\mathbb{N}$ and $B>0$ be given and let $\cJ\subseteq \reals^n$ be a finite set of points with $r =\max_{x_{1:n}\in \cJ} \|x\|$. Further, let $\bm{\sigma}_1,\dots, \bm{\sigma}_n$ be i.i.d.~Rademacher random variables and let $\bm{w}_1,\dots,\bm{w}_n$ be i.i.d.~random variables in $[-B,B]$. Then, we have 
	\begin{align}
\E_{\bm{w}_{1:n}}		\E_{\bsigma_{1:n}} \left[\max_{x_{1:n}\in \cJ} \sum_{i=1}^n  \bsigma_i \cdot x_i \cdot \w_i  \right] \leq rB \sqrt{2 \log |\cJ|}.
	\end{align}
\end{lemma}
\begin{proof}
	Fix $t>0$. By Jensen's inequality, we have 
	\begin{align}
	\exp\left(t \E_{\w_{1:n}}\E_{\bsigma_{1:n}} \left[ \max_{x_{1:n}\in \cJ} \sum_{i\in[n]} \bsigma_i x_i \w_i \right]\right) & \leq  \E_{\w_{1:n}}\E_{\bsigma_{1:n}} \left[ \exp \left({t \max_{x_{1:n}\in \cJ}\sum_{i\in[n]} \bsigma_{i}x_i \w_i}\right)\right], \nn \\
	& \leq \E_{\w_{1:n}}\E_{\bsigma_{1:n}} \left[\sum_{x_{1:n} \in \cJ}  \exp \left({t \sum_{i\in[n]} \bsigma_{i}x_i \w_i}\right) \right], \nn \\
	& = \sum_{x_{1:n} \in \cJ} \E_{\w_{1:n}}\E_{\bsigma_{1:n}} \left[  \exp \left({t \sum_{i\in[n]} \bsigma_{i}x_i \w_i}\right) \right], \nn \\
	& = \sum_{x_{1:n} \in \cJ} \E_{\w_{1:n}}\E_{\bsigma_{1:n}} \left[ \prod_{i\in[n]} \exp \left({t \bsigma_{i}x_i \w_i}\right) \right], \nn \\
	& = \sum_{x_{1:n} \in \cJ} \prod_{i\in[n]} \E_{\w_{1}}\E_{\bsigma_{1}} \left[  \exp \left({t \bsigma_{1}x_i \w_1}\right) \right],  \quad \text{(by the i.i.d.~assumption)}\nn \\
	& \leq  \sum_{x_{1:n} \in \cJ} \prod_{i\in[n]} \exp \left(\frac{(2tB x_i)^2}8\right) ,  \quad \text{(Hoeffding's lemma)}\nn \\
	& = \sum_{x_{1:n} \in \cJ}  \exp \left(\frac{t^2 B^2}2 \sum_{i=1}^n x_i^2\right),\nn \\
	& =|\cJ|  \exp \left(t^2 B^2 r^2/2\right), \quad \left(\text{since $r =\max_{x_{1:n}\in \cJ} \|x\|$}\right).
	\end{align}
	Taking the logarithm of both sides and dividing by $t$, we get
	\begin{align}
		\E_{\w_{1:n}}\E_{\bsigma_{1:n}} \left[ \max_{x_{1:n}\in \cJ} \sum_{i\in[n]} \bsigma_i x_i \w_i \right] \leq \frac{\log |\cJ|}{t} + \frac{tB^2 r^2}{2}.
	\end{align}
	Taking $t= \frac{\sqrt{2\log |\cJ|}}{B r}$ implies the desired result.
	\end{proof}

\begin{lemma}[Union bound]
Let $T,H\in \mathbb{N}$ and $\delta \in (0,1)$ be given.
Further, let $\cB_1$ be an algorithm that runs in $T\in\mathbb{N}$ iterations. At each iteration, $\cB_1$ makes a sequence of $H$ calls to a subroutine $\cB_2$. Let $\mathfrak{S}$ denote the state space of algorithm $\cB_1$; the space capturing the values of all the internal variables of $\cB_1$. Let $\mathbf{S}\ind{t}_{h,-}\in \mathfrak{S}$ denote the random state of $\cB_1$ immediately before the $h$th call to $\cB_2$ during the $t$th iteration; further, let $\mathbf{S}\ind{t}_{h,+} \in \mathfrak{S}$ denote the random state of $\cB_1$ immediately after this call to $\cB_2$. Suppose that for any $S\ind{t}_{h,-}\in \mathfrak{S}$, there is an event $\cE\ind{t}_{h}(S\ind{t}_{h,-}) \subset \mathfrak{S}$ such that $\P[\mathbf{S}\ind{t}_{h,+}\in \cE\ind{t}_{h}(S\ind{t}_{h,-})]\geq 1-\delta$. Then, with probability at least $1-\delta H T$, for all $t\in[T]$ and $h\in[H]$, we have $\mathbf{S}\ind{t}_{h,+}\in \cE\ind{t}_{h}(\mathbf{S}\ind{t}_{h,-})$.   
\end{lemma}
\begin{proof}
  Let $\cE$ be the event defined by 
  \begin{align}
    \cE \coloneqq \left\{\prod_{t=1}^T \prod_{h=1}^H \mathbb{I}\{\mathbf{S}\ind{t}_{h,+} \in  \cE\ind{t}_{h}(\mathbf{S}\ind{t}_{h,-}) \}=1 \right\}.
  \end{align}  
  We need to show that $\P[\cE]\geq 1-\delta H T$. To this end, we note that by the chain rule of probability, we have
  \begin{align}
   \P[\cE] & = \prod_{t=1}^T \prod_{h=1}^H \E\left[ \P[\mathbf{S}\ind{t}_{h,+}\in \cE\ind{t}_{h}(\mathbf{S}\ind{t}_{h,-})\mid \mathbf{S}\ind{t}_{h,-}]\right],\nn\\
   & \geq \prod_{t=1}^T\prod_{h=1}^H \left(1-\delta\right), \quad \label{eq:penul} \\
   & \geq 1- TH \delta,
\end{align}
where \eqref{eq:penul} follows by the fact that $\P[\mathbf{S}\ind{t}_{h,+}\in \cE\ind{t}_{h}(S\ind{t}_{h,-})]\geq 1-\delta$ for all $S\ind{t}_{h,-}\in \mathfrak{S}$, and the last inequality follows by the fact that for any sequence $x_{1},\dots,x_{T}\in (0,1)$, $\prod_{i\in[T]} (1- x_i)\geq 1- \sum_{i\in[T]}x_i$. 

\end{proof}

\begin{lemma}
	\label{lem:unionbound}
	Let $L>0$ be given and consider a collection of random matrices $\{\bm{M}^z\}_{z\in \cZ}$, where $\cZ$ is some abstract set, such that $\|\bm{M}^z\|_{\op}\le L$, for all $z\in \cZ$. Suppose that for some $\veps' ,\delta\in(0,1)$ and $\cK \subseteq \bbB(1)$, we have that for all $u \in \cK$ and $v\in \bbB(1)$, with probability at least $1-\delta$: 
	\begin{align}
	\forall z\in \cZ,\quad u^\top \bm{M}^z v \leq \veps'. \label{eq:even}
	\end{align}
	Then for any $\veps''\in(0,1)$, with probability at least $1-\delta \cdot (3/\veps'')^{2d}$, we have for all $u\in \cK$, $v \in \bbB(1)$, and $z\in \cZ$:
	\begin{align}
	u^\top \bm{M}^z v \leq \veps' + 2 L \veps''.
	\end{align}
	\end{lemma}
	\begin{proof}
	Let $\cN(\bbB(1),\|\cdot\|,\veps'')$ be an $\veps''$-epsilon net of $\cB(1)$ in $\|\cdot\|$. Further, let $\cK' \coloneqq \cK \cap \cN(\bbB(1),\|\cdot\|, \veps'')$. We note that  
	\begin{align}
	|\cK'| \leq |\cN(\bbB(1),\|\cdot\|,\veps'')| \leq (3/\veps'')^d.	
	\end{align}
	By \eqref{eq:even} and the union bound over elements in $\cK'\times \cN(\bbB(1),\|\cdot\|,\veps'')$, there is an event $\cE$ of probability at least $1-\delta \cdot (3/\veps'')^{2d}$ under which for all $u' \in \cK'$, $v' \in \cN(\bbB(1),\|\cdot\|,\veps'')$, and $z\in \cZ$: 
	\begin{align}
	(u')^\top \bm{M}^z v' \leq \veps'. \label{eq:baseevent}
	\end{align} 
	For the rest of the proof, we condition on $\cE$. Since $\cK'$ is an $\veps''$-net of $\cK$, we have that for any $u\in \cK$ and $v\in \bbB(1)$: 
	\begin{align}
		\inf_{u'\in \cK', v'\in \cN(\bbB(1),\|\cdot\|,\veps'')}\|u-u'\|\vee \|v-v'\|\leq \veps''. \label{eq:netcover}
	\end{align} 
	Therefore, we have that for any $u\in\cK$, $v\in \bbB(1)$, and $z\in \cZ$: 
	\begin{align}
		u^\top \bm{M}^z v & = \inf_{u'\in \cK', v'\in \cN(\bbB(1),\|\cdot\|,\veps'')} \left\{ (u')^\top \bm{M}^z v' + (u')^\top \bm{M}^z (v-v') + (u-u')^\top \bm{M}^z v\right\},  \nn \\
		& \leq \sup_{u'\in \cK', v'\in \cN(\bbB(1),\|\cdot\|,\veps'')} (u')^\top \bm{M}^z v' + \inf_{u'\in \cK', v'\in \cN(\bbB(1),\|\cdot\|,\veps'')} \left\{ (u')^\top \bm{M}^z (v-v') + (u-u')^\top \bm{M}^z v\right\},  \nn \\
		& \leq \veps' + \inf_{u'\in \cK', v'\in \cN(\bbB(1),\|\cdot\|,\veps'')}\left\{ L \|v-v'\| + L \|u-u'\|\right\}, \quad \text{(by \eqref{eq:baseevent} and Cauchy Schwarz)} \nn \\
		& \leq \veps' + 2L \veps'',
	\end{align}
where the last inequality follows by \eqref{eq:netcover}. This completes the proof.
	\end{proof}

			We take \cref{ass:pre-bandits-thm,thm:bandits-book-20.5} from \cite{lattimore2020bandit}.

			\begin{assumption}[Prerequisites for Theorem~\ref{thm:bandits-book-20.5}]
				\label{ass:pre-bandits-thm}
			Let $\lambda>0$.
			For $k\in\mathbb{N}$, let $Y_k$ be random variables taking values in $\reals^d$.
			For some $\theta_\star\in\reals^d$, let $Y_k=\inner{X_k}{\theta_\star}+\eta_k$ for all $k\in\mathbb{N}$.
			Here, $\eta_k$ is a conditionally 1-subgaussian random variable; that is, it satisfies:
			\begin{align}%
			\text{for all } \alpha\in\reals \text{ and } t\ge 1,\,\quad\quad \E[\exp(\alpha \eta_k) \,|\, \cF_{k-1}] \le \exp\left(\frac{\alpha^2}{2}\right) \quad\text{a.s.},\,
			\end{align}
			where $\cF_{k-1}$ is such that $X_1, Y_1,\ldots,X_{k-1},Y_{k-1},X_k$ are $\cF_{k-1}$-measurable.
			\end{assumption}
			\begin{theorem}[\cite{lattimore2020bandit}, Theorem 20.5]
				\label{thm:bandits-book-20.5}
			Let $\zeta\in(0,1)$.
			Under Assumption~\ref{ass:pre-bandits-thm}, with probability at least $1-\zeta$, it holds that for all $k\in\mathbb{N}$,
			\[
			\norm{\hat\theta_k-\theta_\star}_{\Sigma_k(\lambda)} < \sqrt{\lambda}\norm{\theta_\star}_2+\sqrt{2\log\left(1/\zeta\right)+\log\left(\frac{\det \Sigma_k(\lambda)}{\lambda^d}\right)}\,,
			\]
			where for $k\in\mathbb{N}$,
			\begin{align}
			\Sigma_k(\lambda)=\lambda I + \sum_{s=1}^k X_s X_s^\top \quad \text{and}\quad
			\hat\theta_k=\Sigma_k(\lambda)^{-1}\sum_{s=1}^k  X_s Y_s.
			\end{align}
			\end{theorem}

			\begin{lemma}
				\label{lem:elliptical}
					Let $\beta>0$ be given. Then, for any sequence of vectors $u\ind{1},u\ind{2},\dots$ in $\bbB(1)$, we have for all $T\in \mathbb{N}$:
					\begin{align}
					\sum_{t\in[T]}\|u\ind{t}\|_{(\beta I + U\ind{t})^{-1/2}} \leq \sqrt{T d \log (1+T/\beta)},	
					\end{align}
					where $U\ind{t} \coloneqq \sum_{\tau \in[t-1]} u\ind{\tau} (u\ind{\tau})^\top$.
			\end{lemma}
			\begin{proof}
				By \cite[Lemma 11]{hazan2007logarithmic}, we have  
\begin{align}
	\sum_{t\in[T]} \|u\ind{t}\|^2_{(\beta I + U\ind{t})^{-1}}&=\sum_{t\in[T]} (u\ind{t})^\top (\beta I + U\ind{t})^{-1} u\ind{t} \leq d \log(1+T/\beta).\label{eq:zoo}
\end{align}
Now, by Jensen's inequality, we have 
\begin{align}
\sum_{t\in[T]}\|u\ind{t}\|_{(\beta I + U\ind{t})^{-1}} &\leq T \cdot \frac{1}{T} \sum_{t\in[T]}\sqrt{\|u\ind{t}\|^2_{(\beta I + U\ind{t})^{-1}}}, \nn \\
 & \leq T \sqrt{\frac{1}{T} \sum_{t\in[T]}\|u\ind{t}\|^2_{(\beta I + U\ind{t})^{-1}}},\nn \\
 & \leq  \sqrt{T d \log (1 + T/\beta)}, \label{eq:build}
\end{align}
where the last inequality follows by \eqref{eq:zoo}. This completes the proof.
			\end{proof}

We require the classical performance difference lemma from \citet{kakade2003sample}.
\begin{lemma}[Performance Difference Lemma]\label{lem:pdl}
	Let $\pistar,\pi \in \Pim$ be arbitrary, and $Q_t^\pi$ be as defined in \eqref{eq:laugh}. Then, for any $h\geq 1$,
	\begin{align}
		\ee^{\pistar}\left[ \sum_{t = 1}^h R(\x_t, \a_t) \right] - \ee^{\pi}\left[ \sum_{t = 1}^h R(\x_t, \a_t) \right] = \sum_{t=  1}^h \ee^{\pistar}\left[Q_t^{\pi}(\x_t, \pistar(\x_t)) - Q_t^{\pi}(\x_t, \pi(\x_t)) \right].
	\end{align}
\end{lemma}

			\begin{lemma}[Skip-step value decomposition]
				\label{lem:skipping}
			Let $\cK_{1}\subseteq \cX_1,\dots, \cK_H\subseteq \cX_H$ be arbitrary subsets and let $\pi'_{1:H}, \pihat_{1:H}, \pihat'_{1:H}\in \Pi$ be such that for all $h\in[H]$ and $x\in \cX_h$, $\pihat'_h(x) = \mathbb{I}\{x\in \cK_h\}\cdot \pihath(x)+ \mathbb{I}\{x\not\in \cK_h\}\cdot  \pi'_h(x)$. Further, let $(\tilde{r}_{h}: \cX_h \times \cA \rightarrow \reals)_{h\in[H]}$ be arbitrary functions and define
			\begin{align}
			V_h(x) \coloneqq  \E^{\pihat'}\left[\sum_{\ell=h}^H  \tilde{r}_\ell(\x_\ell,\a_\ell)  \mid  \x_h = x, \a_h = \pihat'_h(x) \right],\label{eq:doit}
			\end{align} 
			for $h\in[H]$ and $x\in \cX_h$. Then, for all $h\in[H]$ and $x\in \cX_h$, we have
			\begin{align}
				V_h(x)&=  \sum_{\ell=h}^H\E^{\pihat}\left[\mathbb{I}\{\x_\ell\not\in \cK_\ell\}\cdot \prod_{k=h}^{\ell-1} \mathbb{I}\{\x_k\in \cK_k \} \cdot V_{\ell}(\x_{\ell}) \mid \x_h=x,\a_h= \pihat'_h(x)\right]\nn \\
				& \quad + \sum_{\ell=h}^H\E^{\pihat}\left[   \prod_{k=h}^{\ell} \mathbb{I}\{\x_k\in \cK_k \} \cdot \tilde r_\ell(\x_\ell,\a_\ell) \mid \x_h=x,\a_h=\pihat'_h(x) \right],
				\label{eq:raisin}
			\end{align}
			with the convention that $\prod_{k=h}^{h-1} \mathbb{I}\{\x_k\in \cK_k \}=1$.
			\end{lemma}
			\begin{proof}
			We prove the result via backward induction over $h=H,\dots,1$. Suppose that \eqref{eq:raisin} holds for $h\in[H]$. We show that it holds for $h-1$. Fix $x\in \cX_{h-1}$. If $x\not\in \cK_{h-1}$, we trivially have that 
			\begin{align}
				\sum_{\ell=h-1}^H\E^{\pihat}\left[\mathbb{I}\{\x_\ell\not\in \cK_\ell\}\cdot \prod_{k=h-1}^{\ell-1} \mathbb{I}\{\x_k\in \cK_k \} \cdot V_{\ell}(\x_{\ell}) \mid \x_{h-1}=x,\a_{h-1}= \pihatho'(x) \right] &= V_{h-1}(x)
				\shortintertext{and}
				\sum_{\ell=h-1}^H\E^{\pihat}\left[   \prod_{k=h-1}^{\ell} \mathbb{I}\{\x_k\in \cK_k \} \cdot \tilde r_\ell(\x_\ell,\a_\ell) \mid \x_{h-1}=x,\a_{h-1}=\pihatho'(x) \right] & = 0.
			\end{align}
			Therefore,
			\begin{align}
				V_{h-1}(x) = & \sum_{\ell=h-1}^H\E^{\pihat}\left[\mathbb{I}\{\x_\ell\not\in \cK_\ell\}\cdot \prod_{k=h-1}^{\ell-1} \mathbb{I}\{\x_k\in \cK_k \} \cdot V_{\ell}(\x_{\ell}) \mid \x_{h-1}=x,\a_{h-1}= \pihatho'(x) \right]\nn \\
				& \quad + \sum_{\ell=h-1}^H\E^{\pihat}\left[   \prod_{k=h-1}^{\ell} \mathbb{I}\{\x_k\in \cK_k \} \cdot \tilde r_\ell(\x_\ell,\a_\ell) \mid \x_{h-1}=x,\a_{h-1}=\pihatho'(x) \right].
				\end{align} 
			Now, suppose that $x \in \cK_{h-1}$. Then, from \eqref{eq:doit} and the fact that $\pihath'(x) = \mathbb{I}\{x\in \cK_{h-1}\}\cdot \pihatho(x)+ \mathbb{I}\{x\not\in \cK_{h-1}\}\cdot  \piho'(x)$, we have
			\begin{align}
			   & V_{h-1}(x)   \nn \\
			   & = \mathbb{I}\{x\in \cK_{h-1}\}\cdot \tilde r_{h-1}(x,\pihatho(x))  \nn \\
			   & \quad  + \E^{\pihat'} \left[\mathbb{I}\{\x_{h-1}\in \cK_{h-1}\}\cdot\sum_{\ell=h}^H   \tilde r_\ell(\x_\ell,\a_\ell)   \mid  \x_{h-1} = x, \a_{h-1} = \pihatho(x) \right], \nn \\
				& = \mathbb{I}\{x\in \cK_{h-1}\}\cdot \tilde r_{h-1}(x,\pihatho(x)) + \E \left[\mathbb{I}\{\x_{h-1} \in \cK_{h-1}\} \cdot V_h(\x_{h})\mid    \x_{h-1} = x, \a_{h-1} = \pihatho(x)\right], \nn \\
			& = \mathbb{I}\{x\in \cK_{h-1}\}\cdot \tilde r_{h-1}(x,\pihatho(x)) \nn \\
			& \quad+ \E \left[\mathbb{I}\{\x_{h-1} \in \cK_{h-1}\} \cdot \mathbb{I}\{\x_{h} \not\in \cK_{h}\} \cdot V_h(\x_{h}) \mid  \x_{h-1} = x, \a_{h-1} = \pihatho(x)\right]\nn \\
			& \quad + \E\left[\mathbb{I}\{\x_{h-1} \in \cK_{h-1}\} \cdot \mathbb{I}\{\x_{h} \in \cK_{h}\} \cdot V_h(\x_{h}) \mid  \x_{h-1} = x, \a_{h-1} = \pihatho(x)\right]. \label{eq:awe}
			\end{align}
			Now, by the induction hypothesis, i.e.~\eqref{eq:raisin}, we have that for all $x'\in \cK_h$:
			\begin{align}
			V_h(x') &= \sum_{\ell=h+1}^H\E^{\pihat}\left[\mathbb{I}\{\x_\ell\not\in \cK_\ell\}\cdot \prod_{k=h}^{\ell-1} \mathbb{I}\{\x_k\in \cK_k \} \cdot V_{\ell}(\x_{\ell}) \mid \x_h=x',\a_h= \pihath(x') \right]\nn \\
			& \quad + \sum_{\ell=h}^H\E^{\pihat}\left[   \prod_{k=h}^{\ell} \mathbb{I}\{\x_k\in \cK_k \} \cdot \tilde r_\ell(\x_\ell,\a_\ell) \mid \x_h=x',\a_h=\pihath(x') \right]. 
			\end{align}
			Plugging this into \eqref{eq:awe} and using the law of total expectation, we get that 
			\begin{align}
			 &   V_{h-1}(x) \nn \\
			 & = \mathbb{I}\{x\in \cK_{h-1}\}\cdot \tilde r_{h-1}(x,\pihatho(x)) \nn \\
			& \quad + \E^{\pihat} \left[\mathbb{I}\{\x_{h-1} \in \cK_{h-1}\} \cdot \mathbb{I}\{\x_{h} \not\in \cK_{h}\} \cdot V_h(\x_{h}) \mid  \x_{h-1} = x, \a_{h-1} = \pihatho(x)\right]\nn \\
			& \quad +  \sum_{\ell=h+1}^H\E^{\pihat}\left[\mathbb{I}\{\x_{h-1} \in \cK_{h-1}\} \cdot \mathbb{I}\{\x_{h} \in \cK_{h}\}\cdot\mathbb{I}\{\x_\ell\not\in \cK_\ell\}\cdot \prod_{k=h}^{\ell-1} \mathbb{I}\{\x_k\in \cK_k \} \cdot V_{\ell}(\x_{\ell}) \mid \x_{h-1}=x,\a_{h-1}=\pihatho(x) \right]\nn \\
			& \quad + \sum_{\ell=h}^H\E^{\pihat}\left[ \mathbb{I}\{\x_{h-1} \in \cK_{h-1}\} \cdot \mathbb{I}\{\x_{h} \in \cK_{h}\}\cdot  \prod_{k=h}^{\ell} \mathbb{I}\{\x_k\in \cK_k \} \cdot \tilde r_\ell(\x_\ell,\a_\ell) \mid x_{h-1}=x,\a_{h-1}=\pihatho(x)\right],\nn \\
			 & = \mathbb{I}\{x\in \cK_{h-1}\}\cdot \tilde r_{h-1}(x,\pihatho(x)) \nn \\
			& \quad + \E^{\pihat} \left[\mathbb{I}\{\x_{h-1} \in \cK_{h-1}\} \cdot \mathbb{I}\{\x_{h} \not\in \cK_{h}\} \cdot V_h(\x_{h}) \mid  \x_{h-1} = x, \a_{h-1} = \pihatho(x)\right]\nn \\
			& \quad +  \sum_{\ell=h+1}^H\E^{\pihat}\left[\mathbb{I}\{\x_\ell\not\in \cK_\ell\}\cdot \prod_{k=h-1}^{\ell-1} \mathbb{I}\{\x_k\in \cK_k \} \cdot V_{\ell}(\x_{\ell}) \mid \x_{h-1}=x,\a_{h-1}=\pihatho(x) \right]\nn \\
			& \quad + \sum_{\ell=h}^H\E^{\pihat}\left[\prod_{k=h-1}^{\ell} \mathbb{I}\{\x_k\in \cK_k \} \cdot \tilde r_\ell(\x_\ell,\a_\ell) \mid x_{h-1}=x,\a_{h-1}=\pihatho(x)\right],\nn \\
			& = \sum_{\ell=h}^H\E^{\pihat}\left[\mathbb{I}\{\x_\ell\not\in \cK_\ell\}\cdot \prod_{k=h-1}^{\ell-1} \mathbb{I}\{\x_k\in \cK_k \} \cdot V_{\ell}(\x_{\ell}) \mid \x_{h-1}=x,\a_{h-1}=\pihatho(x) \right]\nn \\
			& \quad + \sum_{\ell=h-1}^H\E^{\pihat}\left[   \prod_{k=h-1}^{\ell} \mathbb{I}\{\x_k\in \cK_k \} \cdot \tilde r_\ell(\x_\ell,\a_\ell) \mid x_{h-1}=x,\a_{h-1}=\pihatho(x)\right],\nn \\ 
			& = \sum_{\ell=h-1}^H\E^{\pihat}\left[\mathbb{I}\{\x_\ell\not\in \cK_\ell\}\cdot \prod_{k=h-1}^{\ell-1} \mathbb{I}\{\x_k\in \cK_k \} \cdot V_{\ell}(\x_{\ell}) \mid \x_{h-1}=x,\a_{h-1}=\pihatho(x) \right]\nn \\
			& \quad + \sum_{\ell=h-1}^H\E^{\pihat}\left[   \prod_{k=h-1}^{\ell} \mathbb{I}\{\x_k\in \cK_k \} \cdot r_\ell(\x_\ell,\a_\ell) \mid x_{h-1}=x,\a_{h-1}=\pihatho(x)\right],
			\end{align}
			where the last inequality follows by the fact that $x\in \cK_{h-1}$. This shows \eqref{eq:raisin} with $h$ replaced by $h-1$ and completes the induction.
			\end{proof}

\end{document}